# Deep Learning Activation Functions: Fixed-Shape, Parametric, Adaptive, Stochastic, Miscellaneous, Non-Standard, Ensemble

M. M. Hammad

Department of Mathematics and Computer Science, Faculty of Science, Damanhour University, Egypt
Email: m_hammad@sci.dmu.edu.eg
orcid.org/0000-0003-0306-9719

**Abstract**

In the architecture of deep learning models, inspired by biological neurons, activation functions play a pivotal role. They significantly influence the performance of artificial neural networks. By modulating the non-linear properties essential for learning complex patterns, activation functions are fundamental in both classification and regression tasks. This paper presents a comprehensive review of various types of activation functions, including fixed-shape, parametric, adaptive, stochastic/probabilistic, non-standard, and ensemble/combining types. We begin with a systematic taxonomy and detailed classification frameworks that delineates the principal characteristics of activation functions and organizes them based on their structural and functional distinctions. Our in-depth analysis covers primary groups such as sigmoid-based, ReLU-based, and ELU-based activation functions, discussing their theoretical foundations, mathematical formulations, and specific benefits and limitations in different contexts. We also highlight key attributes of activation functions such as output range, monotonicity, and smoothness. Furthermore, we explore miscellaneous activation functions that do not conform to these categories but have shown unique advantages in specialized applications. Non-standard activation functions are also explored, showcasing cutting-edge variations that challenge traditional paradigms and offer enhanced adaptability and model performance. We examine strategies for combining multiple activation functions to leverage complementary properties. The paper concludes with a comparative evaluation of 12 state-of-the-art activation functions, using rigorous statistical and experimental methodologies to assess their efficacy. This analysis not only aids practitioners in selecting and designing the most appropriate activation functions for their specific deep learning tasks but also encourages continued innovation in activation function development within the machine learning community.

**Abbreviations:**

| | | | |
|---|---|---|---|
| AF | Activation Function | MTLU | Multi-bin Trainable Linear Unit |
| APL | Adaptive Piecewise Linear Unit | MeLU | Mexican Rectified Linear Unit |
| BN | Batch Normalization | MPELU | Multiple Parametric Exponential Linear Unit |
| BLU | Bendable Linear Unit | NLReLU | Natural Logarithm-Rectified Linear Unit |
| BReLU | Bounded Rectified Linear Unit | NN | Neural Network |
| BLReLU | Bounded Leaky Rectified Linear Unit | PTELU | Parametric Tan Hyperbolic Linear Unit |
| BIF | Bi-firing Function | PELU | Parametric Exponential Linear Unit |
| BBIF | Bounded Bi-firing Function | PLU | Piecewise Linear Unit |
| BDAA | Bi-modal Derivative Adaptive Activation | PREU | Parametric Rectified Exponential Unit |
| CNN | Convolutional Neural Network | PDELU | Parametric Deformable Exponential Linear Unit |
| CELU | Continuously differentiable Exponential Linear Unit | PSF | Parametric Sigmoid Function |
| DNN | Deep Neural Network | PReLU | Parametric Rectified Linear Unit |
| DReLU | Displaced Rectified Linear Unit | PTanh | Penalized Tanh |
| DSiLU | Derivative Sigmoid-weighted Linear Unit | ReLU | Rectified Linear Unit |
| ELU | Exponential Linear Unit | ReLTanh | Rectified Linear Tanh |
| EReLU | Elastic Rectified Linear Unit | REU | Rectified Exponential Unit |
| EPReLU | Elastic Parametric Rectified Linear Unit | RTReLU | Random Translation Rectified Linear Unit |
| EELU | Elastic Exponential Linear Unit | RReLU | Randomized Rectified Linear Unit |
| EliSH | Exponential Linear Sigmoid SquasHing | ReSech | Rectified Hyperbolic Secant |
| FFNN | Feedforward Neural Network | RNN | Recurrent Neural Network |
| FReLU | Flexible Rectified Linear Unit | SNN | Self-Normalizing Neural Network |
| FELU | Fast Exponential Linear Unit | STanh | Scaled Hyperbolic Tangent |
| GPU | Graphics Processing Unit | SSigmoid | Scaled Sigmoid |
| GD | Gradient Descent | SiLU | Sigmoid-weighted Linear Unit |
| GELU | Gaussian Error Linear Unit | SGD | Stochastic Gradient Descent |
| HardSigmoid | Hard Sigmoid | SRS | Soft-Root-Sign |
| HardTanh | Hard Tanh | SELU | Scaled Exponential Linear Unit |
| LiSHT | Linearly Scaled Hyperbolic Tangent | SReLU | Shifted Rectified Linear Unit |
| LReLU | Leaky Rectified Linear Unit | SignReLU | Sign Rectified Linear Unit |
| LuTU | Look-up Table Unit | SReLU | S-Shaped Rectified Linear Unit |
| LiSA | Linearized Sigmoidal Activation | SLU | SoftPlus Linear Unit |
| MLP | Multilayer Perceptron | SGELU | Symmetrical Gaussian Error Linear Unit |
| MElliott | Modification Elliott | VReLU | V-shaped Rectified Linear Unit |

## 1. Introduction

The synergy of big data and advanced high-performance hardware, particularly Graphics Processing Units (GPUs), has significantly advanced Deep Neural Networks (DNNs), leading to outstanding performance across diverse domains including computer vision, video analytics, pattern recognition, natural language processing, information retrieval, recommender systems, medical diagnosis, financial forecasting and many more [1-10]. Various Neural Network (NN) architectures have been developed to learn abstract features from data, including Multilayer Perceptron (MLP), Recurrent Neural Networks (RNNs), and Convolutional Neural Networks (CNNs) [1-4]. Critical aspects of NNs include weight initialization strategies, the formulation of loss functions, overfitting prevention techniques, learning rate schedules, and adaptive optimization algorithms [1-10].

Activation Functions (AFs) play a critical role in NNs by introducing non-linearity, enabling the networks to learn complex patterns. However, several challenges are associated with the design and use of AFs. One of the fundamental challenges in NN optimization is AF saturation. Certain AFs, such as the sigmoid or hyperbolic tangent (Tanh) functions, tend to saturate when the input values are too large or too small, where the output of an AF becomes very insensitive to small changes in the input (the change in the output becomes very small or negligible) which hampering the learning process. The vanishing and exploding gradients problem is another significant hurdle faced during NN training. In DNNs, gradients can diminish exponentially or explode during backpropagation, making it challenging to update the weights effectively. This phenomenon hinders the convergence of the model and affects its ability to generalize to unseen data. While non-zero centered AFs like Rectified Linear Unit (ReLU) and its variants indeed alleviate the vanishing gradient problem, they can introduce another issue known as "zig-zag" updates. These updates occur when the neuron's output oscillates between positive and negative values, causing the Gradient Descent (GD) updates to zig-zag back and forth. This oscillatory behavior can potentially slow down the learning process, as the network struggles to converge towards the optimal solution. Moreover, the internal covariate shift problem arises when the distribution of network activations changes due to updates in network parameters during training. This can occur with non-zero centered AFs. The issue is that as the parameters change, the distribution of activations in each layer shifts, which can slow down training and make convergence more difficult. To elaborate, when a NN is being trained, the parameters (weights and biases) are updated after each batch of training data. These updates can cause the inputs to each layer (activations from the previous layer) to change their distribution. If the activations are not zero-centered, these shifts can be more pronounced. For example, ReLU activations are non-zero centered since they only output positive values, leading to shifts in the mean and variance of the activations. These shifts in activation distributions mean that each layer needs to continuously adapt to the changes in the distributions of its inputs, which can slow down the training process and make it harder for the network to converge. Furthermore, the dying ReLU problem is another significant issue encountered with ReLU AF in NNs. This problem occurs when a neuron in a NN becomes inactive and stops responding to any input, effectively "dying." This issue arises because the ReLU function outputs zero for any negative input, and if a neuron's weights are updated in such a way that it consistently receives negative inputs, it will always output zero. Thus, individuating new AFs that can potentially improve the results is still an open field of research.

In the following, we present a brief overview of the development of AFs. During the early 1990s, the Sigmoid AF was extensively utilized in NN research. By the late 1990s, the Tanh function became the prevalent AF. Both Sigmoid and Tanh functions are susceptible to the vanishing gradient problem and exhibit saturation behavior, limiting their effectiveness in training DNNs. Several improvements have been proposed based on the Logistic Sigmoid and Tanh AFs [11-27].

On the other hand, in response to these issues, researchers developed alternative AFs, with the ReLU [28, 29] emerging as a prominent solution. By 2017, ReLU had become the most widely adopted AF, surpassing both Sigmoid and Tanh in popularity. ReLU's widespread adoption can be attributed to its ability to alleviate the vanishing gradient problem, making it particularly effective in applications such as speech recognition and computer vision. However, ReLU is associated with several limitations, including non-differentiability at zero, unbounded output, and the "dying ReLU" phenomenon, wherein neurons can become inactive and cease to learn. To address these limitations, researchers proposed various modifications to the standard ReLU [28-54]. For example, linear variants such as the Leaky ReLU (LReLU) [31] and Parametric ReLU (PReLU) [32] introduce a small, non-zero gradient for negative input values, thereby mitigating the dying ReLU problem.

Logistic Sigmoid and Tanh AFs are limited by their tendency to saturate for large positive and negative inputs, which impedes gradient propagation during training. Similarly, ReLU AFs are constrained by their inability to effectively utilize negative input values. To mitigate these issues, exponential function-based AFs have been introduced in the literature. The Exponential Linear Unit (ELU) [38] employs the exponential function to better accommodate negative input values, thereby enhancing gradient flow. Several variants of the ELU have been proposed [38, 55-64], for example, Scaled ELU (SELU) [55] and Parametric ELU (PELU) [56], etc.

Among the various basic AFs, selecting the most suitable one for a specific task can be challenging. Researchers have addressed this issue by developing adaptable AFs that can evolve to fit specific tasks. These adaptable AFs are governed by trainable parameters, which are optimized using GD algorithms. The learning based adaptive AFs are the recent trends. For example, PReLU is an adaptive AF where the slope parameter $\alpha$ in a LReLU function is trainable. Several variants of adaptive ReLU AF have been proposed, such as PReLU [32], Flexible ReLU (FReLU) [36], S-shaped ReLU (SReLU) [46], etc. Similarly, adaptive ELUs like PELU [56], Continuously Differentiable ELU (CELU) [57], Fast ELU (FELU) [61], Elastic ELU (EELU) [62], etc. are also proposed in the literature. Moreover, the Swish AF, defined as $f(x) = x\sigma(\beta x)$, where $\sigma(.)$ denotes the Sigmoid AF and $\beta$ is an optionally learnable parameter, was introduced [65]. This AF was identified using reinforcement learning techniques. The researchers constructed a network designed to generate a variety of AFs, incorporating a reward mechanism based on the performance of the chosen AF. By employing simple AFs as fundamental building blocks, the network was capable of synthesizing more complex AFs.

In recent years, non-linear variants were developed (Miscellaneous AFs) [65-73], including Softplus, Gaussian Error Linear Units (GELU), probabilistic functions, polynomial functions, and Mish. These variants offer distinct advantages, such as smooth gradients and improved performance, further enhancing the robustness and effectiveness of AFs in deep-learning models.

The other way to introduce adaptivity is through the use of stochastic AFs. Gulcehre et al. [27] proposed a noisy AF that integrates a structured, bounded noise effect, facilitating the optimizer's ability to explore the parameter space more comprehensively and enhancing the learning rate. In their approach, the authors modeled uncertainty in neural information flow by incorporating stochastic samples from a normal distribution into the activations, thereby replicating the stochastic behavior observed in neuronal activity.

Maxout [74] and Softmax [1] are indeed interesting and non-standard AFs, each with unique properties and uses. The Maxout AF selects the maximum value from a set of linear functions. It can approximate any convex function and is particularly useful in combination with dropout due to its ability to form piecewise linear approximations of more complex functions. Softmax function converts a vector of values into a probability distribution. The output values are in the range (0,1) and sum to 1, making them interpretable as probabilities. Softmax is widely used in the output layer of classification networks to produce probabilities for each class, enabling the network to make decisions based on the most probable class. Both Maxout and Softmax differ from standard AFs like ReLU, Sigmoid, and Tanh in their functionality and application. While standard AFs introduce non-linearity into the network, Maxout enhances the network's ability to form complex functions, and Softmax is used for probabilistic interpretation in classification tasks.

Learnable AFs can also be formulated by integrating multiple fixed AFs [75-80], as exemplified by the Adaptive Piecewise Linear (APL) [77] and Mexican ReLU (MeLU) functions [78]. Manessi and Rozza [76] introduced a novel learnable AF through the affine combination of Tanh, ReLU, and the identity function. The authors proposed two methods to automatically learn different combinations of base AFs during the training phase. However, certain adaptive AFs exhibit inherent limitations, such as overfitting and discontinuities. These functions often involve independently constructing individual segments and subsequently joining them to form a complete AF. This approach does not guarantee continuity at the segment joints, as the derivatives of adjacent segments may not align. Furthermore, employing an excessive number of segments can lead to an increased number of trainable parameters, thereby elevating the risk of overfitting.

Good reviews of AFs [81-84] are available for various classification and regression problems.

This survey makes several key contributions to the field of AFs in NNs. It provides comprehensive information on a wide range of AFs, offering meticulous classification and detailed descriptions of various types. The survey emphasizes significant and essential properties of AFs, highlighting their relevance and criticality in NN training. Additionally, performance comparisons are presented, utilizing 12 state-of-the-art AFs, offering valuable insights into their effectiveness and efficiency.

This paper is organized as follows: In Section 2, we discuss the properties and taxonomy of AFs. Section 3 covers Sigmoid-based AFs, detailing their characteristics and applications. Section 4 delves into ReLU-based AFs, examining their benefits and limitations. Section 5 focuses on ELU-based AFs, highlighting their unique features. Section 6 addresses miscellaneous AFs, encompassing a variety of important new AFs. Section 7 explores non-standard AFs, i.e., Maxout and Softmax AFs. Section 8 investigates methods for combining AFs, analyzing how different AFs can be integrated to enhance performance. Finally, A comprehensive performance analysis is conducted in Section 9.

## 2. AF Properties and Taxonomy

The AFs are needed in order to increase the complexity of the NN, without them it would just be a linear sandwich of linear functions (multiplying linear functions = linear function). Although the NN becomes simpler, learning any complex task is impossible, and our model would be just a linear regression model. The choice of AF should be guided by the specific requirements and characteristics of the problem you are trying to solve, as well as empirical testing to determine which AF works best for your NN architecture and data. AFs used in NNs have several common properties that define their behavior and suitability for different tasks. The following are some of the common properties:

1. Non-Linearity: One of the primary properties of AFs is non-linearity. They introduce non-linear transformations to the input data. This non-linearity is crucial for NNs to model complex, non-linear relationships in data.
2. Continuity: AFs are typically continuous functions, which means they do not have abrupt jumps or discontinuities. This continuity is essential for ensuring that the gradients can be calculated and used for gradient-based optimization during training.
3. Differentiability: Many AFs are differentiable, which allows for the calculation of gradients during backpropagation. This property is vital for updating the weights of the NN using optimization algorithms like GD.
4. Smoothness: Smooth AFs ensure a smooth gradient and help in avoiding issues during optimization.
5. Range: Understanding the range of an AF is important when designing NN architectures. Unbounded AFs allow the output to range from negative to positive infinity. This can help capture a wide range of information and facilitate learning complex patterns. On the other hand, bounded AFs restrict the output to a specific range. This can be useful in scenarios where you want to ensure that the output remains within a specific range. The choice between unbounded and bounded AFs depends on the specific requirements of your NN and the task at hand. Unbounded functions are often preferred in hidden layers for their ability to capture diverse patterns, while bounded functions are commonly used in the output layer.
6. Monotonicity: A function is monotonic if it is either entirely non-increasing or non-decreasing. Monotonic AFs are preferred in NNs because they ensure that the output of the model changes consistently with changes in input, making it easier to understand and optimize.
7. Computational Efficiency: AFs with simpler mathematical formulations and fewer parameters can be easier to implement and computationally more efficient. This simplicity is valuable, especially when deploying models in real-world applications. Efficient computation of the function and its derivative is important for training large NNs.
8. Memory Efficiency: AFs that require less memory for computation can be advantageous, especially when dealing with large-scale NNs or resource-constrained environments. Functions with simple calculations or that do not store a significant amount of intermediate values can be more memory-efficient.
9. Parameterization: Some AFs have parameters that can be learned during training which can be advantageous in certain scenarios. Understanding how changes in these parameters affect the behavior of the function allows for more fine-grained control over the network's learning dynamics. However, it is crucial to balance this flexibility with the risk of overfitting to the training data.
10. Sparsity: Some AFs encourage sparsity in the NN, which can lead to more efficient representations and reduced computational requirements.

The design of AFs is an extremely active area of research and does not yet have many definitive guiding theoretical principles [1]. Several AFs have been explored in recent years for deep learning to achieve the above-mentioned properties. It can be difficult to determine when to use which kind. Predicting in advance which will work best is usually impossible. The design process consists of trial and error, intuiting that a kind of hidden unit may work well, and then training a NN with that kind of hidden unit and evaluating its performance on a validation set.

A taxonomy of AFs was proposed in [84] which we also agree with. The classification is primarily based on whether the shape of the AF can be modified during the training phase. There are three main categories:

1. Fixed-shape AFs: These are AFs with a fixed shape. Examples include classic AFs used in NN literature, such as Sigmoid, Tanh, and ReLU. ReLU function has become significant in literature, marking a turning point and contributing to improved NN performance. Further subcategories within fixed-shape AFs include:

- ReLU function: This includes all functions belonging to the rectifier family, such as ReLU, LReLU, etc.
- Classic AF: This includes functions that are not in the rectifier family, such as sigmoid, Tanh, and step function, etc.

2. Trainable AFs: This category includes AFs whose shape is learned during the training phase. Two families of trainable AFs are described:

- Parameterized standard functions: These are trainable functions derived from standard fixed AFs with the addition of a set of trainable parameters. Essentially, they are a parameterized version of a standard fixed function, and the parameter values are learned from data. However, the trained AF shape turns out to be very similar to its corresponding non-trainable version. In the end, the general function shape remains substantially bounded to assume the basic function(s) shape on which it has been built.
- Functions based on ensemble methods: These functions are defined by mixing several distinct functions. A common approach is to combine them linearly, meaning the final AFs are modeled as linear combinations of one-variable functions. These one-variable functions may have additional parameters.

3. Trainable non-standard neuron: These functions can be considered as a different type of computational neuron unit compared with the original computational neuron model.

There are several ways for the taxonomy of AFs. It is possible also to define the taxonomy of AFs based on their scientific applications [83], where the output field states are real or complex-valued.

- The real-valued activations are well-known in literature.
- The complex-valued AFs are another set of AFs whose output is a complex number. These AFs are often required due to the complex-valued output, which has many applications in science and engineering.
- The process of mapping continuous values from an infinite set to discrete finite values is known as quantization. Both real and complex-valued AFs can be quantized in order to reduce memory requirements. The output of the quantized activation is integers rather than floating point values.
- Just like real-valued AFs, both complex-valued as well as quantized activations can be made adaptive by introducing the tunable variables.

In the present work, we categorized the AFs into six distinct groups: sigmoid-based, ReLU-based, ELU-based, miscellaneous, non-standard, and ensemble AFs.

Before discussing AFs in detail, it is important to note that it can be challenging to make a comparison, in terms of network performances, among the AFs that have been proposed in literature, since the experiments are often conducted using different experimental setups. AFs play a crucial role in the training of NNs by introducing non-linearity. However, they are not the sole determinants of a network's performance. The architecture and hyperparameters of a NN, such as the number of neurons, layer arrangement, weight initialization, and hyperparameter values required by the learning algorithms are equally important. The choice of these parameters can significantly influence the network's learning ability and generalization. Differences in the pre-processing of datasets can have a substantial impact on the performance of a NN. Varying pre-processing techniques, such as normalization, augmentation, or handling missing data, can lead to different experimental outcomes. The choice of performance metrics is also critical. Different metrics may emphasize different aspects of model performance (e.g., accuracy, precision, etc.), and the relevance of these metrics depends on the specific task. Moreover, the performance of AFs may vary depending on the nature of the task (classification, regression, etc.) We will discuss the challenges of comparing AFs in detail in Section 9.

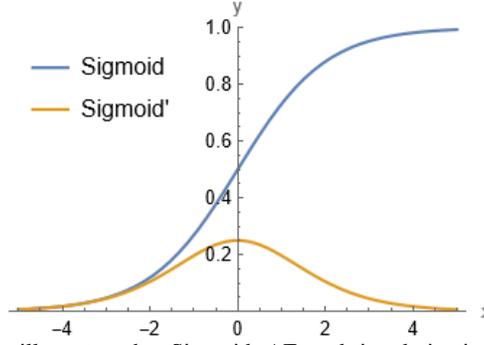

**Fig. 1.** The plot illustrates the Sigmoid AF and its derivative, showcasing the characteristic $S$-shape of the Sigmoid function along with the bell-shaped curve of its derivative. The plot ranges from $x = -5$ to $x = 5$ with the function values constrained between 0 and 1.

## 3. Sigmoid-Based AFs

This section investigates the area of Sigmoid-based AFs, exploring the fundamental concepts behind the Sigmoid function and its variants. The Logistic Sigmoid function is particularly popular in binary classification problems, where the goal is to predict outcomes as either 0 or 1. However, as the field of deep learning has evolved, researchers and practitioners have sought to enhance and modify the Logistic Sigmoid function to address certain limitations and improve performance in different scenarios. These variants aim to overcome issues such as vanishing gradients and extend the applicability of Sigmoid-based activations to different types of NN architectures.

### 3.1 Sigmoid function

Richards [11] developed the Sigmoid AF family that spans the S-shaped curves like the tanh function [12] and the Logistic Sigmoid function. Subsequently, the first NN [13] used the Sigmoid AF for modelling biological neuron firing. The Logistic Sigmoid AF and its derivative are given by

$$\sigma_{\text{Logistic}}(x) = \frac{1}{1 + e^{-x}} = 1 - \sigma_{\text{Logistic}}(-x), \tag{1.1}$$

$$\frac{d}{dx}\sigma_{\text{Logistic}}(x) = \left(1 - \sigma_{\text{Logistic}}(x)\right)\sigma_{\text{Logistic}}(x). \tag{1.2}$$

Fig. 1 depicts the plot of the Logistic Sigmoid function and its derivative.

The Logistic Sigmoid function exhibits several important properties. The graph of the Logistic Sigmoid function resembles the letter "$S$," with the curve starting near 0.5, rising gradually, and then approaching 1 as the input becomes more positive. Similarly, as the input becomes more negative, the function approaches 0. Therefore, the Logistic Sigmoid function is constrained by a pair of horizontal asymptotes as $x \to \pm\infty$, approaching 0 as $x$ approaches negative infinity and approaching 1 as $x$ approaches positive infinity. This function is monotonically increasing, meaning that as the input $x$ increases, the output $\sigma_{\text{Logistic}}(x)$ also increases, and it is infinitely smooth and differentiable for all values of $x$. This differentiability makes it useful for optimization algorithms requiring derivatives, such as GD, though the function's first derivative is bell-shaped and quickly decays to zero away from $x = 0$, which can lead to slow convergence during network training. Another characteristic is that the Logistic Sigmoid function does not have a zero-centered output. Furthermore, the function involves the computation of the exponential function $e^{-x}$, which is computationally expensive and can slow down the training process, especially in DNNs with many parameters.

The output of the Logistic Sigmoid function is always within the range $(0, 1)$, making it useful for modeling probabilities and binary classification problems. When $\sigma_{\text{Logistic}}$ is close to 1, it indicates a high probability that the input $x$ belongs to the positive class, whereas a value close to 0 indicates a high probability that $x$ belongs to the negative class. The midpoint of the Logistic Sigmoid curve is at $x = 0$, with $\sigma(0) = 0.5$, implying that if the resultant value is more than 0.5, the forecasted output is 1; otherwise, it is 0. The predicted output is as follows:

$$\text{Output} = \begin{cases} 1, & \text{Result} > 0.5, \\ 0, & \text{Result} \leq 0.5. \end{cases} \tag{2}$$

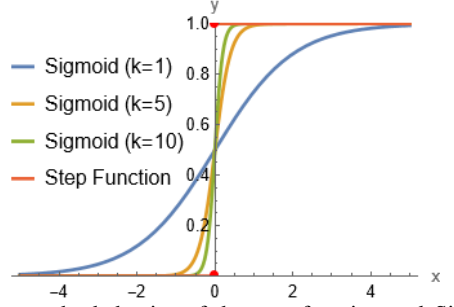

**Fig. 2.** The plot compares the behavior of the step function and Sigmoid functions with varying steepness parameters ($k = 1$, $k = 5$, $k = 10$) over a range of $x$ values from $-5$ to $5$. It illustrates how increasing the steepness parameter $k$ in the Sigmoid function makes it more closely approximate the sharp transition of the step function at $x = 0$. The Sigmoid curves transition smoothly from 0 to 1, with higher $k$ values resulting in steeper transitions. The step function, depicted as a binary switch, activates sharply at $x = 0$.

The Logistic Sigmoid function serves as a continuous and smooth approximation of the step function, which is discontinuous at the threshold and lacks smoothness. It has a parameter that controls its steepness or slope, often denoted as $k$ in the function $\sigma_{\text{Logistic}}(x) = \frac{1}{1+e^{-kx}}$, as illustrated in Fig. 2. The parameter $k$ adjusts the transition speed of the Sigmoid function from 0 to 1 as $x$ passes through zero. As $k$ increases, the transition zone becomes narrower and sharper, and in the limit as $k \to \infty$, the Sigmoid function effectively becomes a step function. Mathematically, this can be expressed as:

$$\lim_{k \to \infty} \sigma_{\text{Logistic}}(x) = \begin{cases} 0, & x < 0, \\ 1, & x > 0. \end{cases} \quad (3)$$

As depicted in Fig. 1 and demonstrated by executing $\sigma_{\text{Logistic}}(0.00001)$, near-0 inputs into the Logistic Sigmoid function will lead it to return values near 0.5. Increasingly large positive inputs will result in values that approach 1. As an extreme example, an input of 10000 results in an output of 1.0. Moving more gradually with our inputs—this time in the negative direction—we obtain outputs that gently approach 0. Any artificial neuron that features the Logistic Sigmoid function as its AF is called a Logistic Sigmoid neuron, and the advantage of these over the perceptron should now be tangible: Small, gradual changes in a given Logistic Sigmoid neuron's parameters **w** or $b$ cause small, gradual changes in pre-activation $z$, thereby producing similarly gradual changes in the neuron's post-activation, $a$. Large negative or large positive values of $z$ illustrate an exception: At extreme $z$ values, Logistic Sigmoid neurons—like perceptrons—will output 0's (when $z$ is negative) or 1's (when $z$ is positive). As with the perceptron, this means that subtle updates to the weights and biases during training will have little to no effect on the output, and thus learning will stall. This situation is neuron saturation.

*3.2 Tanh AF*

Standard Tanh or tangent hyperbolic function is basically the mathematically shifted kind of Logistic Sigmoid AF. Both are analogous and can be derived using each other. The Tanh function has a shape similar to that of the Logistic Sigmoid function, except that it is horizontally stretched and vertically translated/re-scaled to $[-1, 1]$:

$$\sigma_{\text{Tanh}}(x) = \text{Tanh}(x) = \frac{e^x - e^{-x}}{e^x + e^{-x}} = \frac{2}{1 + e^{-2x}} - 1, \quad (4.1)$$

$$\frac{d}{dx}\sigma_{\text{Tanh}}(x) = 1 - \text{Tanh}^2(x). \quad (4.2)$$

Fig. 3 shows the curve for the standard Tanh AF and its derivative.

The standard Tanh and Logistic Sigmoid functions are related as follows:

$$\sigma_{\text{Tanh}}(x) = \frac{e^x - e^{-x}}{e^x + e^{-x}} = 2\sigma_{\text{Logistic}}(2x) - 1. \quad (5)$$

The Logistic Sigmoid and the standard Tanh functions have historically been the primary tools for incorporating nonlinearity in NNs. The Tanh function is infinitely smooth, differentiable, and monotonic, which is crucial for training NNs using gradient-based optimization algorithms like Stochastic Gradient Descent (SGD). Its range is $[-1, 1]$, meaning the function is bounded and maps zero input to zero while pushing positive and negative inputs to $+1$ and $-1$, respectively. The Tanh function is symmetric around the origin, outputting negative values for negative inputs and positive values for positive inputs. Similar to the Logistic Sigmoid function, the derivative of Tanh quickly decays to zero away from $x = 0$, potentially leading to slow convergence during training. However, the negative $x$ inputs corresponding to negative $\sigma_{\text{Tanh}}(x)$ output, $x = 0$ corresponding to $\sigma_{\text{Tanh}}(0) = 0$, and positive $x$ corresponding to positive $\sigma_{\text{Tanh}}(x)$ output, the output from Tanh neurons tends to be centered near 0. These 0-centered outputs usually serve as the inputs $x$ to other artificial neurons in a network, and such 0-centered inputs make neuron saturation less likely, thereby enabling the entire network to learn more efficiently.

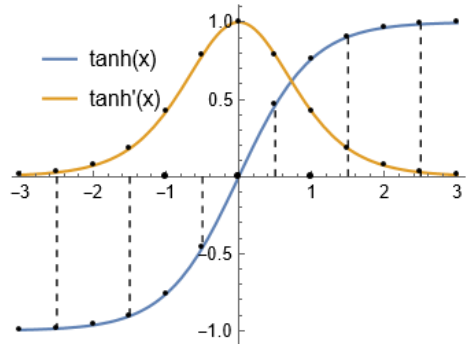

**Fig. 3.** Tanh AF and its derivative. The plot presents the Tanh AF and its derivative plotted over a range of $x$ values from $-3$ to 3. The main plot shows the Tanh function, which smoothly transitions between $-1$ and $1$, and its derivative, which illustrates the rate of change of Tanh and peaks at zero crossings of $x$.

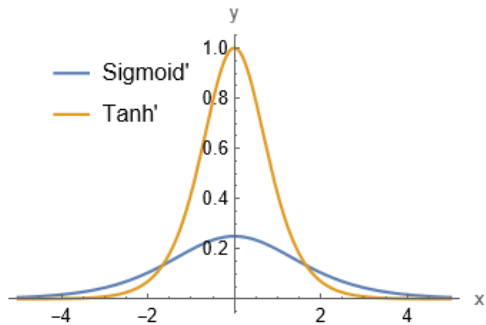

**Fig. 4.** The figure displays the derivatives of the Sigmoid and Tanh functions over a range of $x$ from $-5$ to 5. The Sigmoid derivative, labeled as "Sigmoid'", shows a bell-shaped curve peaking at $x = 0$ and approaching zero as $x$ moves away from zero, reflecting its maximum rate of change at the midpoint of the Sigmoid function. The derivative of the Tanh function, labeled "Tanh'", also presents a similar bell-shaped curve but starts and ends at higher values compared to the Sigmoid derivative, highlighting differences in how these AFs propagate gradients in NNs. The plot effectively visualizes these derivatives, providing insights into the behavior of these commonly used AFs in deep learning.

An important distinction between the Logistic Sigmoid and the Tanh functions is the behavior of their gradients. The gradient of the Logistic Sigmoid function is given by $\frac{d}{dx}\sigma_{\text{Logistic}}(x) = (1 - \sigma(x))\sigma(x)$, while the gradient of the Tanh function is $\frac{d}{dx}\sigma_{\text{Tanh}}(x) = 1 - \text{Tanh}^2(x)$. Fig. 4 shows the curve for the gradient of the Logistic Sigmoid and the Tanh AFs. The Logistic Sigmoid function has a slower rate of change than the Tanh function, meaning its derivative is smaller for a given input. This affects the convergence speed of a NN trained with the Logistic Sigmoid function compared to one trained with the Tanh function. In other words, for Tanh AF the gradient is stronger as compared to the Logistic Sigmoid function. When we are using these AFs in a NN, our data are usually centered around zero. So, we should focus our attention on the behavior of each gradient in the region near zero. We observe that the gradient of Tanh is four times greater than the gradient of the Logistic Sigmoid function. This means that using the Tanh AF results in higher values of gradient during training and higher updates in the weights of the network. So, if we want strong gradients and big learning steps, we should use the Tanh AF.

Despite these advantages, the Tanh function suffers from the vanishing gradient problem. This simply means that as the input function becomes very small or very large, depending on the case, the gradient of the function approaches zero, thus making it difficult for the network to update the weights of the earlier layers to learn from the input data. This is usually a big problem in DNNs having many layers since the gradients can become extremely small by the time they reach the earlier layers. This leads to slow convergence and poor performance.

The Logistic Sigmoid function is often used in the output layer of a binary classification network, while the Tanh function is often used in the hidden layers of a network. The Tanh function is particularly useful when the input data is zero-centered or when negative values are meaningful in the context of the problem, as it can capture both positive and negative features in the data. While traditional AFs like Logistic Sigmoid and Tanh were extensively used in the early days of NNs, they posed challenges for training DNNs due to their saturated output. Numerous attempts have been made to improve these AFs for different networks.

*3.3 Scaled Hyperbolic Tangent AF*

In order to tackle the limited output range and zero gradient problems of Tanh, a Scaled Hyperbolic Tangent (STanh) is used in [14] which is defined as,

$$\sigma_{\text{STanh}}(x) = B \; \text{Tanh}(A \, x), \quad (6)$$

where $B$ is the amplitude of the function and $A$ determines its slope at the origin, with the output range in $[-B, B]$.

The purpose of scaling is to adjust the range of the output values based on the specific requirements of the NN or the layer in which it is used. It is a way to control the magnitudes of the activations, which can affect the training dynamics of the NN.

For convenience, parameters $B = 1.7159$ and $A = 2/3$ are often chosen. With this choice of parameters, the equalities $\sigma_{\text{STanh}}(1) = 1$ and $\sigma_{\text{STanh}}(-1) = -1$ are satisfied. The rationale behind this is that the overall gain of the squashing transformation is around one in normal operating conditions, and the interpretation of the state of the network is simplified. Moreover, the absolute value of the second derivative of $\sigma_{\text{STanh}}(x)$ is a maximum at 1 and $-1$, which improves the convergence toward the end of the learning session. This particular choice of parameters is merely a convenience and does not affect the result.

Some research explores the idea of adaptive scaling during training. This involves dynamically adjusting the scaling factor based on the performance of the network during training, potentially addressing issues related to exploding or vanishing gradients.

*3.4 Parametric Sigmoid AF*

A Parametric Sigmoid Function (PSF) is proposed as a continuous, differentiable, and bounded function. It is defined as,

$$\sigma_{\text{PSF}}(x, m) = \sigma_{\text{Logistic}}^{m} = \frac{1}{(1 + e^{-x})^m}, \quad (7)$$

where $m$ is a hyperparameter [15, 16].

**Lemma 1:** Every member of the class $\sigma_{\text{PSF}}(x, m)$ is a monotonically increasing function [17].

**Proof:** The proof follows from the monotonically increasing property of the Logistic Sigmoid function and the fact that for any real $y_1$ and $y_2$ with $y_1 > 0$ and $y_2 > 0$, and any real $m > 0$, if $y_1 > y_2$ then the relation $(y_1)^m > (y_2)^m$ holds. ∎

**Lemma 2:** Every member of the class $\sigma_{\text{PSF}}$ is bounded above by 1 and below by 0, that is, for any $m > 0$, the following relations are true [17]:

$$\lim_{x \to \infty} \sigma_{\text{PSF}}(x, m) = 1, \; \lim_{x \to -\infty} \sigma_{\text{PSF}}(x, m) = 0. \quad (8)$$

**Proof:** The analytic property of the Logistic Sigmoid function and equivalence of the Logistic Sigmoid function and $\sigma_{\text{PSF}}(x, 1)$, allow us to write

$$\lim_{x \to \infty} \sigma_{\text{PSF}}(x, m) = \lim_{x \to \infty} [\sigma_{\text{PSF}}(x, 1)]^m = 1,$$
$$\lim_{x \to -\infty} \sigma_{\text{PSF}}(x, m) = \lim_{x \to -\infty} [\sigma_{\text{PSF}}(x, 1)]^m = 0,$$

and complete the proof. ∎

**Lemma 3:** Every member of the class $\sigma_{\text{PSF}}(x, m)$, $m > 0$ satisfies the generalized logistic equation given by, [17]:

$$\frac{d}{dx} \sigma_{\text{PSF}}(x, m) = \frac{m}{2} \left( \frac{1}{1 + e^{-\frac{x}{2}}} \right)^m \left( 1 - \frac{1}{1 + e^{-\frac{x}{2}}} \right). \quad (9)$$

From Lemmas 1-3, it follows that every member of $\sigma_{\text{PSF}}(x, m)$ is nonconstant, bounded, and monotonically increasing. Hence, every member of $\sigma_{\text{PSF}}(x, m)$, $m > 0$, satisfies the requirements to act as the AF in hidden layers of Sigmoidal Feedforward Neural Networks (FFNNs), and the networks using them possess the universal approximation property for the approximation of continuous functions [17].

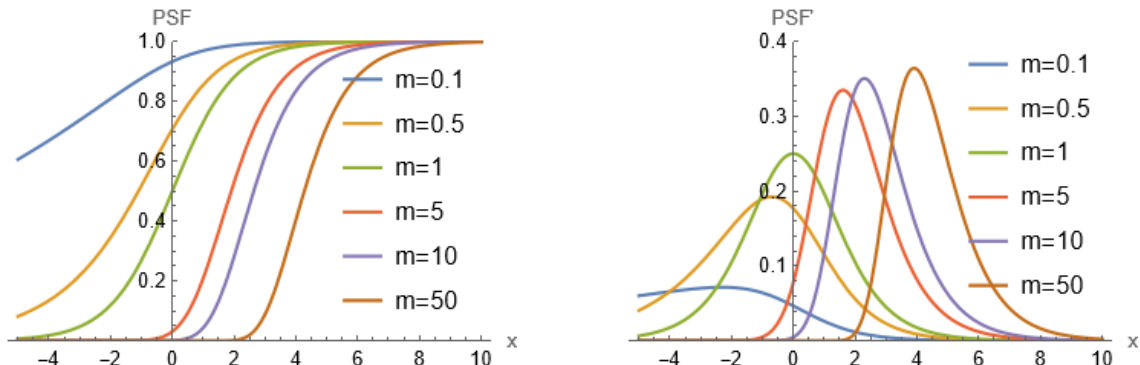

**Fig. 5.** Left panel: The figure illustrates the parametric Sigmoid AF for various values of $m$ (0.1, 0.5, 1, 5, 10, and 50). The $x$-axis represents the input variable $x$ ranging from $-5$ to 10, while the $y$-axis represents the function value $\sigma_{\text{PSF}}(x, m)$, ranging from 0 to 1. As the parameter $m$ increases, the Sigmoid function becomes steeper and more similar to a step function. Right panel: The figure shows the derivative of the parametric Sigmoid function for different values of $m$ (0.1, 0.5, 1, 5, 10, and 50). The $x$-axis represents the input variable $x$ ranging from $-5$ to 10, while the $y$-axis represents the derivative value $\text{PSF}'(x, m)$, ranging from 0 to 0.4. The plot demonstrates how the peak of the derivative shifts and changes magnitude with varying $m$. Higher values of $m$ result in a sharper and more pronounced peak.

One of the features of the commonly used activations (such as Logistic Sigmoid, Tanh, …) is the bell-shaped (symmetric about the $y$-axis) derivative of these AFs. Fig. 5 shows the plot of the $\sigma_{\text{PSF}}$ AFs and the corresponding derivatives for different values of the parameter $m$. It should be noted that the derivatives for $m \neq 1$ are skewed and their maxima's shift from the point corresponding to the input value equal to zero and the envelope of the derivatives is also sigmoidal.

One of the problems with the usually used activations is that the error surface may be very flat near the origin (net input to the activation equal to zero). A skewed derivative (sigmoidal) activation leads to the shift of the maxima of the activation from the origin. The skewed derivative activations are sigmoidal in nature and thus their usage is just justified by the universal approximation theorems. Consequently, we have the following result [15]: "Skewed derivative AFs may be more efficient for FFNN training."

The weight update rule involves the derivatives of the AFs, and the training of FFNNs uses (generally) local optimization techniques. Thus, the training time may depend on the AF used. As a result, we have the following result [15]: "AFs are not equivalent, in the sense of time taken to achieve a particular mean squared error in training." In [15], the "speed of learning" was regressed against the parameter $m$, and it was shown that (for the tasks considered) there is a specific value of the parameter $m$ (the value is different for different tasks and training data sets) for which the speed of learning is the fastest.

### 3.5 Rectified Hyperbolic Secant AF

The Rectified Hyperbolic Secant (ReSech) AF [18] is differentiable, symmetric, and bounded. It is given as,

$$\sigma_{\text{ReSech}}(x) = x\,\text{Sech}(x), \tag{10}$$

where Sech stands for hyperbolic secant function, with the output range in $[-1,1]$.

The $\sigma_{\text{ReSech}}(x)$ function possesses several significant properties. It is defined within the set of real numbers and can have both positive and negative values. Unlike Logistic Sigmoid functions, $\sigma_{\text{ReSech}}(x)$ has a global maximum at approximately $x \simeq 1.19968$ and a global minimum at approximately $x \simeq -1.19968$. This characteristic ensures that its target values are not set at the asymptote, which avoids several drawbacks, see Fig. 6. Defining a function that enjoys the property of having a global maximum and minimum, turned out to be critical during the design process since it is one of the main reasons behind the gap observed in performance between traditional AFs (sigmoidal) and their recently introduced counterparts. Moreover, $\sigma_{\text{ReSech}}(x)$ approaches finite values as $x$ to $\pm\infty$, specifically $\lim_{x \to \pm\infty} \sigma_{\text{ReSech}}(x) = 0$. This ensures that the average output values of the function are smaller and more diffuse no matter the nature of the input data.

The $\sigma_{\text{ReSech}}(x)$ AF is continuously differentiable. This property ensures the existence of the gradient, which is crucial in the derivation of the learning algorithm, such as backpropagation. It is nonlinear in order to provide enough computational power to the system. It is bounded that is, it has a maximum and minimum output value. This property is to ensure that the weights and the activations are bounded, thus keeping the training time limited.

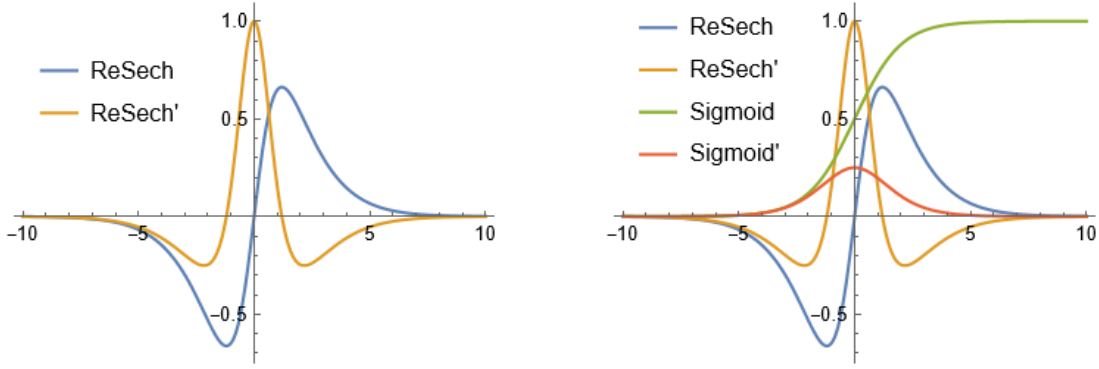

**Fig. 6.** Left panel: The figure shows the ReSech AF and its derivative. The $x$-axis represents the input variable $x$ ranging from $-10$ to $10$. The ReSech AF is shown in one curve, and its derivative in another, with both curves distinguished by a legend positioned in the upper left of the plot. The plot effectively demonstrates the behavior of the ReSech function and its rate of change across the given range of $x$ values. Right panel: The figure illustrates the ReSech AF, its derivative, the Sigmoid function, and its derivative. The $x$-axis represents the input variable $x$ ranging from $-10$ to $10$. This comprehensive plot allows for a comparative analysis of the ReSech and Sigmoid functions, highlighting their respective behaviors and rates of change across the range of $x$ values.

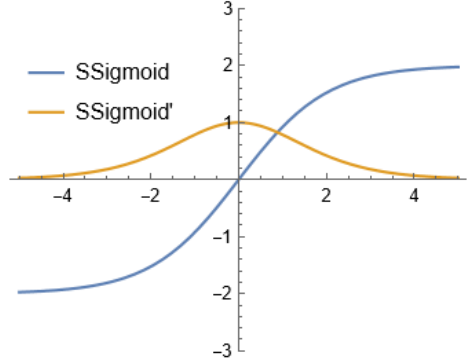

**Fig. 7.** The figure illustrates the SSigmoid AF and its derivative. The $x$-axis represents the input variable $x$ ranging from $-5$ to $5$.

Additionally, $\sigma_{\text{ReSech}}(x)$ is symmetrical about the origin. The symmetry about the origin is crucial, basically for optimization reasons, as it ensures faster convergence when training NN models with learning algorithms such as backpropagation. Furthermore, the AF will be more likely to produce outputs that are on average close to zero. This is an important feature as it is the same reason hidden behind the need to normalize our data before applying a NN to it. Combining these properties together results in a function that has the potential of not only being more useful in practice than sigmoidal functions but also being able to bridge the gap in performance observed between sigmoidal functions and ReLU for instance.

### 3.6 Scaled Sigmoid AF

Consider the Taylor expansions of different AFs,

$$\sigma_{\text{Sigmoid}}(x) = \frac{1}{2} + \frac{x}{4} - \frac{x^3}{48} + O(x^5), \tag{11.1}$$

$$\sigma_{\text{Tanh}}(x) = 0 + x - \frac{x^3}{3} + O(x^5), \tag{11.2}$$

$$\sigma_{\text{ReLU}}(x) = 0 + x. \tag{11.3}$$

It shows that in their linear regimes, $f(x) = \alpha x + \beta$, both Tanh and ReLU have the desirable property that $\alpha = 1$ and $\beta = 0$. Besides the well-known non-zero centred property, the slope of the AF near the origin is another possible reason that makes training DNNs with the Logistic Sigmoid function is difficult to train. For Logistic Sigmoid, its slope in the linear regime is $1/4$ rather than 1, so we need to initialize the weight as a 16 times smaller variance to keep each layer's gradient variance the same. Second, it has a non-zero mean, which makes the output variance increase linearly with the layer [19]. To minimize this limitation, the Scaled Sigmoid (SSigmoid) is defined as,

$$\sigma_{\text{SSigmoid}}(x) = (4\ \sigma_{\text{Sigmoid}}(x) - 2), \tag{12}$$

with the output range in $[-2,2]$.

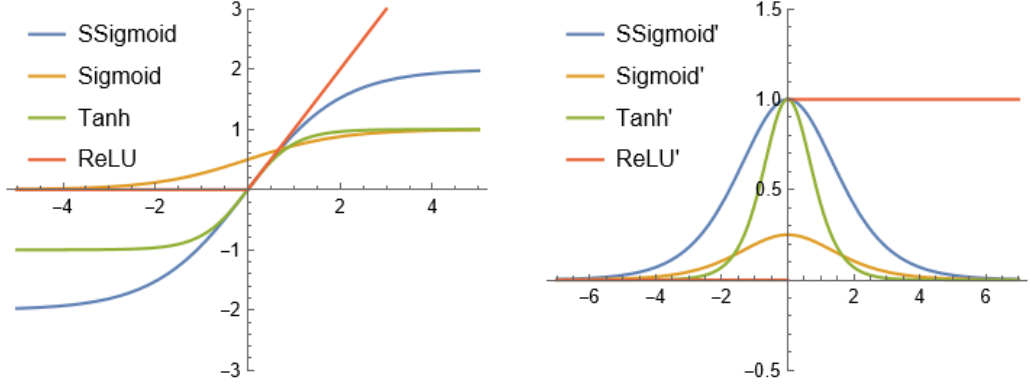

**Fig. 8.** Left panel: The figure shows the SSigmoid AF, the Sigmoid function, the Tanh function, and the ReLU function. The $x$-axis represents the input variable $x$ ranging from $-5$ to 5. The plot includes four distinct curves representing each AF, with the legend indicating which curve corresponds to each function. Right panel: The figure displays the derivatives of the SSigmoid AF, the Sigmoid function, the Tanh function, and the ReLU function. The $x$-axis represents the input variable $x$ ranging from $-7$ to 7.

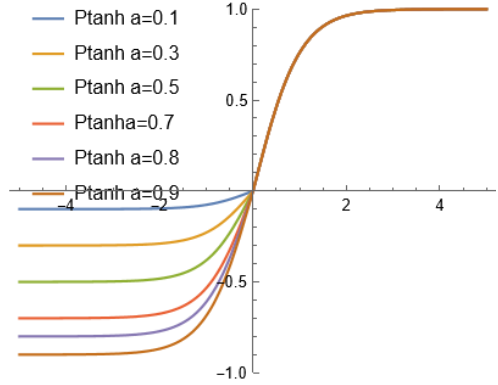

**Fig. 9.** The figure illustrates the PTanh AF for different values of the parameter $a$ (0.1, 0.3, 0.5, 0.7, 0.8, and 0.9) over a range of $x$ from $-5$ to 5. The Ptanh function is piecewise-defined: it behaves like the standard $\tanh(x)$ function for non-negative $x$ values, and as $a \cdot \text{Tanh}(x)$ for negative $x$ values, introducing an adjustable scaling factor $a$. The plot shows how varying $a$ affects the curve's shape, particularly in the negative $x$ region. Lower values of $a$ result in flatter curves for $x < 0$, while higher values make the curve steeper and closer to the standard Tanh function.

The Scaled Sigmoid is illustrated in Figs. 7 and 8. As can be seen, it is similar to Tanh near 0, but the saturation value is two times larger than Tanh.

### 3.7 Penalized Tanh AF

The Penalized Tanh (PTanh) [19] is defined as,

$$\sigma_{\text{PTanh}}(x) = \begin{cases} \text{Tanh}(x), & x > 0, \\ a\,\text{Tanh}(x), & \text{otherwise}, \end{cases} \tag{13}$$

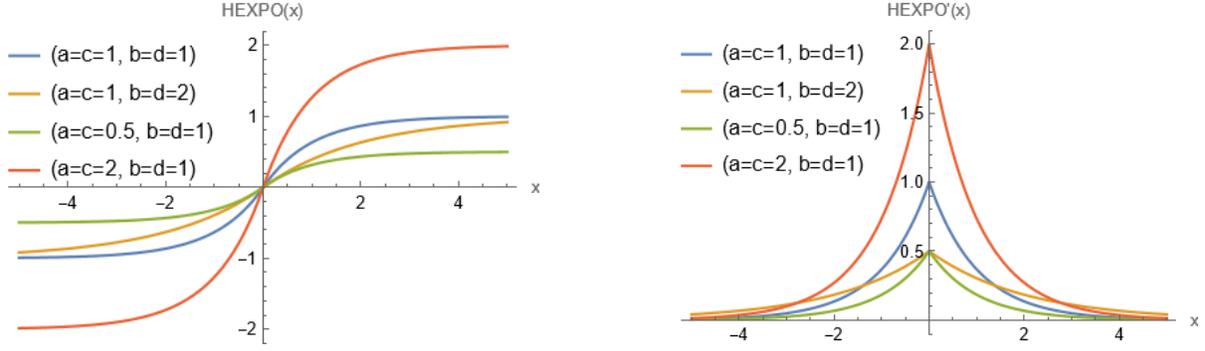

**Fig. 10.** Left panel: The figure illustrates the Hexpo AF. The $x$-axis represents the input variable $x$ ranging from $-5$ to $5$, and the $y$-axis represents the function values. Four different parameter sets are plotted: $(a = c = 1, b = d = 1)$, $(a = c = 1, b = d = 2)$, $(a = c = 0.5, b = d = 1)$ and $(a = c = 2, b = d = 1)$. Each curve corresponds to a different parameter set, with the legend indicating the parameter values. Right panel: The figure shows the derivatives of the Hexpo AF for the same four parameter sets. The $x$-axis represents the input variable $x$ ranging from $-5$ to $5$, and the y-axis represents the derivative values. The plot includes four distinct curves representing the derivatives for each parameter set. This plot provides insight into how the rate of change of the Hexpo AF varies with different parameter sets.

with the output range in $(-a, 1)$ where $a \in (0, 1)$.

In this definition: For positive values of $x$, the function uses the regular $\text{Tanh}(x)$. In this case, the Tanh function squashes the input values between 0 and 1. However, for non-positive values of $x$, the function applies a scaled version of the Tanh function with a scaling factor $a$. This means that for values of $x$ less than or equal to 0, the output is $a$ times the result of $\text{Tanh}(x)$. This allows you to control the behavior of the activation for negative values. In other words, $a$ is a hyperparameter that controls the scaling of the Tanh function for non-positive values of $x$. Fig. 9 compares the PTanh for different values of $a$.

The $\sigma_{\text{PTanh}}(x)$ function possesses several significant properties. PTanh saturates to $-a$ and 1 when moving away from 0. It is saturated outside its linear regime. Both $\sigma_{\text{SSigmoid}}(x)$ and $\sigma_{\text{PTanh}}(x)$ AFs also suffer from the vanishing gradient problem. PTanh is comparable and even outperforms the state-of-the-art non-saturated functions including ReLU on convolution DNNs [19]. This function can be viewed as a saturated version of LReLU. These two functions have similar values near 0 since both functions share the same Taylor expansion up to the first order. This kind of piecewise AF can be useful in scenarios where you want to apply different transformations to the positive and non-positive regions of the input. The choice of $a$ would depend on the specific requirements of your problem and would likely require experimentation and tuning.

### 3.8 Hexpo AF

The vanishing gradient problem is minimized by the Hexpo function [20] which is similar to Tanh with a scaled gradient. It is given as,

$$\sigma_{\text{Hexpo}}(x) = \begin{cases} -a(e^{-x/b} - 1), & x \geq 0, \\ c(e^{x/d} - 1), & \text{otherwise.} \end{cases} \quad (14.1)$$

$$\frac{\partial}{\partial x}\sigma_{\text{Hexpo}}(x) = \begin{cases} \dfrac{a}{b}e^{-x/b}, & x \geq 0, \\ \dfrac{c}{d}e^{x/d}, & \text{otherwise.} \end{cases} \quad (14.2)$$

Hexpo maps the input into range $(-c, a)$. The parameters $a$, $b$, $c$, and $d$ play a crucial role in shaping the behavior of the Hexpo function. The values of these parameters determine the slope, scale, and curvature of the function for positive and negative input values. Parameters of the Hexpo offer the function stronger ability on bias shift correction, which speeds up the learning process. Function (14.2) is the gradient of the function with respect to $x$. The gradient can be scaled according to the parameters. Therefore, if parameters satisfy $a > b$ or $c > d$, there exists some domain near zero that mapped into the output of greater than one. Having a gradient of greater than one on some neurons can prevent the gradient from vanishing.

Generally, increasing $b$ and $d$ or decreasing $a$ and $b$ would both push the gradient away from zero, though the former is more critical than the later. For example, as shown in Fig. 10, as the $a/b$ or $c/d$ ratio increases, the rate at which the gradient decays to zero would decrease. Alternatively, by solely decreasing parameters $b$ and $d$, the shape of the curve of the function would be flattened, and hence the gradient can be more difficult to reach zero. Conversely, solely increasing $a$ and $c$ would scale the gradient around the origin up. Hence the input around the origin would have a relatively large gradient. By tuning the ratio $a/b$ or $d/c$, the gradient on the neurons could even be scaled up to a value larger than one, which prevents the gradient flow from vanishing.

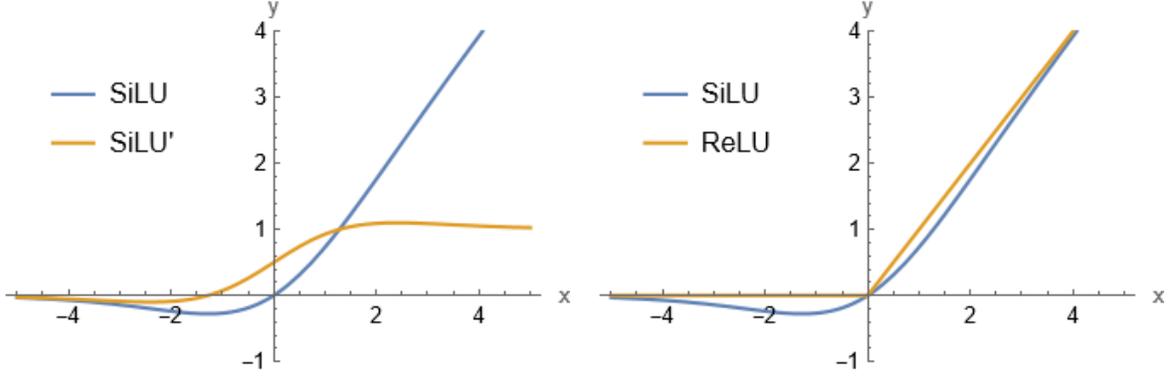

**Fig. 11.** Left panel: This plot shows the SiLU AF and its derivative. The SiLU function smoothly transitions between linear behavior for large positive values and exponential decay for negative values. The derivative curve demonstrates how the gradient varies, providing insights into the function's smooth gradient properties. Right panel: This plot compares the SiLU function with the ReLU function. While ReLU outputs zero for negative inputs and increases linearly for positive inputs, SiLU offers a smooth, continuous alternative that avoids the abrupt transition at zero, blending linearity and Sigmoid behavior.

In comparison with Sigmoid, which always has a maximum gradient of 0.25, and Tanh, which has a maximum gradient of 1, Hexpo scales its gradient by manipulating its parameters as shown in Fig. 10. Hence, Hexpo weakens the vanishing gradient problem by its ability to prevent gradient flow from vanishing.

*3.9 Sigmoid-Weighted Linear Unit*

The output of the Sigmoid function is multiplied by its input in the Sigmoid-weighted Linear Unit (SiLU) [21] as
$$\sigma_{\text{SiLU}}(x) = x\, \sigma_{\text{Sigmoid}}(x), \tag{15}$$
in the output range of $(-0.5, \infty)$.

The $\sigma_{\text{SiLU}}(x)$ function possesses several significant properties. The $\sigma_{\text{SiLU}}(x)$ AF looks like a continuous and "undershooting" version of the ReLU AF, as illustrated in Fig. 11. For $x$-values of large magnitude, the activation of the SiLU is approximately equal to the activation of the ReLU, i.e., the activation is approximately equal to zero for large negative $x$-values and approximately equal to $x$ for large positive $x$-values. Unlike the ReLU (and other commonly used activation units such as Sigmoid and Tanh units), the activation of the SiLU is not monotonically increasing, meaning it does not strictly increase or decrease across its entire domain. This non-monotonic behavior has been suggested to improve learning dynamics in certain cases. It has a global minimum value of approximately $-0.28$ for $x \approx -1.28$. The SiLU function is smooth and differentiable everywhere. This smoothness can be beneficial for gradient-based optimization algorithms used in training NNs, such as SGD.

SiLU has been proposed as a way to mitigate the vanishing gradient problem. SiLU's non-linearity and the fact that it allows information to flow through the network more freely than traditional AFs like Sigmoid or hyperbolic tangent can help alleviate this issue. SiLU has been compared with other popular AFs like ReLU and has shown competitive performance in various scenarios [21]. It has been reported to achieve similar or better accuracy in certain deep-learning tasks.

The Derivative SiLU (DSiLU) AF is computed by the derivative of the SiLU [21]:
$$\sigma_{\text{DSiLU}}(x) = \sigma_{\text{Sigmoid}}(x)\bigl[1 + x\bigl(1 - \sigma_{\text{Sigmoid}}(x)\bigr)\bigr]. \tag{16}$$
The AF of the $\sigma_{\text{DSiLU}}(x)$ looks like a steeper and "overshooting" Sigmoid function (see Fig. 12). The DSiLU has a maximum value of approximately 1.1 and a minimum value of approximately $-0.1$ for $x \approx \pm 2.4$.

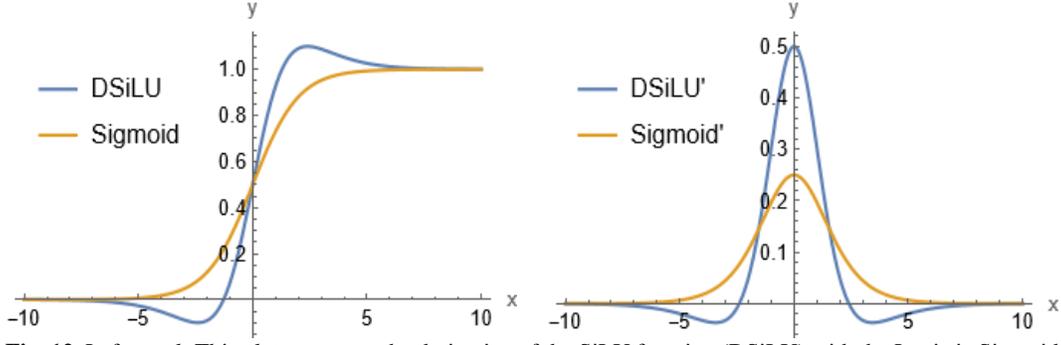

**Fig. 12.** Left panel: This plot compares the derivative of the SiLU function (DSiLU) with the Logistic Sigmoid function itself. Right panel: This plot shows the second derivative of SiLU (DSiLU') compared with the first derivative of the Logistic Sigmoid function. The curves illustrate the more complex gradient behavior of DSiLU, providing a deeper understanding of its smooth and adaptive nature compared to the simpler gradient of the Sigmoid function.

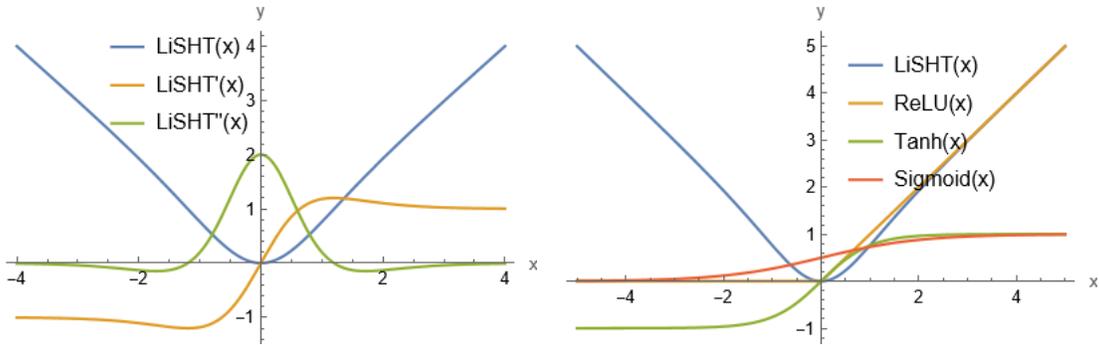

**Fig. 13.** Left panel: The figure illustrates the LiSHT AF, its first derivative, and its second derivative. The $x$-axis represents the input variable $x$ ranging from $-4$ to $4$. Right panel: The figure compares the LiSHT AF with the ReLU function, the Tanh function, and the Sigmoid function. The $x$-axis represents the input variable $x$ ranging from $-5$ to $5$. The legend, placed in the upper right of the plot, identifies each curve. The plot range is set to fully encompass the behavior of these AFs across the specified range of $x$ values, allowing for a clear comparison of their different characteristics.

### *3.10 Linearly Scaled Hyperbolic Tangent AF*

The Linearly Scaled Hyperbolic Tangent (LiSHT) AF scales the Tanh in a linear fashion to overcome the vanishing gradient issue [22]. The LiSHT can be defined as,

$$\sigma_{\text{LiSHT}}(x) = x \, \text{Tanh}(x), \tag{17.1}$$

in the output range of $[0, \infty)$.

The first derivative of LiSHT is given as follows,

$$\frac{\partial}{\partial x} \sigma_{\text{LiSHT}}(x) = \frac{\partial}{\partial x} x \, \text{Tanh}(x)$$
$$= x + \text{Tanh}(x) \, (1 - \sigma_{\text{LiSHT}}(x)). \tag{17.2}$$

LiSHT is a smooth function that shares similarities with ReLU, particularly in having unbounded upper limits on the right-hand side of the activation curve. However, because of the symmetry-preserving property of LiSHT, the left-hand side of the activation is in the upwardly unbounded direction, hence it satisfies non-monotonicity (see Fig. 13).

For the large positive inputs, the behavior of the LiSHT is close to the ReLU, i.e., the output is close to the input as depicted in Fig. 13. Whereas, unlike ReLU and other commonly used AFs, the output of LiSHT for negative inputs is symmetric to the output of LiSHT for positive inputs as illustrated in Fig. 13.

It can be observed from the derivatives of LiSHT in Fig. 13 that the amount of non-linearity is very high near zero as compared to the existing activations which can boost the learning of a complex model.

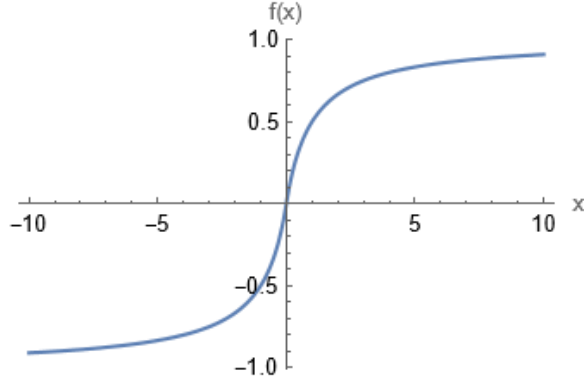

**Fig. 14.** The figure illustrates the Elliott AF. The $x$-axis represents the input variable $x$ ranging from $-10$ to $10$, while the $y$-axis represents the function values. The Elliott function smoothly transitions from negative to positive values, saturating as $x$ moves away from zero. This characteristic makes it useful for NN AFs due to its simplicity and smooth gradient properties.

**Lemma 4:** If $\sigma_{\text{LiSHT}}(x) = x\,\text{Tanh}(x)$ then the local gradient $\frac{\partial}{\partial x}\sigma_{\text{LiSHT}}(x) = 0$ iff $x = 0$.

**Proof:**

$$\sigma_{\text{LiSHT}}(x) = x\,\text{Tanh}(x).$$

The local gradient for $\sigma_{\text{LiSHT}}(x)$ activation is given as,

$$\frac{\partial}{\partial x}\sigma_{\text{LiSHT}}(x) = x + \text{Tanh}(x)\,(1 - x\,\text{Tanh}(x)).$$

For $x < -2$, $\text{Tanh}(x) \approx -1$, thus $\frac{\partial}{\partial x}\sigma_{\text{LiSHT}}(x) \approx -1$. For $x > 2$, $\text{Tanh}(x) \approx 1$, thus $\frac{\partial}{\partial x}\sigma_{\text{LiSHT}}(x) \approx 1$. For $-2 \leq x \leq 2$, $-1 \leq \text{Tanh}(x) \leq 1$, thus $2x - 1 \leq \frac{\partial}{\partial x}\sigma_{\text{LiSHT}}(x) \leq 2x + 1$. The $\sigma_{\text{LiSHT}}(x)$ can lead to gradient-diminishing problem iff $\frac{\partial}{\partial x}\sigma_{\text{LiSHT}}(x) = x + \text{Tanh}(x)\,[1 - x\,\text{Tanh}(x)] = 0$. It means $x = \frac{\text{Tanh}(x)}{\text{Tanh}^2(x) - 1}$ which is only possible iff $x = 0$.

Thus, the $\sigma_{\text{LiSHT}}(x)$ AF exhibits a non-zero gradient for all positive and negative inputs and solves the gradient-diminishing problem of $\text{Tanh}(x)$ AF. ∎

### 3.11 Elliott and Modified Elliott AFs

The Elliott AF [23, 24] is similar to the Sigmoid function in terms of the characteristics diagram and defined as,

$$\sigma_{\text{Elliott}}(x) = \frac{x}{1 + |x|}, \qquad (18.1)$$

or

$$\sigma_{\text{Elliott}}(x) = \frac{0.5\,x}{1 + |x|} + 0.5, \qquad (18.2)$$

in the output range of $[-1,1]$ or $[0,1]$. It can be calculated much faster than Sigmoid, see Fig. 14.

The modification of the Elliott AF (MElliott) [25] is defined as

$$\sigma_{\text{MElliott}}(x) = \frac{x}{\sqrt{1 + x^2}}. \qquad (19)$$

Fig. 15 presents a comparison between the Sigmoid and MElliott activation, illustrating that the MElliott function is significantly steeper around 0 and reaches its threshold at much lower input values. Additionally, a comparison between the MElliott and ReLU functions is also shown in Fig. 15. While ReLU offers a similar gradient to MElliott, it lacks gradients in the negative input spectrum, unlike MElliott.

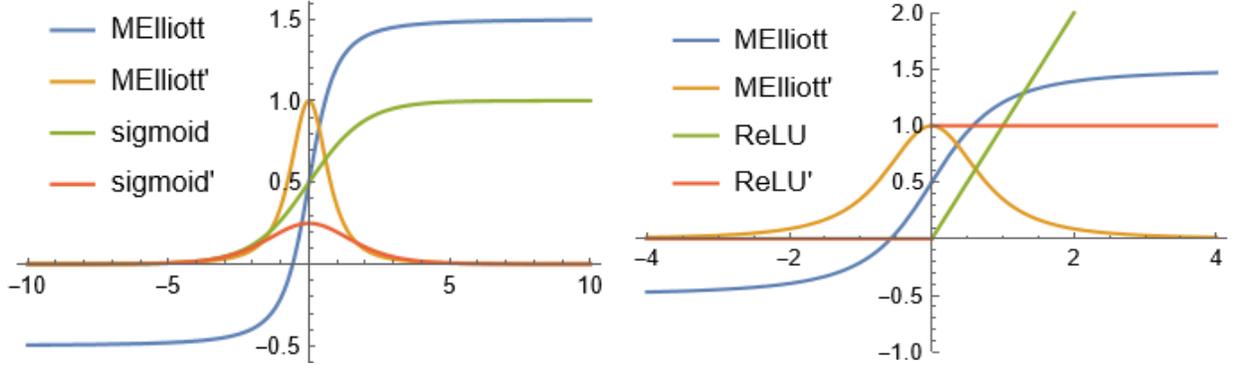

**Fig. 15.** Left panel: The figure compares the Modified Elliott AF and its derivative with the Sigmoid AF and its derivative. The $x$-axis represents the input variable $x$ ranging from $-10$ to $10$. The legend, positioned in the upper left of the plot, identifies each curve. The plot provides a comprehensive comparison of the behaviors and rates of change of these AFs across the specified range of $x$ values. Right panel: The figure compares the Modified Elliott AF and its derivative with the ReLU AF and its derivative. The $x$-axis represents the input variable $x$ ranging from $-4$ to $4$.

*3.12 Soft-Root-Sign AF*

An effective AF is required to have

1. negative and positive values for controlling the mean toward zero to speed up learning;
2. saturation regions (derivatives approaching zero) to ensure a noise-robust state;
3. a continuous-differential curve that helps with effective optimization and generalization.

Based on the above insights, the Soft-Root-Sign (SRS) AF [26] was defined as,

$$\sigma_{SRS}(x) = \frac{x}{\frac{x}{\alpha} + e^{-\frac{x}{\beta}}},$$

(20.1)

where $\alpha$ and $\beta$ are a pair of trainable non-negative parameters. Fig. 16 shows the graph of the $\sigma_{SRS}$ AF. The derivative of SRS is defined as

$$\frac{\partial}{\partial x}\sigma_{SRS}(x) = \frac{\left(1+\frac{x}{\beta}\right)e^{-\frac{x}{\beta}}}{\left(\frac{x}{\alpha} + e^{-\frac{x}{\beta}}\right)^2},$$

(20.2)

Fig. 16 illustrates the first derivative of $\sigma_{SRS}(x)$, which gives nice continuity and effectiveness.

The $\sigma_{SRS}(x)$ function possesses several significant properties. SRS is a smooth, non-monotonic, and bounded AF. As shown in Fig. 16, the $\sigma_{SRS}(x)$ AF is bounded output with a range $[\frac{\alpha\beta}{\beta-\alpha e}, \alpha)$. Specifically, the minimum of $\sigma_{SRS}(x)$ is observed to be at $x = -\beta$ with a magnitude of $\frac{\alpha\beta}{\beta-\alpha e}$; and the maximum of $\sigma_{SRS}(x)$ is $\alpha$ when the network input $x \to +\infty$. The maximum value and slope of the function can be controlled by changing the parameters $\alpha$ and $\beta$, respectively. Through further setting $\alpha$ and $\beta$ as trainable, $\sigma_{SRS}(x)$ can not only control how fast the first derivative asymptotes to saturation, but also adaptively adjust the output to a suitable distribution, which avoids gradient-based problems and ensures fast as well as robust learning for multiple layers.

The choice of AF is essential for building state-of-the-art NNs. At present, the most widely used AF with effectiveness is ReLU. However, ReLU has the problems of non-zero mean, negative missing, and unbounded output, thus it has potential disadvantages in the optimization process. In contrast to ReLU, SRS has a non-monotonic region when $x < 0$ which helps capture negative information and provides zero-mean property. Meanwhile, SRS is bounded output when $x > 0$ which avoids and rectifies the output distribution to be scattered in the non-negative real number space.

Moreover, SRS adaptively adjusts a pair of independent trainable parameters to provide a zero-mean output, resulting in better generalization performance and faster learning speed.

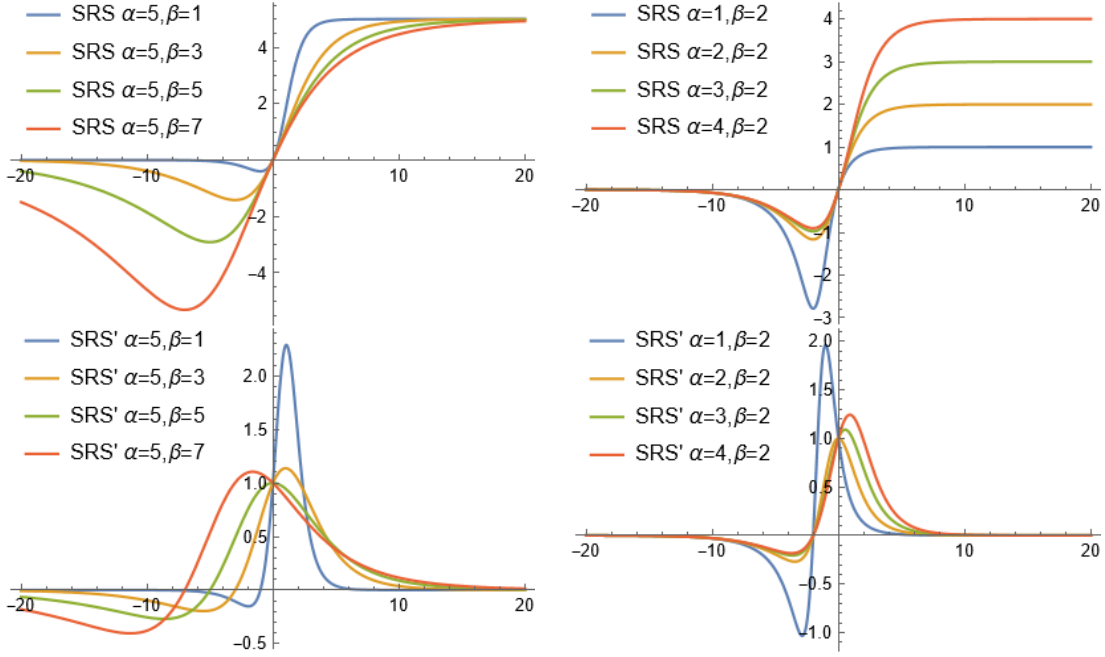

**Fig. 16.** The figure consists of four plots illustrating the SRS AF and its derivatives for various parameter values over a range of $x$ from $-20$ to $20$. Top left panel: The plot shows the AF for a fixed $\alpha = 5$ with different $\beta$ values $(1, 3, 5, 7)$, highlighting how $\beta$ affects the function's smoothness and steepness. Top right panel: The plot presents the AF for a fixed $\beta = 2$ with varying $\alpha$ values $(1, 2, 3, 4)$, demonstrating the impact of $\alpha$ on the function's scaling and shape. Bottom left panel: The plot displays the derivatives of the AF for $\alpha = 5$ and $\beta$ values $(1, 3, 5, 7)$, showing how the gradient changes across the input range. Bottom right panel: The plot shows the derivatives for $\beta = 2$ and $\alpha$ values $(1, 2, 3, 4)$, illustrating the effect of $\alpha$ on the gradient behavior.

### *3.13 HardSigmoid and HardTanh*

**Definition (Hard and Soft Saturation):** Let $h(x)$ be AF with derivative $h'(x)$, and $c$ be a constant. We say that $h(\cdot)$ right hard saturates when $x > c$ implies $h'(x) = 0$ and left hard saturates when $x < c$ implies $h'(x) = 0$, $\forall x$. We say that $h(\cdot)$ hard saturates (without qualification) if it is both left and right hard saturates. An AF that saturates but achieves zero gradient only in the limit is said to be soft saturate.

We can construct hard-saturating versions of soft-saturating AFs [27] by taking a first-order Taylor expansion about zero and clipping the results to an appropriate range. For example, expanding Tanh and Sigmoid around 0, with $x \approx 0$, we obtain linearized functions $u^t$ and $u^s$ of Tanh and Sigmoid, respectively:

$$\sigma_{\text{Sigmoid}}(x) \approx u^s(x) = 0.25x + 0.5, \tag{21}$$
$$\sigma_{\text{Tanh}}(x) \approx u^t(x) = x. \tag{22}$$

Clipping the linear approximations results in,

$$\sigma_{\text{HardSigmoid}}(x) = \max(\min(u^s(x), 1), 0), \tag{23}$$
$$\sigma_{\text{HardTanh}}(x) = \max(\min(u^t(x), 1), -1). \tag{24}$$

The motivation behind this construction is to introduce linear behavior around zero to allow gradients to flow easily when the unit is not saturated while providing a crisp decision in the saturated regime.

The hard Sigmoid is a non-smooth function used in place of a Sigmoid function. These retain the basic shape of a Sigmoid, rising from 0 to 1, but using simpler functions, especially piecewise linear functions or piecewise constant functions. Hard Sigmoid is defined as

$$\begin{aligned}\sigma_{\text{HardSigmoid}}(x) &= \text{Clip}(0.2x + 0.5, 0, 1) \\ &= \max(0, \min(1, (0.2x + 0.5))) \\ &= \begin{cases} 0, & x < -2.5, \\ 1, & x > 2.5, \\ 0.2x + 0.5, & -2.5 \leq x \leq 2.5. \end{cases}\end{aligned} \tag{25}$$

In this definition, the function is linear in the "transition" region between $-2.5$ and $2.5$, and it quickly saturates to 0 or 1 outside of this range (see Fig. 17).

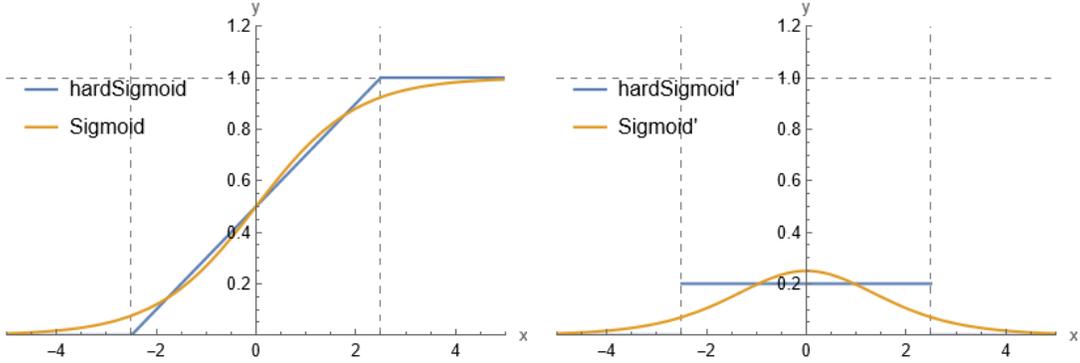

**Fig. 17.** Left panel: The plot visually illustrates the Sigmoid AF, which exhibits a smooth, S-shaped curve, with the HardSigmoid AF, a piecewise linear approximation. HardSigmoid is defined with flat segments at the extremes and a linear segment in the middle, clearly delineated by grid lines at $x = -2.5$ and $x = 2.5$. Right panel: This plot illustrates the derivatives of both the Sigmoid and HardSigmoid functions across a range of $x$ from $-5$ to $5$. The derivative of the Sigmoid function forms a bell-shaped curve, highlighting its smooth and continuous nature. In contrast, the derivative of the HardSigmoid is piecewise constant, zero for most of the range except in the middle transition region between $x = -2.5$ and $x = 2.5$, where it is constant (non-zero).

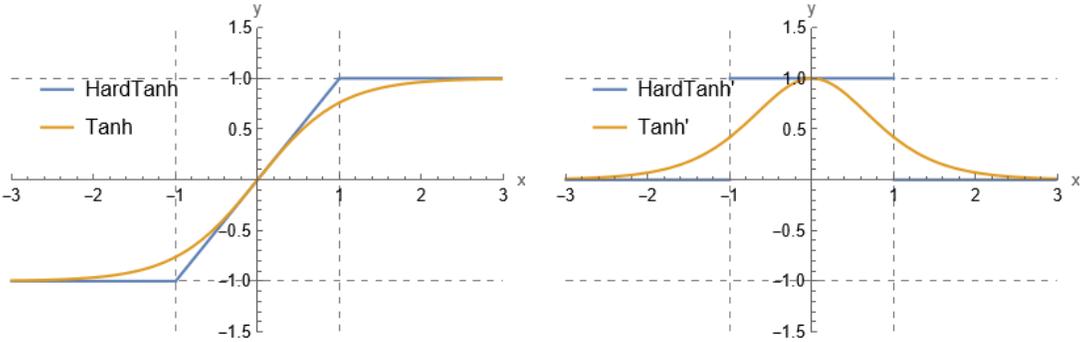

**Fig. 18.** Left panel: The plot compares the Tanh AF with the HardTanh AF over a range of $x$ from $-3$ to $3$. Tanh smoothly transitions between $-1$ and $1$, while HardTanh abruptly clips values outside this range, creating a piecewise linear effect. The grid lines at $x = -1$ and $x = 1$ highlight these transition points. Right panel: The plot displays the derivatives of the Tanh and HardTanh AFs over the same range. The derivative of Tanh, "Tanh'", shows a bell-shaped curve, illustrating its non-zero gradient everywhere except the asymptotes. In contrast, "HardTanh'" has a derivative of zero almost everywhere except at $x = -1$ and $x = 1$, where it is undefined.

One of the main advantages of the Hard Sigmoid function is its computational efficiency. Unlike the standard Sigmoid function, which involves exponentiation and can be computationally expensive, the Hard Sigmoid function relies on simple linear operations (addition and multiplication), making it faster to compute. This efficiency can be particularly beneficial in scenarios where computational resources are limited or when training large NNs.

The "HardTanh" AF is a piecewise linear approximation of the standard hyperbolic tangent function. It is often used in NNs as an alternative to the Tanh. It is a cheaper and more computationally efficient version of the Tanh activation. HardTanh is defined as [27]

$$\sigma_{\text{HardTanh}}(x) = \text{Clip}[x, -1, 1]$$
$$= \max(-1, \min(1, x))$$
$$= \begin{cases} -1, & x < -1, \\ 1, & x > 1, \\ x, & -1 \leq x \leq 1. \end{cases} \quad (26.1)$$

The derivative can also be expressed in a piecewise functional form (see Fig. 18):

$$\frac{\partial}{\partial x} \sigma_{\text{HardTanh}} = \begin{cases} 1, & -1 \leq x \leq 1, \\ 0, & \text{otherwise}. \end{cases} \quad (26.2)$$

The Hard Tanh function is piecewise linear and consists of three linear segments: one for inputs less than $-1$, one for inputs between $-1$ and 1, and one for inputs greater than 1. This linearity simplifies the training process. A primary advantage of the Hard Tanh function is that it bounds the output within a specific range, typically between $-1$ and 1. This bounded output range can be beneficial in certain applications where the network needs to produce values within a specific interval, such as in reinforcement learning or when modeling probabilities. The gradient of the Hard Tanh function is well-behaved almost everywhere, except at the points where the function changes from one linear segment to another (i.e., at $-1$ and 1).

## 4. ReLU Based AFs

In the realm of NNs, the ReLU AF stands as a fundamental AF known for its ability to efficiently handle the vanishing gradient problem and expedite learning through non-saturation properties. However, the need for variant AFs arises from the shortcomings of standard ReLU, such as the "dying ReLU" problem, where neurons can become inactive and stop learning during training. Additionally, the unbounded nature of ReLU can lead to issues like exploding gradients, making it challenging to optimize DNNs effectively. These challenges prompt researchers to explore modifications and alternatives to enhance the performance and stability of NNs.

One such adaptation is the LReLU, which mitigates the dying ReLU effect by incorporating a fixed negative slope, ensuring the preservation of small negative signals through non-zero gradients. Taking the concept further, the PReLU emerges as a dynamic alternative, allowing the learning of negative slopes from training samples. The flexibility of PReLU lies in its ability to share parameters across all channels or assign them independently to each channel within a hidden layer. The exploration doesn't stop there, as the introduction of randomness into ReLU-based architectures introduces new dimensions of adaptability. RReLU adds an element of unpredictability by altering the slope of the negative part during training, drawing from a uniform distribution. Delving into the positive domain, EReLU and its composite EPReLU modify the positive parts of the input during the training stage. EPReLU, in particular, combines the positive slope variation of EReLU with the parameter learning capability of PReLU, demonstrating its efficacy without introducing additional parameters during testing. A novel approach unfolds with Random Translation ReLU (RT-ReLU), which introduces an offset to the input sampled from a Gaussian distribution. Not only does this technique prove robust against small jitters and noise, but it also exhibits resilience to overfitting.

This section provides an overview of the key issues associated with standard ReLU and introduces the reader to the main objectives of exploring its variants. Each variant introduces unique characteristics, advantages, and potential drawbacks, contributing to the rich landscape of AF in modern deep learning. From LReLU to RT-ReLU, and beyond, we will explore several prominent variants (20 variants) of the ReLU AF.

### *4.1 ReLU AF*

The ReLU AF is an AF defined as the positive part of its argument [28]:

$$\sigma_{\text{ReLU}}(x) = \max(0, x)$$
$$= \frac{x + |x|}{2}$$
$$= \begin{cases} x, & x \geq 0, \\ 0, & x < 0. \end{cases} \quad (27.1)$$

$$\frac{\partial}{\partial x}\sigma_{\text{ReLU}}(x) = \begin{cases} 1, & x > 0, \\ 0, & x < 0, \end{cases} \quad (27.2)$$

where $x$ is the input to a neuron. This is also known as a ramp function. This AF was introduced by Kunihiko Fukushima in 1969 [28]. In 2011 [29], it was found to enable better training of deeper networks, compared to the widely used AFs before 2011, e.g., the Sigmoid and the hyperbolic tangent. The ReLU function is one of the simplest functions to imagine that is nonlinear. That is, like the Sigmoid and Tanh functions, its output does not vary uniformly linearly across all values of $x$. The ReLU is in essence two distinct linear functions combined (one at negative $x$ values returning 0, and the other at positive $x$ values returning $x$, as is visible in Fig. 19), to form a straightforward, nonlinear function overall. It provides output as $x$ if $x$ is positive and 0 if the value of $x$ is negative.

The ReLU function possesses several significant properties. ReLU AF introduces non-linearity to the NNs, allowing it to model complex, non-linear relationships in the data. This is crucial for the network's ability to learn and represent a wide range of functions. Unlike the linear and Sigmoid AFs, which are implemented at the output layer of the NN, ReLU is implemented at the hidden layers of the NN. ReLU involves simpler mathematical operations compared to Tanh and Sigmoid, thereby boosting its computational performance further. It is computationally efficient and easy to implement, making it a popular choice in deep learning models. The range of the function is $[0, \infty)$ and it does not saturate for positive input values. Additionally, ReLU does not suffer from the vanishing gradient problem for positive inputs, a common issue with AFs like Tanh and Sigmoid, and it converges much faster in practice.

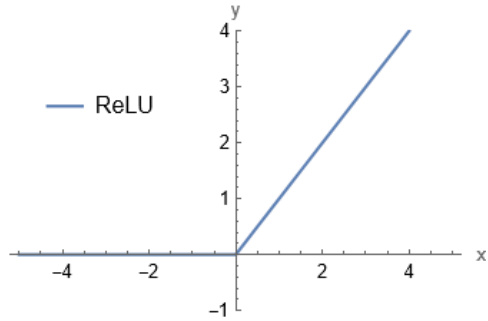

**Fig. 19.** The plot illustrates the ReLU AF plotted over a range of $x$ from $-5$ to 5. ReLU is defined as $\max(0, x)$, which results in a graph where all negative values are clamped to zero, and positive values remain unchanged. The function transitions sharply from 0 to a linear increase at $x = 0$.

ReLU is unbounded AF. One may thus need to use a regularizer to prevent potential numerical problems. It does not produce zero-centered output. The hyperbolic tangent absolute value non-linearity $|\text{Tanh}(x)|$ enforces sign symmetry. A $\text{Tanh}(x)$ non-linearity enforces sign antisymmetry. ReLU is one-sided and therefore does not enforce a sign symmetry or antisymmetry, instead, the response to the opposite of an excitatory input pattern is 0 (no response). However, we can obtain symmetry or antisymmetry by combining two rectifier units sharing parameters. To efficiently represent symmetric or antisymmetric behavior in the data, a ReLU network would need twice as many hidden units as a network of symmetric or antisymmetric AFs.

The ReLU function is continuous, while its derivative will be piecewise constant with a jump at $x = 0$. The second derivative will be a Dirac function concentrated at $x = 0$. In other words, the higher-order derivates (greater than 1) are not well-defined.

*4.1.1 Network Sparse Representation*

Studies on brain energy expense suggest that neurons encode information in a sparse and distributed way, estimating the percentage of neurons active at the same time to be between 1 and 4%. This corresponds to a trade-off between the richness of representation and small action potential energy expenditure. Without additional regularization, ordinary FFNNs do not have this property. For example, the Sigmoid activation has a steady state regime around $1/2$, therefore, after initializing with small weights, all neurons fire at half their saturation regime. This is biologically implausible and hurts gradient-based optimization. ReLU units are sparsely activated, which can lead to more efficient training and generalization, meaning that they are active (output a non-zero value) for positive inputs and inactive (output zero) for negative inputs. This can be seen as a form of regularization that helps prevent overfitting in some cases. For example, after uniform initialization of the weights, around 50% of hidden units' continuous output values are real zeros, and this fraction can easily increase with sparsity-inducing regularization.

As illustrated in Fig. 20, the only non-linearity in the network comes from the path selection associated with individual neurons being active or not. For a given input only a subset of neurons is active. Computation is linear on this subset: once this subset of neurons is selected, the output is a linear function of the input (although a large enough change can trigger a discrete change of the active set of neurons). The function computed by each neuron or by the network output in terms of the network input is thus linear by parts. We can see the model as an exponential number of linear models that share parameters [30]. Because of this linearity, gradients flow well on the active paths of neurons (there is no gradient vanishing effect due to activation non-linearities of Sigmoid or Tanh units), and mathematical investigation is easier.

*4.1.2 Dying ReLU problem*

ReLU neurons can sometimes be pushed into states in which they become inactive for essentially all inputs. Neurons that output zero for all inputs are considered "dead" and do not contribute to the learning process. This can happen when the input to a ReLU neuron is consistently negative, causing the gradient during backpropagation to be zero, preventing weight updates. In this state, no gradients flow backward through the neuron, and so the neuron becomes stuck in a perpetually inactive state and "dies". This is a form of the vanishing gradient problem. In some cases, large numbers of neurons in a network can become stuck in dead states,

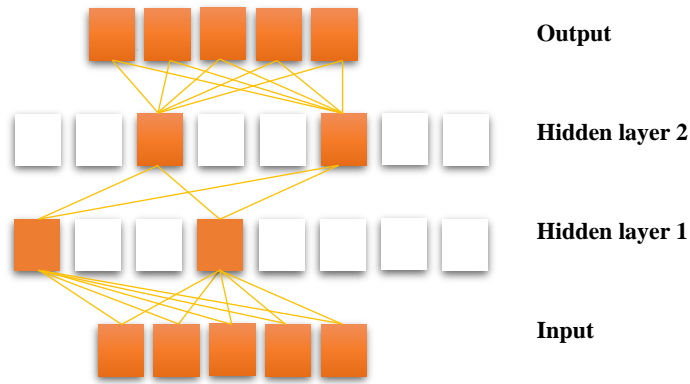

**Fig. 20.** In a network comprised of ReLU units, the process of activation and gradient propagation occurs sparsely. This means that only a subset of neurons, determined by the input, becomes active, and the computational load remains linear within this selected subset.

effectively decreasing the model capacity. The "dying ReLU" problem can be exacerbated by high learning rates and large negative biases. Let us examine how each of these factors can contribute to the issue:

High Learning Rates: In NNs, the weights are updated using the following equation:

$$\left(W_{ij}^{(k)}\right)_{new} = \left(W_{ij}^{(k)}\right)_{old} - \alpha \frac{\partial}{\partial W_{ij}^{(k)}} \text{Loss}. \tag{28}$$

When a high learning rate, $\alpha$, is used in training a NN, the weight updates during backpropagation are more substantial, and this can lead to rapid changes in the model's parameters. i.e., if the $\alpha$ is too high, when setting the learning rate, the new weights can have a negative range. If a ReLU neuron's output is pushed into a negative region (i.e., below zero) due to the weight updates (these negative values become the new inputs to the ReLU), it may become inactive (outputting zero) for many inputs, effectively leading to the dying ReLU problem. To address this, it's common to reduce the learning rate or apply techniques like learning rate scheduling, which gradually reduces the learning rate as training progresses. Lower learning rates allow for more stable weight updates and can help prevent the dying ReLU problem.

Large Negative Bias: Neurons in a NN often have biases associated with them. A large negative bias term can shift the entire input to a neuron to the left on the ReLU AF's curve. This can make it more likely for the neuron to be pushed into a region where it always outputs zero (i.e., if a neuron's bias is set too negatively, it effectively forces the neuron to be inactive for most inputs, and it contributes to the dying ReLU problem). To mitigate the impact of a large negative bias, you can either initialize biases closer to zero or use AFs like LReLU, PReLU, or ELU, which are less sensitive to bias values far from zero.

In practice, it is often recommended to fine-tune the learning rate, use proper weight initialization techniques (e.g., Xavier/Glorot initialization), and avoid large negative biases to ensure smoother training and avoid the dying ReLU problem. Moreover, techniques like Batch Normalization (BN) can also help stabilize activations during training and mitigate the issues associated with high learning rates and biases.

*4.1.3 Understanding the Derivative of ReLU*

Some of the AF units are not actually differentiable at some input points. For example, the ReLU AF is not differentiable at $x = 0$. This may seem like it invalidates AF for use with a gradient-based learning algorithm. In practice, GD still performs well enough for these models to be used for machine learning tasks.

ReLU is not differentiable at $x = 0$. In mathematical terms, it means that the derivative of the ReLU function is undefined at $x = 0$. This can be seen when you consider the slope of the function at $x = 0$. The slope on the left side is 0 (since the function is flat at that point), and the slope on the right side is 1 (since the function is a straight line with a slope of 1 for $x > 0$). So, the function has a sharp corner at $x = 0$, and the derivative does not exist at this point.

Now, when it comes to gradient-based learning algorithms, like SGD, backpropagation, and various optimization techniques used for training NNs, the differentiability of the AFs plays a crucial role. The gradient of an AF is used to compute the gradient of the loss function with respect to the parameters of the model. This gradient is essential for adjusting the weights of the model during training to minimize the loss and improve the performance of the model. In the case of ReLU, since it is not differentiable at $x = 0$, you might wonder how gradient-based learning algorithms work with it. In practice, a common approach is to consider the subderivative or subgradient of ReLU at $x = 0$. The subgradient is a generalization of the derivative that can handle non-differentiable points. To understand the subgradient, let us delve into the mathematics and properties of the ReLU function and its subdifferential. At $x = 0$, the ReLU function has a sharp corner, which is the point of non-differentiability. The subdifferential of a function at a point is a set of slopes or gradients that describe the behavior of the function around that point, see Fig. 21. In the case of the ReLU function at $x = 0$, the subdifferential contains all possible values between 0 and 1 because there is a range of possible slopes at this point.

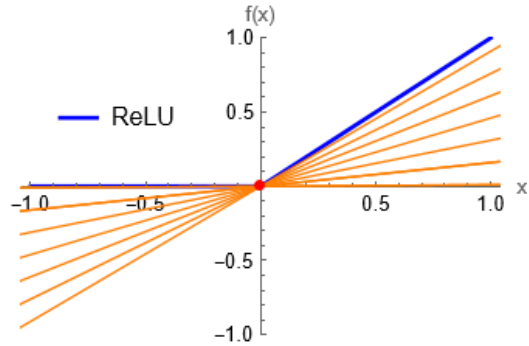

**Fig. 21.** The figure showcases the ReLU function plotted over a range of $x$ from $-1$ to $1$, with the function $\text{ReLU}(x) = \max(0, x)$. It highlights the ReLU function in blue, which sharply transitions from 0 to a linear increase at $x = 0$. Additionally, the plot includes seven possible slopes, visualized in orange, representing different elements of the subdifferential at $x = 0$. These lines illustrate the concept that at the point of non-differentiability (where $x = 0$), multiple tangent slopes could potentially describe the function. A red point marks the exact location $x = 0$ on the function to draw attention to this critical point of non-differentiability.

Mathematically, the subdifferential of the ReLU function at $x = 0$ can be expressed as:

$$\partial f(0) = [0,1].$$

Here, the square brackets denote a closed interval, meaning that the subdifferential includes both 0 and 1. This means that any value between 0 and 1 is a valid subgradient for the ReLU function at $x = 0$. In practice, the subgradient can be any value between 0 and 1, and the choice of a specific subgradient value is often based on the specific requirements of the optimization algorithm being used.

Common choices for the subgradient at $x = 0$ include:

- Subgradient = 0: This choice simplifies the calculation and is often used in practice. It effectively means that the ReLU function is treated as a straight line with a slope of 0 for $x \leq 0$. This makes the derivative 0 at $x = 0$, and the subgradient is also set to 0. In certain platforms of deep learning, when $x = 0$, the derivative of ReLU with respect to $x$ is computed as 0. Their justification is that we should favor more sparse outputs for a better training experience.
- Subgradient = 1: This choice also simplifies the calculation. It assumes that the ReLU function is a straight line with a slope of 1 for $x \geq 0$, treating it as if it were a continuous function with a derivative of 1 at $x = 0$.
- Some would choose the exact value of 0.5 as the gradient of ReLU with respect to $x$ at $x = 0$. Because the expected value of the sub-gradient over an infinite number of sub-gradients is 0.5
- Random subgradient: In some cases, for added robustness or exploration in optimization algorithms, a random subgradient within the interval [0,1] may be chosen. This can introduce some stochasticity into the training process.

The choice of subgradient at $x = 0$ can impact the convergence and behavior of gradient-based optimization algorithms. It is often chosen based on the specific problem and the needs of the training process. Regardless of the choice, it allows gradient-based optimization algorithms to continue making updates to the weights of the model, even in the presence of non-differentiable points like $x = 0$ in the ReLU function. So, while ReLU is not differentiable at $x = 0$, it can still be used in gradient-based learning algorithms by considering the subgradient or by approximating it as 0 or 1. This has proven to be effective in training DNNs and has contributed to the widespread use of ReLU in modern deep-learning architectures.

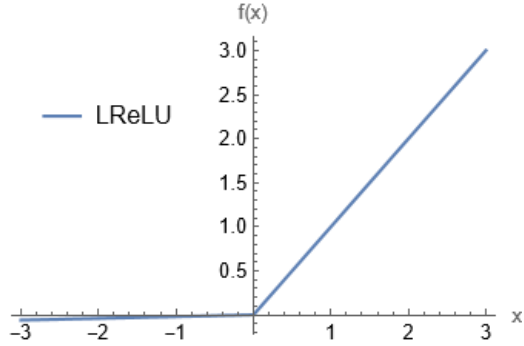

**Fig. 22.** The figure illustrates the LReLU AF, plotted over a range of $x$ from $-3$ to $3$. LReLU is designed to address the "dying ReLU" problem by allowing a small, non-zero gradient ($\alpha = 0.01$) when the input is less than zero. This is achieved by using the function LReLU, resulting in a linear relationship with a very gentle slope for negative values and a normal linear relationship for positive values.

Moreover, hidden units that are not differentiable are usually nondifferentiable at only a small number of points. The functions used in the context of NNs usually have defined left derivatives and defined right derivatives. In the case of ReLU AF, the left derivative at $x = 0$ is 0, and the right derivative is 1. Software implementations of NN training usually return one of the one-sided derivatives rather than reporting that the derivative is undefined or raising an error. This may be heuristically justified by observing that gradient-based optimization on a digital computer is subject to numerical error anyway. When a function is asked to evaluate $\sigma_{\text{ReLU}}(0)$, it is very unlikely that the underlying value truly was 0. Instead, it was likely to be some small value $\epsilon$ that was rounded to 0. For a given neuron with ReLU AF, the chances of the pre-activation value $x$ becoming exactly 0 is infinitesimally low (0.0000000000 is seldom reached).

### 4.2 Leaky Rectified Linear Unit (LReLU)

The LReLU [31] is defined as,

$$\sigma_{\text{LReLU}}(x) = \max(\alpha x, x) = \begin{cases} x, & x \geq 0, \\ \alpha x, & x < 0. \end{cases} \tag{29.1}$$

$$\frac{\partial}{\partial x}\sigma_{\text{LReLU}}(x) = \begin{cases} 1, & x > 0, \\ \alpha, & x < 0. \end{cases} \tag{29.2}$$

The amount of leak is determined by the value of hyper-parameter $\alpha$. Its value is small and generally varies between 0.01 to 0.1

LReLU is an attempt to fix the Dying ReLU problem. Instead of the function being zero when $x < 0$, a LReLU will instead have a small negative slope of 0.01, as $0.01x, x < 0$. This small slope for negative values allows the neurons to have some gradient even when the input is less than zero. This means that during backpropagation, the gradients are non-zero for both positive and negative inputs, making it easier for the network to learn. Fig. 22 shows the plot for the LReLU AF.

In a standard ReLU, the output is zero for all negative inputs, effectively "turning off" certain neurons. The sparsity property can be beneficial in some scenarios, as it reduces the overall computational load and makes the network more efficient. On the other hand, the LReLU allows a small, non-zero output for negative inputs, preventing neurons from being completely turned off. While this helps mitigate the dying ReLU problem and allows for the flow of gradients during backpropagation, it comes at the cost of losing the sparsity property.

The loss of sparsity might lead to a denser representation in the network, potentially increasing the computational requirements. However, the impact of this loss depends on the specific characteristics of the data and the goals of the model. In some cases, the benefits of avoiding dead neurons and maintaining a continuous flow of gradients during training outweigh the loss of sparsity.

Some features of the LReLU AF include its ability to allow a small gradient for negative inputs, which helps prevent neurons from becoming inactive during training. Like the standard ReLU, LReLU is simple to implement and computationally efficient. The range of the function is $(-\infty, \infty)$, and it does not saturate for positive input values, thus avoiding issues related to exploding or vanishing gradients during GD. LReLU converges much faster than Sigmoid or Tanh in practice but does not produce zero-centered output. At $x = 0$, the left-hand derivative of LReLU is $0.01$ while the right-hand derivative is $1$. Since the left-hand derivative and the right-hand derivative are not equal at $x = 0$, the LReLU function is not differentiable at $x = 0$. In the positive part of the LReLU function where the gradient is always $1$, there is no vanishing gradient problem. But on the negative part, the gradient is always $0.01$ which is close to zero. It leads to a risk of vanishing gradient problem.

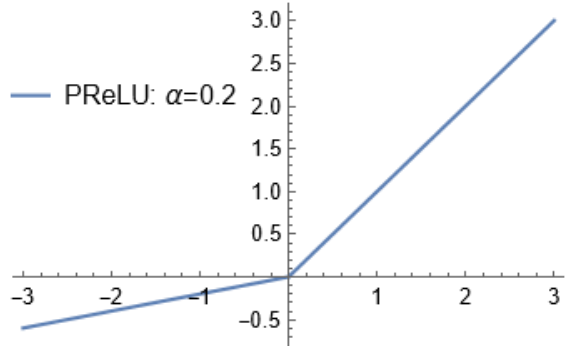

**Fig. 23.** The figure depicts the PReLU AF, plotted over a range of $x$ from $-3$ to $3$ with the parameter $\alpha = 0.2$. PReLU enhances the ReLU function by introducing a learnable parameter $\alpha$ that scales the negative part of the input, thereby allowing the gradient to flow through and avoid dead neurons for negative inputs. The function transitions from a gentle negative slope ($y = 0.2x$ for $x < 0$) to a positive linear relationship ($y = x$ for $x \geq 0$), clearly showing the adaptive nature of PReLU.

In the traditional LReLU AF, the value of $\alpha$ is indeed a fixed hyperparameter that needs to be chosen prior to the training process. This fixed value of $\alpha$ may not be optimal for all scenarios, and finding the right value often involves a degree of trial and error.

*4.3 Parametric ReLU (PReLU)*

One major problem associated with LReLU is the finding of the right slope in linear function for negative inputs. Different slopes might be suited for different problems and different networks. Thus, it is extended to PReLU by considering the slope for negative input as a trainable parameter [32]. The PReLU is given as,

$$\sigma_{\text{PReLU}}(x) = \max(0, x) + \alpha \min(0, x) = \begin{cases} x, & x \geq 0, \\ \alpha x, & x < 0. \end{cases} \quad (30.1)$$

$$\frac{\partial}{\partial x}\sigma_{\text{PReLU}}(x) = \begin{cases} 1, & x > 0, \\ \alpha, & x < 0, \end{cases} \quad (30.2)$$

in the output range of $(-\infty, \infty)$ where $\alpha$ is a learnable (trainable) parameter that determines the slope for negative values. Fig. 23 depicts the plot of the PReLU AF.

The PReLU AF possesses several significant properties. PReLU provides a form of adaptive activation, where $\alpha = 0$ makes it equivalent to ReLU, and a small, fixed $\alpha$ (e.g., $\alpha = 0.01$) makes it equivalent to LReLU. The primary goal of PReLU is to address limitations of the standard ReLU, such as the "dying ReLU" problem, where neurons can become inactive (output zero) for all inputs during training, leading to dead paths and potential loss of information. By allowing a small negative slope, PReLU seeks to mitigate this issue and enhance the learning capability of NNs.

During the training process, the parameter $\alpha$ is adjusted using GD or other optimization algorithms, allowing the network to learn an optimal value for the negative slope. This adaptability can be beneficial in scenarios where different parts of the input space might require different slopes for effective learning. PReLU introduces a very small number of extra parameters. The number of extra parameters is equal to the total number of channels, which is negligible when considering the total number of weights. There is no extra risk of overfitting. It uses a channel-shared variant: $\sigma_{\text{PReLU}}(x) = \max(0, x) + \alpha \min(0, x)$ where the coefficient is shared by all channels of one layer. This variant only introduces a single extra parameter into each layer. Hence, the computational cost is usually not significantly higher compared to standard ReLU.

The added flexibility can lead to better model performance, especially in scenarios where the data distribution varies across different dimensions.

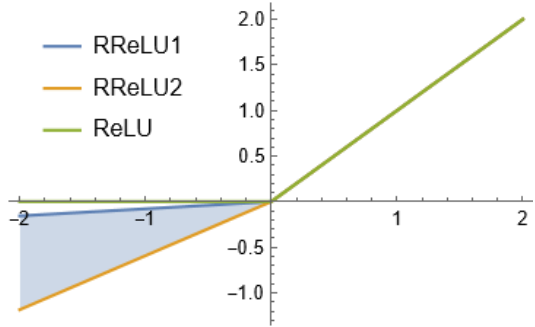

**Fig. 24.** The figure visualizes two instances of the RReLU AF alongside the standard ReLU function, plotted over a range of $x$ from $-2$ to 2. Each RReLU curve is generated with a specific, randomly chosen slope parameter $r$, resulting in two unique AFs (denoted as "RReLU1" and "RReLU2") that apply different linear transformations for negative input values. The standard ReLU function, represented here as "ReLU", acts as a reference by clamping all negative inputs to zero.

### *4.4 Randomized ReLU*

The Randomized ReLU (RReLU) introduces randomness into the AF by using a random slope for negative inputs during training. This randomness can act as a form of regularization and help prevent overfitting. The RReLU considers the slope of LReLU randomly during training sampled from a uniform distribution $U(l, u)$ [33]. The RReLU is defined as,

$$\sigma_{\text{RReLU}}(x) = \begin{cases} x, & x \geq 0, \\ r\,x, & x < 0. \end{cases} \quad (31)$$

Here, "$r$" is a random number sampled from a uniform distribution [34] within a specified range, typically [$l$ =lower, $u$ =upper], where both "lower" and "upper" are small positive constants. These values are usually set before training and remain fixed throughout the training process.

$$r \sim U(l, u), \quad l < u \quad \text{and} \quad l, u \in [0, 1). \quad (32)$$

It uses a deterministic value $\frac{x}{\frac{l+u}{2}}$ during test time. The random nature of "$r$" introduces variability into the AF. The output range is $(-\infty, \infty)$. Fig. 24 depicts the plot of the RReLU AF.

The choice of $l$ and $u$ defines the range from which $r$ is sampled. The specific values for these parameters depend on the characteristics of the problem and should be chosen through experimentation. As suggested by the Kaggle National Data Science Bowl (NDSB) competition winner, $r$ is sampled from $U(3,8)$.

The random slope adds robustness to the model by preventing neurons from shutting down completely. This ensures that even if a neuron receives consistently negative inputs, it can still contribute to the learning process. This can be beneficial for capturing subtle patterns in the data. RReLU does not significantly increase computational complexity compared to standard ReLU. The additional randomness in the slope is usually a lightweight operation in terms of computation. Moreover, the introduction of randomness in the AF can be viewed as a form of regularization. It adds noise to the learning process, which can help prevent overfitting and improve generalization to unseen data.

### *4.5 Parametric Tan Hyperbolic Linear Unit*

A Parametric Tan Hyperbolic Linear Unit (PTELU) is also used as an AF [35]. The PTELU AF is inspired from, PReLU and ELU. The PTELU is defined as,

$$\sigma_{\text{PTELU}}(x) = \begin{cases} x, & x > 0, \\ \alpha\,\text{Tanh}(\beta x), & x \leq 0, \end{cases} \quad (33)$$

where $\alpha \geq 0$, $\beta \geq 0$ are the learnable parameters. Similar to RELU, LRELU, and PRELU, PTELU uses identity mapping for positive inputs. This ensures that the gradient for positive inputs is '1' and hence, takes care of the problem of exploding/vanishing gradients. For the negative inputs, PTELU uses a tan hyperbolic function that is somewhat similar to the exponential function of ELU. Fig. 25 depicts the plot of the PTELU AF.

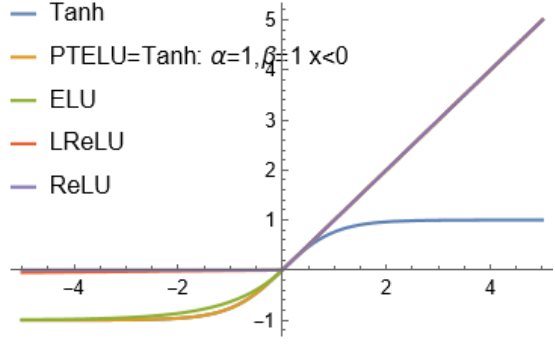

**Fig. 25.** The figure compares the Tanh, PTELU, ELU, LReLU, and ReLU AFs. The $x$-axis represents the input variable $x$ ranging from $-5$ to 5. This plot allows for a detailed comparison of the behaviors of these AFs, showcasing how each function handles positive and negative input values.

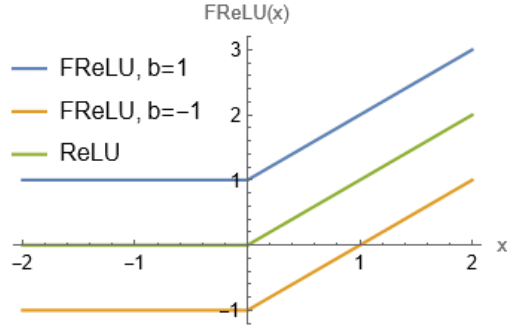

**Fig. 26.** The figure illustrates the FReLU. The $x$-axis represents the input variable $x$ ranging from $-2$ to 2. Three curves are plotted: (FReLU, $b = 1$)) showing the FReLU function with a bias value of ($b = 1$), (FReLU, $b = -1$)) showing the FReLU function with a bias value of ($b = -1$), and (ReLU) representing the standard ReLU function without any bias. The plot demonstrates how the addition of a bias term $b$ to the ReLU function shifts the function values up or down, allowing for more flexibility in modeling.

The specific design of PTELU offers the advantage of tuning parameters based on the data distribution, similar to PReLU. Geometrically, the parameter $\alpha$ controls the saturation value, while $\beta$ controls the convergence rate. The decision to use the Tanh function for negative inputs, rather than the exponential function as in ELU, is based on the intuition that Tanh has a higher gradient for small negative inputs. These larger gradients enable the AF to exit saturation more quickly. From a computational perspective, PTELU introduces twice as many learnable parameters compared to PReLU. However, this increase is still insignificant compared to the total number of weights learned in the DNN.

### *4.6 Flexible ReLU (FReLU)*

The FReLU [36] captures the negative values with a rectified point. By redesigning the rectified point of ReLU as a learnable parameter, FReLU improves flexibility on the horizontal and vertical axis, which is expressed as:

$$\sigma_{\text{FReLU}}(x) = \sigma_{\text{ReLU}}(x) + b = \begin{cases} x + b, & x > 0, \\ b, & x \leq 0, \end{cases} \quad (34.1)$$

$$\frac{\partial}{\partial x}\sigma_{\text{FReLU}}(x) = \begin{cases} 1, & x > 0, \\ 0, & x \leq 0, \end{cases} \quad (34.2)$$

$$\frac{\partial}{\partial b}\sigma_{\text{FReLU}}(x) = 1, \quad (34.3)$$

where $b$ is learnable variables. The output range is $[b, \infty)$. FReLU is non-differentiable at $x = 0$. Fig. 26 depicts the plot of the FReLU AF.

There are three output states represented by FReLU with $b < 0$:

$$\sigma_{\text{FReLU}}(x) = \begin{cases} \text{positive}, & \text{if } x > 0 \text{ and } x + b > 0, \\ \text{negative}, & \text{if } x > 0 \text{ and } x + b < 0, \\ \text{inactivation}, & x \leq 0. \end{cases} \quad (35)$$

Considering a layer with $n$ units, FReLU with $b = 0$ (equal to ReLU) or $b > 0$ can only generate $2^n$ output states, while FReLU with $b < 0$ can generate $3^n$ output states.

The FReLU function is an extension of ReLU by adding a learnable bias term $b$. Therefore, FReLU retains the same non-linear and sparse properties as ReLU and extends the output range from $[0, +\infty)$ to $[b, +\infty)$. Here, $b$ is a learnable parameter for adaptive selection by training. When $b = 0$, FReLU generates ReLU. When $b > 0$, FReLU tends to move the output distribution of ReLU to larger positive areas. When $b < 0$, FReLU expands the states of the output to increase the expressiveness of the AF.

The AF PReLU is defineded as $\sigma_{\text{PReLU}}(x) = \max(0, x) + k \min(0, x)$, where $k$ is the learnable parameter. When $k$ is a small fixed number, PReLU becomes LReLU. To avoid zero gradients, PReLU and LReLU propagate negative input with penalization, thus avoiding negative missing. However, PReLU and LReLU probably lose sparsity, which is an important factor in achieving good performance for NNs. Note that FReLU also can generate negative outputs but in a different way. FReLU obstructs the negative input as same as ReLU, the backward gradient of FReLU for the negative part is zero and retains sparsity.

The AF ELU is defined as $\text{elu}(x) = \max(x, 0) + \min((\exp(x) - 1), 0)$. FReLU and ELU have similar shapes and properties in some extent. Different from ELU, FReLU uses the bias term instead of exponential operation and reduces the computation complexity.

The FReLU AF brings several benefits, including (1) fast convergence and higher performance in NN training. (2) It has a low computation cost since it does not involve exponential operations, making it more efficient. (3) FReLU is compatible with BN, which is essential for stabilizing and accelerating the training process. (4) Additionally, it operates under weak assumptions and exhibits self-adaptation, allowing it to effectively adjust to different data distributions and scenarios, further enhancing its versatility and robustness in various applications.

### *4.7 Random Translation ReLU*

A similar arrangement is also followed by Random Translation ReLU (RTReLU) [37] by utilizing an offset, sampled from a Gaussian distribution, given as,

$$\sigma_{\text{RTReLU}}(x) = \begin{cases} x + a, & x + a > 0, \\ 0, & x + a \leq 0. \end{cases} \quad (36)$$

The output range is $[0, \infty)$ where $a$ is a random number. Fig. 27 depicts the plot of the RTReLU AF.

In the training stage, the offset of RTReLU (i.e., $a$) is randomly sampled from Gaussian distribution at each iteration. Namely,

$$a \sim N(0, \sigma), \quad (37)$$

where $\sigma$ is the standard deviation of Gaussian distribution.

In the test stage, the average value of $a$ is used for RTReLU. Because $a$ is sampled from Gaussian distribution, the average value of $a$ is 0. Thus, RTReLU in the test stage is the same as ReLU.

Based on PReLU, the randomly translational non-linear activation is called RTPReLU, which can be expressed as:

$$\sigma_{\text{RTReLU}}(x) = \begin{cases} x + a, & x + a > 0, \\ k(x + a), & x + a \leq 0, \end{cases} \quad (38)$$

where $a$ is the offset of RTPReLU on $x$-axis, which is sampled from Gaussian distribution.

The advantages of RTReLU can be summarized as follows: (1) In the training stage, incorporating the random translation near zero (i.e., $a$) will result that the output of non-linear activation is sensitive to the hard threshold zero if the input is close to zero. However, if the input is far from zero, the output is not sensitive to the hard threshold of zero. (2) Randomly translational non-linear activation can be seen as the regularization of ReLU to reduce the overfitting in the training stage; (3) Because the randomly translational non-linear activation is the same as the original non-linear activation in the test stage, it can improve accuracy with no increase in computation cost.

### *4.8 Shifted ReLU and Displaced ReLU*

The ReLU AFs are commonly used in NNs due to their simplicity and effectiveness in mitigating the vanishing gradient problem. The BN is another technique that is often employed to improve the training stability and convergence of DNNs. While both ReLU and BN have been successful in practice, there are certain interactions and complications when using them together. Here are some considerations. In ML, normalizing the distribution of the input data decreases the training time and improves test accuracy. Consequently, normalization also improves NNs performance. A standard approach to normalizing input data distributions is the mean standard technique. The input data is transformed to present zero mean and standard deviation of one. However, if instead of working with shallow NNs, we are dealing with DNNs; the problem becomes more sophisticated. Indeed, in a DNN, the output of a layer works as input data to the next. Therefore, in this sense, each layer of a deep model has his own "input data" that is composed of the previous layer output. The only exception is the first layer, for which the input is the original data. Considering each layer has its own "input data" (the output of the previous layer), normalizing only the actual input data of a DNN produces a limited effect in enhancing learning speed and test accuracy. Moreover, during the training process, the distribution of the input of each layer changes, which makes training even harder. Indeed, the parameters of a layer are updated while its input (the activations of the previous layer) is modified.

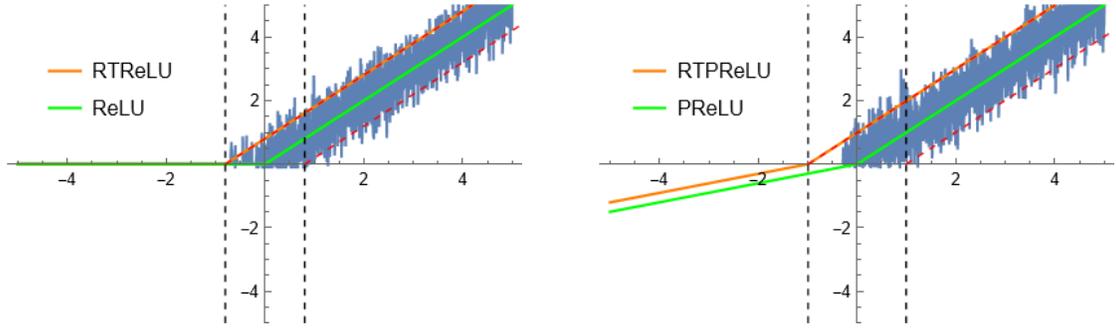

**Fig. 27.** Left panel: The figure illustrates the behavior of the RTReLU AF compared with the standard ReLU function over a range of $x$ from $-5$ to $5$. RTReLU introduces a translation based on a random variable $a$, sampled from a normal distribution with standard deviation $\sigma = 0.5$. The RTReLU curve demonstrates how the random translation affects the activation, introducing variability compared to the fixed threshold of the standard ReLU. Dashed lines indicate the regions of translation, marking the boundaries at $-a$ and $a$, and extending these lines to show the impact on the activation thresholds. Right panel: The figure compares two AF: RTPReLU, and PReLU over a range of $x$ from $-5$ to $5$. The RTReLU function incorporates a random translation sampled from a normal distribution with $\sigma = 0.5$, demonstrating variability in its activation threshold. The RTPReLU function further modifies RTReLU by applying a parameterized linear function with a slope of 0.3 for negative inputs. The PReLU function, shown with $\alpha = 0.3$, serves as a reference, applying a fixed slope for negative inputs and linear for positive. Dashed lines indicate the regions of translation, showing the boundaries at $-a$ and $a$, and extending these lines to illustrate the impact on activation thresholds. The plot highlights the flexible behavior of RTReLU and RTPReLU, compared to the more deterministic PReLU, showcasing the potential benefits of incorporating randomness and parameterization in AFs for NNs.

This phenomenon is called internal covariant shift, which is a major factor that hardens the training of DNNs. In fact, while the data of shallow NNs are normalized and static during training, the input of a deep model layer, which is the output of the previous one, is neither a priori normalized nor static throughout training. BN is an effective method to mitigate the internal covariant shift. This approach, which significantly improves training speed and test accuracy, proposes normalizing the inputs of the layers when training DNNs. Despite BN, in the case of ReLU AF, the activations are not perfectly normalized since these outputs present neither zero mean nor unit variance. Consequently, regardless of the presence of a BN layer, after the ReLU, the inputs passed to the next composed layer have neither mean of zero nor variance of one that was the objective in the first place. In this sense, ReLU skews an otherwise previously normalized output. In other words, ReLU reduces the correction of the internal covariance shift promoted by the BN layer. The ReLU bias shift effect is directly related to the drawback ReLU generates to the BN procedure.

Consequently, Shifted ReLUs (SReLU) [38] and Displaced ReLU (DReLU) [39], aiming to mitigate the mentioned problem, which is essentially a diagonally displaced ReLU were proposed. SReLU can be written as,
$$\sigma_{\text{SReLU}}(x) = \max(-1, x). \tag{39}$$
Fig. 28 depicts the plot of the SReLU AF. The DReLU is designed as a generalization of Shifted ReLU. The DReLU displaces the rectification point to consider the negative values, given as,
$$\sigma_{\text{DReLU}}(x) = \begin{cases} x, & x \geq -\delta, \\ -\delta, & x < -\delta, \end{cases} \tag{40}$$
having the output range in $[-\delta, \infty]$, see Fig. 28.

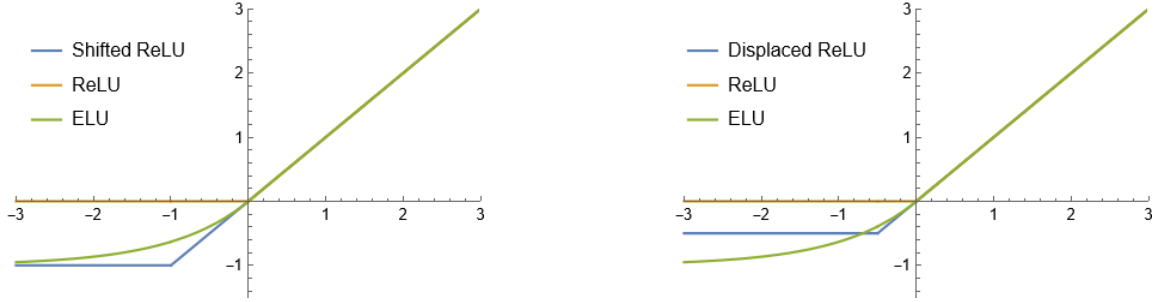

**Fig. 28.** The figure consists of two plots comparing different AFs: SReLU, ReLU, and ELU in the first plot, and DReLU, ReLU, and ELU in the second plot, all over a range of $x$ from $-3$ to $3$. Left panel: This plot visualizes the SReLU, ReLU, and ELU AFs. The SReLU function is defined as $\max(-1, x)$, shifting the ReLU function downward by 1. The ReLU function outputs zero for negative inputs and linearly increases for positive inputs. The ELU function, with $\alpha = 1$, exponentially decays for negative inputs and linearly increases for positive inputs. The plot ranges from $-1.5$ to $3$ on the $y$-axis, showing how each function behaves across the input range. Right panel: This plot shows the DReLU function, defined as $\max(x, -\delta)$ with $\delta = 0.5$, alongside the standard ReLU and ELU ($\alpha = 1$) functions. The DReLU introduces a displacement, shifting the threshold to $-0.5$, compared to the standard ReLU and ELU functions.

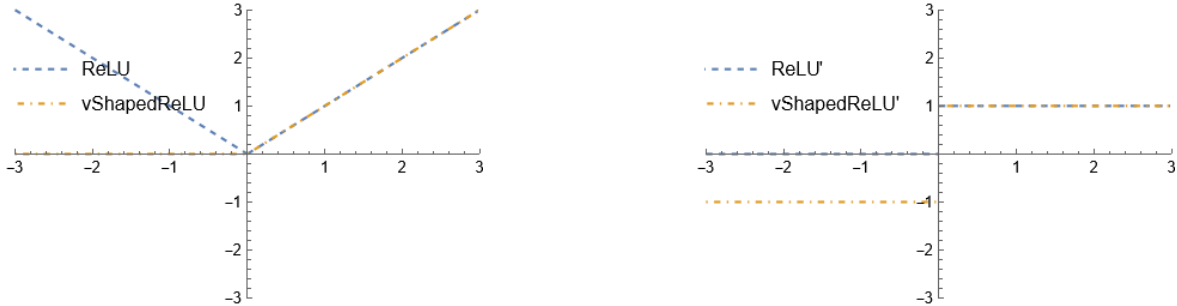

**Fig. 29.** Left panel: The figure illustrates the ReLU AF and the vReLU AF. The $x$-axis represents the input variable $x$ ranging from $-3$ to $3$. The plot range is set to fully capture the behavior of these functions across the specified range of $x$ values. Right panel: The figure shows the derivatives of the ReLU and the vReLU AFs. The $x$-axis represents the input variable $x$ ranging from $-3$ to $3$. This plot highlights the distinct differences in the rate of change between the standard ReLU and its V-Shaped variant.

DReLU is essentially a ReLU diagonally displaced into the third quadrant. Different from LReLU, PReLU, and ELU AFs, the inflection of DReLU does not happen at the origin, but in the third quadrant. It generalizes both ReLU and SReLU by allowing its inflection to move diagonally from the origin to any point of the form $(-\delta, -\delta)$. If $\delta = 0$, DReLU becomes ReLU. If $\delta = 1$, DReLU becomes SReLU. Therefore, the slope zero component of the AF provides negative activations, instead of null ones. Unlike ReLU, in DReLU learning can happen for negative inputs since the gradient is not necessarily zero. Note that, DReLU is less computationally complex than LReLU, PReLU, and ELU. In fact, since DReLU has the same shape as ReLU, it essentially has the same computational complexity.

### *4.9 V-shaped ReLU*

In another attempt, V-shaped ReLU (vReLU) AF [40] is defined as,

$$\sigma_{\text{vReLU}}(x) = \begin{cases} x, & x \geq 0, \\ -x, & x < 0, \end{cases} \tag{41}$$

having the output range in $[0, \infty]$. Fig. 29 depicts the plot of the vReLU AF.

For non-negative inputs, the function returns the input value itself $f(x) = x$. This is similar to the behavior of the standard ReLU. For negative inputs, the function returns the negative of the input value $f(x) = -x$. This creates a V-shaped pattern when plotted, hence the name. The V-shaped ReLU introduces a degree of symmetry in its response to positive and negative inputs. The negative slope for negative inputs $x < 0$ can help prevent neurons from becoming completely inactive during training (the "dying ReLU" problem). This can be especially relevant when dealing with DNNs.

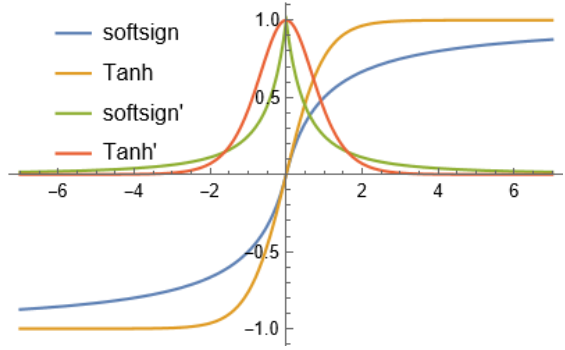

**Fig. 30.** The figure compares the Softsign AF and its derivative with the Tanh AF and its derivative. The $x$-axis represents the input variable $x$ ranging from $-7$ to $7$. This comparison highlights the similarities and differences between the Softsign and Tanh AFs, as well as their respective rates of change.

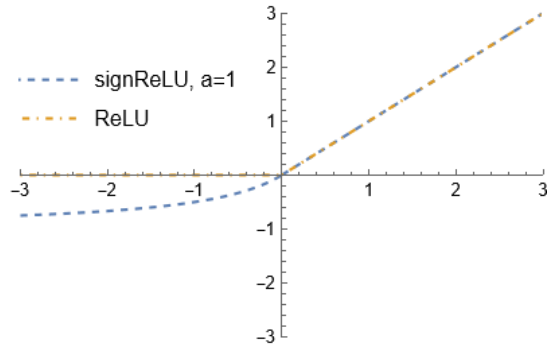

**Fig. 31.** The figure illustrates the ReLU AF and the SignReLU AF with $a = 1$. The $x$-axis represents the input variable $x$ ranging from $-3$ to $3$. This comparison highlights the differences between the standard ReLU function and its SignReLU variant, which modifies the behavior for negative input values.

### *4.10 Softsign and SignReLU AFs*

The expression of the Softsign function [41] (or called Elliott [23]) is

$$\sigma_{\text{Softsign}}(x) = \sigma_{\text{Elliott}}(x) = \frac{x}{|x| + 1}, \tag{42.1}$$

$$\frac{\partial}{\partial x}\sigma_{\text{Softsign}}(x) = \frac{1}{(|x| + 1)^2}, x \neq 0. \tag{42.2}$$

The function and its derivative are shown in Fig. 30.

The Softsign function is supposed to be a continuous approximation to the sign function (gives $-1$, $0$, or $1$ depending on whether $x$ is negative, zero, or positive.). As shown in Fig. 30, the Softsign function compresses data into the interval $(-1,1)$, similar to the Tanh function. However, unlike Tanh, Softsign's tails are quadratic polynomials rather than exponentials, causing it to approach its asymptotes much more slowly. The output of the Softsign function is centered on 0, with a relatively wide range, making the initialization process more robust. The middle part of the Softsign function is broad, and it has a high degree of non-linearity near $x = 0$, making it easier to delineate more complex boundaries. This characteristic means that the Softsign activation does not saturate as quickly as Tanh.

Softsign function is a polynomial non-linear AF, which is relatively mid in the nonlinear part. As such, there is a greater range of input values for which the Softsign assigns an output of strictly between $-1$ and $1$. It is simple in calculation with soft saturation and can reduce the number of iterations, easy convergence.

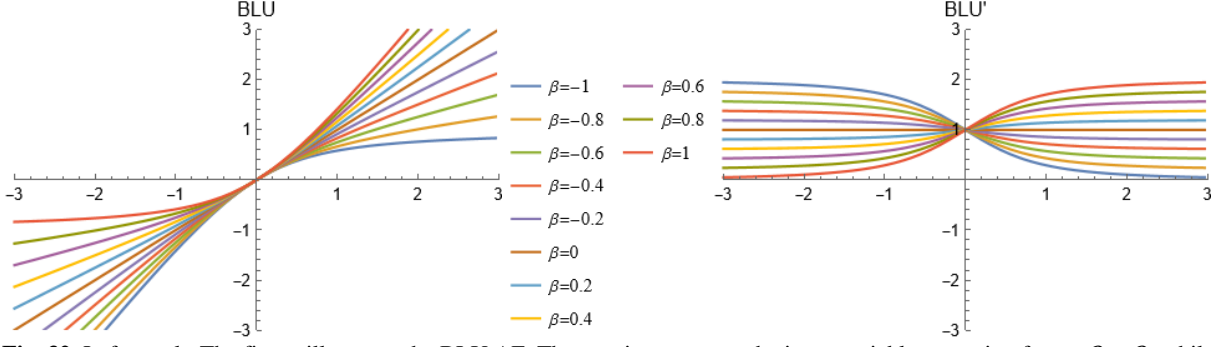

**Fig. 32.** Left panel: The figure illustrates the BLU AF. The $x$-axis represents the input variable $x$ ranging from $-3$ to 3, while the $y$-axis represents the function values. Eleven curves are plotted, each corresponding to a different value of the parameter $\beta$ ranging from $-1$ to 1 in increments of 0.2. This plot demonstrates how the BLU AF varies with different $\beta$ values, showcasing its flexibility in modeling different activation behaviors. Right panel: The figure shows the derivatives of the BLU AF. The $x$-axis represents the input variable $x$ ranging from $-3$ to 3, while the $y$-axis represents the derivative values. Eleven curves are plotted, each corresponding to a different value of the parameter $\beta$ ranging from $-1$ to 1 in increments of 0.2. This plot provides insight into how the rate of change of the BLU AF varies with different $\beta$ values, highlighting the function's adaptability.

Based on the characteristics of the ReLU function and Softsign function, a Sign ReLU (SignReLU) of unsaturated segment neuron AF was proposed [41]. The SignReLU AF utilizes the negative values using the Softsign function. The positive part of SignReLU is the same as the ReLU.

$$\sigma_{\text{SignReLU}}(x) = \begin{cases} x, & x \geq 0, \\ a\dfrac{x}{|x|+1}, & x < 0, \end{cases} \tag{43}$$

where $x$ represents the input of the nonlinear AF $\sigma_{\text{SignReLU}}$; $a$ represents the variable superparameter. Fig. 31 depicts the plot of the SignReLU AF. When $a = 0$, the function is the ReLU function.

### *4.11 Bendable Linear Unit*

A Bendable Linear Unit (BLU) has one parameter, $\beta$, where $-1 \leq \beta \leq 1$ is a learnable parameter to adapt the shape between the identity function and a rectifier function [42]. The output of a BLU is

$$\sigma_{\text{BLU}}(x) = \beta\left(\sqrt{x^2 + 1} - 1\right) + x. \tag{44.1}$$

Fig. 32 depicts the plot of the BLU AF. When $\beta$ is 0, BLU is exactly the identity function. When $0 < |\beta| \leq 1$, the function produces a nonlinear bend; higher magnitudes produce a steeper bend. The derivatives of BLU with respect to its input $x$ and parameter $\beta$ are

$$\frac{\partial}{\partial x}\sigma_{\text{BLU}}(x) = \frac{\beta x}{\sqrt{x^2+1}} + 1, \tag{44.2}$$

$$\frac{\partial}{\partial \beta}\sigma_{\text{BLU}}(x) = \sqrt{x^2} - 1. \tag{44.3}$$

BLU has the following useful properties: First, it contains the identity function. This is one of the most important advantages of a BLU. NNs that use AFs with this property can learn more effectively in deep networks than those that do not, for two reasons: 1) the gradient can neither explode nor vanish through the identity function, and 2) layers that are not needed to represent training data in a very deep network can be "skipped" using the identity function. Second, it is a non-saturating nonlinearity. Perhaps the greatest disadvantage of saturating nonlinearities like hyperbolic tangent is that their derivatives asymptotically approach 0, with the result that training can slow down or even halt as the gradient approaches 0. Non-saturating nonlinearities, like BLU and ReLU, cannot get stuck in these regions of slow learning and are thus able to learn more quickly than saturating AFs. Except in the case where $\beta = 1$, the derivative of a BLU is non-zero for both positive and negative inputs, like LReLU, as opposed to ReLU, the derivatives of which have regions of saturation for negative inputs. Third, it can have a slope greater than 1 for positive inputs. BLU can model the identity function, but it also has the capacity for growth for positive inputs; when $\beta$ is 1, the positive-side slope of BLU is 2. A slope greater than 1 for positive inputs avoids a vanishing gradient in deep networks. Fourth, it is $C^\infty$ continuous. Every order of the derivative is well-defined and smooth, in contrast with piecewise activations like ReLU and related functions. This is particularly useful in the first derivative, which yields a smoother error surface than nonlinearities that are not $C^\infty$ continuous.

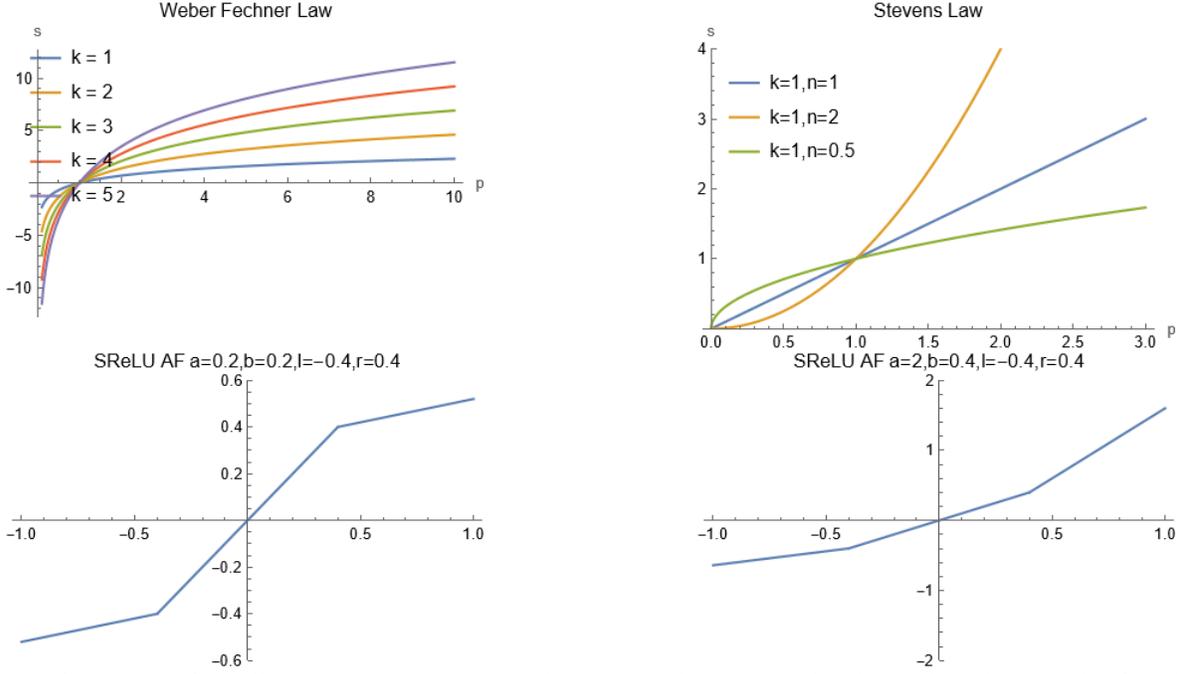

**Fig. 33.** Top left panel: The figure illustrates the Weber-Fechner law. The $x$-axis represents the stimulus intensity $p$ ranging from 0.1 to 10, while the $y$-axis represents the perceived stimulus $s$. Five curves are plotted for different values of $k$ from 1 to 5. This plot demonstrates how the perceived stimulus varies logarithmically with the physical intensity for different values of $k$. Top right panel: The figure illustrates Stevens Law. The $x$-axis represents the stimulus intensity $p$ ranging from 0 to 3, while the $y$-axis represents the perceived stimulus $s$. Three curves are plotted for different values of $n$ (1, 2, and 0.5) with $k$ fixed at 1. This plot shows how the perceived stimulus varies non-linearly with the physical intensity for different values of $n$. Bottom left panel: The figure shows the SReLU AF. The $x$-axis represents the input variable $x$ ranging from $-1$ to 1, while the $y$-axis represents the function values ranging from $-0.6$ to 0.6. The parameters are set to $a = 0.2$, $b = 0.2$, $l = -0.4$, and $r = 0.4$. This plot demonstrates the SReLU function's behavior across the specified range of $x$ values, showing smooth transitions between the linear segments. Bottom right panel: The figure illustrates the SReLU AF with parameters $a = 2$, $b = 0.4$, $l = -0.4$, and $r = 0.4$. The $x$-axis represents the input variable $x$ ranging from $-1$ to 1, while the $y$-axis represents the function values ranging from $-2$ to 2. This plot shows the SReLU function's behavior with a steeper slope due to the larger value of $a$, highlighting how changing the parameters affects the function's shape and range.

### *4.12 S-Shaped ReLU (SReLU)*

The Webner-Fechner law [43] and the Stevens law [44] are two basic laws in neural sciences [45]. The Weber-Fechner law describes the relationship between the intensity of a physical stimulus and the perceived intensity of the stimulus. The law was formulated by Ernst Heinrich Weber and later refined by Gustav Theodor Fechner in the 19th century. The Weber-Fechner law suggests that the perceived intensity of a stimulus is not directly proportional to its physical intensity but rather to the logarithm of its intensity. The Webner-Fechner law holds that the perceived magnitude $s$ is a logarithmic function of the stimulus intensity $p$ multiplied by a modality and a dimension specific constant $k$, see Fig. 33. That is,

$$s = k \log p. \tag{45}$$

Let us consider an example using the sense of touch and the perception of weight. According to the Weber-Fechner law, the perceived difference in weight is proportional to the ratio of the actual weights rather than the absolute difference in weights.

Stevens' power law is a psychological principle that describes the relationship between the intensity of a stimulus and the perceived sensation or perception associated with that stimulus. The law was proposed by psychologist S. S. Stevens in the 1950s and is a part of psychophysics, which is the branch of psychology that deals with the quantitative relationship between physical stimuli and perceptual experiences. The Stevens law explains the relationship through a power function, i.e.,

$$s = kp^n, \tag{46}$$

where all the parameters have the same definitions as in the Webner-Fechner law, except for an additional parameter $n$ which is an exponent depending on the type of the stimulus, see Fig. 33. The two laws have been verified through lots of experiments [45]. For example, in vision, the amount of change in brightness with respect to the present brightness accords with the Webner-Fechner law. In neural sciences, these two laws also explain many properties of sensory neurons and the response characteristics of receptor cells [45].

Motivated by the previous research on these two laws, the S-shaped rectified linear activation unit (SReLU) was proposed which imitates the logarithm function and the power function given by the Webner-Fechner law and the Stevens law, respectively, and uses piecewise linear functions to approximate non-linear convex and non-convex functions.

None of ReLU, LReLU, and PReLU can learn the non-convex functions since they are essentially all convex functions. The SReLU was proposed to learn both convex and non-convex functions [46]. SReLU consists of three piecewise linear functions, which are formulated by four learnable parameters. SReLU is essentially defined as:

$$\sigma_{\text{SReLU}}(x) = \begin{cases} r + a(x - r), & x \geq r, \\ x, & l < x < r, \\ l + b(x - l), & x \leq l, \end{cases} \tag{47.1}$$

$$\frac{\partial}{\partial x}\sigma_{\text{SReLU}}(x) = \begin{cases} a, & x \geq r, \\ 1, & l < x < r, \\ b, & x \leq l, \end{cases} \tag{47.2}$$

$$\frac{\partial}{\partial r}\sigma_{\text{SReLU}}(x) = I(x \geq r)(1 - a), \tag{47.3}$$

$$\frac{\partial}{\partial a}\sigma_{\text{SReLU}}(x) = I(x \geq r)(x - r), \tag{47.4}$$

$$\frac{\partial}{\partial l}\sigma_{\text{SReLU}}(x) = I(x \leq l)(1 - b), \tag{47.5}$$

$$\frac{\partial}{\partial b}\sigma_{\text{SReLU}}(x) = I(x \leq l)(x - l), \tag{47.6}$$

where $\{r, a, l, b\}$ are four learnable parameters used to model an individual SReLU activation unit and $I(\cdot)$ is an indicator function and $I(\cdot) = 1$ when the expression inside holds true, otherwise $I(\cdot) = 0$. As shown in Fig. 33, in the positive direction, $a$ is the slope of the right line when the inputs exceed the threshold $r$. Symmetrically, $l$ is used to represent another threshold in the negative direction. When the inputs are smaller than $l$, the outputs are calculated by the left line. When the inputs of SReLU fall into the range of $(l, r)$, the outputs are linear functions with unit slope 1 and bias 0.

By designing SReLU in this way, it can imitate the formulations of multiple non-linear functions, including the logarithm function (45) and the power function (46) given by the Webner-Fechner law and the Stevens law, respectively. As shown in Fig. 33, when $r > 1, a > 0$, the positive part of SReLU imitates the power function with the exponent $n$ larger than 1; when $1 > r > 0$, $r > 0$, the positive part of SReLU imitates the logarithm function; when $r = 1, a > 0$, SReLU follows the power function with the exponent 1. For the negative part of SReLU, we have a similar observation except for the inverse representation of the logarithm function and the power function as analyzed for its positive counterpart. The reason for setting the middle line to be a linear function with slope 1 and bias 0 when the input is within the range $(l, r)$ is that it can better approximate both (45) and (46) using such a function, because the change of the outputs with respect to the inputs is slow when the inputs are in small magnitudes.

The usage of SReLU brings the following advantages to the DNN: (1) SReLU can learn both convex and non-convex functions, without imposing any constraints on its learnable parameters, thus the DNN with SReLU has a stronger feature learning capability. (2) Since SReLU utilizes piecewise linear functions rather than saturated functions, thus it shares the same advantages of the non-saturated AFs. (3) It does not suffer from the "exploding/vanishing gradient" problem and has a high computational speed during the forward and backpropagation of deep networks. (4) It can be easily concluded that ReLU, LReLU, and PReLU can be seen as special cases of SReLU. Specifically, when $r \geq 0, a = 1, l = 0, b = 0$, SReLU is degenerated into ReLU; when $r \geq 0, a = 1, l = 0, b > 0$, SReLU is transformed to LReLU and PReLU. However, ReLU, LReLU, and PReLU can only approximate convex functions, while SReLU is able to approximate both convex and non-convex functions.

### 4.13 Elastic ReLU and Elastic Parametric ReLU

An Elastic ReLU (EReLU) considers a slope randomly drawn from a uniform distribution during the training for the positive inputs to control the amount of non-linearity [47]. The EReLU is defined as,

$$\sigma_{\text{EReLU}}(x) = \max(R\,x, 0) = \begin{cases} R\,x, & x > 0, \\ 0, & x \leq 0, \end{cases} \tag{48}$$

in the output range of $[0, \infty)$ where $R$ is randomly selected from a uniform distribution denoted by $R \sim U(1 - \alpha, 1 + \alpha)$ with $\alpha \in (0, 1)$. And $\alpha$ is a variable representing the degree of response fluctuation. It is noted that $R$ only changes during the training stage. Fig. 34 depicts the plot of the EReLU AF. At test time, $R$ equal to the expectation value of $R$, $\mathbb{E}[R]$. Obviously, $\mathbb{E}[R]$ equals to 1 given $R \sim U(1 - \alpha, 1 + \alpha)$. Therefore, at test time, EReLU becomes ReLU.

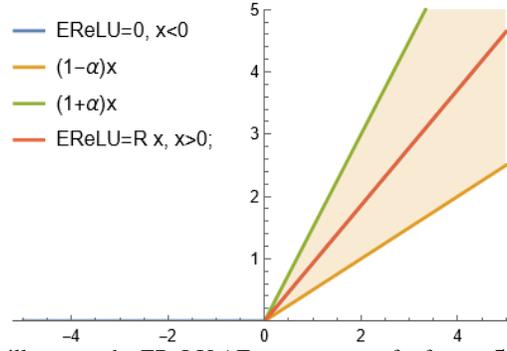

**Fig. 34.** The figure illustrates the EReLU AF over a range of $x$ from $-5$ to 5, highlighting its adaptive nature by introducing a random scaling factor $R$ for positive inputs, sampled from a uniform distribution between $(1 - \alpha)$ and $(1 + \alpha)$, $\alpha = 0.5$. The plot displays the EReLU function (depicted as $R\,x$), along with the lower bound $(1 - \alpha)x$ and upper bound $(1 + \alpha)x$, filling the area between these bounds to emphasize the range of possible values.

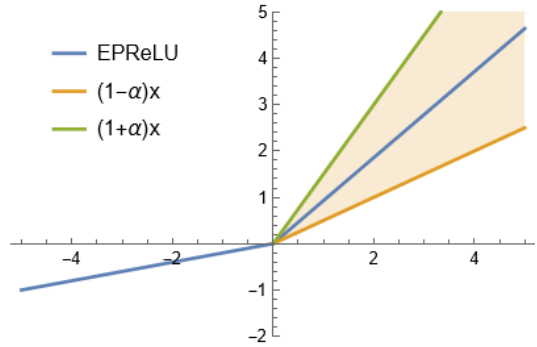

**Fig. 35.** The figure illustrates the EPReLU AF over a range of $x$ from $-5$ to 5, showcasing its flexible behavior by incorporating a random scaling factor $R$ for positive inputs and a fixed scaling factor $a = 0.2$ for negative inputs. The plot includes the EPReLU function, with $R$ sampled from $[1 - \alpha, 1 + \alpha]$ where $\alpha = 0.5$, and also displays the lower bound $(1 - \alpha)\,x$ and upper bound $(1 + \alpha)\,x$, filling the area between these bounds to emphasize the range of possible values.

An illustration of EReLU is shown in Fig. 34. It is seen that the positive outputs of EReLU do not lie in one line. Instead, they lie in the region determined by two rays from the origin, which are denoted by $\sigma_{\text{EReLU 1}}(x) = (1 - \alpha)x$ and $\sigma_{\text{EReLU 2}} = (1 + \alpha)x$, respectively. Given an input $x > 0$, ReLU outputs an exact value $\sigma_{\text{ReLU}}(x) = x$ whereas EReLU generates a random value selected uniformly from $[(1 - \alpha)x, (1 + \alpha)x]$.

EReLU has several useful properties. It improves model fitting without introducing any extra parameters and poses little risk of overfitting. During the training stage, EReLU causes the positive part of the AF to fluctuate within a moderate range in each epoch, enhancing the robustness of the network model. This characteristic helps maintain a stable learning process, contributing to the overall performance and reliability of the NN.

PReLU and EReLU are two quite different techniques. PReLU mainly improves the negative part of the AF while EReLU reforms the positive part of the AF. Elastic Parametric ReLU (EPReLU) combines the advantages of EReLU and PReLU. Formally, the EPReLU was defined as:

$$\sigma_{\text{EPReLU}}(x) = \max(R\,x, a\,x) = \begin{cases} R\,x, & x > 0, \\ a\,x, & x \leq 0. \end{cases} \tag{49.1}$$

$$\frac{\partial}{\partial x} \sigma_{\text{EPReLU}}(x) = \begin{cases} R, & x > 0, \\ a, & x \leq 0. \end{cases} \tag{49.2}$$

It is seen that EPReLU utilizes EReLU and PReLU to process the positive part and negative part, respectively. Fig. 35 depicts the plot of the EPReLU AF. It is important to note that the two variables $R$ and $a$ are processed quite differently. $R$ is randomly chosen from a uniform distribution whereas $a$ needs to be updated using backpropagation. Both R and $a$ have an effect on the updating of the weights $W$ of networks during the training stage.

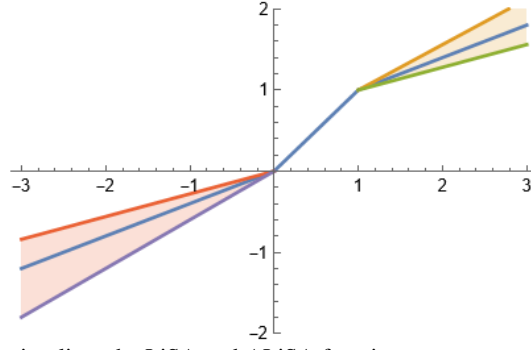

**Fig. 36.** The figure visualizes the LiSA and ALiSA functions over a range of $x$ from $-3$ to $3$, illustrating their piecewise linear behavior across different input regions. The LiSA and ALiSA functions are defined with three segments: a scaled linear function for negative inputs, a linear region for inputs between 0 and 1, and another scaled linear function for positive inputs greater than 1. The plot includes the LiSA and ALiSA functions with parameters $\alpha_1 = 0.4$ and $\alpha_2 = 0.4$, as well as upper and lower bounds for the positive and negative segments. The areas between the bounds are filled to emphasize the range of variability.

## *4.14 Linearized Sigmoidal Activation*

The Linearized Sigmoidal Activation (LiSA) [48] was defined as:
$$\sigma_{\text{LiSA}}(x) = \begin{cases} \alpha_1 x - \alpha_1 + 1, & 1 < x < \infty, \\ x, & 0 \leq x \leq 1, \\ \alpha_2 x, & -\infty < x < 0. \end{cases} \quad (50.1)$$
$$\frac{\partial}{\partial x}\sigma_{\text{LiSA}}(x) = \begin{cases} \alpha_1, & 1 < x < \infty, \\ 1, & 0 \leq x \leq 1, \\ \alpha_2, & -\infty < x < 0. \end{cases} \quad (50.2)$$
The function considers the input data range in three activity regions, as shown in Fig. 36: the positive activity region for $1 < x < \infty$, the linear activity region for $0 \leq x \leq 1$, and the negative activity region for $-\infty < x < 0$.

Instead of considering data into only two segments (positive or negative), LiSA function divides the data range into multiple segments. The data within a single range segment hold a linear relation, whereas the data points from different range segments hold a non-linear relationship. In other words, all the data points falling in the single range segment (e.g., within linear, positive, or negative activity regions) will hold a linear association. Hence, the model can exploit a wide range of linear and non-linear structural associations in data.

Above mentioned behavior brings qualities of saturating and non-saturating AFs into LiSA. Similar to non-saturating AF, LiSA provides a smooth gradient flow even in models with much higher depth and hence does not suffer from vanishing gradient problem. On the other hand, LiSA is able to model higher-order non-linear data associations because all the segments exhibit different activation behavior. In other words, the LiSA AF shows segment-wise linear behavior similar to non-saturating AFs. But instead of modeling the activity of neurons with a completely linear function, LiSA makes smaller non-linear range segments; which provides LiSA capability to learn higher-order non-linear transformations similar to saturating AFs.

This nonlinearity is of paramount importance while modeling higher-complexity data. The LiSA function considers three linear functions to increase the non-linearity characteristics.

LiSA exhibits linear behavior for the middle range segment of input (linear activity region), hence data passes unaffected (similar to ReLU); whereas input values with high variance (value outside linear activity region) are suppressed.

If the coefficients $\alpha_1$ and $\alpha_2$ are set as follows: $\alpha_1 = 1$ and $\alpha_2 = 0$, the LiSA function transforms into ReLU AF.
$$\begin{aligned}\sigma_{\text{LiSA}}(x) &= \begin{cases} x, & 1 < x < \infty \\ x, & 0 \leq x \leq 1 \\ 0, & -\infty < x < 0 \end{cases} \\ &= \begin{cases} x, & 0 \leq x \leq \infty \\ 0, & -\infty < x < 0 \end{cases} \\ &= \sigma_{\text{ReLU}}(x).\end{aligned} \quad (51)$$

Similarly, if slope coefficients of LiSA function are set as: $\alpha_1 = 1$ and $\alpha_2 = a$. Then, the LiSA function takes the shape of LReLU AF.

$$\sigma_{\text{LiSA}}(x) = \begin{cases} x, & 1 < x < \infty \\ x, & 0 \leq x \leq 1 \\ ax, & -\infty < x < 0 \end{cases}$$
$$= \begin{cases} x, & 0 \leq x \leq \infty \\ ax, & -\infty < x < 0 \end{cases}$$
$$= \sigma_{\text{LReLU}}(x). \tag{52}$$

The Adaptive Linearized Sigmoidal Activation (ALiSA) function uses trainable slope parameters which can adjust themselves to the required value according to the task at hand. These trainable slope parameters are trained alongside the model parameters during the training process. The basic structure of ALiSA remains the same as LiSA. Hence, the same figure can be used to represent the function (Fig. 36).

$$\sigma_{\text{ALiSA}}(x) = \begin{cases} \alpha_1 x - \alpha_1 + 1, & 1 < x < \infty, \\ x, & 0 \leq x \leq 1, \\ \alpha_2 x, & -\infty < x < 0. \end{cases} \tag{53.1}$$

$$\frac{\partial}{\partial x}\sigma_{\text{ALiSA}}(x) = \begin{cases} \alpha_1, & 1 < x < \infty, \\ 1, & 0 \leq x \leq 1, \\ \alpha_2, & -\infty < x < 0. \end{cases} \tag{53.2}$$

$$\frac{\partial}{\partial \alpha_1}\sigma_{\text{ALiSA}}(x) = \begin{cases} x - 1, & 1 < x < \infty, \\ 0, & 0 \leq x \leq 1, \\ 0, & -\infty < x < 0. \end{cases} \tag{53.3}$$

$$\frac{\partial}{\partial \alpha_2}\sigma_{\text{ALiSA}}(x) = \begin{cases} 0, & 1 < x < \infty, \\ 0, & 0 \leq x \leq 1, \\ x, & -\infty < x < 0. \end{cases} \tag{53.4}$$

*4.15 Bounded ReLU and Bounded LReLU*

The conventional ReLU function has unbounded outputs for non-negative inputs. The unbounded outputs of ReLU and many of its variants may lead to training instability. According to the universal approximation theorem, a function should be bounded within a range of inputs. Therefore, an upper boundary condition is added to $\sigma_{\text{ReLU}}(x) = \max(0, x)$ to produce a bounded version of the ReLU function (BReLU) [49]:

$$\sigma_{\text{BReLU}}(x) = \min(\max(0, x), A) = \begin{cases} 0, & x \leq 0, \\ x, & 0 < x \leq A, \\ A, & x > A. \end{cases} \tag{54.1}$$

$$\frac{\partial}{\partial x}\sigma_{\text{BReLU}}(x) = \begin{cases} 1, & 0 < x \leq A, \\ 0, & \text{otherwise}, \end{cases} \tag{54.2}$$

where $A$ defines the maximum output value the function can produce. This only requires an additional minimum operator, and has an additional interesting effect of creating a sigmoidal-like activation pattern. The training stability is improved in BReLU due to two rectifications (i.e., at 0 and $A$).

The LReLU function has non-zero gradient values for all negative input values. Similar to the bounded ReLU function, another limiting condition is introduced as its output approaches to a positive value $A$. Instead of hard-limiting any output values greater than $A$ to be equal to $A$, a slightly oblique line with a constant small positive gradient is added to create the odd symmetricity about the origin, and its gradient value is set to be identical to the line region where $x \leq 0$. The Bounded Leaky ReLU (BLReLU) [49] is defined as,

$$\sigma_{\text{BLReLU}}(x) = \begin{cases} 0.01x, & x \leq 0, \\ x, & 0 < x \leq A, \\ 0.01x + c, & x > A. \end{cases} \tag{55}$$

In order to create a continuous function over all regions, $c$ is set as $0.99A$ by solving the equation $x = 0.01x + c$ when $x = A$, which produces the following equation:

$$\sigma_{\text{BLReLU}}(x) = \begin{cases} 0.01x, & x \leq 0, \\ x, & 0 < x \leq A, \\ 0.01x + 0.99A, & x > A. \end{cases} \tag{56.1}$$

$$\frac{\partial}{\partial x}\sigma_{\text{BLReLU}}(x) = \begin{cases} 1, & 0 < x \leq A, \\ 0.01, & \text{otherwise}. \end{cases} \tag{56.2}$$

The BLReLU function has a similar sigmoidal-like activation pattern as the bounded ReLU function, except that small gradients exist at both sides of the region where input is not within the region of $[0, A]$. Fig. 37 illustrates the activation and gradient curves of the BReLU and BLReLU functions.

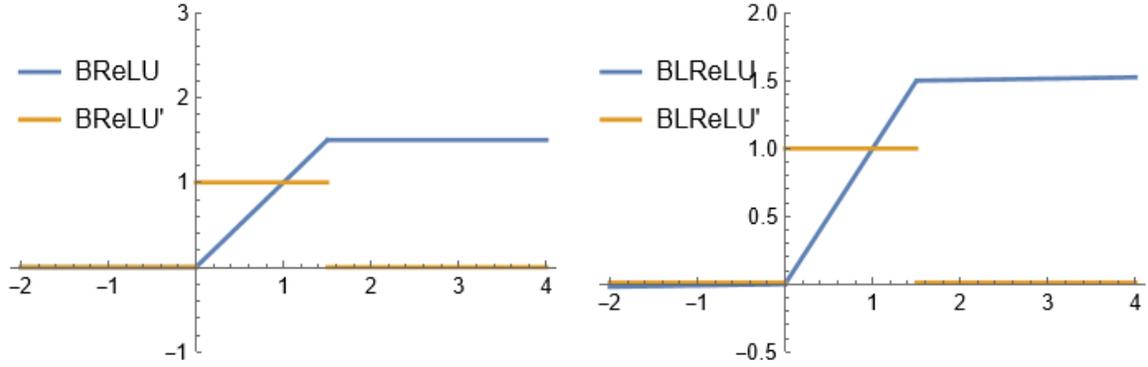

**Fig. 37.** Left panel: The figure illustrates the BReLU AF and its derivative. The $x$-axis represents the input variable $x$ ranging from $-2$ to 4. The plot range is set to fully capture the behavior of these functions across the specified range of $x$ values. Right panel: The figure illustrates the BLReLU AF and its derivative. The $x$-axis represents the input variable $x$ ranging from $-2$ to 4.

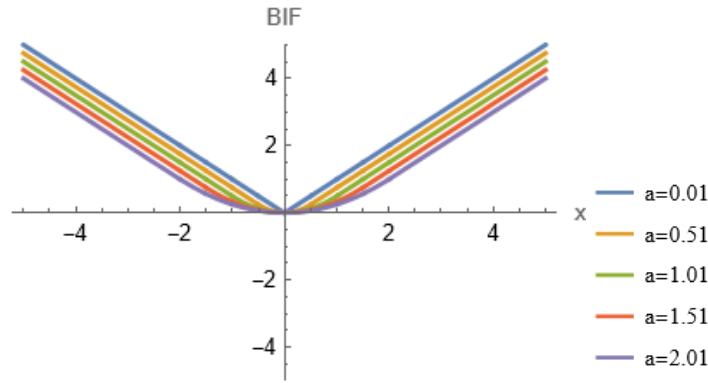

**Fig. 38.** The figure illustrates the BIF AF. The $x$-axis represents the input variable $x$ ranging from $-5$ to 5, while the $y$-axis represents the function values ranging from $-5$ to 5. Five curves are plotted for different values of $a$ (0.01, 0.51, 1.01, 1.51, and 2.01). The legend, positioned in the lower right of the plot, identifies each curve by its corresponding $a$ value. This plot demonstrates how the BIF AF changes its shape depending on the parameter $a$, highlighting its flexibility and different response characteristics for varying $a$.

Both BReLU and BLReLU functions are non-constant, limited, continuous, monotonically increasing, and piecewise differentiable for all input values $x$. The gradient diffusion problem primarily arises due to the large saturation region of an AF, where smaller gradients are produced, leading to inefficient error propagation. However, bounded AFs like BReLU and BLReLU are designed to alleviate this problem by maintaining constant gradient values within their active regions. This design ensures that gradient values are not suppressed during backpropagation. Additionally, as the active region is dependent on the choice of the hyperparameter value for the upper output boundary, the idea is that as long as the output boundary is large enough to alleviate the vanishing gradient problem, but at the same time small enough to avoid the numerical instability problem, then the NN training will produce faster and more stable convergence.

### 4.16 Bi-firing and Bounded Bi-firing AFs

The symmetric and piecewise Bi-firing (BIF) [50] AF, $\sigma_{\text{PBIF}}$, is defined as follows:

$$\sigma_{\text{BIF}}(x) = \begin{cases} -x - \dfrac{a}{2}, & x < -a, \\ x - \dfrac{a}{2}, & x > a, \\ \dfrac{x^2}{2a}, & \text{else,} \end{cases} \tag{57}$$

where $a > 0$ is a smoothing parameter. Fig. 38 shows the BIF AFs with different values of the smooth parameter.

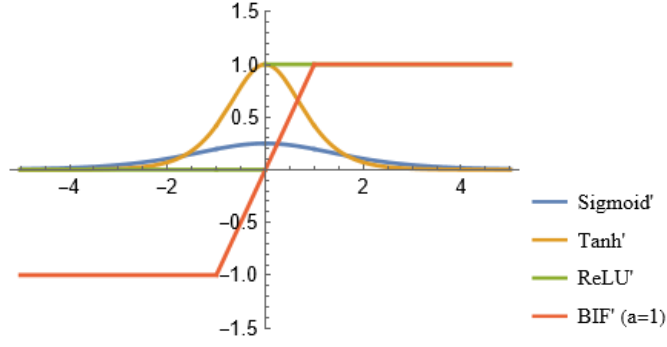

**Fig. 39.** The figure illustrates the derivatives of various AFs, including the Sigmoid, Tanh, ReLU, and BIF AF with $a = 1$. The $x$-axis represents the input variable $x$ ranging from $-5$ to $5$. This plot allows for a comprehensive comparison of the different behaviors of the AF derivatives across the specified range of $x$ values, highlighting their distinct characteristics.

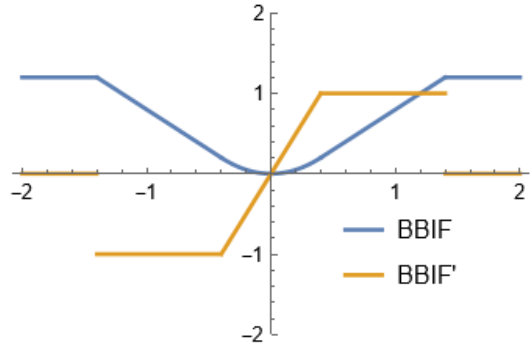

**Fig. 40.** The figure illustrates the BBIF AF and its derivative. The $x$-axis represents the input variable $x$ ranging from $-2$ to $2$. The plot demonstrates the BBIF's behavior and its derivative across the specified range of $x$ values, highlighting the different regions (hard limits, linear, and quadratic) of the AF.

A small value of $a$ leads to a sharper AF near the point which input is equal to zero. The size of the quadratic region increases when the value of $a$ increases. The BIF AF is a differentiable function whose first order derivative is continuous. The gradient magnitude of the BIF function within the region between $-a \leq x \leq a$ is $\left|\frac{x}{a}\right|$ and is equal to 1 elsewhere. On the other hand, the BIF function is an unbounded function which may cause numerical problems.

The BIF AF has several advantages: Firstly, most of the hidden activations are in linear regions of the function. As defined in (57), regions $x > a$ and $x < -a$ are linear regions of the BIF functions while the region $a < x < -a$ is a quadratic region. So, for a small $a$ (e.g., $a = 0.01$) and random initialization of network weights, the number of input $x$ falling into the quadratic region is much smaller than that falling in linear regions. Secondly, the BIF function has a very small saturation region. Therefore, gradient diffusion will not occur when the BIF function is used as AF for DNN. Thirdly, the computational cost is cheaper than the Sigmoid and Tanh functions because no exponential function is involved in the computation of activation value. Finally, in comparison to rectifier function, the BIF function is differentiable and symmetric which provides a better representation to symmetric data.

The saturation region is defined as the input region with a gradient magnitude near or equal to zero, e.g. less than $10^{-3}$. When the input of a neuron falls in this region, the error can not be propagated to the next layer efficiently. Fig. 39 shows the gradient magnitude of four different AFs in the real value domain. The derivative of the Sigmoid function is $\sigma'_{\text{sigmoid}}(x) = 1/(2 + e^x + e^{-x})$. When input $x$ is larger than 6.9 or smaller than $-6.9$, the gradient magnitude of the Sigmoid function is smaller than $10^{-3}$, i.e., saturated. The derivative of the Tanh function is $\sigma'_{\text{Tanh}}(x) = 2/(e^x + e^{-x})$. The Tanh function saturates when the input $x$ is larger than 4 or smaller than $-4$. The gradient magnitude of the rectifier function is zero when input $x < 0$. The derivative of BIF is:

$$\frac{\partial}{\partial x}\sigma_{\text{BIF}}(x) = \begin{cases} -1, & x < a, \\ 1, & x > a, \\ \dfrac{x}{a}, & -a \leq x \leq a. \end{cases} \tag{58}$$

The gradient magnitude of BIF function equals 1 when the absolute value of input $x$ is larger than $a$. The gradient magnitude of the BIF function is smaller than $10^{-3}$ only when the input $x$ falls in the range $-10^{-3}a < x < 10^{-3}a$, e.g. $-10^{-5} < x < 10^{-5}$ for $a = 0.01$. In comparison to other three widely used AFs, the saturation region of the BIF function is very small.

The BIF has unbounded outputs and thus requires some weight penalty techniques during the training process to reduce the risk of training instability. The Bounded Bi-firing (BBIF) [49] AF was proposed to avoid the necessity of using any weight penalty techniques when training an NN model with the BIF AF. This is accomplished by restricting the maximum activation value that the function can produce by adding hard limits on both sides using an upper boundary value $b$ to preserve the even symmetricity of the function:

$$\sigma_{\text{BBIF}}(x) = \begin{cases} b, & x < -b - a/2, \\ -x - \dfrac{a}{2}, & -b - \dfrac{a}{2} \leq x < -a, \\ \dfrac{x^2}{2a}, & -a \leq x \leq a, \\ x - a/2, & a < x \leq b + a/2, \\ b, & x > b + a/2. \end{cases} \tag{59.1}$$

$$\frac{\partial}{\partial x}\sigma_{\text{BBIF}}(x) = \begin{cases} -1, & -b - \dfrac{a}{2} \leq x < -a, \\ x/a, & -a \leq x \leq a, \\ 1, & a < x \leq b + a/2, \\ 0 & \text{otherwise.} \end{cases} \tag{59.2}$$

The BBIF function has a near inverse bell-shaped activation curve which is even symmetrical about the origin (as illustrated in Fig. 40). BBIF AF has non-constant outputs for a range of inputs, bounded within a range of input values, continuous for all ranges of inputs, monotonically increasing only for $x \geq 0$ and piecewise differentiable over all input values.

### *4.17 Rectified Linear Tanh*

In order to diminish the vanishing gradient problem that perplexes Tanh and reduce bias shift and noise-sensitiveness that torments the ReLU family, Rectified Linear Tanh (ReLTanh) AF was proposed [51]. The AF ReLTanh is defined as follows.

$$\sigma_{\text{ReLTanh}}(x) = \begin{cases} \text{Tanh}'(\lambda^+)(x - \lambda^+) + \text{Tanh}(\lambda^+), & x \geq \lambda^+, \\ \text{Tanh}(x), & \lambda^- < x < \lambda^+, \\ \text{Tanh}'(\lambda^-)(x - \lambda^-) + \text{Tanh}(\lambda^-), & x \leq \lambda^-. \end{cases} \tag{60.1}$$

$$\sigma'_{\text{ReLTanh}}(x) = \begin{cases} \text{Tanh}''(\lambda^+), & x \geq \lambda^+, \\ \text{Tanh}'(x), & \lambda^- < x < \lambda^+, \\ \text{Tanh}''(\lambda^+), & x \leq \lambda^-, \end{cases} \tag{60.2}$$

where $\lambda^+_{\text{lower}} \leq \lambda^+ \leq \lambda^+_{\text{upper}}$ and $\lambda^-_{\text{lower}} \leq \lambda^- \leq \lambda^-_{\text{upper}}$. Fig. 41 depicts the plot of the ReLTanh AF.

ReLTanh is constructed by replacing the saturated waveforms of Tanh in the positive and negative inactive regions with two straight lines. The slopes of these lines are calculated using Tanh's derivatives at two learnable thresholds, resulting in a function that consists of a nonlinear Tanh center and two linear parts on both ends. The positive line is steeper than the negative one, and both lines start at their respective learnable thresholds. The middle Tanh waveform provides ReLTanh with the ability for nonlinear fitting, while the linear parts help alleviate the vanishing gradient problem. The thresholds, $\lambda^+$ and $\lambda^-$, determine the start positions and slopes of the straight lines and can be trained using the backpropagation algorithm. Besides, both $\lambda^+$ and $\lambda^-$ can be trained by backpropagation algorithm, so it can tolerate the variation of inputs and help to minimize the cost function and maximize the data fitting performance. Additionally, it is worth noting there are extra limiting conditions for both $\lambda^+$ and $\lambda^-$: $\lambda^+_{\text{lower}} \leq \lambda^+ \leq \lambda^+_{\text{upper}}$ and $\lambda^-_{\text{lower}} \leq \lambda^- \leq \lambda^-_{\text{upper}}$, and they are mainly used to constrain the learnable range of slopes to avoid unreasonable waveform and guarantee the gradient-vanishing proof capacity. These thresholds can be trained and updated along the GD direction of cost function. So ReLTanh can improve the vanishing gradient problem just like the ReLU family, and its mean outputs are closer to zero so that it is affected less bias shift than the ReLU family.

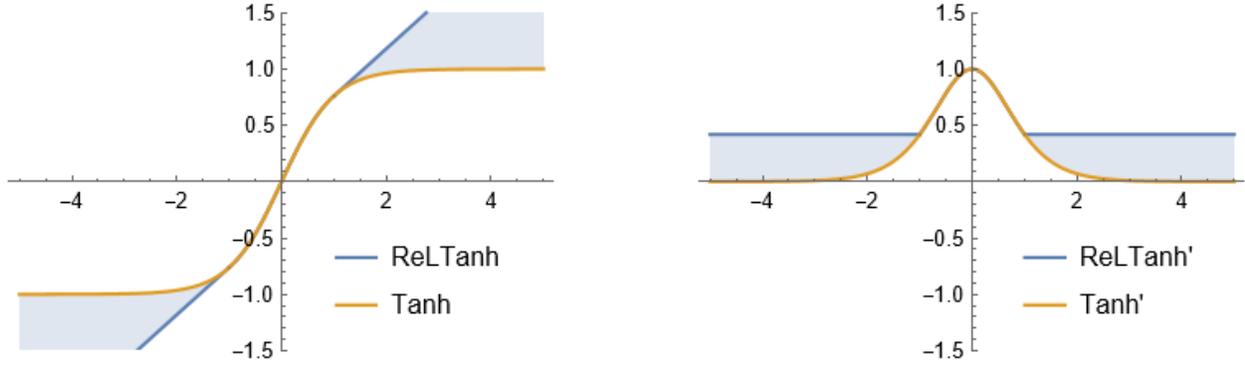

**Fig. 41.** The figure includes two plots comparing the ReLTanh AF and its derivative with the standard Tanh AF and its derivative over a range of $x$ from $-5$ to $5$. Left panel: This plot shows the ReLTanh function, which combines three regions: a right linear region for $x \geq \lambda^+$ (with $\lambda^+ = 1$), a hyperbolic tangent region for $\lambda^- < x < \lambda^+$ (with $\lambda^- = -1$), and a left linear region for $x \leq \lambda^-$. The Tanh function is included for comparison, and the area between the ReLTanh and Tanh functions is filled to highlight their differences. Right panel: This plot presents the derivatives of the ReLTanh and Tanh functions. The ReLTanh derivative consists of constant values in the linear regions $\text{Tanh}'(\lambda^+)$ and $\text{Tanh}'(\lambda^-)$ and the derivative of the hyperbolic tangent $\text{sech}[x]^2$ in the middle region. The derivative of Tanh is shown as $1 - \text{Tanh}[x]^2$, and the area between the ReLTanh and Tanh derivatives is filled to emphasize the differences in their gradient behaviors.

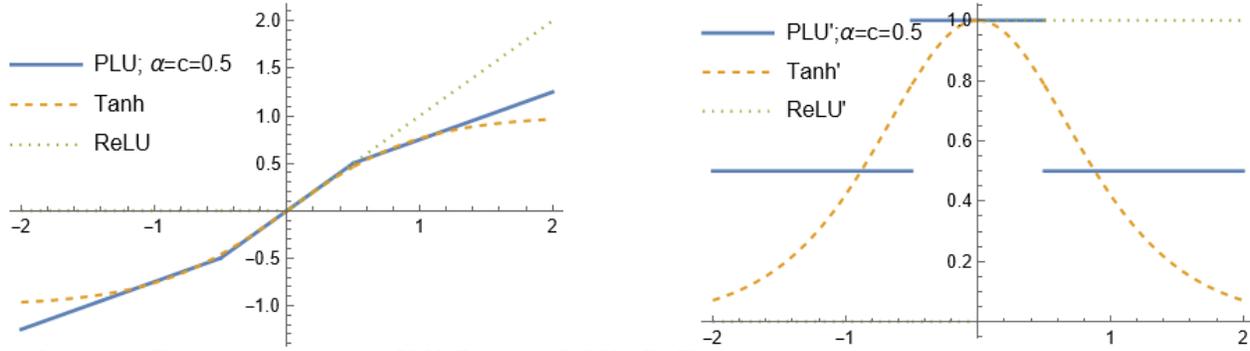

**Fig. 42.** Left panel: The figure illustrates the PLU, Tanh, and ReLU AFs. The $x$-axis represents the input variable $x$ ranging from $-2$ to $2$. This plot allows for a comparison of the behaviors of these AFs across the specified range of $x$ values. Right panel: The figure illustrates the derivatives of the PLU, Tanh, and ReLU AFs. The $x$-axis represents the input variable $x$ ranging from $-2$ to $2$. This plot allows for a detailed comparison of the rates of change of these AFs across the specified range of $x$ values.

The advantages of ReLTanh compared to other common AFs are noteworthy. ReLTanh exhibits superior derivative performance compared to Tanh and mitigates the vanishing gradient problem similarly to the ReLU family. For mean activation, the outputs of ReLTanh are closer to zero, thus it affected less by bias shift than ReLU family. Compared to ReLTanh, ReLU, and Softplus are absolutely non-negative, and LReLU and Swish have negligible negative outputs compared to their positive ones. With lighter bias shift influence, ReLTanh can speed up and smooth training process. The advantage of learnable thresholds helps ReLTanh to approach more closely to the global minimum. As training goes on, ReLTanh can adjust automatically the learnable thresholds in the descending direction of loss. It is worth noting that ReLTanh have a similar waveform to ELU, but ReLTanh can outperform ELU, not only because ReLTanh can update thresholds to help to search the minimize of cost function, but also because ELU still suffers from vanishing gradient problem in the negative interval.

*4.18 Piecewise Linear Unit*

A Piecewise Linear Unit (PLU) [52] is defined as,
$$\sigma_{\text{PLU}}(x) = \max\bigl(\alpha(x + c) - c, \min(\alpha\,(x - c) + c, x)\bigr), \tag{61}$$
having the output range in $[-\infty, +\infty]$, where $\alpha$ and $c$ are constants. Basically, the PLU AF consists of three linear functions in pieces, but continuous. Hence, it avoids the saturation and leads to a good amount of gradient flow through the AF during backpropagation in order to resolve the vanishing gradient problems of ReLU and Tanh. However, the PLU activation is unbounded in both positive and negative directions. Fig. 42 depicts the plot of the PLU AF and its derivative.

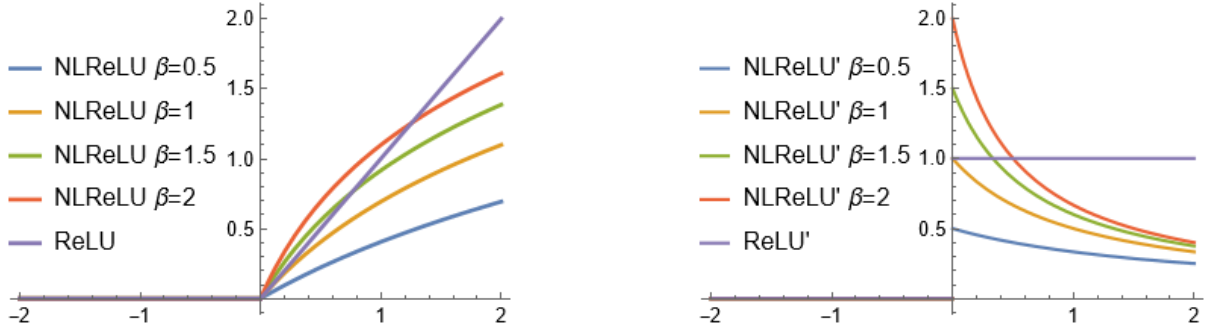

**Fig. 43.** Left panel: The figure illustrates the NLReLU AF for different $\beta$ values and the standard ReLU AF. The $x$-axis represents the input variable $x$ ranging from -2 to 2. Five curves are plotted with $\beta = 0.5, 1, 1.5, 2$ and ReLU. The legend, positioned in the upper left of the plot, identifies each curve by the corresponding $\beta$ value. This plot demonstrates how the NLReLU function varies with different $\beta$ values in comparison to the standard ReLU function. Right panel: The figure illustrates the derivatives of the NLReLU AF for different $\beta$ values and the standard ReLU AF. The $x$-axis represents the input variable $x$ ranging from $-2$ to 2. Five curves are plotted with $\beta = 0.5, 1, 1.5, 2$ and the derivative of ReLU. This plot allows for a detailed comparison of the rates of change of the NLReLU function with different $\beta$ values and the standard ReLU function across the specified range of $x$ values.

*4.19 Natural-Logarithm ReLU*

The Natural Logarithm-Rectified Linear Unit (NLReLU) [53] uses the parametric natural logarithmic transform to improve the $x > 0$ portion of ReLU and is simply defined as

$$\sigma_{\text{NLReLU}}(x) = \ln(\beta \max(0, x) + 1.0), \tag{62.1}$$

$$\frac{\partial}{\partial x} \sigma_{\text{NLReLU}}(x) = \begin{cases} \dfrac{\beta}{\beta x + 1}, & x \geq 0, \\ 0, & \text{otherwise.} \end{cases} \tag{62.2}$$

having the output range in $[0, \infty)$ where $\beta$ is a constant. The NLReLU does not affect the negative regime and thus suffers from vanishing gradient. Fig. 43 shows plots of NLReLU and its first derivative for different values of $\beta$.

The scale of $\beta$ controls the speed at which the response of the AF rises as the input grows. If $\beta < 1$, the function value and derivative of NLReLU are both less than that of ReLU. A larger $\beta$ value means that NLReLU neurons are more sensitive to small values because the activation response rises faster. The mean activation is closer to zero when $\beta$ is smaller. $\beta$ increases the adaptability of NLReLU and allows it to be fine-tuned to different NNs.

A logarithmic transformation is often used in econometrics to convert skewed data before further analysis. Since the raw data from the real world is often heteroscedastic, remedial measures like a logarithmic transformation can decrease this heteroscedasticity. Thus, the data can be analyzed with conventional parametric analysis. Each hidden layer in the ReLU network is easy to bias shift, and the distribution of activated neurons between each layer of the network is prone to be heteroscedastic. Although these problems can be improved by improving the initialization method, ReLU still cannot effectively control the change in the data distribution during training. It is difficult for the 30-layer ReLU network to converge if BN is not used; this arises even if the network is initialized with Xavier. Introducing the logarithmic transformation into the AF can shift the mean activation in each layer closer to 0 while reducing the heteroscedasticity in the data distribution among layers.

Although NLReLU is right-soft saturated, the first derivative of NLReLU is less likely to fall into the saturation regime where the derivative is very close to 0 compared with Sigmoid and Tanh, thereby alleviating the vanishing gradient problem.

The advantages of NLReLU can be summarized as follows. NLReLU can be fine-tuned to different networks, providing adaptability across various NN architectures. It pushes the mean activation of each hidden layer close to 0, and reduces the variance, as well as reduces heteroscedasticity in the data distribution among layers. This reduction in bias shift effect speeds up the learning process. NLReLU minifies most gradients and helps prevent gradients from falling into the saturation regime, therefore it helps solve the "dying ReLU" and vanishing gradient problems to some extent, thus improving convergence. Additionally, NLReLU provides more obvious discrimination of small activation values, thus weakening the effect of some large activation values (e.g., due to noise in the data).

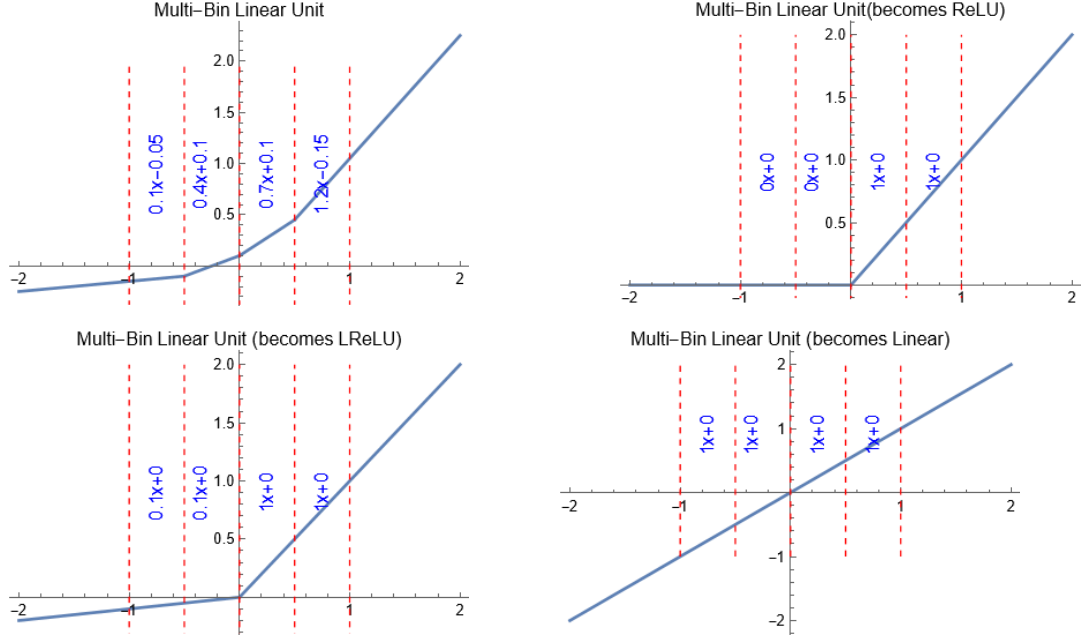

**Fig. 44.** The figure showcases four plots of the MBLU AF, illustrating its versatility in mimicking different types of AFs over a range of $x$ from $-2$ to $2$. Top left panel: The plot presents the MBLU with piecewise linear segments having varying slopes and intercepts. Top right panel: The plot demonstrates how the MBLU can resemble a ReLU function, outputting zero for negative inputs and linearly increasing for positive inputs. Bottom left panel: The plot shows the MBLU configured like a LReLU, introducing a small slope for negative inputs. Bottom right panel: The plot displays the MBLU as a purely linear function across all input values. Each plot includes dashed lines and labels to indicate the specific equations for each segment, highlighting the adaptable nature of the MBLU in NN applications.

*4.20 Multi-bin Trainable Linear Unit*

The nonlinearity of NNs comes from the non-linear AFs, by stacking some simple operations, e.g. ReLU, the networks are able to model any nonlinear function. However, in many real applications, the computational resources are limited and, thus, we are not able to deploy very deep models to fully capture the nonlinearity. The Multi-bin Trainable Linear Unit (MTLU) is used to improve the capacity of AFs for better nonlinearity modeling.

Instead of designing fixed AFs, MTLU was proposed to parameterize the AFs and learn optimal functions for different stage of networks [54]. The MTLU AF simply divides the activation space into multiple equidistant bins and uses different linear functions to generate activations in different bins. The MTLU can be written as,

$$\sigma_{\text{MTLU}}(x) = \begin{cases} a_0 x + b_0, & x \leq c_0, \\ a_k x + b_k, & c_{k-1} < x \leq c_k, \\ \quad \cdots \\ a_K x + b_K, & c_{K-1} < x, \end{cases} \tag{63}$$

having the output range in $(-\infty, \infty)$. The number of bins and the range of bins are the hyperparameters, whereas the linear function of a bin is trainable (i.e., $a_0, ..., a_K$ $b_0, ..., b_K$ are the learnable parameters). Since the anchor points $c_k$ in the model are uniformly assigned, they are defined by the number of bins, $K$, and the bin width. Furthermore, given the input value $x$, a simple dividing and flooring function can be utilized to find its corresponding bin index. Having the bin index, the activation output can be achieved by an extra multiplication and addition function.

Fig. 44 presents 4-bin MTLU AF and, for reference, some other commonly used AFs are also included. One can see that in this simple case, MTLU divides the activation space into 4 parts, $(-\infty, -0.5]$, $(-0.5, 0]$, $(0, 0.5]$ and $(0.5, \infty)$, and adopts different linear functions in different parts to form the non-linear AF. PReLU can be seen as a special case of the proposed parameterization in which the input space is divided into two bins $(-\infty, 0]$ and $(0, \infty)$ and only the parameter $a_0$ is learned, the other parameters $b_0$, $a_1$ and $b_1$ being fixed to 0, 1, and 0, respectively.

You can initialize MTLU as a ReLU function, see Fig. 44. With other initializations, such as random initialization of $\{a_k, b_k\}, k = 0, \dots, K$ and initialization MTLU as a identity mapping function $f(x) = x$, Fig. 44, MTLU is still trainable.

It is important to note the following: (1) Activations are carefully parameterized within the range of $[-1, 1]$, as most inputs to MTLU fall within this interval. (2) After fixing the parameterization range, the choice of bin-width is crucial. The number of bins can be determined as 2 divided by the bin width. (3) There is a trade-off between bin width and the number of bins. A smaller bin width is intuitively expected to enhance the parameterization accuracy of MTLU, thereby improving the nonlinearity modeling capacity of the network. However, this improvement comes at the cost of introducing more parameters (a larger number of bins) to cover the range between $(-1, 1)$.

## 5. ELU Based AFs

### 5.1 Exponential Linear Unit (ELU)

Why ELU AF? Besides producing sparse codes, the main advantage of ReLU is that it alleviates the vanishing gradient problem since the derivative of 1 for positive values is not contractive. However, ReLU is non-negative and, therefore, has a mean activation larger than zero. Units with non-zero mean activations effectively act as biases for the next layer. If these non-zero means are not canceled out or balanced, it can lead to a bias shift effect as learning progresses through the network. The more correlated the units with non-zero mean activations are, the higher the potential for bias shift. Correlated activations can amplify each other's impact, leading to a cumulative effect on the bias shift in subsequent layers. AFs play a crucial role in shaping the distribution of activations. Good AFs aim to push activation means closer to zero. This property helps in reducing bias shift effects and contributes to more stable and efficient learning in NNs.

Centering activations around zero is known to speed up learning in NNs. This is because weight updates during training are influenced by the gradients of the AFs. When activations are centered, gradients are more balanced, and the optimization process is generally more effective. BN is a technique that aims to address issues related to internal covariate shift. It normalizes the activations in a mini-batch, centering them around zero and scaling them to have a certain variance. This normalization can contribute to mitigating bias shift effects and accelerating training. Moreover, Projected Natural Gradient Descent (PRONG) is an optimization algorithm that leverages the natural gradient, which is the gradient of the loss function with respect to the model parameters, adjusted by the Fisher information matrix. The use of the natural gradient helps in more efficient optimization. Additionally, PRONG involves projecting the parameters to ensure that the activations are centered, contributing to faster learning.

Using an AF that naturally pushes the mean activation toward zero is an alternative approach to achieve the benefits of centered activations. AFs that inherently maintain activations around zero can help mitigate bias shift and facilitate more stable learning in NNs. For example: (1) Tanh has been preferred over logistic functions. (2) LReLUs that replace the negative part of the ReLU with a linear function have been shown to be superior to ReLUs. (3) PReLUs generalize LReLUs by learning the slope of the negative part which yielded improved learning behavior on large image benchmark data sets. (4) RReLUs which randomly sample the slope of the negative part raised the performance on image benchmark datasets and convolutional networks.

The ELU has negative values to allow for mean activations close to zero, but which saturates to a negative value with smaller arguments. ELU with $0 < \alpha$ [38] is given as,

$$\sigma_{\text{ELU}}(x) = \begin{cases} x, & x > 0, \\ \alpha(e^x - 1), & x \leq 0. \end{cases} \tag{64.1}$$

$$\frac{\partial}{\partial x}\sigma_{\text{ELU}}(x) = \begin{cases} 1, & x > 0 \\ \alpha e^x, & x \leq 0 \end{cases} = \begin{cases} 1, & x > 0, \\ \sigma_{\text{ELU}}(x) + \alpha, & x \leq 0, \end{cases} \tag{64.2}$$

having the output range in $[-1, \infty)$ where $\alpha$ is a learnable parameter. The ELU hyperparameter $\alpha$ controls the value to which an ELU saturates for negative net inputs (see Fig. 45). If $x$ keeps reducing past zero, eventually, the output of the ELU will be capped at $-1$, as the limit of $e^x$ as $x$ approaches negative infinity is 0. The value for $\alpha$ is chosen to control what we want this cap to be regardless of how low the input gets. This is called the saturation point. At the saturation point, and below, there is very little difference in the output of this function, and hence there's little to no variation (differential) in the information delivered from this node to the other node in the forward propagation.

The formulation of ELU involves an exponential term for negative inputs, which allows the network to capture information from both positive and negative sides of the input space, contributing to its ability to maintain mean activations close to zero. Note that, the $\alpha$ parameter in the ELU equation controls the slope of the function for negative inputs. It's often set to a small positive value like 1.0, but it can be tuned to suit the specific problem. The choice of $\alpha$ can impact the behavior of ELU, and in practice, it's often chosen through experimentation.

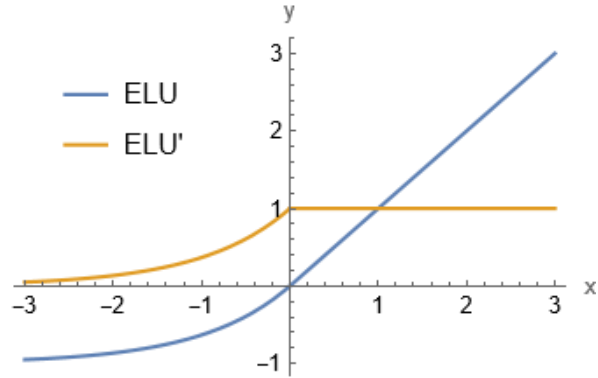

**Fig. 45.** The figure illustrates the ELU AF alongside its derivative, plotted over a range of $x$ from $-3$ to $3$. The ELU function is designed to improve the convergence rate during training and reduce the vanishing gradient problem compared to traditional ReLU. It transitions from an exponential growth for negative values (where $\sigma_{ELU}(x) = \alpha(\exp(x) - 1)$ for $x < 0$) to a linear relationship for non-negative values (where $\sigma_{ELU}(x) = x$ for $x \geq 0$). The derivative of ELU shows how the rate of change adapts across different values of $x$.

Key characteristics and benefits of the ELU AF include its ability to alleviate the vanishing gradient problem, similar to ReLUs, LReLUs, and PReLUs, by maintaining identity for positive values. ELUs offer improved learning characteristics compared to other AFs, as their negative values push mean unit activations closer to zero, akin to batch normalization but with lower computational complexity. This shift towards zero speeds up learning. While LReLUs and PReLUs also have negative values, they do not ensure a noise-robust deactivation state. ELUs, on the other hand, saturate to a negative value with smaller inputs, reducing forward propagated variation and information.

ELUs help mitigate the "dying ReLU" problem, which occurs when ReLU units output zero for certain inputs, causing neurons to become inactive and impeding learning. ELU neurons avoid this issue by producing non-zero outputs for negative inputs. Moreover, ELU can handle outliers better than ReLU, as the exponential term allows it to capture extreme values without saturating (Robust to outliers). Additionally, ELU can lead to faster convergence and potentially better generalization performance compared to some other AFs, especially when dealing with complex datasets.

*5.2 Scaled ELU (SELU)*

Self-Normalizing Neural Networks (SNNs) are a type of NN designed to address the challenges of training DNNs. They were introduced to overcome some of the limitations associated with traditional AFs and help stabilize the training process in DNNs. The key idea behind SNNs is the use of AFs that promote self-normalization of the network's activations. The concept of self-normalization refers to the ability of the network to maintain a stable distribution of activations across layers during training, which can lead to more efficient and faster convergence.

For a NN with AF $f$, let us consider two consecutive layers that are connected by a weight matrix $\mathbf{W}$. Since the input to a NN is a random variable, the activations in the lower layer $\mathbf{x}$, network inputs $\mathbf{z} = \mathbf{W}\mathbf{x}$, and activations in the higher layer $\mathbf{y} = f(\mathbf{z})$ are treated as random variables due to the variability in the data and weights. The mean $\mu$ and variance $\nu$ of activations $x_i$ in the lower layer are denoted by $\mathbb{E}(x_i)$ and $\text{Var}(x_i)$, respectively. Similarly, the mean $\tilde{\mu}$ and variance $\tilde{\nu}$ of activations $y$ in the higher layer are denoted by $\mathbb{E}(y)$ and $\text{Var}(y)$, respectively. Hence, the net input $z$ for a single activation $y = f(z)$ is given by $z = \mathbf{w}^T\mathbf{x}$, $\mathbf{w} \in \mathbb{R}^n$. For $n$ units in the lower layer with activations $x_i$, $1 \leq i \leq n$, the mean of the weight vector $\mathbf{w}$ is defined as $\omega = \sum_{i=1}^{n} w_i$ and the second moment as $\tau = \sum_{i=1}^{n} w_i^2$.

The mapping function $g$ is introduced, which maps the mean and variance of activations from one layer to the mean and variance of activations in the next layer. Formally, $g: (\mu, \nu) \to (\tilde{\mu}, \tilde{\nu})$. Normalization techniques (batch, layer, or weight normalization) ensure that $g$ keeps $(\mu, \nu)$ and $(\tilde{\mu}, \tilde{\nu})$ close to predefined values, typically $(0,1)$.

**Definition (SNN):** A NN is considered SNN if it possesses a mapping $g: \Omega \to \Omega$ for each activation $y$ that maps mean and variance from one layer to the next, and has a stable and attracting fixed point depending on $(\omega, \tau)$ in $\Omega$, where $\Omega = \{(\mu, \nu): \mu \in [\mu_{\min}, \mu_{\max}], \nu \in [\nu_{\min}, \nu_{\max}]\}$. When iteratively applying the mapping $g$, each point within $\Omega$ converges to this fixed point.

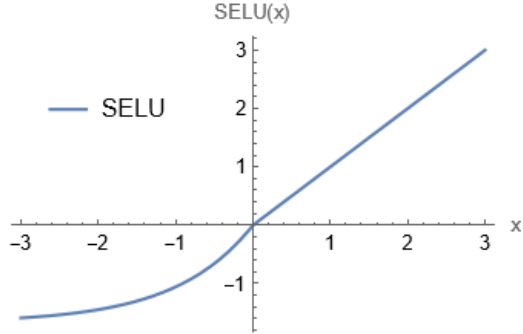

**Fig. 46.** The figure displays the SELU AF, plotted over a range of $x$ from $-3$ to 3. SELU is an adaptation of the ELU function, scaled by the parameters $\lambda$ and $\alpha$ to ensure self-normalizing properties in NNs. It behaves linearly for positive inputs ($x > 0$, where $\sigma_{\text{SELU}}(x) = \lambda x$) and transitions to an exponential growth minus one for negative inputs ($x \leq 0$, where $\sigma_{\text{SELU}}(x) = \lambda\alpha(\exp(x) - 1)$), designed to keep the mean and variance of the outputs close to zero and one, respectively, across layers.

Note that, activations of a NN are considered normalized if both their mean and their variance across samples are within predefined intervals. If the mean and variance of **x** are already within these intervals, then the mean and variance of **y** also remain in these intervals, indicating transitivity across layers. Within these intervals, both the mean and variance converge to a fixed point if the mapping $g$ is applied iteratively. Therefore, SNNs keep normalization of activations when propagating them through layers of the network. This convergence property of SNNs allows training deep networks with many layers, employs strong regularization schemes, and makes learning highly robust.

SNNs cannot be derived with (scaled) ReLUs, Sigmoid units, Tanh units, and LReLUs. Requirements for the AF used in SNNs are: (1) The AF must have both negative and positive values to control the mean of the activations. (2) It should have saturation regions where derivatives approach zero. This helps dampen the variance if it is too large in the lower layer. (3) The AF should have a slope larger than one to increase the variance if it is too small in the lower layer. (4) It should exhibit a continuous curve to ensure the existence of a fixed point, where variance damping is balanced by variance increasing.

The ELU is extended to SELU [55] by using a scaling hyperparameter to make the slope larger than one for positive inputs (see Fig. 46). The SELU AF has specific properties that encourage self-normalization. The SELU can be defined as,

$$\sigma_{\text{SELU}}(x) = \lambda \begin{cases} x, & x > 0, \\ \alpha(e^x - 1), & x \leq 0, \end{cases} \qquad (65)$$

having the output range in $[-\lambda, \infty)$ where $\alpha$ is a hyperparameter. $\lambda$ is a scaling factor, typically set to values close to 1 (e.g., 1.0507).

In (65), the ELU is chosen as the base AF due to its advantageous properties. ELU is a type of ReLU variant that allows negative values, addressing the requirement for both negative and positive values. The ELU is multiplied by a scaling factor $\lambda > 1$, which ensures that the slope of the AF is larger than one for positive net inputs, satisfying the requirement for a slope larger than one. This modification enhances the AF's ability to increase the variance when needed, achieving a balance between damping and increasing variance. This balance promotes stability during the training of DNNs, making ELU a suitable choice.

When properly initialized, SELU neurons transform their inputs in such a way that the activations tend to have a mean of approximately zero and a standard deviation of approximately one during training. This self-normalization property helps in addressing the vanishing and exploding gradient problems, making it easier to train DNNs.

### 5.3 The Parametric ELU (PELU)

The standard ELU is defined as identity for positive arguments and $a(e^x - 1)$ for negative arguments ($x < 0$). Although the parameter $a$ can be any positive value, we need $a = 1$ to have a fully differentiable function. For other values $a \neq 1$, the function is non-differentiable at $x = 0$. Directly learning parameter $a$ would break differentiability at $x = 0$, which could impede back-propagation. For this reason, let us first start by adding two additional parameters to ELU:

$$\sigma_{\text{PELU}}(x) = \begin{cases} c\, x, & x \geq 0, \\ a\left(e^{\frac{x}{b}} - 1\right), & x < 0, \end{cases} \qquad (66)$$

where $a, b, c > 0$. The original ELU can be recovered when $a = b = c = 1$. Each parameter controls different aspects of the activation.

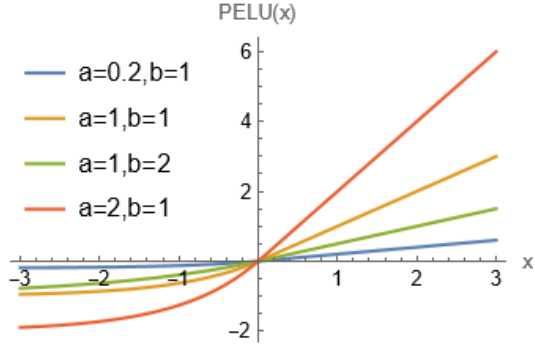

**Fig. 47.** The figure showcases the PELU AF with various parameter settings, plotted over a range of $x$ from $-3$ to $3$. The PELU function generalizes the ELU by introducing parameters $a$ and $b$, which control the output for negative inputs more flexibly. The plot includes four curves, each representing different combinations of the parameters $a$ and $b$: $(a = 0.2, b = 1)$, $(a = 1, b = 1)$, $(a = 1, b = 2)$, and $(a = 2, b = 1)$. These variations illustrate how adjusting $a$ and $b$ affects the slope and curvature of the PELU function for negative $x$, demonstrating its adaptability to different activation characteristics.

Note that, the parameter $c$ changes the slope of the linear function in the positive quadrant (the larger $c$, the steeper the slope). However, the parameter $b$ affects the scale of the exponential decay (the larger $b$, the smaller the decay). While $a$ acts on the saturation point in the negative quadrant (the larger $a$, the lower the saturation point). Constraining the parameters in the positive quadrant forces the activation to be a monotonic function, such that reducing the weight magnitude during training always lowers the neuron contribution.

Using this parameterization, the network can control its non-linear behavior throughout the course of the training phase. It may increase the slope with $c$, the decay with $b$, or lower the saturation point with $a$. However, a standard gradient update on parameters $a, b, c$ would make the function non-differentiable at $x = 0$ and impair backpropagation. By equalling the derivatives on both sides of zero, solving for $c$ gives $c = a/b$ as a solution. The PELU is as follows [56]:

$$\sigma_{\text{PELU}}(x) = \begin{cases} \dfrac{a}{b} x, & x \geq 0, \\ a\left(e^{\frac{x}{b}} - 1\right), & x < 0. \end{cases} \tag{67.1}$$

$$\frac{\partial}{\partial x}\sigma_{\text{PELU}}(x) = \begin{cases} \dfrac{a}{b}, & x \geq 0, \\ \dfrac{a}{b} e^{\frac{x}{b}}, & x < 0. \end{cases} \tag{67.2}$$

$$\frac{\partial}{\partial a}\sigma_{\text{PELU}}(x) = \begin{cases} \dfrac{x}{b}, & x \geq 0, \\ e^{\frac{x}{b}} - 1, & x < 0. \end{cases} \tag{67.3}$$

$$\frac{\partial}{\partial b}\sigma_{\text{PELU}}(x) = \begin{cases} -\dfrac{a}{b^2} x, & x \geq 0, \\ -\dfrac{a}{b^2} e^{\frac{x}{b}}, & x < 0. \end{cases} \tag{67.4}$$

With this parameterization, in addition to changing the saturation point and exponential decay respectively, both $a$ and $b$ adjust the slope of the linear function in the positive part to ensure differentiability at $x = 0$, (see Fig. 47). PELU is trained simultaneously with all the network parameters during backpropagation.

The PELU AF is related to other parametric approaches in the literature. For example, PReLU learns a parameterization of the LReLU activation. PReLU learns a leak parameter in order to find a proper positive slope for negative inputs. This prevents negative neurons from dying. Based on the empirical evidence learning the leak parameter $a$ rather than setting it to a pre-defined value (as done in LReLU) improves performance.

### *5.4 Continuously differentiable ELU (CELU)*

ELU is a useful rectifier for constructing deep learning architectures, as they may speed up and otherwise improve learning by virtue of not have vanishing gradients and by having mean activations near zero. However, the ELU activation as parametrized is not continuously differentiable with respect to its input when the shape parameter $\alpha$ is not equal to 1. The PELU is explored in CELU [57] for the negative inputs. The CELU is given as,

$$\sigma_{\text{CELU}}(x) = \begin{cases} x, & x \geq 0, \\ \alpha\left(e^{\frac{x}{\alpha}} - 1\right), & x < 0. \end{cases} \tag{68.1}$$

$$\frac{\partial}{\partial x}\sigma_{\text{CELU}}(x) = \begin{cases} 1, & x \geq 0, \\ e^{\frac{x}{\alpha}}, & x < 0. \end{cases} \tag{68.2}$$

$$\frac{\partial}{\partial \alpha}\sigma_{\text{CELU}}(x) = \begin{cases} 0, & x \geq 0, \\ e^{\frac{x}{\alpha}}\left(1 - \frac{x}{\alpha}\right) - 1, & x < 0, \end{cases} \tag{68.3}$$

having the output range in $[-\alpha, \infty)$ where $\alpha$ is a learnable parameter (see Fig. 48). The CELU also converges to ReLU as $\alpha$ approaches 0 from the right and converges to a linear activation as $\alpha$ approaches $\infty$:

$$\lim_{\alpha \to 0^+} \sigma_{\text{CELU}}(x, \alpha) = \max(0, x), \tag{69.1}$$

$$\lim_{\alpha \to \infty} \sigma_{\text{CELU}}(x, \alpha) = x. \tag{69.2}$$

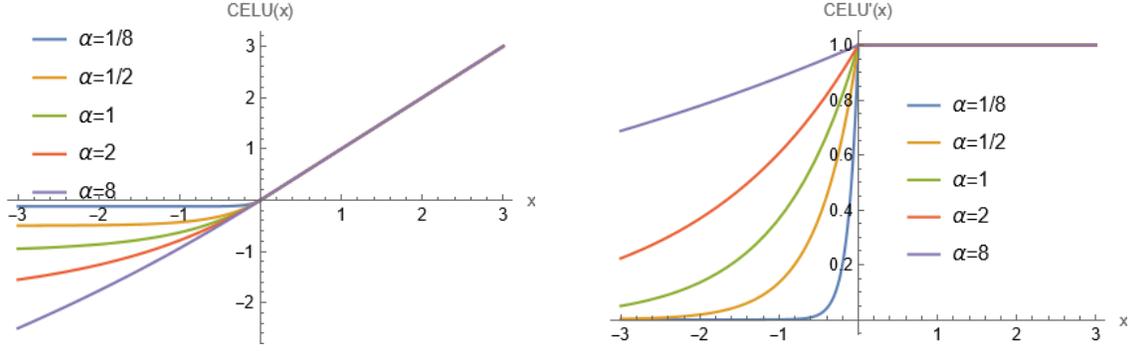

**Fig. 48.** Left panel: The figure illustrates the CELU AF for different values of $\alpha$. The $x$-axis represents the input variable $x$ ranging from $-3$ to $3$, while the $y$-axis represents the function values. Five curves are plotted for different $\alpha$ values: ($\alpha = 1/8, 1/2, 1, 2, 8$). This plot demonstrates how the CELU function behaves with different $\alpha$ values, showcasing its flexibility and response characteristics for varying $\alpha$. Right panel: The figure illustrates the derivatives of the CELU AF for different values of $\alpha$. The $x$-axis represents the input variable $x$ ranging from $-3$ to $3$, while the $y$-axis represents the derivative values. Five curves are plotted for different $\alpha$ values: ($\alpha = 1/8, 1/2, 1, 2, 8$).

This alternative parametrization has several other useful properties that the original parametrization of ELU does not: (1) its derivative with respect to $x$ is bounded, (2) it contains both the linear transfer function and ReLU as special cases, and (3) it is scale-similar with respect to $\alpha$.

*5.5 Multiple PELU*

The rectified and exponential linear units are commonly adopted by the recent deep learning architectures to achieve good performance. However, there exists a gap in representation space between the two types of AFs. For the negative part, ReLU or PReLU are able to represent the linear function family but not the non-linear one, while ELU is able to represent the non-linear function family but not the linear one. The representation gap to some extent undermines the representational power of those architectures using a particular AF.

The PELU is extended to Multiple PELU (MPELU) [58] by using two learnable parameters to represent MPELU as either rectified, exponential or combined. The MPELU can be expressed as,

$$\sigma_{\text{MPELU}}(x) = \begin{cases} x, & x > 0, \\ \alpha(e^{\beta x} - 1), & x \leq 0. \end{cases} \tag{70.1}$$

$$\frac{\partial}{\partial x}\sigma_{\text{MPELU}}(x) = \begin{cases} 1, & x > 0, \\ \alpha\beta e^{\beta x}, & x \leq 0. \end{cases} \tag{70.2}$$

$$\frac{\partial}{\partial \alpha}\sigma_{\text{MPELU}}(x) = \begin{cases} 0, & x > 0, \\ e^{\beta x} - 1, & x \leq 0. \end{cases} \tag{70.3}$$

$$\frac{\partial}{\partial \beta}\sigma_{\text{MPELU}}(x) = \begin{cases} 0, & x > 0, \\ x\alpha e^{\beta x}, & x \leq 0, \end{cases} \tag{70.4}$$

having the output range in $[-\alpha, \infty)$, where $\alpha$ and $\beta$ are the trainable parameters. MPELU is unifying the existing ReLU, LReLU, PReLU, and ELU.

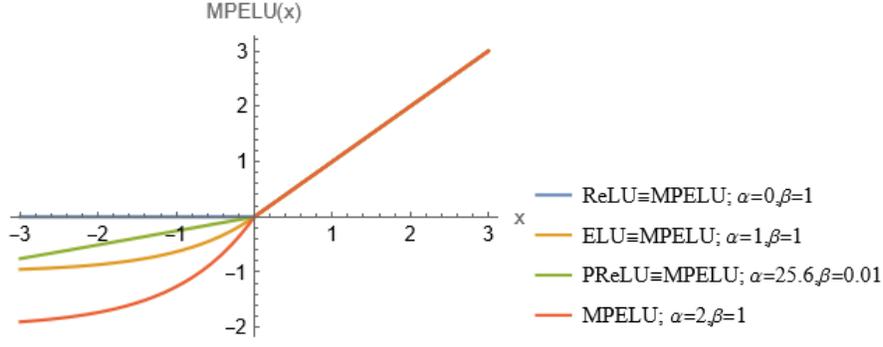

**Fig. 49.** The plot illustrates the MPELU AF for different values of $\alpha$ and $\beta$. The $x$-axis represents the input variable $x$ ranging from -3 to 3, while the $y$-axis represents the function values. Four curves are plotted for different combinations of $\alpha$ and $\beta$ values: ($\alpha = 0$, $\beta = 1$), which approximates the ReLU function. ($\alpha = 1$, $\beta = 1$), which approximates the ELU function. ($\alpha = 25.6$, $\beta = 0.01$), which approximates the PReLU function. ($\alpha = 2$, $\beta = 1$), representing the MPELU function. This plot demonstrates how the MPELU function behaves with different parameter values, highlighting its flexibility and ability to approximate other popular AFs.

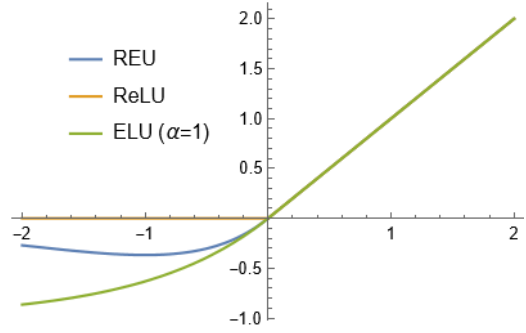

**Fig. 50.** The figure compares three AFs: REU, ReLU, and ELU with $\alpha = 1$, plotted over a range of $x$ from $-2$ to 2. The REU function is defined as $x$ for $x > 0$ and $xe^x$ for $x \leq 0$, blending linear and exponential behaviors. The ReLU function outputs zero for negative inputs and linearly increases for positive inputs. The ELU function, with $\alpha = 1$, exponentially decays for negative inputs and linearly increases for positive inputs.

By adjusting $\beta$, MPELU can switch between the rectified and exponential linear units. To be specific, if $\beta$ is set to a small number, for example, 0.01, the negative part of MPELU approximates to a linear function. In this case, MPELU becomes the PReLU. On the other side, if $\beta$ takes a large value, for example, 1.0, the negative part of MPELU is a non-linear function, making MPELU turn back into the exponential linear units.

Introducing $\alpha$ helps further control the form of MPELU, as shown in Fig. 49. If $\alpha$ and $\beta$ are set to 1, MPELU reduces to ELU. Decreasing $\beta$ in this case lets MPELU go to LReLU. Finally, MPELU is exactly equivalent to ReLU when $\alpha = 0$. From the above analysis, it is easy to see that the flexible form of MPELU makes it cover the solution space of its special cases, and therefore grants it more powerful representation.

*5.6 Rectified Exponential and Parametric Rectified Exponential Units*

The design choices in Rectified Exponential Unit (REU) [59] are motivated by the desire to combine the positive aspects of identity mapping for positive values with a non-monotonic behavior in the negative values. The definition of REU is given by (see Fig. 50):

$$\sigma_{\text{REU}}(x) = \begin{cases} x, & x > 0, \\ xe^x, & x \leq 0. \end{cases} \tag{71}$$

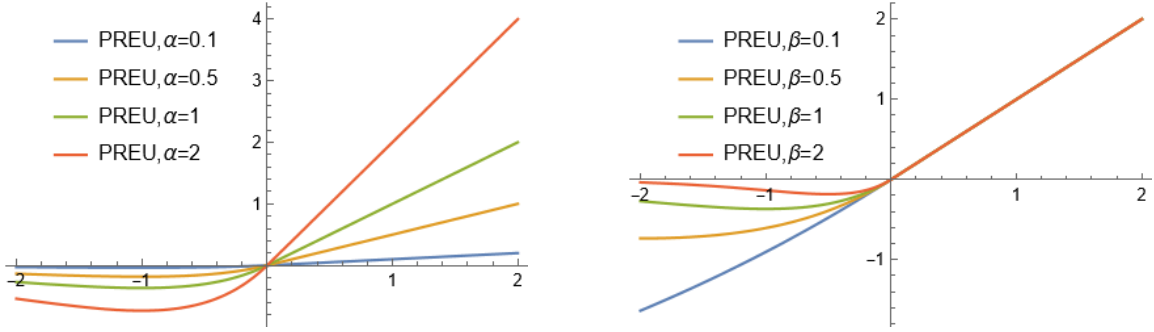

**Fig. 51.** The figure contains two plots illustrating the behavior of the PREU AF over a range of $x$ from $-2$ to $2$, highlighting the effects of varying parameters $\alpha$ and $\beta$. Left panel: This plot shows the PREU function for a fixed $\beta = 1$ and different values of $\alpha$ (0.1, 0.5, 1, 2). The PREU function is defined as $\alpha x$ for $x > 0$ and $\alpha x e^{\beta x}$ for $x \leq 0$. The plot demonstrates how increasing $\alpha$ affects the slope of the function for positive inputs and the curvature for negative inputs. Right panel: This plot presents the PREU function for a fixed $\alpha = 1$ and varying values of $\beta$ (0.1, 0.5, 1, 2). Here, the PREU function's behavior for positive inputs remains linear, while the curvature for negative inputs changes with different $\beta$ values. The plot shows how increasing $\beta$ intensifies the exponential growth for negative inputs.

Similar to other popular AFs like ReLU, LReLU, PReLU, and ELU, REU maintains an identity mapping for positive input values ($x > 0$). This means that if the input is positive, the output is the same as the input, allowing the positive information to pass through unchanged. The introduction of a non-monotonic property in the negative part of the function is highlighted as a key feature. Non-monotonic functions have points where the slope changes direction, providing a more complex behavior compared to monotonic functions. REU employs an exponential function $e^x$ in the negative part $x \leq 0$. The rationale behind this choice is to retain more information in the negative part of the input. Exponential functions grow rapidly, and this characteristic may capture and amplify certain patterns or features in the negative range.

The Parametric Rectified Exponential Unit (PREU) [59] is a flexible AF with adjustable parameters $\alpha$ and $\beta$, providing control over the slope in the positive quadrant and the saturation in the negative quadrant. A PREU is designed as,

$$\sigma_{\text{PREU}}(x) = \begin{cases} \alpha x, & x > 0, \\ \alpha x e^{\beta x}, & x \leq 0, \end{cases} \quad (72.1)$$

having the output range in $[-1, \infty)$. Here, $\alpha$ and $\beta$ can be fixed constants or trainable parameters. Each parameter controls a different aspect of PREU, as shown in Fig. 51: $\alpha$ mainly controls the slope in the positive quadrant, influencing how quickly the function increases for positive values of $x$. However, $\beta$ controls the saturation in the negative quadrant, determining how quickly the function approaches zero for negative values of $x$.

The derivatives of PREU with respect to $x$, $\alpha$, and $\beta$ are given by

$$\frac{\partial}{\partial x} \sigma_{\text{PREU}}(x) = \begin{cases} \alpha, & x > 0, \\ \alpha(1 + \beta x)e^{\beta x}, & x \leq 0. \end{cases} \quad (72.2)$$

$$\frac{\partial}{\partial \alpha} \sigma_{\text{PREU}}(x) = \begin{cases} x, & x > 0, \\ x e^{\beta x}, & x \leq 0. \end{cases} \quad (72.3)$$

$$\frac{\partial}{\partial \beta} \sigma_{\text{PREU}}(x) = \begin{cases} 0, & x > 0, \\ \alpha x^2 e^{\beta x}, & x \leq 0. \end{cases} \quad (72.4)$$

*5.7 Fast ELU (FELU)*

The fast exponential calculation method, based on the floating-point representation, is a technique used to approximate the exponential function $e^x$. This approximation is often employed in NN implementations to speed up the computation, especially since the standard mathematical library's exp function can be computationally expensive.

The exponential function is the most typical nonlinear function in NN. The typical exponential function $e^x$ is calculated by using the power series expansion, and the approximate calculation of summing the first $n$ terms. It is defined as

$$e^x = 1 + x + \frac{x^2}{2!} + \frac{x^3}{3!} + \frac{x^4}{4!} \ldots + \frac{x^n}{n!}. \quad (73)$$

The standard math library of exp function has a high precision. The $e$ is an infinite loop number, which value is 2.718281828459......, and the exp function takes a long time to calculate, so we need to find a fast exponential function to speed up the calculation. Schraudolph [60] proposed a fast exponential calculation method based on floating-point number, and the mathematical formula of exp macro operation is

$$e^{x+\gamma} = 2^k \left(1 + \frac{x}{\ln 2} - k\right), \qquad k = \left\lfloor \frac{x}{\ln 2} \right\rfloor, \tag{74}$$

where $\gamma = \frac{c \ln 2}{2^{20}}$, $\lfloor u \rfloor$ represents the largest integer less than or equal to $u$, $c$ is an optimal constant obtained after multiple calculations. Using floating point calculation [61], the $e^x$ can be reduced to a power-based operation form of base 2 as

$$e^x = 2^{x/\ln(2)} \tag{75}$$

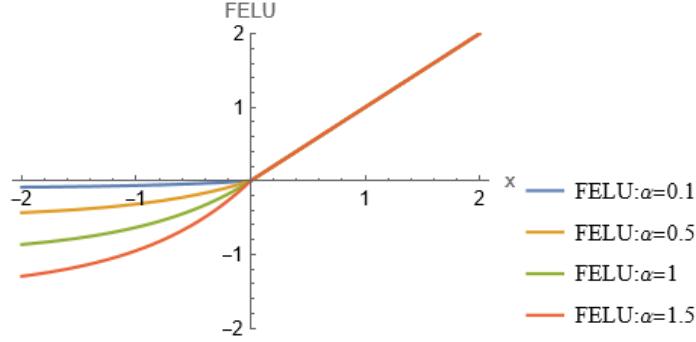

**Fig. 52.** The plot illustrates the FELU AF for different values of $\alpha$. The $x$-axis represents the input variable $x$ ranging from $-2$ to $2$, while the $y$-axis represents the function values ranging from $-2$ to $2$. Four curves are plotted for different $\alpha$ values: 0.1, 0.5, 1, and 1.5. This plot demonstrates how the FELU function's behavior changes with different $\alpha$ values, showcasing its flexibility and response characteristics for varying $\alpha$. The FELU function provides a smooth transition between the exponential and linear regions, depending on the chosen $\alpha$ value.

The expansion of $2^{x/\ln(2)}$ is the same as the expansion of $e^x$, however, due to the use of the idea of floating-point calculations, the new function not only can accelerate the exponential calculation, but also achieve the purpose of quickly approximating $e^x$. Inspired by this idea, it can be considered that $2^{x/\ln(2)}$ is an exponential function that can achieve the fast exponential calculation. Therefore, the value of $e^x$ of the AF ELU on the negative part is optimized to $2^{x/\ln(2)}$, generating a new AF, FELU [61]. The FELU is defined as,

$$\sigma_{\text{FELU}}(x) = \begin{cases} x, & x > 0, \\ \alpha(2^{x/\ln(2)} - 1), & x \le 0. \end{cases} \tag{76.1}$$

$$\frac{\partial}{\partial x}\sigma_{\text{FELU}}(x) = \begin{cases} 1, & x > 0, \\ \alpha 2^{x/\ln(2)}, & x \le 0. \end{cases} \tag{76.2}$$

having the output range in $[-\alpha, \infty)$ with $\alpha$ as a learnable parameter. It can be seen from Fig. 52 that the curve of this function is smooth respect to the linear function, and it has the advantage of being differentiable everywhere and output zero mean, which can effectively avoid the occurrence of output offset. So, you can find the direction in which the gradient drops faster. At the same time, the parameter $\alpha$ is a variable parameter learned through training, which is used to control the soft saturation region of the negative part to determine the activation degree of the feature in the negative region. Fig. 52 shows the curve of the FELU corresponding to the case that the value of $\alpha$ is 0.1, 0.5, 1, 1.5 in the negative part.

### 5.8 Elastic ELU (EELU)

Since the ReLU was introduced, many modifications have been proposed to avoid overfitting. ELU and their variants, with trainable parameters, have been proposed to reduce the bias shift effects which are often observed in ReLU-type AFs. The EELU [62] combines the advantages of both types of AFs in a generalized form. EELU changes the positive slope to prevent overfitting, as do EReLU and RReLU and also preserves the negative signal to reduce the bias shift effect, as does ELU. However, the positive slope of EELU is modified from the Gaussian distribution with a randomized standard deviation, instead using a simple uniform distribution to determines the scale of random noise, like EReLU and RReLU. The EELU is defined as,

$$\sigma_{\text{EELU}}(x) = \begin{cases} k\, x, & x > 0, \\ \alpha(e^{\beta x} - 1), & x \le 0, \end{cases} \tag{77}$$

having the output range in $[-\alpha, \infty)$ where $\alpha$ and $\beta$ are the trainable parameters.

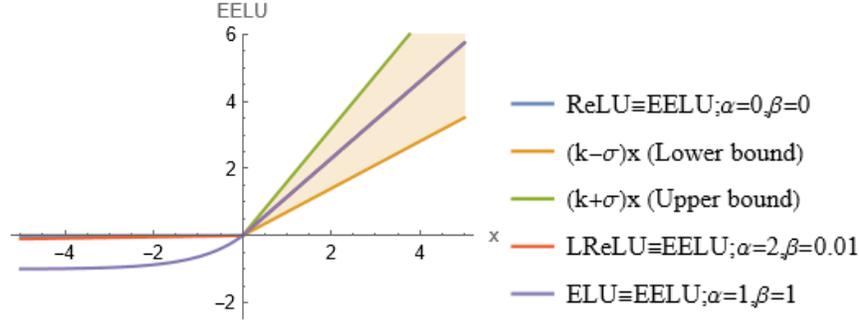

**Fig. 53.** The figure depicts the EELU AF, illustrating its versatility through five different curves across an $x$ range of $-5$ to $5$. The plot demonstrates the function's linear behavior similar to ReLU for positive $x$ values, modifiable through parameters $\alpha$, $\beta$, and $k$, and includes bounds to show the variability due to the sampled $k$ value. Additionally, the figure showcases variations that mimic LReLU and ELU for negative $x$ values, thereby highlighting the function's adaptability to different neuron behaviors in a NN. The filled area between the lower and upper bounds emphasizes the potential variability in the linear portion.

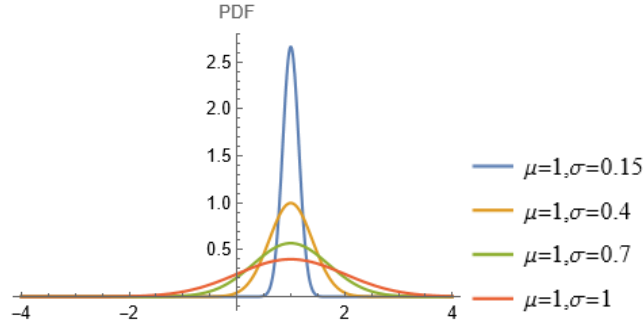

**Fig. 54.** The figure displays the probability density functions (PDFs) of four normal distributions, each with a mean ($\mu$) of 1 but different standard deviations ($\sigma$) of 0.15, 0.4, 0.7, and 1, plotted over a range of $x$ from $-4$ to $4$. As $\sigma$ increases, the curves become wider and flatter, indicating greater variability around the mean.

The coefficient $k$ plays an important role in EELU. In the training stage, the coefficient is sampled from a Gaussian distribution with a random standard deviation and a fixed mean; the sampled coefficient is truncated from 0 to 2, because the Gaussian distribution ranges from $-\infty$ to $+\infty$. The standard deviation $\sigma$ of the Gaussian distribution is randomly chosen from a uniform distribution, instead of using a fixed parameter. It is denoted by:

$$k = \max(0, \min(s, 2)), \quad s \sim N(1, \sigma), \tag{78.1}$$
$$\sigma \sim U(0, \epsilon), \quad \epsilon \in (0, 1]. \tag{78.2}$$

Here, $\epsilon$ is a hyperparameter. In the test stage, EELU replaces $k$ with $\mathbb{E}(k)$ (the expectation of $k$) in the positive part. If $\mathbb{E}(k)$ is equal to 1, then EELU becomes MPELU, as shown in Fig. 53.

Fig. 53 shows how EELU works in the training stage and the test stage. EELU is divided into two parts, one for positive and one for negative inputs. The slope of the positive part is varied using a Gaussian distribution with a randomized standard deviation. This approach is similar to inserting noise, because it outputs similar input features to various output features. The negative input values are determined by $\alpha$ and $\beta$. $\alpha$ and $\beta$ are the learning parameters determined from the training samples and are constrained to be greater than zero. The advantage of EELU is that it can represent various output features with random noise. The random noise is independent from the inputs and can confer sensitivity of the NNs to a wide variety of inputs, in a way which is similar to data augmentation. Moreover, EELU can represent various AFs such as ReLU, LReLU, PReLU, ELU, and MPELU.

Gaussian distributions with different standard deviations are shown in Fig. 54. The probabilities of a negative $k$ with different standard deviations are shown in Table 1. The larger the standard deviation, the greater the probability that the sampled parameter is less than zero. If the sampled value is negative, input signals are discarded, and the hidden unit is deactivated. The neuronal noise, which is simulated by various scales and random deactivation, works to regularizing DNNs to learn various latent representations from training samples.

Table 1: Probability of a value of less than zero from the Gaussian distribution

| Standard Deviation | Probability |
|---|---|
| 0.1 | 0.00% |
| 0.2 | 0.00% |
| 0.3 | 0.04% |
| 0.4 | 0.62% |
| 0.5 | 2.28% |
| 0.6 | 4.78% |
| 0.7 | 7.66% |
| 0.8 | 10.56% |
| 0.9 | 13.33% |
| 1.0 | 15.87% |

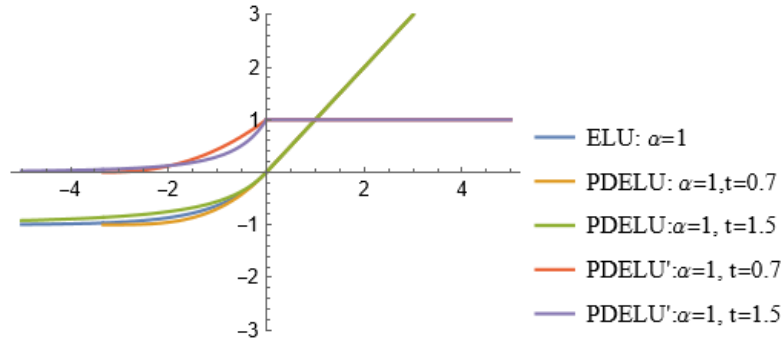

**Fig. 55.** The plot illustrates the ELU and PDELU AFs for different values of $t$, along with their derivatives. The $x$-axis represents the input variable $x$ ranging from -5 to 5. Five curves are plotted: $\text{ELU}(x, \alpha = 1)$, $\text{PDELU}(x, \alpha = 1, t = 0.7)$, $\text{PDELU}(x, \alpha = 1, t = 1.5)$, the derivative of $\text{PDELU}(x, \alpha = 1, t = 0.7)$, and the derivative of $\text{PDELU}(x, \alpha = 1, t = 1.5)$. This plot demonstrates how the PDELU function and its derivative change with different $t$ values, in comparison to the standard ELU function. The derivatives highlight the smooth transitions and the varying slopes introduced by different $t$ values in the PDELU function.

*5.9 Parametric Deformable ELU*

To reduce the undesired bias shift effect without the natural gradient, either the (i) activation of incoming units can be centered at zero or (ii) AFs with negative values can be used. Based on the basic theory of zero-mean activation, ELU uses the AFs with negative values and prunes the negative part to exponential output. However, it fails to push the mean value of the activation units' responses to zero. The statistical results show that the majority of mean of activation is 0.8 when using ELU as the AF. In order to improve the performance of pushing the mean of activation to zero, Parametric Deformable ELU (PDELU) AF [63] was proposed based on ELU.

A PDELU AF tries to shift the mean value of output closer to zero using the flexible map shape. The PDELU is defined as,

$$\sigma_{\text{PDELU}}(x) = \begin{cases} x, & x > 0, \\ \alpha \left( [1 + (1-t)x]^{\frac{1}{1-t}} - 1 \right), & x \leq 0, \end{cases} \quad (79)$$

having the output range in $[-\alpha, \infty)$ where $\alpha$ is a learnable parameter, $t \neq 1$. As shown in Fig. 55, the blue curve is ELU, and the other lines are PDELU with different $t$.

The PDELU has several notable properties. Here, $x$ is the input of the nonlinear activation $\sigma_{\text{PDELU}}$ on the $i$th channel, and $\alpha$ is a coefficient controlling the slope of the negative part. The $\alpha$ indicates that the nonlinear activation is allowed to vary on different channels. The parameter $\alpha$ is jointly learned by the loss function. Different channels have different $\alpha$, which makes the NN more flexible to learn the optimized weights. The PDELU has clear geometric flexibility. $t$ controls the degree of deformation. With bigger $t$, the response of the negative semi-axis becomes much larger and flatter. With, $t \to 1$, the PDELU becomes identical to the original ELU. Furthermore, flexibility makes the NNs more robust to noise and the change of different input. The statistical results show that the majority of mean of activation of PDELU is 0.45 which is much better than ELU (0.80 for ELU). An adaptive AF can effectively avoid overfitting and take full advantage of the strong learning ability from deep and wide architectures.

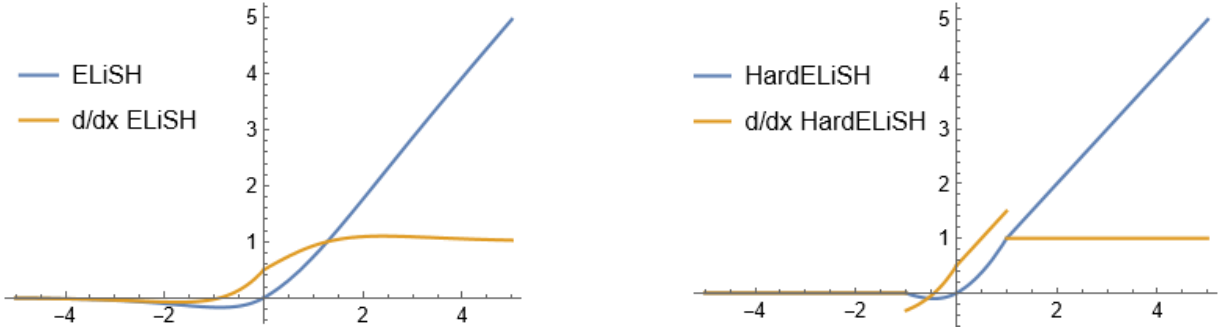

**Fig. 56.** Left panel: The plot illustrates the ELiSH AF and its derivative. The $x$-axis represents the input variable $x$ ranging from $-5$ to $5$. This plot demonstrates the behavior of the ELiSH function and its rate of change across the specified range of $x$ values, highlighting its smooth transitions and non-linear characteristics. Right panel: The figure illustrates the HardELiSH AF and its derivative. The $x$-axis represents the input variable $x$ ranging from $-5$ to $5$. This plot demonstrates the behavior of the HardELiSH function and its rate of change across the specified range of $x$ values, highlighting its piecewise linear characteristics and non-linear transitions.

### 5.10 Exponential Linear Sigmoid SquasHing

An Exponential Linear Sigmoid SquasHing (ELiSH) is defined in [64] as,

$$\sigma_{\text{ELiSH}}(x) = \begin{cases} \dfrac{x}{(1+e^{-x})}, & x \geq 0, \\ \dfrac{(e^x - 1)}{(1+e^{-x})}, & x < 0. \end{cases} \quad (80)$$

Its negative part is a multiplication of ELU and Sigmoid while the positive part is a multiplication of $x$ and Sigmoid. Moreover, it is also extended to HardELiSH which is a multiplication of HardSigmoid and Linear in the positive part and HardSigmoid and ELU in the negative part. Here, HardSigmoid is defined as,

$$\sigma_{\text{HardSigmoid}}(x) = \max\left(0, \min\left(1, \dfrac{x+1}{2}\right)\right), \quad (81)$$

and HardELiSH is defined as,

$$\sigma_{\text{HardELish}}(x) = \begin{cases} x\, \sigma_{\text{HardSigmoid}}(x), & x \geq 0, \\ (e^x - 1)\, \sigma_{\text{HardSigmoid}}(x), & x < 0. \end{cases} \quad (82)$$

Both ELiSH and HardELiSH AFs and their derivatives are shown in Fig. 56.

## 6. Miscellaneous AFs

### 6.1 Swish

The introduction of Swish was part of a broader effort to explore and discover new AFs. The authors of the paper "Searching for AFs" [65] used a neural architecture search approach to automatically discover AFs that performed well on certain tasks. The paper contributed to the ongoing research in the deep learning community by highlighting the potential benefits of automatic methods for discovering NN architectures and components. The introduction of Swish and similar approaches demonstrated that there might be room for improvement beyond traditional AFs, and automatic search methods could be valuable in this exploration.

Swish [65] is defined as

$$\sigma_{\text{Swish}}(x) = x\, \sigma_{\text{sigmoid}}(\beta x), \quad (83)$$

where $\sigma_{\text{sigmoid}}(x) = (1 + \exp(-x))^{-1}$ is the Sigmoid function and $\beta$ is either a constant or a trainable parameter. The output range of Swish is $(-\infty, \infty)$. Swish is a smooth and differentiable AF. Smoothness and differentiability are desirable properties in NN AFs because they facilitate gradient-based optimization during the training process. Fig. 57 plots the graph of Swish for different values of $\beta$. When $\beta = 1$, Swish is equivalent to the SiLU (15). If $\beta = 0$, Swish becomes the scaled linear function $f(x) = \dfrac{x}{2}$. As $\beta \to \infty$, the Sigmoid component approaches a $0 - 1$ function, so Swish becomes like the ReLU function.

This suggests that Swish can be loosely viewed as a smooth function which nonlinearly interpolates between the linear function and the ReLU function. The degree of interpolation can be controlled by the model if $\beta$ is set as a trainable parameter.

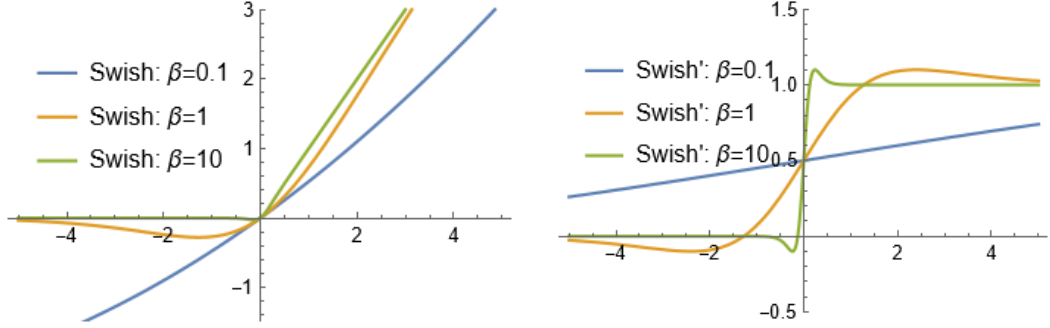

**Fig. 57.** Left panel: This plot visualizes the Swish AF, a smooth, non-monotonic function, for three different values of the parameter $\beta$ (0.1, 1, and 10) over a range of $x$ from $-5$ to 5. The function is defined as $x\,\sigma_{\text{sigmoid}}(\beta x)$. As $\beta$ increases, the Swish function transitions from a nearly linear behavior ($\beta = 0.1$) to a more pronounced ReLU-like curve ($\beta = 10$), illustrating its adaptability and the impact of $\beta$ on the activation output. Right panel: The plot shows the first derivatives of the Swish function for the same $\beta$ values (0.1, 1, and 10). Each derivative curve illustrates the rate of change of the Swish function at different points, highlighting how the responsiveness of the function varies with changes in $\beta$. For lower $\beta$, the derivative remains close to constant, while for higher $\beta$, the curve exhibits sharper changes, especially near zero.

Like ReLU, Swish is unbounded above and bounded below. Unlike ReLU, Swish is smooth and non-monotonic. In fact, the non-monotonicity property of Swish distinguishes it from most common AFs. The derivative of Swish is

$$\frac{\partial}{\partial x}\sigma_{\text{Swish}}(x) = \sigma_{\text{Sigmoid}}(\beta x) + x\,\beta\,\sigma_{\text{Sigmoid}}(\beta x)\left(1 - \sigma_{\text{Sigmoid}}(\beta x)\right)$$
$$= \beta\,\sigma_{\text{Swish}}(x) + \sigma_{\text{Sigmoid}}(\beta x)(1 - \sigma_{\text{Swish}}(x)). \tag{84}$$

The first derivative of Swish is shown in Fig. 57 for different values of $\beta$. The scale of $\beta$ controls how fast the first derivative asymptotes to 0 and 1.

When $\beta = 1$, the derivative has magnitude less than 1 for inputs that are less than around 1.25. Thus, the success of Swish with $\beta = 1$ implies that the gradient preserving property of ReLU (i.e., having a derivative of 1 when $x > 0$) may no longer be a distinct advantage in modern architectures.

The most striking difference between Swish and ReLU is the non-monotonic "bump" of Swish when $x < 0$. The shape of the bump can be controlled by changing the $\beta$ parameter. The smaller and higher values of $\beta$ lead towards the linear and ReLU functions, respectively. Thus, it can control the amount of non-linearity based on the dataset and network complexity.

While Swish showed promise in terms of performance improvements, it is important to consider the computational cost. Swish involves the computation of the Sigmoid function, which might be more computationally expensive compared to simpler AFs like ReLU.

*6.2 E-Swish*

Swish is also extended to E-Swish by multiplying the Swish with a learnable parameter to control the slope in the positive direction [66]. The E-Swish is defined as,

$$\sigma_{\text{E-Swish}}(x) = \beta\,x\,\sigma_{\text{Sigmoid}}(x), \tag{85.1}$$

having the output the range in $(-\infty, \infty)$ and $\beta$ is trainable parameter and $1 \leq \beta \leq 2$. The properties of E-swish are very similar to the ones of Swish, see Fig. 58. In fact, when $\beta = 1$, E-swish reverts to Swish. Like both ReLU and Swish, E-swish is unbounded above and bounded below. Like Swish, it is also smooth and non-monotonic. The property of non-monotonicity is almost exclusive of Swish and E-swish. Another exclusive feature of both, Swish and E-swish, is that there is a region where the derivative is greater than 1.

The derivative of E-swish is:

$$\frac{\partial}{\partial x}\sigma_{\text{E-Swish}}(x) = \beta\sigma_{\text{Sigmoid}}(x) + x\,\beta\,\sigma_{\text{Sigmoid}}(x)\left(1 - \sigma_{\text{Sigmoid}}(x)\right)$$
$$= \sigma_{\text{E-Swish}}(x) + \sigma_{\text{Sigmoid}}(x)(\beta - \sigma_{\text{E-Swish}}(x)). \tag{85.2}$$

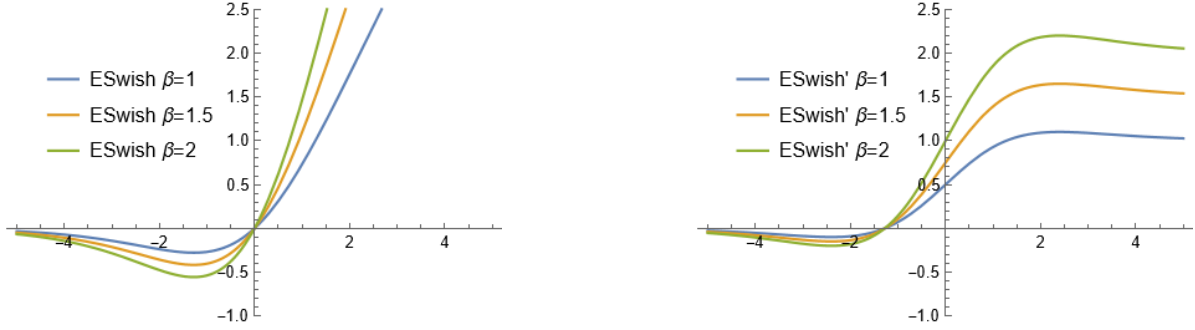

**Fig. 58.** Left panel: This plot displays the E-Swish AF, which is a variant of the Swish function enhanced by a scaling parameter $\beta$, plotted for three values of $\beta$ (1, 1.5, and 2) over a range of $x$ from $-5$ to 5. The E-Swish function is defined as $\beta\, x\, \sigma_{\text{Sigmoid}}(x)$. As $\beta$ increases, the amplitude of the E-Swish function also increases, showing a more pronounced curve that enhances the activation's capability to manage different signal strengths in NNs. Right panel: This plot illustrates the first derivatives of the E-Swish function for the same set of $\beta$ values. These derivatives highlight how the rate of change of the E-Swish function varies with $\beta$, indicating the function's sensitivity at different points. For higher $\beta$, the derivative shows more pronounced peaks, suggesting a more responsive behavior at certain ranges of $x$.

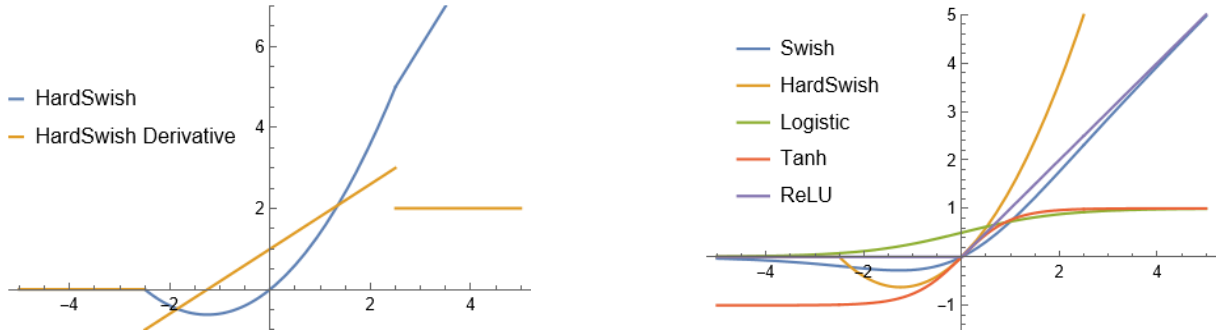

**Fig. 59.** Left panel: The figure displays the HardSwish AF and its derivative, plotted over a range of $x$ from $-5$ to 5. HardSwish is a piecewise linear approximation of the Swish function, designed to provide computational benefits while retaining non-linear characteristics beneficial for deep learning models. The AF (HardSwish) shows a smoother transition compared to traditional ReLU, incorporating a non-linear region around zero, and it smoothly transitions to linear regions for positive and negative values. Its derivative, plotted alongside, highlights how the gradient varies across different values of $x$, with distinct segments showing where the function is actively adjusting its slope. Right panel: The figure provides a comprehensive comparison of five popular AFs used in NNs: Swish, HardSwish, Logistic Sigmoid, Tanh, and ReLU, plotted over a range of $x$ from $-5$ to 5. Each function demonstrates distinct characteristics: Swish shows a smooth, Sigmoid-like response; HardSwish offers a piecewise linear, computationally efficient approximation of Swish; Logistic Sigmoid illustrates a classic S-shaped curve; Tanh presents a symmetric Sigmoid centered around zero; and ReLU activates linearly for positive inputs while clamping negative values to zero.

The fact that the gradients for the negative part of the function approach zero can also be observed in Swish and ReLU activations. However, the particular shape of the curve described in the negative part, which gives both Swish and E-swish the non-monotonicity property, improves performance since they can output small negative numbers, unlike ReLU.

### *6.3 HardSwish*

Hard-Swish [67] is closely related to AF Swish. It is defined as

$$\sigma_{\text{HardSigmoid}}(x) = \max(0, \min(1, (0.2x + 0.5))), \tag{86}$$

$$\begin{aligned}\sigma_{\text{HardSwish}}(x) &= 2x\, \sigma_{\text{HardSigmoid}}(\beta x) \\ &= 2x\, \max(0, \min(1, (0.2\beta x + 0.5))) \\ &= x \min \frac{[\max[x+3, 0], 6]}{6} \\ &= \begin{cases} 0, & x \leq -3, \\ x, & x \geq 3, \\ x(x+3)/6, & \text{otherwise,} \end{cases}\end{aligned} \tag{87}$$

where $\beta$, is either a trainable parameter or a constant. As $\beta \to \infty$, the hard-Sigmoid component approaches $0 - 1$, and Hard-Swish will act like the ReLU AF. This indicates that Hard-Swish interpolates non-linearly between the ReLU function and linear function smoothly. Setting $\beta$, as a trainable parameter can be used to control the degree of interpolation in the model.

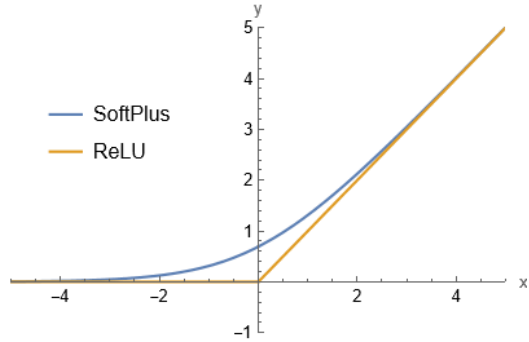

**Fig. 60.** The figure compares the SoftPlus and ReLU AFs, plotted over a range of $x$ from $-5$ to $5$. SoftPlus, a smooth approximation of ReLU, gradually transitions from near zero to a linearly increasing function, displaying a continuous and differentiable curve. In contrast, ReLU features a sharp threshold at $x = 0$, where it transitions from zero for negative inputs to a linear response for positive inputs.

The properties of HardSwish are similar to Swish because both are unbounded above and bounded below. It is non-monotonic. It is faster in computation compared to swish because it doesn't involve any exponential calculation. The particular shape of the curve in negative part improves performance as they can output small negative numbers.

The non-monotonic bump is the most striking difference between Hard-Swish and other AF when $x$ is less than 0 as shown in Fig. 59. Inside the domain of the bump ($2.5 \leq x \leq 0$), a large percentage of pre-activations fall leading to better convergence and improvement on benchmarks.

### 6.4 SoftPlus

The SoftPlus function [68] was proposed in 2001 and is mostly used in statistical applications. SoftPlus unit-based AF is also used in DNNs [69]. As a smoothing version of the ReLU function, Fig. 60, the SoftPlus function is defined as:

$$\sigma_{\text{SoftPlus}}(x) = \log(e^x + 1). \tag{88.1}$$

The derivative of the SoftPlus function with respect to its input $x$ can be computed as follows:

$$\frac{\partial}{\partial x} \sigma_{\text{SoftPlus}}(x) = \frac{e^x}{1 + e^x} = \sigma_{\text{Sigmoid}}(x). \tag{88.2}$$

This derivative can be helpful in training NNs using gradient-based optimization algorithms like backpropagation.

The SoftPlus has a number of advantages. It is smooth and differentiable everywhere, which makes it well-suited for gradient-based optimization techniques like backpropagation. This property makes the SoftPlus function more stable no matter when being estimated from the positive and negative directions, while ReLU has a discontinuous gradient at point 0. The SoftPlus function maps its input to a range between 0 and positive infinity, which can be useful in certain situations. It is a monotonic function, meaning that as the input $x$ increases, the output also increases.

Unlike some other AFs, such as the Sigmoid or Tanh, the SoftPlus function does not saturate for large positive inputs. Saturated AFs can lead to vanishing gradients and slow down the learning process. Additionally, the SoftPlus function has a non-zero gradient even when the input is negative. Unlike ReLU, which propagates no gradient for $x < 0$, SoftPlus can propagate gradients across all real inputs. This feature prevents the "dying ReLU" problem, where neurons become inactive and do not update their weights during training. Furthermore, the SoftPlus unit also outperforms the Sigmoid unit with the following aspects. The derivative of SoftPlus is a Sigmoid function. It means that the gradient of the SoftPlus unit approaches 1 when the input increases, which largely reduces the bad effects of the vanishing gradient problem.

It is important to note that, while there may be concerns about the hard saturation of ReLU and the resulting zero gradients, experimental evidence [29] suggests that this characteristic might not hinder supervised training as much as initially thought. The use of SoftPlus as an alternative does not necessarily provide a clear advantage in all cases, and the benefits of hard zeros in ReLU might outweigh the drawbacks in certain scenarios, particularly when some hidden units remain active during training. The hard non-linearities do not hurt so long as the gradient can propagate along some paths, i.e., that some of the hidden units in each layer are non-zero.

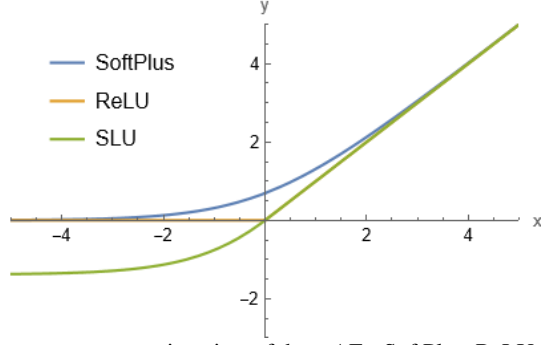

**Fig. 61.** The figure presents a comparative view of three AFs: SoftPlus, ReLU, and SLU, plotted over a range of $x$ from $-5$ to $5$. SoftPlus offers a smooth, gradual transition, acting as a continuously differentiable approximation of the ReLU function, which directly clamps all negative values to zero and linearly passes positive values. SLU introduces a unique behavior where it behaves linearly for non-negative inputs and transitions to a Sigmoid-logarithmic form for negative inputs, blending properties of the Sigmoid and logarithmic functions for a softer response compared to ReLU.

*6.5 SoftPlus Linear Unit (SLU)*

A SoftPlus Linear Unit (SLU) was also proposed by considering SoftPlus with rectified unit [70]. It is known that a zero gradient problem and a bias shift exist when ReLU is used in networks. Based on the theory that "zero mean activations improve learning ability", the SLU was proposed as an adaptive AF. The SLU AF is defined as,

$$\sigma_{SLU}(x) = \begin{cases} \beta x, & x \geq 0, \\ \alpha \log(e^x + 1) + \gamma, & x < 0, \end{cases} \tag{89.1}$$

where $\beta$ is the slope factor in the positive part. The larger the slope factor $\beta$, the steeper the slope of the positive part is. The parameter $\alpha$ relates to the location of the saturation point in the negative quadrant. The larger the $\alpha$, the lower the saturation point is. The parameter $\gamma$ denotes the distance to the horizontal axis in the negative part, and the bigger the value of $\gamma$, the larger the distance is. Obviously, the negative part of the SLU is always less than zero. The average value of the SLU reduces the bias shift compared to the ReLU.

The derivative of SLU can be calculated as:

$$\frac{\partial}{\partial x}\sigma_{SLU}(x) = \begin{cases} \beta, & x \geq 0, \\ \dfrac{\alpha}{e^{-x} + 1}, & x < 0. \end{cases} \tag{89.2}$$

The negative part is the Sigmoid function multiplied by a constant $\alpha$. In order to ensure that the function is continuous and differentiable at zero, the parameters were constrained as follows:

$$\lim_{x \to 0^+} \sigma_{SLU}(x) = 0, \tag{90.1}$$

$$\lim_{x \to 0^-} \sigma_{SLU}(x) = \alpha \log 2 - \gamma, \tag{90.2}$$

$$\lim_{x \to 0^+} \sigma_{SLU}(x) = \lim_{x \to 0^-} \sigma_{SLU}(x), \tag{90.3}$$

$$\lim_{x \to 0^+} \sigma'_{SLU}(x) = \beta, \tag{90.4}$$

$$\lim_{x \to 0^-} \sigma'_{SLU}(x) = \frac{\alpha}{2}, \tag{90.5}$$

$$\lim_{x \to 0^+} \sigma'_{SLU}(x) = \lim_{x \to 0^-} \sigma'_{SLU}(x). \tag{90.6}$$

By solving (90.3) and (90.6), the following is obtained:

$$\gamma = \alpha \log 2, \tag{91.1}$$

$$\beta = \frac{\alpha}{2}. \tag{91.2}$$

In order to avoid a vanishing or exploding gradient during backpropagation, the slope factor $\beta$ should be constrained by: $\beta = 1$. Hence, $\alpha = 2$ and $\gamma = 2 \log 2$. Therefore, the precise definition of SLU, Fig. 61, is:

$$\sigma_{SLU}(x) = \begin{cases} x, & x \geq 0, \\ 2\log\left(\dfrac{e^x + 1}{2}\right), & x < 0. \end{cases} \tag{92}$$

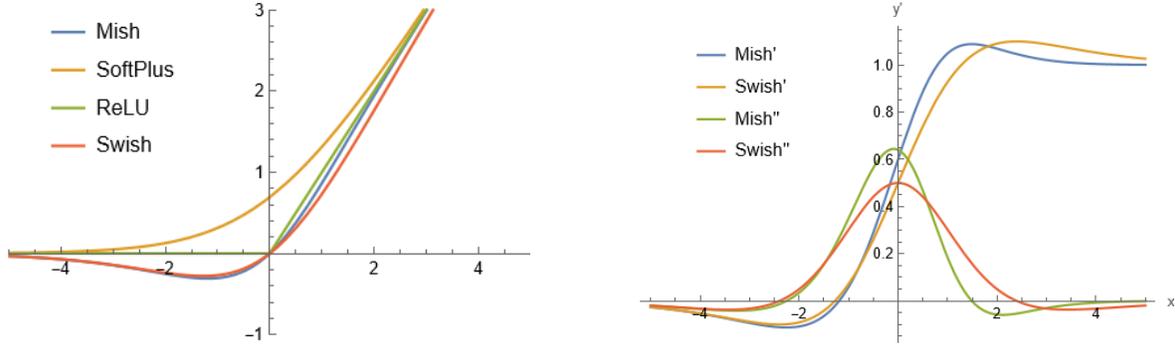

**Fig. 62.** Left panel: The figure illustrates a comparison of four advanced AFs: Mish, SoftPlus, ReLU, and Swish, each plotted over a range of $x$ from $-5$ to $5$. Mish, defined as $x \operatorname{Tanh}(\ln(1 + e^x))$, showcases a smooth and non-monotonic curve that closely resembles the behavior of Swish but with subtle differences in negative input handling. SoftPlus, a smooth approximation of ReLU, shows a gentle logistic-like transition, emphasizing its continuous and differentiable nature. ReLU remains the simplest, with a direct zero-clamping behavior for negative inputs and linear response for positive values. Swish, defined as $x \cdot \sigma_{\text{Sigmoid}}(\beta x)$ with $\beta = 1$, combines aspects of linear and sigmoidal responses, providing a flexible shape that varies with the input value. Right panel: The figure presents a detailed comparison of the first and second derivatives of the Mish and Swish AFs, plotted over a range of $x$ from $-5$ to $5$. The first derivatives, labeled as "Mish'" and "Swish'", show the rate of change of the respective AFs, illustrating how they respond to different input values. The second derivatives, labeled as "Mish''" and "Swish''", indicate the curvature or the rate of change of the slope of the AFs, providing deeper insights into their dynamic behavior.

### *6.6 Mish*

The SoftPlus function is also used with the Tanh function in Mish AF [71]. Mish, as visualized in Fig. 62, is a smooth, continuous, self-regularized, non-monotonic AF. Mathematically, it was defined as:

$$\sigma_{\text{Mish}}(x) = x \operatorname{Tanh}(\sigma_{\text{SoftPlus}}(x))$$
$$= x \operatorname{Tanh}(\ln(1 + e^x)). \quad (93.1)$$

Similar to Swish, Mish is bounded below and unbounded above with a range of $[\approx -0.31, \infty)$. The 1st derivative of Mish, as shown in Fig. 62, can be defined as:

$$\frac{\partial}{\partial x}\sigma_{\text{Mish}}(x) = \frac{\omega e^x}{\delta}, \quad (93.2)$$

where, $\omega = 4(x + 1) + 4e^{2x} + e^{3x} + e^x(4x + 6)$ and $\delta = 2e^x + e^{2x} + 2$.

The Mish AF offers several advantages. It preserves a small amount of negative information, thereby eliminating the preconditions necessary for the Dying ReLU phenomenon and improving expressivity and information flow. Being unbounded above, Mish avoids saturation, which typically slows down training due to near-zero gradients. Additionally, its boundedness below results in strong regularization effects. Unlike ReLU, Mish is continuously differentiable, avoiding singularities and undesired side effects during gradient-based optimization.

Mish's non-monotonic nature helps capture complex and non-linear relationships in data. Similar to the SoftPlus function, Mish is smooth and differentiable everywhere, making it well-suited for gradient-based optimization methods like backpropagation. The function also has a built-in self-regularization property, resisting very large input values and helping to avoid exploding gradients during training. It has demonstrated competitive performance in DNNs compared to other popular AFs, such as ReLU and LReLU, in various experimental settings. However, the increased complexity of Mish, due to its multiple functions, can be a limitation for DNNs.

### *6.7 Gaussian Error Linear Unit*

The motivation behind Gaussian Error Linear Unit (GELU) [72] is to bridge stochastic regularizers, such as dropout, with non-linearities. Dropout is a regularization technique commonly used in NNs to prevent overfitting. The idea behind dropout is to randomly "drop out" (i.e., set to zero) a fraction of the units/neurons in a layer during training. This prevents individual neurons from becoming overly specialized and encourages more robust learning. When dropout is applied stochastically during training, it means that for each training example, a random subset of neurons is dropped out. In other words, dropout regularization stochastically multiplies a neuron's inputs with 0, randomly rendering them inactive. This suggests a more probabilistic view of a neuron's output. On the other hand, ReLU activation deterministically multiplies inputs with 0 or 1 dependent upon the input's value.

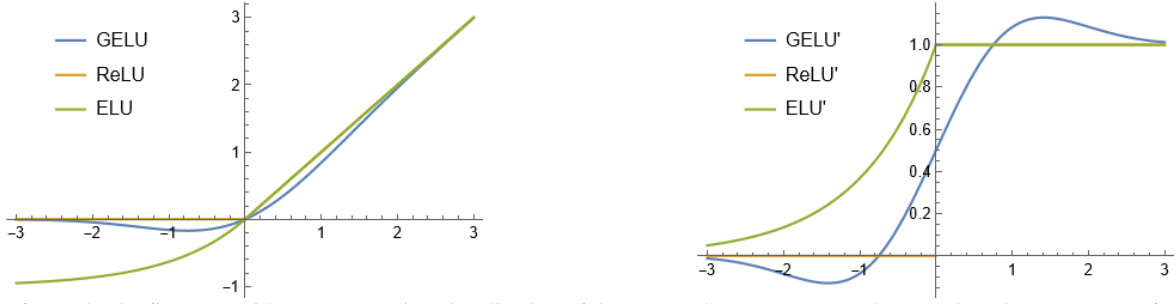

**Fig. 63.** Left panel: The figure provides a comparative visualization of three AFs: GELU, ReLU, and ELU, plotted over a range of $x$ from $-3$ to 3. The GELU function, defined as $\frac{x}{2}\left(1+\text{erf}\left(\frac{x}{\sqrt{2}}\right)\right)$, offers a smooth, probabilistic approach to activation by incorporating the error function (erf). ReLU, the simplest function, activates linearly for positive $x$ values and clamps negative values to zero, illustrating a sharp transition at zero. ELU, with $\alpha=1$, combines linear behavior for positive $x$ values and exponential decay for negative $x$, providing a smooth transition that aims to mitigate the dying ReLU problem. Right panel: The figure displays the first derivatives of three AFs: GELU, ReLU, and ELU, plotted over a range of $x$ from $-3$ to 3. The derivative of the GELU function, labeled as "GELU'", demonstrates a smooth, S-shaped curve reflecting its probabilistic nature and gradual transitions. The ReLU derivative, labeled as "ReLU'", is a piecewise function that is zero for negative $x$ and one for positive $x$, showing a sharp change at zero. The ELU derivative, labeled as "ELU'" with $\alpha=1$, combines a constant positive slope for $x>0$ and an exponential rise for $x\leq 0$, offering a smoother gradient flow compared to ReLU.

GELU merges both functionalities by multiplying inputs by a value from 0 to 1. However, the value of this zero-one mask, while stochastically determined, is also dependent upon the input's value. The GELU AF is

$$\sigma_{\text{GELU}}(x) = x\,\Phi(x) = x\,P(X \leq x), \tag{94.1}$$

where $\Phi(x)$ is the standard Gaussian cumulative distribution function. The GELU AF is defined as $x$ times the standard Gaussian cumulative distribution function of $x$, which can be written as $x$ times the probability that a random variable from a normal distribution with mean 0 and variance 1 is less than or equal to $x$.

In this setting, inputs have a higher probability of being "dropped" as $x$ decreases, so the transformation applied to $x$ is stochastic yet depends upon the input. The GELU nonlinearity weights inputs by their value, rather than gates inputs by their sign as in ReLU. So, the transformation applied by GELU is stochastic, yet it depends upon the input's value through $\Phi(x)$.

In fact, the GELU can be viewed as a way to smooth a ReLU. To see this, recall that $\sigma_{\text{ReLU}}(x) = \max(x,0) = x\mathbf{1}(x>0)$ (where $\mathbf{1}$ is the indicator function), while the GELU is $\sigma_{\text{GELU}}(x) = x\,\Phi(x)$ if $\mu=0, \sigma=1$. Then the CDF is a smooth approximation to the binary function the ReLU uses, like how the Sigmoid smoothed binary threshold activations. Unlike the ReLU, the GELU and ELU can be both negative and positive.

In Fig. 63, observe how $\sigma_{\text{GELU}}(x)$ starts from zero for small values of $x$ since the CDF is almost equal to 0. However, around the value of $-2$, CDF starts increasing. Hence, we see $\sigma_{\text{GELU}}(x)$ deviating from zero. For the positive values, since CDF moves closer to a value of 1, $\sigma_{\text{GELU}}(x)$ starts approximating $\sigma_{\text{ReLU}}(x)$.

Since the cumulative distribution function of a Gaussian is often computed with the error function, GELU was defined as

$$\sigma_{\text{GELU}}(x) = x\,\Phi(x) = \frac{x}{2}\left(1+\text{erf}\left(\frac{x}{\sqrt{2}}\right)\right). \tag{94.2}$$

It can be approximated using

$$\sigma_{\text{GELU}}(x) \approx 0.5\,x\left(1+\text{Tanh}\left[\sqrt{\frac{2}{\pi}}(x+0.044715x^3)\right]\right). \tag{95.1}$$

Alternatively, the function can be approximated using the Sigmoid function and a scaling parameter, as shown below:

$$\sigma_{\text{GELU}}(x) \approx x\sigma_{\text{Sigmoid}}(1.702x). \tag{95.2}$$

Both are sufficiently fast, easy-to-implement approximations.

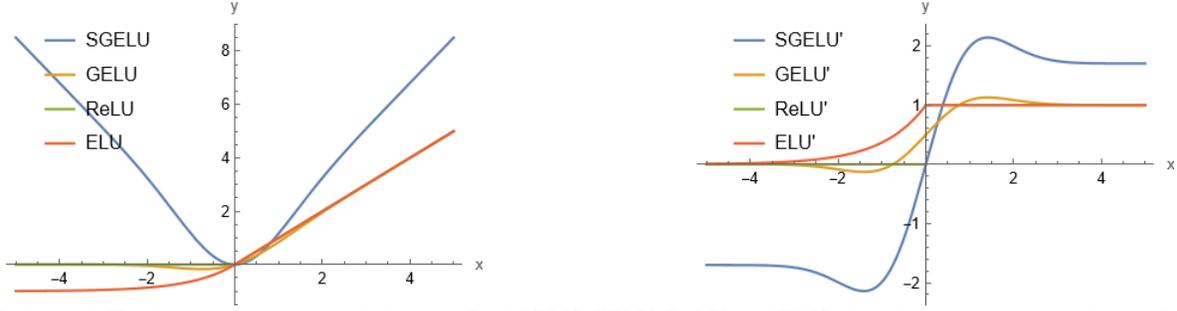

**Fig. 64.** Left panel: The figure compares four different AFs: SGELU, GELU, ReLU, and ELU, plotted over a range of $x$ from $-5$ to $5$. The SGELU function, scaled by a parameter $\alpha$ (default 1.702), shows a smooth curve enhanced by the error function, providing a flexible and smooth transition. The GELU function, with its probabilistic nature, also exhibits a smooth transition similar to SGELU but without the scaling factor. ReLU is characterized by its linear increase for positive values and zero output for negative values, indicating a sharp change at zero. The ELU function, with $\alpha = 1$, combines an exponential approach for negative inputs with a linear response for positive inputs, providing a more gradual transition compared to ReLU. Right panel: The figure presents the first derivatives of four AFs: SGELU, GELU, ReLU, and ELU, plotted over a range of $x$ from $-5$ to $5$.

The choice of the GELU AF is motivated by several desirable properties that make it suitable for NNs. GELU introduces a non-linearity in the network, allowing it to capture complex relationships in the data. The cubic term in the GELU function contributes to this non-linearity, and the combination of different elements helps shape the AF's curve. GELU is a smooth function, meaning that it has continuous derivatives across its entire domain. This smoothness can be advantageous for optimization algorithms that rely on gradients, such as GD. GELU is differentiable everywhere, which is crucial for backpropagation during the training of NNs. The ability to compute gradients allows optimization algorithms to update the model parameters in the direction that minimizes the loss function.

Unlike some other AFs, such as ReLU, GELU is zero-centered, which can help the optimization process by reducing the risk of weights consistently updating in a positive or negative direction, contributing to more stable learning dynamics. The incorporation of the Gaussian cumulative distribution function $\Phi(x)$ in the GELU formulation aligns its behavior with aspects of the standard normal distribution, beneficial for capturing statistical properties in the data and improving generalization.

GELU differs from other AFs in notable ways. This non-convex, non-monotonic function is not linear in the positive domain and exhibits curvature at all points, unlike ReLUs and ELUs, which are convex, monotonic, and linear in the positive domain. The increased curvature and non-monotonicity may allow GELUs to approximate complicated functions more easily than ReLUs or ELUs. Additionally, since $\sigma_{\text{ReLU}}(x) = x\mathbf{1}(x > 0)$ and $\sigma_{\text{GELU}}(x) = x\Phi(x)$ if $\mu = 0$, $\sigma = 1$, we can see that the ReLU gates the input depending upon its sign, while the GELU weights its input depending upon how much greater it is than other inputs.

Significantly, GELU has a probabilistic interpretation, as it represents the expectation of a stochastic regularizer, adding a unique dimension to its functionality in NNs.

### *6.8 Symmetrical Gaussian Error Linear Unit*

Since the GELU function represents the nonlinearity using the stochastic regularizer on an input, which is the cumulative distribution function derived from the Gaussian error function, it has shown the advantage over other functions, e.g., ReLU, ELU. However, most AFs do not fully exploit the negative value. Taking this into account, Symmetrical Gaussian Error Linear Unit (SGELU) [73] was proposed to combine the advantage of stochastic regularizer on the input and exploit the negative value, which can be represented by

$$\sigma_{\text{SGELU}}(x) = \alpha x \, \text{erf}\left(\frac{x}{\sqrt{2}}\right), \tag{96}$$

in which $\alpha$ represents the hyper-parameter that can be tuned in the computation to obtain the optimum solution. The properties of different AFs have been shown in Fig. 64. In Fig. 64, the SGELU AF is different from GELU, ReLU, and ELU, which shows the symmetrical characteristics together with Gaussian regularizer. For better illustration, the derivatives of ReLU, ELU, GELU, and SGELU are plotted as shown in Fig. 64.

The biggest difference of SGELU from other AFs is what happens in the negative half-axis. Instead of forcing the output to be zero like ReLU, deviating the output from a true value like GELU until it stops converging, or dragging the output from negative to positive like ELU, SGELU can update its weight symmetrically towards to two directions in both positive and negative half axis.

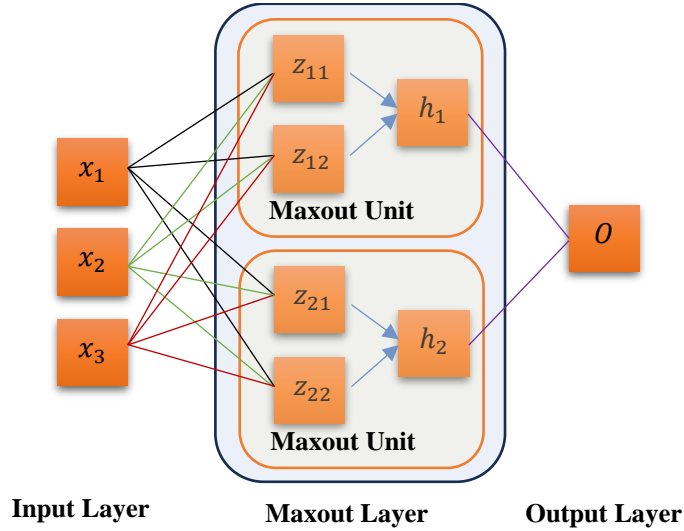

**Fig. 65.** NN Architecture with Maxout Layer. The figure illustrates a NN architecture that includes a Maxout layer. The network consists of an input layer, a hidden Maxout layer, and an output layer. The input layer receives three inputs, $x_1, x_2$, and $x_3$. The Maxout layer contains two Maxout units. Each Maxout unit takes two inputs, $z_{11}$ and $z_{12}$ for the first unit, and $z_{21}$ and $z_{22}$ for the second unit, and computes the maximum value between them, denoted as $h_1$ and $h_2$ respectively. The outputs from the Maxout layer are then fed into the output layer to produce the final output $O$. This configuration enables the network to learn more complex functions by implementing a piecewise linear AF through the Maxout units.

In other words, the function of SGELU is a two-to-one mapping between the input and the output, while the others are a one-to-one mapping. Meanwhile, different from LiSHT, the Gaussian stochastic characteristics is employed to weight the input, resulting in further enhancement of the model capability. Therefore, these features of SGELU make it more efficient and robust.

## 7. Non-Standard AFs

So far, we have taken care of AFs in the classic meaning given by literature, i.e., a function that builds the output of the neuron using as input the value returned by the internal transformation $\mathbf{w}^T\mathbf{x} + b$ made by the classic computational neuron model. In this section, we will review works which change the standard definition of neuron computation but are considered as NN models with trainable AFs in the literature. In other terms, these functions can be considered as a different type of computational neuron unit compared with the original computational neuron model.

### 7.1 Maxout

In NNs, the Maxout activation [74] takes the maximum value of the pre-activations. Fig. 65 shows two pre-activations per Maxout unit, each of these pre-activations has a different set of weights from the inputs denoted as "$x$". Each hidden unit takes the maximum value over the $j$ units of a group:

$$h_i = \max_j z_{ij}, \tag{97}$$

where $z$ is the linear pre-activation value, $i$ is the number of Maxout units and $j$ the number of pre-activation values.

In other words, a layer of linear nodes is added to every hidden unit, each connected to all the input nodes, see Fig. 65. These nodes use the AF $y(x) = x$ and send a signal forward to the hidden node, which uses the maxout unit. The hidden node simply chooses the largest of the values produced by the linear nodes and sends it forward to the output node. Hence, the Maxout AF works as follows:

Input: Maxout takes multiple inputs (usually two or more) instead of a single scalar value like other AFs. Each input represents the weighted sum of the inputs to a neuron before the AF is applied.

Computation: For each neuron, Maxout computes the maximum value of its input values. In other words, it takes the maximum value among the inputs and returns that as the output. Mathematically, for a Maxout unit with '$k$' inputs, the output is the maximum of these inputs: $\max(z_1, z_2, \ldots, z_k)$.

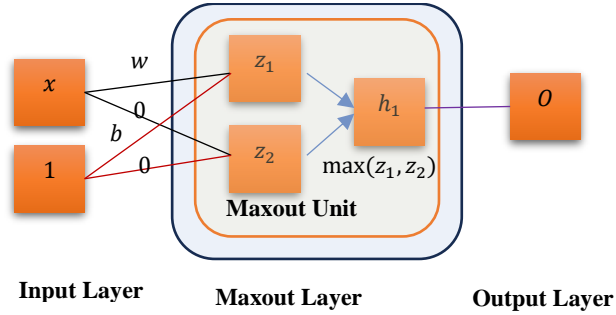

**Fig. 66.** ReLU as a special case of Maxout in a NN. This figure demonstrates how the ReLU AF can be seen as a special case of the Maxout AF within a NN. The network consists of an input layer, a Maxout layer, and an output layer. The input layer receives a single input, $x$, which is processed into two linear combinations, $z_1$ and $z_2$. The Maxout layer includes a Maxout unit that computes the maximum value between $z_1$ and $z_2$, represented as $h_1 = \max(z_1, z_2)$. This value is then passed to the output layer to produce the final output $O$. By setting one of the linear combinations, $z_2$, to zero, and the bias term $b$ to zero, the Maxout unit effectively implements the ReLU function, $\max(0, z_1)$, showing that ReLU is a specific form of the more general Maxout function.

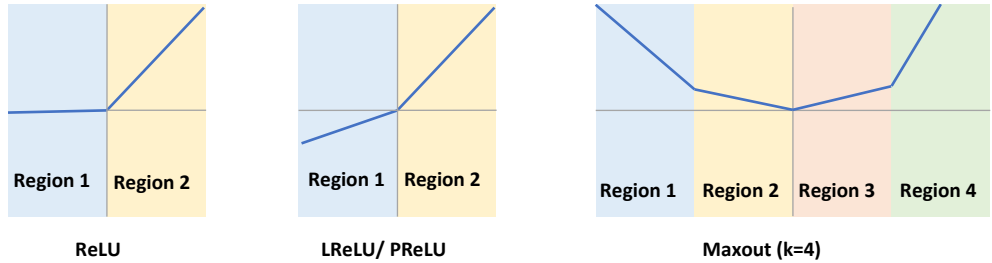

**Fig. 67.** The figure demonstrates how the Maxout AF can generalize ReLU and LReLU/PReLU. In the ReLU case, the Maxout function reduces to $\max(z_1, 0)$ by setting $z_2$ to zero and bias $b$ to zero, creating two regions (Region 1 and Region 2). For LReLU/PReLU, different slopes for the negative part of the activation are introduced, still dividing the function into two regions. The Maxout function with $k = 4$ shows the division into multiple regions (Regions 1-4), indicating its flexibility in approximating various piecewise linear functions by taking the maximum of multiple linear combinations. This versatility makes Maxout a powerful tool for learning complex activation patterns in NNs.

Maxout chooses the maximum of $n$ input features to produce each output feature in a network, the simplest case of maxout is the Max-Feature-Map (MFM), where $n = 2$. The MFM Maxout computes the function

$$\sigma_{\text{Maxout}}(\mathbf{x}) = \max(\mathbf{w}_1^T \mathbf{x} + b_1, \mathbf{w}_2^T \mathbf{x} + b_2), \tag{98}$$

and both the ReLU and LReLU are a special case of this form. When specific weight values $\mathbf{w}_1$, $b_1$, $\mathbf{w}_2$ and $b_2$ of the MFM inputs are learned, MFM can emulate ReLU and other rectified linear variants, see Fig. 66. The Maxout unit is helpful for tackling the problem of vanishing gradients because the gradient can flow through every Maxout unit.

In the ReLU, LReLU and PReLU, there must be two sets of points: one lying in the negative side and another on the positive side of the AF domain (see region 1 covering the negative side and region 2 on the positive side in PReLU/ LReLU cases of Fig. 67). Moreover, in the Maxout activation units, there must be $k$ sets of points - each set lying in one of the $k$ regions of the AF domain (see Maxout case in Fig. 67). Both ReLU and LReLU can be seen as the special cases of Maxout.

Maxout units take the maximum value over a subspace of $k$ trainable linear functions of the same input $\mathbf{x}$, obtaining a piece-wise linear approximator capable of approximating any convex function. The AF created by the Maxout nodes will always be a convex function, however, with enough Maxout nodes, it can approximate any convex function arbitrarily well. In theory, Maxout can approximate any convex function, but a large number of extra parameters introduced by the $k$ linear functions of each hidden Maxout unit result in large RAM storage memory cost and a considerable increase in training time, which affect the training efficiency of very DNNs.

*7.2 Softmax*

The Softmax AF [1] is a widely used AF in NNs, particularly in multi-class classification problems. It takes as input a vector of real numbers and transforms them into a probability distribution, where the values are in the range [0,1], and they sum up to 1.

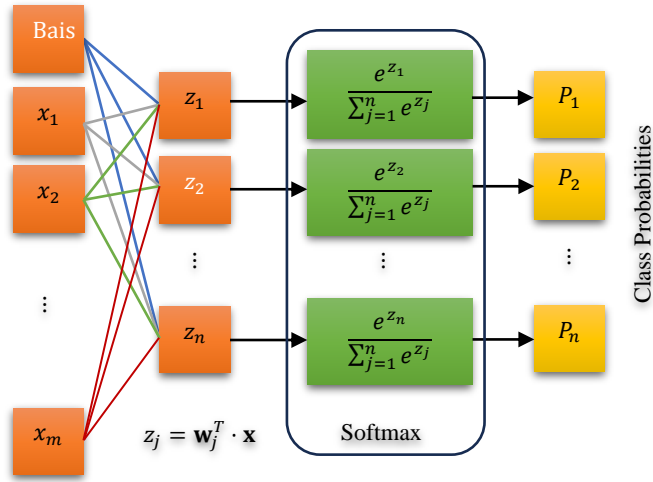

**Fig. 68.** The figure illustrates the Softmax AF used in the output layer of a NN for multi-class classification. The network consists of an input layer, a fully connected layer, and a Softmax output layer. The input layer receives $m$ input features $x_1, x_2, \ldots, x_m$, which are linearly combined with weights and biases to compute the logits $z_1, z_2, \ldots, z_n$ for each class $j$. The logits are calculated as $z_j = \mathbf{w}_j^T \cdot \mathbf{x} + b_j$, where $\mathbf{w}_j$ is the weight vector for class $j$, and $\mathbf{x}$ is the input feature vector. The Softmax function then transforms these logits into class probabilities. For each class $i$, the probability $P_i$ is computed using the Softmax formula: $P_i = e^{z_i} / \sum_j e^{z_j}$. This formula ensures that the probabilities for all classes sum to 1, allowing the network to assign a probability distribution over the possible classes. In the figure, the computation for each class probability is depicted, showing the exponential of each logit $e^{z_1}, e^{z_2}, \ldots, e^{z_n}$ and the normalization by the sum of all exponentials. The resulting class probabilities $P_1, P_2, \ldots, P_n$ indicate the likelihood of the input belonging to each respective class. This Softmax layer is commonly used in the final layer of NNs for classification tasks to provide a probabilistic interpretation of the model's predictions.

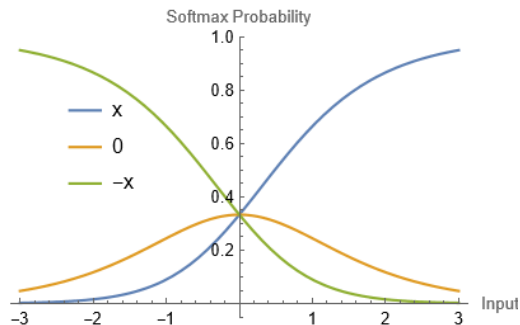

**Fig. 69.** The figure visualizes the Softmax function applied to a range of inputs, showcasing how it transforms raw input values into probabilities that sum to 1. The plot demonstrates the Softmax probabilities for three inputs $(x, 0, -x)$ as $x$ varies from $-3$ to 3. The Softmax function, which is often used in classification tasks in NNs, converts these inputs into a set of probabilities. As $x$ increases, the probability associated with the input $x$ rises, while the probabilities for the inputs 0 and $-x$ adjust accordingly.

Given an input vector $\mathbf{z} = [z_1, z_2, \ldots, z_n]$, the Softmax function calculates the output vector $\mathbf{a} = [a_1, a_2, \ldots, a_n]$ as follows:

$$\sigma_{\text{Softmax}}(z_i) = a_i = \frac{e^{z_i}}{\sum_j e^{z_j}} \quad for\ i = 1, 2, \ldots, n. \tag{99}$$

The Softmax AF works by taking a vector of real numbers as input and transforming these numbers into a probability distribution. It assigns probabilities to each element in the input vector, such that the values are in the range [0,1], and they sum up to 1. This makes it suitable for multi-class classification problems where you want to determine the likelihood of an input belonging to each class.

Here's a step-by-step explanation of how the Softmax AF works, see Figs. 68 and 69:

1. The final hidden layer preceding the Softmax layer might use linear (identity) activations. The identity activation means that the output of the layer is simply a linear combination of its inputs, without applying any non-linear AF. "Logits" is defined as the numerical output of the final linear layer of a multi-class NN.
2. You start with an input vector $\mathbf{z} = [z_1, z_2, \ldots, z_n]$, where each $z_i$ is a real number. These values are typically the raw scores or logits obtained from the previous layer of a NN. The goal is to convert these scores into a probability distribution.
3. For each element $z_i$ in the input vector, you compute the exponential ($e^{z_i}$) of that element. This step amplifies the differences between the values in the input vector, emphasizing the larger values and diminishing the smaller ones.
4. Next, you sum up all the exponentiated values to get the denominator of the Softmax formula. This step involves calculating the sum of all $e^{z_i}$ for $i$ in the range from 1 to $n$, $\sum_j e^{z_j}$.
5. For each element in the input vector, you divide the exponential of that element by the denominator calculated in the above step. This calculates the probability of each element being the most likely class. The result is an output vector $\mathbf{a}$, where each $a_i$ is a probability.

$$a_i = \frac{e^{z_i}}{\sum_j e^{z_j}} \quad \text{for } i = 1, 2, \ldots, n. \tag{100}$$

6. The output vector now contains the probabilities for each class. Each element, $a_i$, represents the probability that the input belongs to class $i$. The class with the highest probability is considered the predicted class, and the Softmax function ensures that all probabilities sum up to 1.
7. For example, if you have an input vector $\mathbf{z} = [2.0, 1.0, 0.1]$, applying the Softmax function would yield an output vector $\mathbf{a} = [0.659001, 0.242433, 0.0985659]$. This means that the first class is the most likely class for the given input, with a probability of approximately 0.659.
8. Before applying the Softmax function, class labels are typically represented using one-hot encoding, where each class is represented as a binary vector with a 1 at the index corresponding to the class and 0s everywhere else. Softmax then provides probabilities for each class.

**Lemma 5:** The derivative of the Softmax function with respect to the logit ($z_i = \mathbf{W}_i^T \cdot \mathbf{x}$) is

$$\frac{\partial}{\partial z_i} \sigma_{\text{Softmax}}(z_j) = \sigma_{\text{Softmax}}(z_j)\left(\delta_{ij} - \sigma_{\text{Softmax}}(z_i)\right). \tag{101}$$

**Proof:** Let

$$\sigma_{\text{Softmax}}(z_j) = \frac{e^{\mathbf{W}_j^T \cdot \mathbf{x}}}{\sum_{k=1}^K e^{\mathbf{W}_k^T \cdot \mathbf{x}}} = \frac{e^{z_j}}{\sum_{k=1}^K e^{z_k}}.$$

Computing the

$$\frac{\partial}{\partial z_i} \sigma_{\text{Softmax}}(z_j) = \frac{\partial}{\partial z_i} \frac{e^{z_j}}{\sum_{k=1}^K e^{z_k}}.$$

The derivative of $\sum_{k=1}^K e^{z_k}$ with respect to any $z_i$ will be $e^{z_i}$. As for the numerator, $e^{z_j}$ the derivative will be $e^{z_i}$ if and only if $z_i = z_j$; otherwise, the derivative is 0.

If $i = j$, and using the quotient rule,

$$\begin{aligned}
\frac{\partial}{\partial z_i} \frac{e^{z_j}}{\sum_{k=1}^K e^{z_k}} &= \frac{e^{z_j} \sum_{k=1}^K e^{z_k} - e^{z_i} e^{z_j}}{[\sum_{k=1}^K e^{z_k}]^2} \\
&= \frac{e^{z_j}(\sum_{k=1}^K e^{z_k} - e^{z_i})}{\sum_{k=1}^K e^{z_k} \sum_{k=1}^K e^{z_k}} \\
&= \frac{e^{z_j}}{\sum_{k=1}^K e^{z_k}} \frac{\sum_{k=1}^K e^{z_k} - e^{z_i}}{\sum_{k=1}^K e^{z_k}} \\
&= \frac{e^{z_j}}{\sum_{k=1}^K e^{z_k}} \left(1 - \frac{e^{z_i}}{\sum_{k=1}^K e^{z_k}}\right) \\
&= \sigma_{\text{Softmax}}(z_j)\left(1 - \sigma_{\text{Softmax}}(z_i)\right).
\end{aligned}$$

If on the other hand, $i \neq j$:

$$\frac{\partial}{\partial z_i} \frac{e^{z_j}}{\sum_{k=1}^{K} e^{z_k}} = \frac{0 - e^{z_i} e^{z_j}}{[\sum_{k=1}^{K} e^{z_k}]^2}$$

$$= \frac{-e^{z_i} e^{z_j}}{\sum_{k=1}^{K} e^{z_k} \sum_{k=1}^{K} e^{z_k}}$$

$$= -\frac{e^{z_j}}{\sum_{k=1}^{K} e^{z_k}} \frac{e^{z_i}}{\sum_{k=1}^{K} e^{z_k}}$$

$$= -\sigma_{\text{Softmax}}(z_j) \sigma_{\text{Softmax}}(z_i)$$

$$= \sigma_{\text{Softmax}}(z_j) \big(0 - \sigma_{\text{Softmax}}(z_i)\big).$$

These two scenarios can be brought together as

$$\frac{\partial}{\partial z_i} \sigma_{\text{Softmax}}(z_j) = \sigma_{\text{Softmax}}(z_j) \big(\delta_{ij} - \sigma_{\text{Softmax}}(z_i)\big).$$

∎

## 8. Combining AFs

### *8.1 Mixed, Gated, and Hierarchical AFs*

First, let us consider the mixed activation and gated activation strategies [75]. The mixed activation strategy involves combining basic AFs linearly. This means taking a weighted sum of different AFs. The combination coefficients are learned from the data, meaning the NN adapts and adjusts these weights during training. In contrast, in gated activation strategy, basic AFs are combined nonlinearly. This often involves the use of gating mechanisms to control the flow of information. Similar to the mixed strategy, the coefficients for combining AFs are learned from the data.

We begin with AFs with predefined parameters (LReLU and ELU). Note that, LReLU AFs $\sigma_{\text{LReLU}}(x) = (x \text{ if } x > 0, \alpha x \text{ if } x \leq 0)$ (considering the negative part) do not become saturated (i.e., they do not reach extreme values) regardless of how small the input is. However, ELU AFs $\sigma_{\text{ELU}}(x) = (x \text{ if } x > 0, \beta(e^x - 1) \text{ if } x \leq 0)$ saturate to a negative value when the input is small. Both types of AFs can change forward and backward propagated information, but they do so in different ways. The variables $\alpha$ and $\beta$ are used to represent the degree of information change, and they provide a way to quantify how much the information is altered by the AFs.

Consider the above two strategies for combining these basic AFs. The first strategy, mixed activation, does not adapt to specific inputs. In this approach, the learning process results in a fixed mixture of LReLU and ELU. The second strategy, gated activation, adapts to specific inputs. Here, the learning process produces a learned gating mask that determines an adaptive mixture of LReLU and ELU, allowing the network to adjust based on the specific inputs. Both of these strategies involve combining AFs with predefined parameters. A further extension of these strategies is hierarchical activation, where the AFs themselves can be learned. This means that their parameters are determined in a data-driven way, and the combination is organized within a hierarchical structure.

These strategies offer several advantages. They make the designed AFs more flexible, enhancing the NN's ability to learn non-linear transformations. Additionally, the designed AFs retain the characteristic of information change, modeling the degree of this change both qualitatively and quantitatively.

The mixed activation was defined as [75]:
$$\sigma_{\text{Mix}}(x) = \rho \sigma_{\text{LReLU}}(x) + (1 - \rho) \sigma_{\text{ELU}}(x), \tag{102}$$
where $\rho \in [0,1]$ is a combination coefficient specifying the specific combination of LReLU and ELU. The specific combination coefficient $\rho$ is learned from the data. However, once each combination coefficient is learned in mixed activation strategy, then the mixed activation will be kept constant. In other words, it is not adapted to the specific inputs after learning because it keeps fixed no matter what characteristics of the inputs appear.

Instead of directly learning a fixed combination coefficient, a gating mask is learned during training. The combination coefficient is not fixed; it depends on both the learned gating mask and the input values. The product of the gating mask and inputs is passed through a Sigmoid function, providing a dynamic and adaptive way to generate the combination coefficient between different AFs.

The AF for the gated strategy can be formulated as follows: Let $x$ be the input, $\omega$ be the gating mask, $\sigma_{\text{Sigmoid}}(.)$ be the Sigmoid function. The combination coefficient $\tau$ is defined using the Sigmoid function applied to the product of $\omega$ and $x$:

$$\tau = \sigma_{\text{Sigmoid}}(\omega x) = 1/(1 + \exp(-\omega x)). \tag{103}$$

Now, the AF for the gated strategy can be expressed as a weighted combination of different AFs (LReLU and ELU, in this case) using the learned combination coefficient $\tau$:

$$\sigma_{\text{Gate}}(x) = \tau \sigma_{\text{LReLU}}(x) + (1-\tau)\sigma_{\text{ELU}}(x). \tag{104}$$

This formulation allows the AF to dynamically adjust its behavior based on both the input $x$ and the learned gating mask $\omega$. The Sigmoid function ensures that the combination coefficient $\tau$ is within the range of [0,1], providing a weighted combination of the two AFs.

The mixed strategy and the gated strategy adopt linear and nonlinear methods, respectively. By comparing the above two strategies, the main difference is that the mixed strategy is not adaptive to the specific inputs, but the gated strategy is adaptive in adjusting the mixture of LReLU and ELU for the specific inputs. From the perspective of information change as described above, the degree of information change can be further quantitatively adjusted by $\rho$ or $\tau$. Specifically, the gated strategy learns $\omega$ which indicates the degree of information change for the whole dataset. After learning $\omega$, $\tau$ determines the specific proportion of information change for the specific $x$. In addition, when $x$ varies from zero to a smaller value in $\tau$, the degree of information change will tend to be determined by one of the AF types instead of both.

The two strategies detailed above focus on the ways of combining basic AFs with predefined parameters. In order to further improve the ability of learning non-linear transformation and has the adaptability to the inputs, the basic AFs being combined are organized in a more complex hierarchical structure. The hierarchical structure involves three levels, where low-level nodes are associated with learnable AFs, middle-level nodes combine pairs of low-level nodes, and high-level nodes integrate outputs from all middle-level nodes using the winner-take-all principle.

In other words, the hierarchical structure introduces a more complex organization of basic AFs with learnable parameters, aiming to enhance the network's capacity for learning non-linear transformations and improving adaptability to diverse inputs. The winner-take-all integration principle at the high level suggests a mechanism for selecting the most relevant information from the lower levels.

To be more specific in the hierarchical structure, for a low-level node with index $n$, the basic AF is denoted by $\sigma_{\text{low}}^n(x)$. For middle-level nodes, we proceed in the same way to the gated strategy. At each middle-level node with index $m$, a pair of child low-level nodes are combined into a single parent value $\sigma_{\text{mid}}(x)$ with a learned gating mask denoted by $\omega_m$. From the perspective of the middle-level node, the pair of low-level nodes can also be denoted by $\sigma_{\text{mid}}^{m\,\text{left}}(x)$ and $\sigma_{\text{mid}}^{m\,\text{right}}(x)$. The combination of child nodes into a parent value is given by:

$$\sigma_{\text{mid}}^m(x) = \tau\sigma_{\text{mid}}^{m\,\text{left}}(x) + (1-\tau)\sigma_{\text{mid}}^{m\,\text{right}}(x). \tag{105}$$

The overall activation operation $\sigma_{\text{high}}(x) = \sigma_{\text{Hierarchical}}(x)$ involves selecting the maximum value across $k$ middle-level nodes based on the winner-take-all principle. Neurons in the high-level node compete with each other, with only the neuron having the highest activation being activated while others are inhibited. Overall, the activation result at each level node is

$$\sigma_{\text{Hierarchical}}(x) = \begin{cases} \sigma_{\text{low}}^n(x), & \text{low} - \text{level nodes,} \\ \sigma_{\text{mid}}^m(x), & \text{middle} - \text{level nodes,} \\ \sigma_{\text{high}}(x), & \text{high} - \text{level node,} \end{cases} \tag{106.1}$$

where

$$\sigma_{\text{high}}(x) = \max_{m \in [1,k]} \sigma_{\text{mid}}^m(x). \tag{106.2}$$

Inspired by Maxout, the winner-take-all principle, $\max_{m \in [1,k]} \sigma_{\text{mid}}^m(x)$, introduces competition among activation neurons in a layer. Only the neuron with the highest activation is allowed to be activated, enhancing nonlinearity and adaptability to specific inputs.

*8.2 Combinations of AFs*

Assuming we have an input vector **x**, a NN $N_d$ made of $d$ hidden layers can be seen as the functional composition of $d$ functions $L_i$ followed by a final mapping $\bar{L}$ that depends on the task at hand (e.g. classification, regression). Mathematically, this can be expressed as:

$$N_d = \bar{L}(L_d(\ldots(L_1(\mathbf{x}))). \tag{107}$$

In each layer, there are weights and biases associated with the connections between neurons. If $\mathbf{W}_i$ represents the weight matrix and $\mathbf{b}_i$ represents the bias vector for the $i$-th layer, the transformation $L_i$ in the $i$-th layer can be expressed as the composition of two functions, $\Sigma_i$ followed by $\sigma_i$, the former being a suitable remapping of the layer input neurons, the latter being the AF of the layer:
$$L_i = \sigma_i \circ \Sigma_i = \sigma_i(\mathbf{W}_i \cdot \mathbf{x} + \mathbf{b}_i). \tag{108}$$
In the most general case, both $\sigma_i$ and $\Sigma_i$ are parameterized and belong to some hypothesis spaces $H_{\sigma_i}$ and $H_{\Sigma_i}$, respectively. The composition of these transformations through all $d$ layers gives the final output of the NN, $N_d(\mathbf{x})$. Hence, the learning procedure of $L_i$ amounts to an optimization problem over the layer hypothesis space
$$H_i = H_{\sigma_i} \times H_{\Sigma_i}. \tag{109}$$
The choice of AF $\sigma_i$ in a NN is often predetermined and considered non-learnable in many scenarios. In such cases, the hypothesis space for the AF $H_{\sigma_i}$ is a singleton set, meaning it contains only one fixed function. This is because the parameters of the AF are typically not adjusted during the training process but are rather predefined. Common choices for AFs include ReLU, Sigmoid, Tanh, and others. Each of these AFs serves a specific purpose in introducing non-linearity to the network, allowing it to learn complex patterns and relationships in the data.
$$H_i = \{\sigma_i\} \times H_{\Sigma_i}. \tag{110}$$
For example, for a fully connected layer with ReLU activation that takes an input from $\mathbb{R}^{n_i}$ and produces an output in $\mathbb{R}^{m_i}$, we have that $H_{\Sigma_i}$ is the set of all affine transformations from $\mathbb{R}^{n_i}$ to $\mathbb{R}^{m_i}$. An affine transformation is a mathematical function that transforms points in one space to points in another space. It consists of two main components: 1- Linear Transformation: This part involves scaling, rotating, shearing, or any combination of these operations. In the context of NNs, it is typically represented by a matrix multiplication. If $\mathbf{W}_i$ is the weight matrix, and $\mathbf{x}$ is the input vector, the linear transformation is given by $\mathbf{W}_i \cdot \mathbf{x}$. 2- Translation: This involves adding an offset or shift to the transformed points. In the context of NNs, this is often represented by a bias vector. So, mathematically, the hypothesis space for $\Sigma_i$ can be expressed as follows:
$$H_{\Sigma_i} = \text{Lin}(\mathbb{R}^{n_i}, \mathbb{R}^{m_i}) \times K(\mathbb{R}^{m_i}) = \mathbf{W}_i \cdot \mathbf{x} + \mathbf{b}_i, \tag{111}$$
$$H_i = \{ReLU\} \times \text{Lin}(\mathbb{R}^{n_i}, \mathbb{R}^{m_i}) \times K(\mathbb{R}^{m_i}), \tag{112}$$
where $\text{Lin}(A, B)$ and $K(B)$ are the sets of linear maps between $A$ and $B$, and the set of translations of $B$, respectively.

Two techniques were proposed for defining learnable AFs [76] that can be used in all hidden layers of a NN architecture. Both of them are based on the following idea: (1) A finite set of AFs $\Phi = \{\phi_1, \ldots, \phi_N\}$, is selected as base elements; (2) The learnable AF $\sigma_i$ is defined as a linear combination of the elements of $\Phi$; (3) A suitable hypothesis space $H_{\sigma_i}$ is identified; (4) The overall network is optimized, where the hypothesis space of each hidden layer is defined as $H_i = \{\sigma_i\} \times H_{\Sigma_i}$. This suggests that the optimization involves tuning both the AF parameters and the parameters of the other components of each hidden layer.

Given a vector space $V$ and a finite subset $S \subseteq V$, we can define the following two subsets of $V$:

**Definition (Convex Hull)**: The convex hull of a finite subset $S$ of the vector space $V$, denoted as $\text{conv}(S)$, is the set of all convex combinations of points in $S$. Mathematically, $\text{conv}(S)$ is defined as:
$$\text{conv}(S) = \left\{ \sum_i c_i \mathbf{s}_i : \sum_i c_i = 1, c_i \geq 0, \mathbf{s}_i \in S \right\}. \tag{113}$$
Here, $c_i$ are coefficients such that their sum is equal to 1, and each is non-negative. The resulting sum represents a convex combination of points from $S$.

**Definition (Affine Hull)**: The affine hull of a finite subset $S$ of the vector space $V$, denoted as $\text{aff}(S)$, is the set of all affine combinations of points in $S$. Mathematically, $\text{aff}(S)$ is defined as:
$$\text{aff}(S) = \left\{ \sum_i c_i \mathbf{s}_i : \sum_i c_i = 1, \mathbf{s}_i \in S \right\}, \tag{114}$$
here, $c_i$ are coefficients such that their sum is equal to 1. Unlike in the convex hull, there is no requirement that the coefficients be non-negative. The resulting sum represents an affine combination of points from $A$. In both cases, the convex hull and the affine hull provide a way to describe the set of points that can be obtained as combinations of the points in the original finite subset $S$.

Let $\Phi = \{\phi_0, \phi_1, \ldots, \phi_N\}$ be a finite collection of AFs $\phi_i$ from $\mathbb{R}$ to $\mathbb{R}$. We can define a vector space $\mathbf{V}$ from $\Phi$ by taking all linear combinations:
$$\sum_i c_i \phi_i. \tag{115}$$
Note that, despite $\Phi$ is (by definition) a spanning set of $\mathbf{V}$, it is not generally a basis; indeed $|\Phi| \geq \dim \mathbf{V}$.

*Differentiability*

Differentiability (almost everywhere) is a property preserved by finite linear combinations. This is a key property in the context of NNs because it implies that if the original functions are differentiable, any finite linear combination of them will also be differentiable. Assuming $\Phi$ contains (almost everywhere) differentiable AFs, both $\text{conv}(\Phi)$ and $\text{aff}(\Phi)$ are made up of (almost everywhere) differentiable functions. This is because the property of differentiability is preserved under finite linear combinations, and both convex hull and affine hull involve such combinations. Hence, we have
$$\text{conv}(\Phi) \subset \text{aff}(\Phi) \subset \mathbf{V}. \tag{116}$$

*Monotonic increasing functions*

Activations commonly used in real-world scenarios are described as monotonic increasing functions. Monotonicity is a property where the function consistently moves in one direction (either increasing or decreasing) as its input varies. However, monotonicity is not necessarily preserved under arbitrary linear combinations. In other words, even if all individual functions $\phi_i \in \Phi$ are non-decreasing (monotonic increasing), the result of an arbitrary linear combination $\phi \in \mathbf{V}$ might not be guaranteed to be non-decreasing or non-increasing. This has implications for the design and use of AFs in NNs. While individual AFs may exhibit desired properties, the combination of these functions through linear combinations may not always guarantee the same properties. As a matter of fact, the monotonicity of the elements in $\Phi$ is preserved by all the elements in the convex hull, $\text{conv}(\Phi)$, which means that any linear combination of functions in $\Phi$ while staying within the convex hull will also be monotonic. On the other hand, monotonicity is not preserved by the elements of the affine hull, $\text{aff}(\Phi)$. This implies that while convex combinations ensure monotonicity, affine combinations (which include both convex combinations and translations) may not guarantee monotonicity.

Actually, non-monotonic AFs can approximate arbitrarily complex functions for sufficiently large NNs. Therefore, also $\text{aff}(\Phi)$ is a proper candidate for $H_{\sigma_i}$.

**Lemma 6:** Let all $\phi_i \in \Phi$ be linearly independent and approximate the identity function near the origin (i.e. $\phi_i(0) = 0$ and $\phi_i'(0) = 1$), then, $\phi \in \mathbf{V}$ approximates the identity if and only if $\phi \in \text{aff}(\Phi)$. Since $\text{conv}(\Phi) \subset \text{aff}(\Phi)$, also $\text{conv}(\Phi)$ enjoys the same property.

**Proof:** By hypothesis the $\phi_i$ form a basis (linearly independent and span $\mathbf{V}$), we have, $\phi \in \mathbf{V}$,
$$\phi = \sum_i c_i \phi_i, \qquad \phi' = \sum_i c_i \phi_i',$$

and prove the two-way implication.

$\implies$ By hypothesis, $\phi(0) = 0$ and $\phi'(0) = 1$. The relation on the derivative reads $\sum_i c_i \phi_i'(0) = \sum_i c_i = 1$. In turn, this implies that $\phi = \sum_i c_i \phi_i$ with $\sum_i c_i = 1$, namely $\phi \in \text{aff}(\Phi)$.

$\impliedby$ By hypothesis, $\phi = \sum_i c_i \phi_i$ with $\sum_i c_i = 1$. Hence, $\phi(0) = \sum_i c_i \phi_i(0) = 0$ and $\phi'^{(0)} = \sum_i c_i \phi_i'(0) = \sum_i c_i = 1$, namely $\phi$ approximates the identity near the origin. ∎

We can now formalize the two techniques to build learnable AFs as follows [76]: (i) choose a finite set $\Phi = \{\phi_1, \ldots, \phi_N\}$, where each $\phi_i$ is a (almost everywhere) differentiable AF approximating the identity near origin (at least from one side); (ii) define a new AF $\phi$ as a linear combination of all the $\phi_i \in \Phi$, $\phi = \sum_i c_i \phi_i$; (iii) select as hypothesis space $H_\sigma = \text{aff}(\Phi)$ or $H_\sigma = \text{conv}(\Phi)$.

These techniques provide a structured approach to creating learnable AFs within a NN architecture. The use of a finite set and the combination of functions allow for adaptability and flexibility in capturing complex relationships in the data.

Each choice of $\Phi$ offers a different combination of AFs, and the linear combination of these functions will result in a new AF with properties inherited from its components. The specific choice of $\Phi$ can influence the expressiveness and learning capacity of the resulting AF in a NN. Examples of the choices of $\Phi$ are: $\{\text{id}, \text{ReLU}\}$, $\{\text{id}, \text{Tanh}\}$, $\{\text{ReLU}, \text{Tanh}\}$, and $\{\text{id}, \text{ReLU}, \text{Tanh}\}$,

Moreover, let the set $\Phi$ is chosen to include the identity function id and the ReLU AF. The convex hull of $\Phi$, $\text{conv}(\Phi)$, is formed by taking all convex combinations of the functions in $\Phi$. In this case $\text{conv}(\Phi)$ is expressed as
$$\text{conv}(\Phi) = \{\phi = \alpha \, \text{id} + (1 - \alpha) \, \text{ReLU with } 0 \leq \alpha \leq 1\}. \tag{117}$$
The convex hull-based technique together with $\Phi = \{\text{id}, \text{ReLU}\}$ recovers the learnable LReLU AF. Since ReLU = id for $x \geq 0$ and ReLU = 0 otherwise, we have that $\phi = \alpha \, \text{id} + (1 - \alpha) \, \text{id} = \text{id}$ for $x \geq 0$ and $\phi = \alpha \, \text{id} + (1 - \alpha) \, 0 = \alpha \, \text{id}$ otherwise, i.e. LReLU.

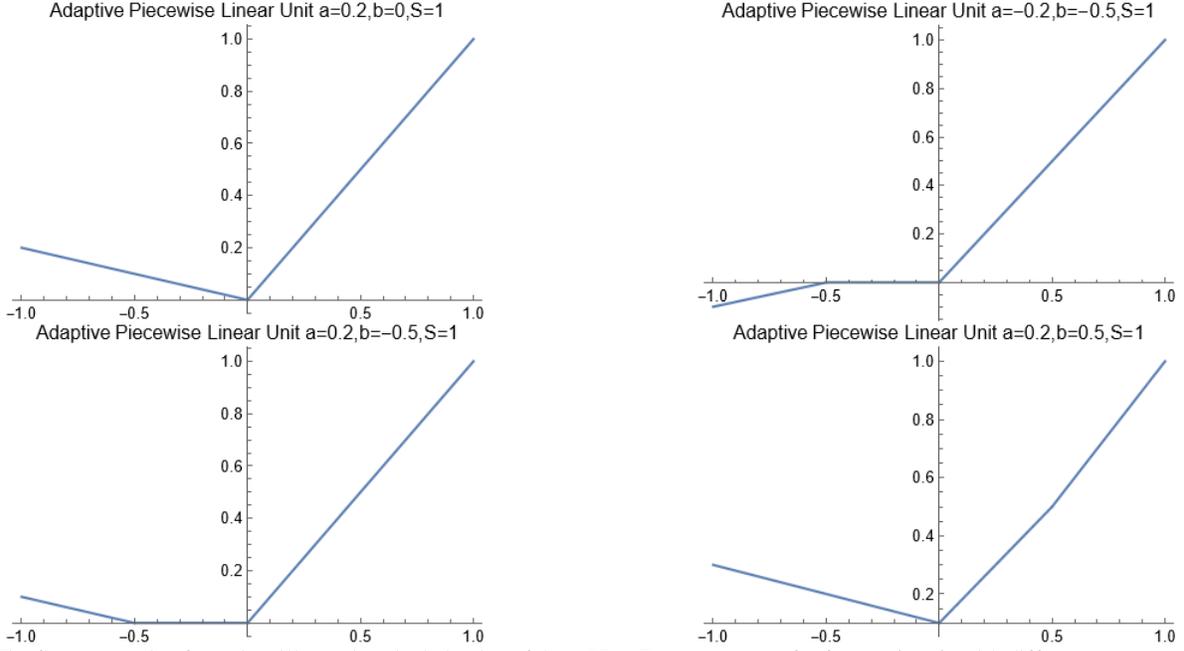

Fig. 70. The figure contains four plots illustrating the behavior of the APL AF over a range of $x$ from $-1$ to $1$, with different parameter settings for $a$ and $b$ with $S = 1$. APL ($a = 0.2, b = 0, S = 1$): The function combines a ReLU component $\max(0, x)$ with a linear adjustment term, resulting in a slight modification for negative inputs. APL ($a = -0.2, b = -0.5, S = 1$): The negative $a$ value introduces a downward adjustment for inputs less than $b$, significantly modifying the activation for negative inputs. APL ($a = 0.2, b = -0.5, S = 1$): This configuration introduces a positive adjustment for inputs less than $b$, altering the AF's slope in the negative region. APL ($a = 0.2, b = 0.5, S = 1$): The positive $b$ value shifts the adjustment term to a different input range, affecting the slope for inputs around $b$.

## 8.3 Adaptive Piecewise Linear Units (APL)

While the type of AF can have a significant impact on learning, the space of possible functions has hardly been explored. One way to explore this space is to learn the AF during training. The APL [77] activation unit uses the summation of ReLU-like units to increase the capacity of the AF. The method formulates the AF $\sigma_{\text{APL}}(x)$ of an APL unit $i$ as a sum of hinge-shaped functions,

$$\sigma_{\text{APL}}(x) = \max(0, x) + \sum_{s=1}^{S} (a_i^s \max(0, -x + b_i^s)). \tag{118}$$

The function consists of two parts. The first part is $\max(0, x)$, which is the ReLU AF. This function returns the maximum of zero and $x$, introducing non-linearity by zeroing out negative values. The second part is $\max(0, -x + b_i^s)$, another ReLU activation applied to the quantity $-x + b_i^s$. Like the first part, it introduces non-linearity by zeroing out negative values. The result is a piecewise linear AF. Fig. 70 shows sample AFs obtained from changing the parameters, $S = 1$ for all plots.

The number of hinges, $S$, is a hyperparameter set in advance, while the variables $a_i^s$, $b_i^s$ for $i \in 1, \ldots, S$ are learned using standard GD during training. The $a_i^s$ variables control the slopes of the linear segments, while the $b_i^s$ variables determine the locations of the hinges. The number of additional parameters that must be learned when using these APL units is $2SM$, where $M$ is the total number of hidden units in the network. This number is small compared to the total number of weights in typical networks. This parametrized, piecewise linear AF is learned independently for each neuron and can represent both convex and non-convex functions of the input.

## 8.4 Mexican ReLU (MeLU)

The "Mexican hat type" function is defined as follows:
$$\phi_{a,\lambda}(x) = \max(\lambda - |x - a|, 0). \tag{119}$$
Here, $a$ and $\lambda$ are real numbers, and $x$ is a variable. The parameter $a$ determines the center of the function along the $x$-axis. It represents the location where the function has its peak or central point resembling the top of a Mexican hat. Shifting the value of $a$ will move the entire function horizontally along the $x$-axis. A larger $a$ shifts the function to the right, and a smaller $a$ shifts it to the left (see Fig. 71). The parameter $\lambda$ controls the width of the function. It is associated with the spread or width of the "Mexican hat" shape.

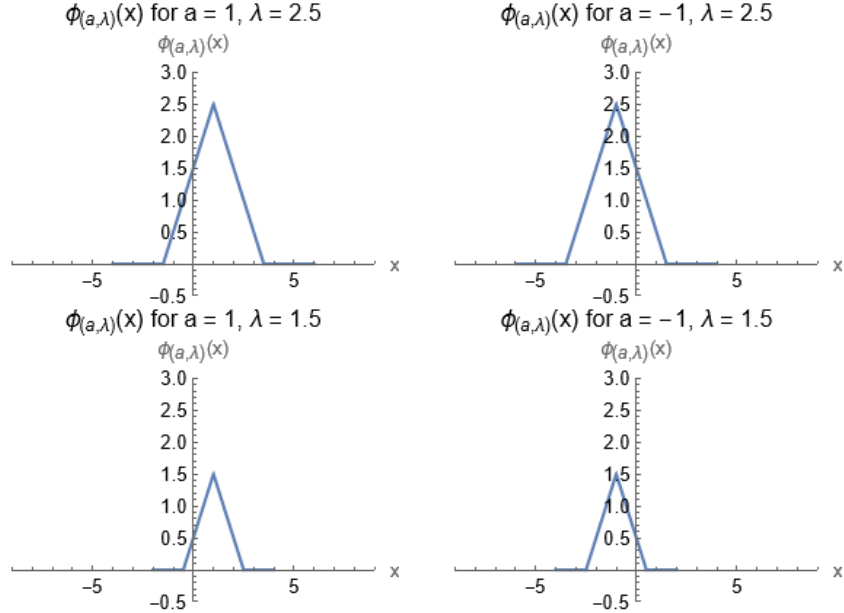

**Fig. 71.** The figure includes four plots illustrating the Mexican hat-type function $\phi_{a,\lambda}(x) = \max(\lambda - |x - a|, 0)$ for various parameter settings. Each plot demonstrates how the function changes based on different values of $a$ and $\lambda$.

A larger $\lambda$ results in a broader "Mexican hat," meaning that the function will have a wider range of values for which it is non-zero. Conversely, a smaller $\lambda$ leads to a narrower hat with a more localized impact. The width of the function is related to the scale of features that the function can detect or represent. Smaller values of $\lambda$ are associated with higher-frequency components, while larger values correspond to lower-frequency components.

The function has a shape resembling a Mexican hat, which is a term often used in mathematics to describe functions with a characteristic peaked shape. The function is null (equal to 0) when the absolute difference between $x$ and $a$ is greater than $\lambda$, i.e., $|x - a| > \lambda$. It increases with a derivative of 1 in the interval $(a - \lambda, a)$. This means that in this interval, the function rises steadily, and its rate of increase is constant. It decreases with a derivative of $-1$ in the interval $(a, a + \lambda)$. In this interval, the function decreases steadily, and its rate of decrease is constant. These functions are the building blocks of MeLU AF [78]. MeLU is defined as

$$\sigma_{\text{MeLU}}(x) = \sigma_{\text{PReLU}}(x) + \sum_{j=1}^{k-1} c_j \, \phi_{a_j, \lambda_j}(x), \tag{120}$$

for each channel of the hidden layer. $c_j$ is a learnable parameter associated with the $j$-th term in the sum. The parameters $a_j$ and $\lambda_j$ are chosen recursively. This means that the values of $a_j$ and $\lambda_j$ for each term in the sum depend on previous terms, likely through a recursive relationship. The total number of learnable parameters in every channel is given by $k$. This includes the $k-1$ coefficients, $c_j$, associated with the "Mexican hat type" functions, and one learnable parameter in the PReLU AF.

MeLU is a flexible AF that combines the piecewise linearity of PReLU with the localized and potentially non-linear effects of "Mexican hat type" functions. This combination can enhance the model's ability to capture and represent both linear and non-linear features in the data. The model will learn the values of the parameters during the training process to optimize the network for a specific task.

The Mexican hat functions are continuous and piecewise differentiable. MeLU inherits these properties because it is defined as a combination of PReLU and the sum of "Mexican hat type" functions. If all the coefficients $c_i$ in the MeLU AF are initialized to zero, MeLU reduces to the PReLU AF. This is an interesting observation, as it implies that MeLU can smoothly transition from PReLU behavior by adjusting the values of the $c_i$ coefficients during training. The same holds for networks trained with LReLU or PReLU. It can exhibit different AF behaviors based on initialization and adapt its characteristics during training. Since it has more parameters, it has a higher representation power, but it might overfit easily, see Fig. 72.

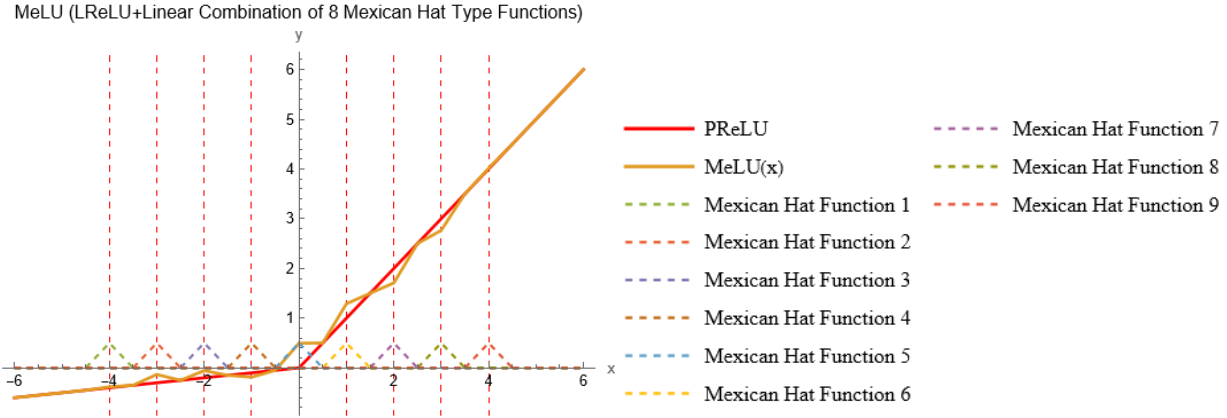

**Fig. 72.** The figure illustrates the MeLU AF, which is a combination of a PReLU and a series of nine Mexican Hat-type functions, plotted over a range of $x$ from $-6$ to $6$. The PReLU function is shown as a red, thick line, representing the baseline AF. The MeLU function, plotted as a thick solid line, incorporates the PReLU function and adds a linear combination of the Mexican Hat type functions, each centered at different points ($x = -4$ to $4$) with widths of $0.5$ and randomly generated coefficients. The individual Mexican Hat-type functions are displayed as dashed lines, showing their contributions to the overall MeLU function.

Note that, the derivatives of "Mexican hat type" functions form a Hilbert basis on a compact set with the $L^2$ norm. This is a notable mathematical property indicating that the derivatives of these functions can be used to approximate every function in $L^2$ space on a compact set.

The structure of a hidden layer in a NN is defined as $\sigma(\mathbf{W}^T \cdot \mathbf{x} + b)$, where $\mathbf{x}$ is the input of the hidden layer, $\mathbf{W}$ is the weight matrix, $\mathbf{b}$ is the bias and $\sigma$ is the activation. Through joint optimization of weights $\mathbf{W}$, biases $\mathbf{b}$, and activation parameters (parameters of the MeLU AF), (i.e., simultaneous optimization of all these parameters during the training process), NNs with MeLU activation can approximate any continuous function on a compact set.

The most similar activations to MeLU in the literature are SReLU and APL. Both APL and MeLU have the ability to approximate the same set of functions: piecewise linear functions that behave like the identity for sufficiently large $x$. While APL and MeLU can both approximate the same set of functions, they do so in distinct ways. APL achieves its flexibility through learnable points of non-differentiability. The positions of these points are adjusted during training. MeLU, on the other hand, achieves its expressive power through the joint optimization of the weights matrix and biases. MeLU is noted to use only half of the parameters of APL while having the same representation power. This suggests that MeLU may offer a more parameter-efficient way to achieve similar function approximation capabilities.

*8.5 Look-up Table Unit*

In traditional methods, feature extraction typically involves designing specific rules or using handcrafted features. In contrast, DNNs automatically learn hierarchical and abstract representations of features through the multiple layers of the network. This enables them to adapt to the complexity of the underlying data distribution. AFs introduce non-linearity to the NN, allowing it to model complex relationships. Instead of using pre-defined AFs (e.g., Sigmoid or hyperbolic tangent), the Look-up Table Unit (LuTU) [79] suggests learning the shape of the AF itself. This implies that the NN will determine the optimal AF during the learning process, offering a high degree of flexibility.

This approach aims to strike a balance between allowing the model to learn the optimal AF shape and ensuring some level of control or constraint over the range of possible shapes. This may lead to more adaptive and data-driven AFs tailored to the specific characteristics of the problem at hand. Moreover, this approach aligns with the broader philosophy of leveraging the power of deep learning to automatically discover complex patterns and representations in the input data.

The LuTU method employs a look-up table-style structure to store a set of anchor points. This structure is likely a data structure, possibly an array or a table, where each entry corresponds to an anchor point for the AF. The AF is interpolated from the anchor points using either simple linear interpolation or cosine smoothing. Interpolation is a method of estimating values between two known values. Linear interpolation is a straightforward method, while cosine smoothing suggests a more sophisticated interpolation technique that likely aims to produce a smoother transition between anchor points. When the distance between adjacent anchor points is small enough, the proposed function can approximate any univariate function. This implies that the look-up table, along with the chosen interpolation method, is capable of capturing a wide range of AF shapes. Visualization of the learned functions demonstrates high diversity, indicating that the LuTU method is capable of capturing a variety of AF shapes. This diversity is considered different from previous AFs, suggesting that the learned functions may have unique characteristics.

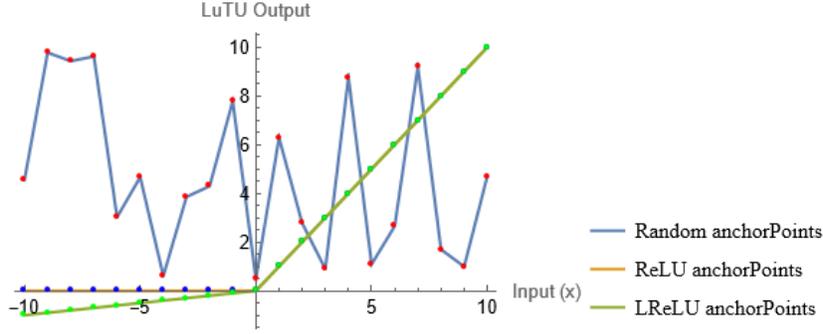

**Fig. 73.** The figure displays the LuTU AF with linear interpolation applied to three sets of anchor points (21 anchor points): randomly generated points (red), ReLU points (blue), and LReLU points (green). Each set of anchor points defines a distinct LuTU function through linear interpolation, calculated as $\sigma_{\text{LulU-Interp}}(x) = \frac{1}{s}(y_i(x_{i+1} - x) + y_{i+1}(x - x_i))$, for $x$ values between consecutive anchor points. The plot shows how the LuTU activation varies based on the different anchor point definitions, with the $x$-axis representing input values from $-10$ to $10$ and the $y$-axis showing the corresponding LuTU output. The legends distinguish the three types of anchor points.

The LuTU AF is a type of AF for NNs that instead of being defined by a mathematical expression, LuTU is controlled by a look-up table. The look-up table consists of a set of anchor points $\{x_i, y_i\}$, where $i = 0, 1, \ldots, n$. The $x$-values $\{x_i\}$ are pre-defined and uniformly spaced with a step size of $s$, while the $y$-values $\{y_i\}$ are learnable parameters. These anchor points are crucial as they determine the rough shape of the LuTU AF. The anchor points $\{x_i, y_i\}$ are configured in such a way that

$$x_i = x_0 + s * i, \tag{121}$$

where $x_0$ is a starting point, $s$ is the step size, and $i$ is the index. This configuration ensures that the $x$-values are uniformly spaced with the specified step size. The AF is generated from the anchor points using two methods: linear interpolation and cosine smoothing.

*Linear Interpolation:*

This method likely involves connecting the anchor points with straight line segments. The LuTU AF is then interpolated between these points using linear interpolation. Linear interpolation is a simple method that estimates values between two known points based on a linear equation. The AF based on linear interpolation is defined as

$$\sigma_{\text{LulU-Interp}}(x) = \frac{1}{s}\big(y_i(x_{i+1} - x) + y_{i+1}(x - x_i)\big), \qquad \text{if } x_i \le x \le x_{i+1}. \tag{122.1}$$

For any input value between $x_i$ and $x_{i+1}$, the output is linearly interpolated from $y_i$ and $y_{i+1}$. When $s \to 0$, $x_0 \to -\infty$ and $n \to \infty$, this AF can approximate any univariate function, see Fig. 73.

The derivative of $\sigma_{\text{LulU-Interp}}(x)$ over $y_i$ and input $x$ are straightforward:

$$\frac{\partial}{\partial y_i}\sigma_{\text{LulU-Interp}}(x) = \frac{x_{i+1} - x}{s}, \tag{122.2}$$

$$\frac{\partial}{\partial y_{i+1}}\sigma_{\text{LulU-Interp}}(x) = \frac{x - x_i}{s}, \tag{122.3}$$

$$\frac{\partial}{\partial x}\sigma_{\text{LulU-Interp}}(x) = \frac{y_{i+1} - y_i}{s}, \qquad x_i \le x \le x_{i+1}. \tag{122.4}$$

*Cosine Smoothing:*

This method involves a more sophisticated interpolation technique known as cosine smoothing. Cosine smoothing suggests that the transition between anchor points is made smoother, possibly to avoid abrupt changes and improve the continuity of the LuTU. The smoothing function is defined as:

$$r(x, \tau) = \begin{cases} \frac{1}{2\tau}\left(1 + \cos\left(\frac{\pi}{\tau}x\right)\right), & -\tau \le x \le \tau, \\ 0, & \text{otherwise}. \end{cases} \tag{123}$$

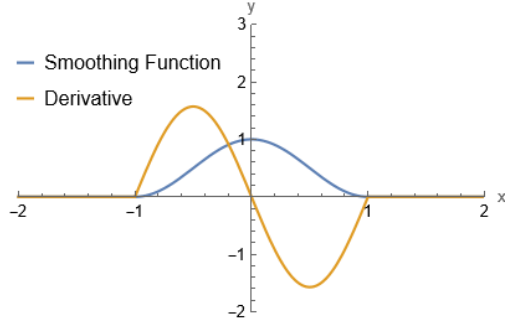

**Fig. 74.** The figure shows a plot of a smoothing function and its derivative for $\tau = 1$, highlighting their behavior over the range $x = -2$ to $x = 2$. The smoothing function is defined as $\frac{1}{2\tau}\left(1 + \cos\left(\frac{\pi}{\tau}x\right)\right)$ within $[-\tau, \tau]$ and zero elsewhere, while the derivative captures the rate of change of the smoothing function.

The function is defined piecewise. For values of $x$ within the interval $[-\tau, \tau]$, it is given by $\frac{1}{2\tau}\left(1 + \cos\left(\frac{\pi}{\tau}x\right)\right)$, and for values outside this interval, the output is zero, Fig. 74. The smoothing function is constructed by shifting and scaling one period of a cosine function. The hyper-parameter $\tau$ controls the period ($2\tau$) of the cosine function. This allows flexibility in adjusting the width of the smoothing function. This function has several favorable properties. First, it is differentiable in all its input domain. This property is advantageous when used as part of AFs in NNs, as differentiability is essential for gradient-based optimization algorithms like backpropagation. Second, the valid input domain for the smoothing function is limited to one period of the cosine function, i.e., $[-\tau, \tau]$. Outside this interval, the output is zero. This limitation reduces the computational workload when calculating the output of the smoothed AF, as only values within the specified interval contribute to the result. Lastly, the integral of this function over its entire domain is equal to one. This property is important when using the smoothing function to convolve or filter other functions. The integral being one means that smoothing a discrete function with this mask will not change its scale, preserving the overall magnitude of the function.

The overall AF $\sigma_{\text{LulU-Cos}}(x)$ is defined as the sum of individual terms, each of which involves multiplying a learnable parameter $y_i$ with a smoothing mask $r(x - x_i, t\,s)$. The function is defined as follows:

$$\sigma_{\text{LulU-Cos}}(x) = \sum_{i=0}^{n} y_i\, r(x - x_i, t\,s). \tag{124}$$

The smoothing mask $r(x - x_i, t\,s)$ is generated using the cosine smoothing function, see Fig. 75. It is centered at $x_i$, and the parameter $t$ determines the ratio between $\tau$ and $s$. The input domain of the smoothing mask is truncated, limited to the interval $[x_i - t\,s, x_i + t\,s]$. This means that for each input value $x$, the calculation of the smoothing masks only involves those masks whose input domain includes $x$. This is a computational optimization, as it reduces the number of masks that need to be evaluated for a given input. The parameter $t$ is an integer that defines the ratio between $\tau$ (the hyper-parameter controlling the period of the cosine function) and $s$ (the step size between anchor points). It influences the width of the smoothing mask and can be adjusted to control the smoothness of the overall AF.

In summary, the AF is a weighted sum of smoothing masks, where each mask is generated by multiplying a learnable parameter with a cosine smoothing function centered at a specific anchor point. The truncated input domain of the smoothing mask allows for efficient computation, considering only the relevant masks for a given input value $x$. The integer parameter $t$ provides a means of controlling the width and smoothness of the individual smoothing masks.

*Mixture of Gaussian Unit:*

Based on the observed learned shapes, the mixture of Gaussian unit (MoGU) AF is designed to capture shapes similar to those learned with LuTU cosine smoothing but with a reduced number of parameters. This can lead to more efficient learning and potentially faster convergence during training. The AF is defined as the sum of individual Gaussian functions:

$$\sigma_{\text{MoGU}}(x) = \sum_{i=1}^{n} \frac{\lambda_i}{\sqrt{2\pi\sigma_i^2}} e^{-\frac{(x-\mu_i)^2}{2\sigma_i^2}}. \tag{125}$$

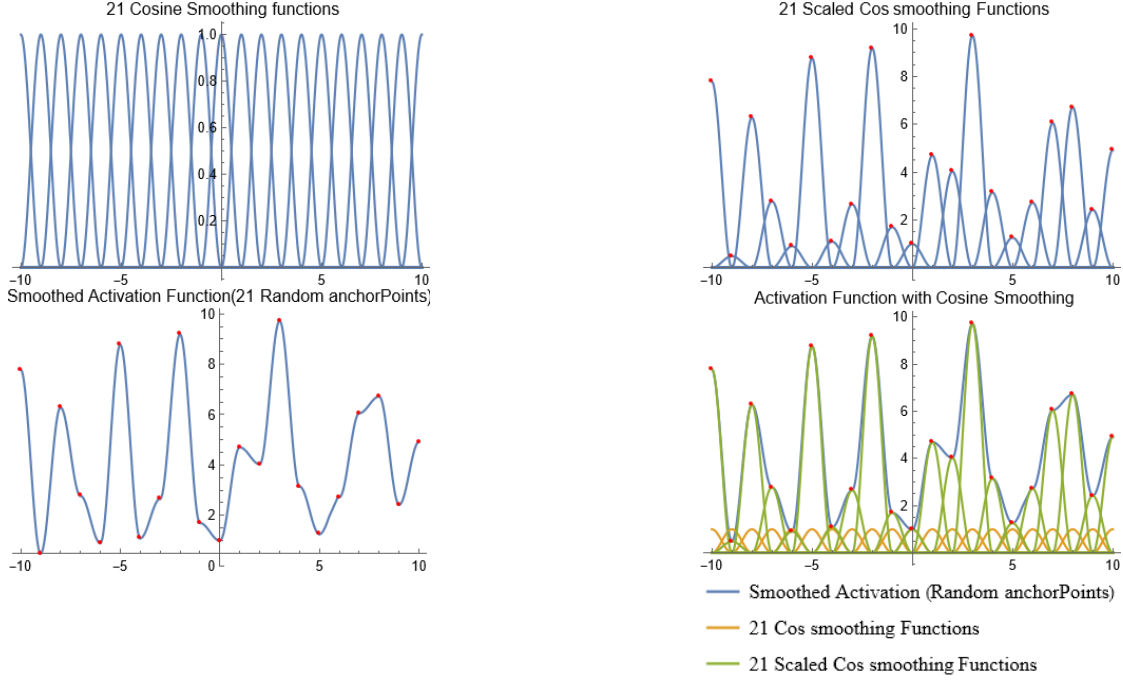

**Fig. 75.** The figures illustrate the construction of the LuTU AF using cosine smoothing. Top left panel: The figure shows 21 individual cosine smoothing functions centered around anchor points from −10 to 10. Top right panel: The figure scales these functions by random values associated with each anchor point, highlighting their contribution to the overall activation. Bottom left panel: The figure sums these scaled functions to form the smoothed AF, with red points indicating the anchor locations. Bottom right panel: The figure combines all components, demonstrating the integration process of individual and scaled smoothing functions into the final LuTU AF.

where, $\lambda_i$ is a learnable parameter controlling the scale of the i-th Gaussian. $\mu_i$ is a learnable parameter representing the mean of the i-th Gaussian. $\sigma_i$ is a learnable parameter controlling the standard deviation of the i-th Gaussian. Each term in the summation corresponds to a Gaussian function. Gaussian functions are bell-shaped curves determined by their mean $\mu_i$ and standard deviation $\sigma_i$. The scale of each Gaussian is controlled by $\lambda_i$.

### *8.6 Bi-Modal Derivative Sigmoidal AFs*

A Bi-modal Derivative Adaptive Activation (BDAA) function uses twin maxima derivative sigmoidal function [80] by controlling the maxima's position with an adaptive parameter. The BDAA is given as,

$$\sigma_{\text{BDAA1}}(x) = \frac{1}{2}\left(\sigma_{\text{Logistic}}(x) + \sigma_{\text{Logistic}}(x+a)\right), \tag{126.1}$$

$$\frac{\partial}{\partial x}\sigma_{\text{BDAA1}}(x) = \frac{1}{2}\left(\sigma_{\text{Logistic}}(x)\left(1 - \sigma_{\text{Logistic}}(x)\right) + \sigma_{\text{Logistic}}(x+a)\left(1 - \sigma_{\text{Logistic}}(x+a)\right)\right), \tag{126.2}$$

$$\frac{\partial}{\partial a}\sigma_{\text{BDAA1}}(x) = \frac{\sigma_{\text{Logistic}}(x+a)\left(1 - \sigma_{\text{Logistic}}(x+a)\right)}{2}, \tag{126.3}$$

where $a \in \mathbb{R}$ and $\sigma_{\text{Logistic}} = 1/(1 + e^{-x})$ is the Logistic Sigmoid function. The Sigmoid function has an S-shaped curve. It maps any real-valued number to the range (0,1). The term $\sigma_{\text{Logistic}}(x + a)$ implies a horizontal shift of the Sigmoid function by $a$ units. The function $\sigma_{\text{BDAA1}}$ takes the average of the original Sigmoid function $\sigma_{\text{Logistic}}(x)$ and the shifted Sigmoid function $\sigma_{\text{Logistic}}(x + a)$. The scaling factor 1/2 ensures that the resulting function stays within the range (0,1) since each Sigmoid function individually is in that range.

The function, $\sigma_{\text{BDAA1}}$, is not symmetric in $a$, thus, the possible values of $a$ to be considered lie in the interval $(-\infty, \infty)$. The function and its derivatives are shown in Fig. 76 for a few values of $a$. As seen from the figure, the derivatives are not symmetric about the $y$-axis, except for the value $a = 0$. The derivative of the Logistic Sigmoid function, $\frac{\partial}{\partial x}\sigma_{\text{Logistic}}(x)$, has a maximum value at the points where $\sigma_{\text{Logistic}}(x)$ is 0.5, and it approaches 0 as $x$ goes to $-\infty$ or $+\infty$. When you combine this with the derivative of the shifted Sigmoid function $\sigma_{\text{Logistic}}(x + a)$, you introduce another set of points where the derivative has a maximum value.

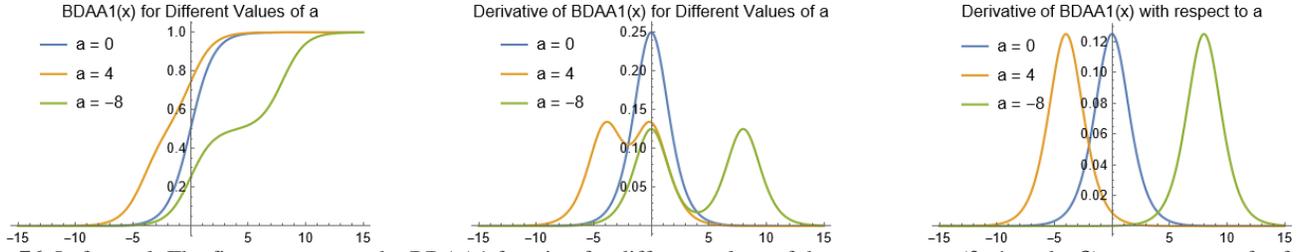

**Fig. 76.** Left panel: The figure compares the BDAA1 function for different values of the parameter $a$ (0, 4, and $-8$) over a range of $x$ from $-15$ to 15. The BDAA1 function combines two Logistic Sigmoid functions shifted by $a$. The plot demonstrates how varying $a$ shifts and blends the sigmoidal curves, influencing the function's output. When $a = 0$, the curve resembles a standard Logistic Sigmoid, while positive and negative $a$ values create asymmetry, modifying the steepness and position of the transition. Middle panel: The figure displays the derivatives of the BDAA1 function for different values of the parameter $a$ (0, 4, and $-8$) over a range of $x$ from $-15$ to 15. The BDAA1 derivative combines the derivatives of two Logistic Sigmoid functions shifted by $a$. The plot illustrates how the value of $a$ affects the derivative's shape and position. When $a = 0$, the derivative resembles the standard Logistic Sigmoid derivative, while positive and negative values of $a$ create variations in the peaks and transition points. Right panel: The figure displays the derivatives of the BDAA1 function with respect to the parameter $a$, plotted for three different values of $a$ (0, 4, and $-8$) over a range of $x$ from $-15$ to 15.

The derivative of the shifted Sigmoid also approaches 0 as $x$ goes to $-\infty$ or $+\infty$, but its maximum value occurs at the points where $\sigma_{\text{Logistic}}(x + a)$ is 0.5. As a result, you get two peaks or modes in the derivative of $\sigma_{\text{BDAA1}}(x)$ with respect to $x$. The positions of these modes depend on the values of $a$ and the locations of the points where the Sigmoid functions have a value of 0.5. Hence, the derivatives of the AF in this case are bi-modal.

This bi-modal behavior can have interesting implications in the context of NNs, especially in terms of learning and convergence dynamics. It indicates that there are two regions in the input space where the network's weights are updated more significantly. Understanding such behaviors can provide insights into the learning dynamics of NNs with this specific AF.

The BDAA2 is given as,

$$\sigma_{\text{BDAA2}}(x) = \sigma_{\text{BDAA1}} - \frac{1}{2} = \frac{1}{2}\big(\sigma_{\text{Logistic}}(x) + \sigma_{\text{Logistic}}(x + a) - 1\big), \tag{127.1}$$

$$\frac{\partial}{\partial x}\sigma_{\text{BDAA2}}(x) = \frac{\partial}{\partial x}\sigma_{\text{BDAA1}}(x), \qquad \frac{\partial}{\partial a}\sigma_{\text{BDAA2}}(x) = \frac{\partial}{\partial a}\sigma_{\text{BDAA1}}(x). \tag{127.2}$$

The subtracted constant $1/2$ shifts the function downwards, affecting the baseline. The subtraction does not affect the shape of the Sigmoid functions themselves but influences their interaction and the overall position of the function. The entire curve of $\sigma_{\text{BDAA1}}(x)$ is shifted downward by $1/2$. This means that the function's values are lowered, and the new baseline (the level at which the function approaches for large negative or positive $x$ is set to $\pm \frac{1}{2}$). Before the subtraction, the baseline was 1. This function is not symmetric in $a$, thus, the values of $a$ to be considered lie in the interval $(-\infty, \infty)$. The function and its derivatives are shown in Fig. 77 for a few values of $a$, as seen from the figure, the derivatives are asymmetric about the $y$-axis, and except for $a = 0$, the function $\sigma_{\text{BDAA2}}(x)$ is not antisymmetric.

$\sigma_{\text{BDAA3}}(x)$ is a combination of two Sigmoid functions with positive and negative shifts, resulting in a function that is influenced by both shifts with equal weight. The symmetry of the Sigmoid terms plays a role in determining the overall symmetry of $\sigma_{\text{BDAA3}}(x)$.

The BDAA3 is given as,

$$\sigma_{\text{BDAA3}}(x) = \frac{1}{2}\big(\sigma_{\text{Logistic}}(x + a) + \sigma_{\text{Logistic}}(x - a)\big), \tag{128.1}$$

$$\frac{\partial}{\partial x}\sigma_{\text{BDAA3}}(x) = \frac{1}{2}\Big(\sigma_{\text{Logistic}}(x + a)\big(1 - \sigma_{\text{Logistic}}(x + a)\big) + \sigma_{\text{Logistic}}(x - a)\big(1 - \sigma_{\text{Logistic}}(x - a)\big)\Big), \tag{128.2}$$

$$\frac{\partial}{\partial a}\sigma_{\text{BDAA3}}(x) = \frac{1}{2}\Big(\sigma_{\text{Logistic}}(x + a)\big(1 - \sigma_{\text{Logistic}}(x + a)\big) - \sigma_{\text{Logistic}}(x - a)\big(1 - \sigma_{\text{Logistic}}(x - a)\big)\Big). \tag{128.3}$$

The positive shift $(x + a)$ and negative shift $(x - a)$ in the Sigmoid functions introduce translations along the $x$-axis. The value of $a$ determines the extent of these shifts. This function is symmetric in $a$, hence the set of values to be considered for $a$ may be taken as $[0, \infty)$. The function and its derivatives are shown in Fig. 78 for a few values of $a$, as seen from the figure, the derivatives are symmetric about the $y$-axis.

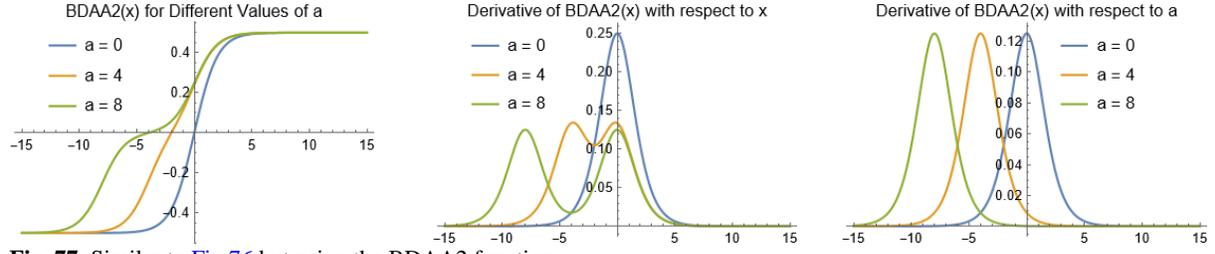

**Fig. 77.** Similar to Fig.76 but using the BDAA2 function.

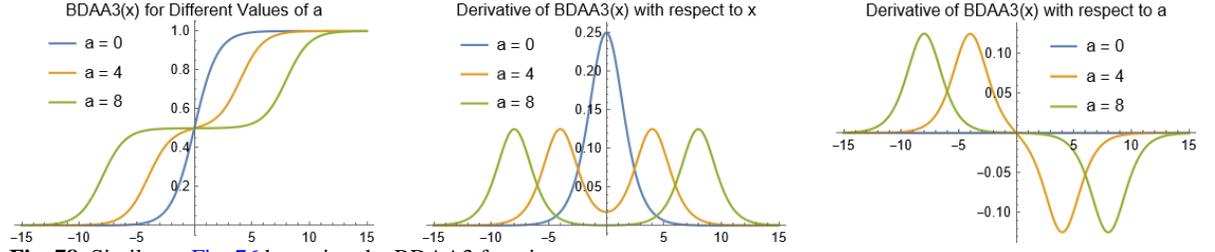

**Fig. 78.** Similar to Fig. 76 but using the BDAA3 function.

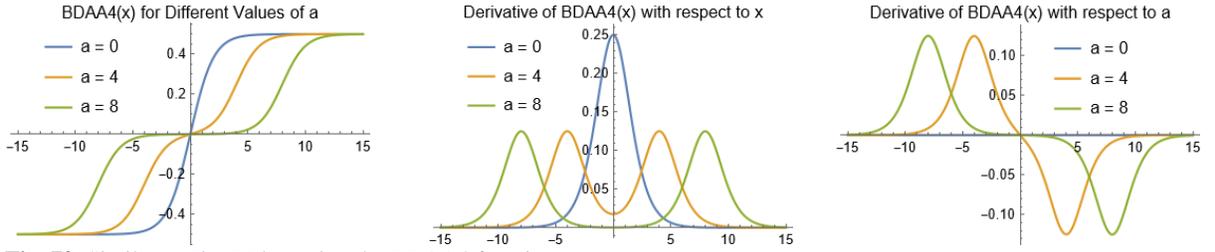

**Fig. 79.** Similar to Fig. 76 but using the BDAA4 function.

The BDAA4 is given as,

$$\sigma_{BDAA4}(x) = \sigma_{BDAA3}(x) - \frac{1}{2},$$  (129.1)

$$\frac{\partial}{\partial x}\sigma_{BDAA4}(x) = \frac{\partial}{\partial x}\sigma_{BDAA3}(x), \qquad \frac{\partial}{\partial a}\sigma_{BDAA4}(x) = \frac{\partial}{\partial a}\sigma_{BDAA3}(x).$$  (129.2)

This function is symmetric in $a$, thus, the possible values of $a$ to be considered lie in the interval $[0, \infty)$. In this case, the derivatives are symmetric about the $y$-axis. The function and its derivatives are shown in Fig. 79 for a few values of $a$, as seen from the figure, the derivatives are symmetric about the $y$-axis, and the function $\sigma_{BDAA4}(x)$ is anti-symmetric for all values of $a$.

The functions $\sigma_{BDAA1}(x)$, $\sigma_{BDAA2}(x)$, $\sigma_{BDAA3}(x)$ and $\sigma_{BDAA4}(x)$ are bounded, continuous, differentiable and sigmoidal functions satisfying the relations:

$$\lim_{x \to \infty} \sigma_{BDAA1}(x) = \lim_{x \to \infty} \sigma_{BDAA3}(x) = 1, \qquad \lim_{x \to -\infty} \sigma_{BDAA1}(x) = \lim_{x \to -\infty} \sigma_{BDAA3}(x) = 0,$$  (130.1)

$$\lim_{x \to \infty} \sigma_{BDAA2}(x) = \lim_{x \to \infty} \sigma_{BDAA4}(x) = \frac{1}{2}, \qquad \lim_{x \to -\infty} \sigma_{BDAA2}(x) = \lim_{x \to -\infty} \sigma_{BDAA4}(x) = -\frac{1}{2}.$$  (130.2)

Note that, the functions $\sigma_{BDAA1}(x)$, $\sigma_{BDAA2}(x)$, $\sigma_{BDAA3}(x)$ and $\sigma_{BDAA4}(x)$ are monotonically increasing functions for any value of $a$. The derivative functions $\sigma'_{BDAA1}(x)$, $\sigma'_{BDAA2}(x)$, $\sigma'_{BDAA3}(x)$ and $\sigma'_{BDAA4}(x)$ have two local maximas for $a \neq 0$, that is, these functions are bi-modal. The functions $\sigma'_{BDAA1}(x)$, and $\sigma'_{BDAA2}(x)$ have local maxima at the point $x = 0$ and $x = -a$, while the functions $\sigma'_{BDAA3}(x)$, and $\sigma'_{BDAA4}(x)$ have the local maxima at $x = \pm a$.

## 9. Performance Comparison and Analysis

Comparing the performance of AFs in NNs requires a systematic and comprehensive approach to ensure fairness and reliability. The following best practices are recommended for making such comparisons:

1. Use a standardized experimental setup by employing the same network architecture across all experiments. This includes keeping the number of layers, number of neurons per layer, and the type of layers used consistent. Additionally, maintain constant hyperparameters like learning rate, batch size, and number of epochs. Furthermore, apply the same weight initialization method to ensure that initial conditions do not bias the results.
2. Utilize the same training and testing datasets for all experiments and apply identical pre-processing techniques such as normalization, augmentation, and handling of missing data. This ensures uniformity in the data input across experiments.
3. Implement $k$-fold cross-validation to ensure that the results are not dependent on a particular train-test split. This provides a more robust measure of performance by averaging results over multiple data partitions.
4. Evaluate the models using multiple metrics. For classification tasks, use (for example) accuracy, precision, and F1-score, and for regression tasks, use mean squared error and $R^2$. It is crucial to choose metrics that are most relevant to the specific task being addressed.
5. Apply statistical analysis using tests such as T-tests or ANOVA to determine if differences in performance are statistically significant. Report confidence intervals for performance metrics to provide a range within which the true performance likely falls.
6. Measure and compare the computational efficiency of each AF by assessing the time taken to train the network and the time required for inference.
7. Conduct multiple runs of each experiment to ensure consistency of results and eliminate the possibility of outcomes being influenced by random variation. Test with different random initializations to confirm the robustness of the results.
8. Visualize the training process by plotting learning curves to illustrate the training and validation performance over epochs. This comprehensive approach will ensure that the comparisons of AFs are fair, reliable, and insightful.

In order to compare the AFs, two experiments are conducted in this paper, one for regression and another for classification. The Mathematica framework is used in all the experiments. All experiments are performed over a desktop system. The code used for experimental comparison is released at: https://github.com/MMHammad24/Activation-Functions-Compared-

## 9.1 Regression Experiment

*Objective:*

The primary objective of this experiment is to systematically evaluate the performance of various AFs in a NN. This will be achieved by closely monitoring several key metrics throughout the training process:

- Gradients Root Mean Square (RMS) per Batch for Each Layer: This metric will provide insights into the stability and magnitude of gradients at each layer during individual training batches.
- Gradients RMS per Epoch (or Round) for Each Layer: By examining this metric, we can observe the overall trend and stability of gradient magnitudes across all layers over each training epoch (or round).
- Weights RMS per Epoch for Each Layer: This will allow us to track the evolution and scaling of weights for each layer as training progresses, offering insights into weight dynamics.
- Round Loss: This metric reflects the loss incurred during each training round, helping us gauge the immediate impact of different AFs on training efficacy.
- Validation Loss: Monitoring validation loss will enable us to assess how well the network generalizes to unseen data, providing a measure of the practical performance of each AF.

By systematically analyzing these metrics, we aim to gain a comprehensive understanding of how different AFs influence the training dynamics and overall performance of the NN.

*Experiment Design:*

- Generating Training Data: The synthetic training data is generated using a Gaussian-modulated exponential function, $f(x) = e^{-x^2}$, which produces a smooth, bell-shaped curve. To simulate real-world data variability, Gaussian noise is added to the function values. The noise level is set to 0.15, meaning the noise is sampled from a normal distribution with a mean of 0 and a standard deviation of 0.15. The training data is generated for $x$ values ranging from $-3$ to 3 with a step size of 0.01, resulting in a dense set of data points.
- AFs: The experiment evaluates 12 different AFs: ReLU, ELU, SELU, GELU, Swish, HardSwish, Mish, SoftPlus, HardTanh, HardSigmoid, Sigmoid, and Tanh. Each AF has unique properties affecting gradient flow and training stability.
- NN Architecture: The NN consists of three hidden layers, each followed by an AF from the set being evaluated. Each hidden layer has 10 neurons followed by the specified AF. The output layer is a linear layer with one neuron, suitable for regression tasks.

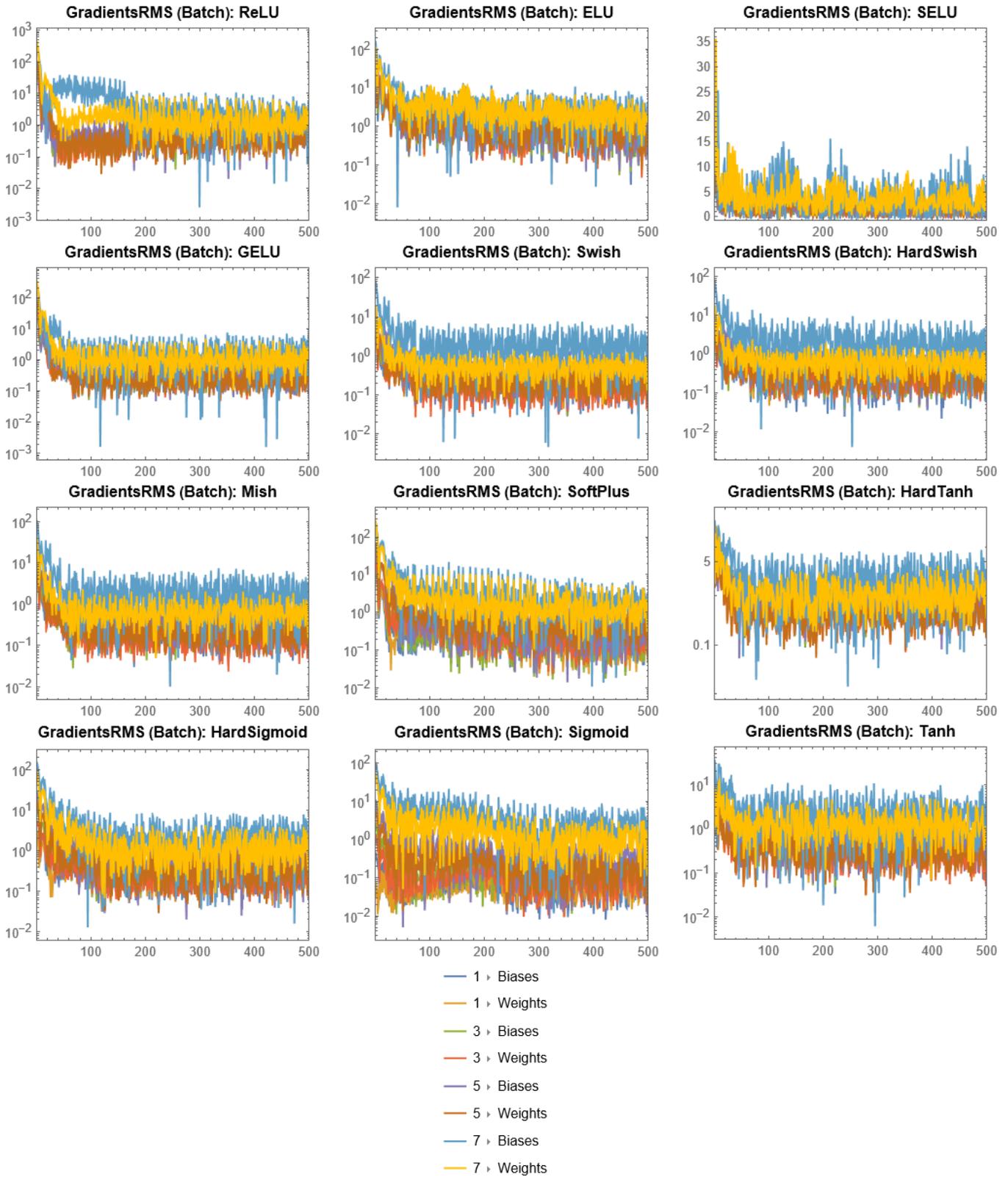

**Fig. 80. (Regression Experiment)** This set of subplots illustrates the Gradients RMS evolution for NNs trained with different AFs: ReLU, ELU, SELU, GELU, Swish, HardSwish, Mish, SoftPlus, HardTanh, HardSigmoid, Sigmoid, and Tanh. Each subplot presents the Gradients RMS monitored at each training batch over a maximum of 50 training rounds (epochs) for each layer. The $x$-axis represents the training batches, while the $y$-axis shows the Gradients RMS values. Monitoring GradientsRMS helps understand the stability and convergence of the training process.

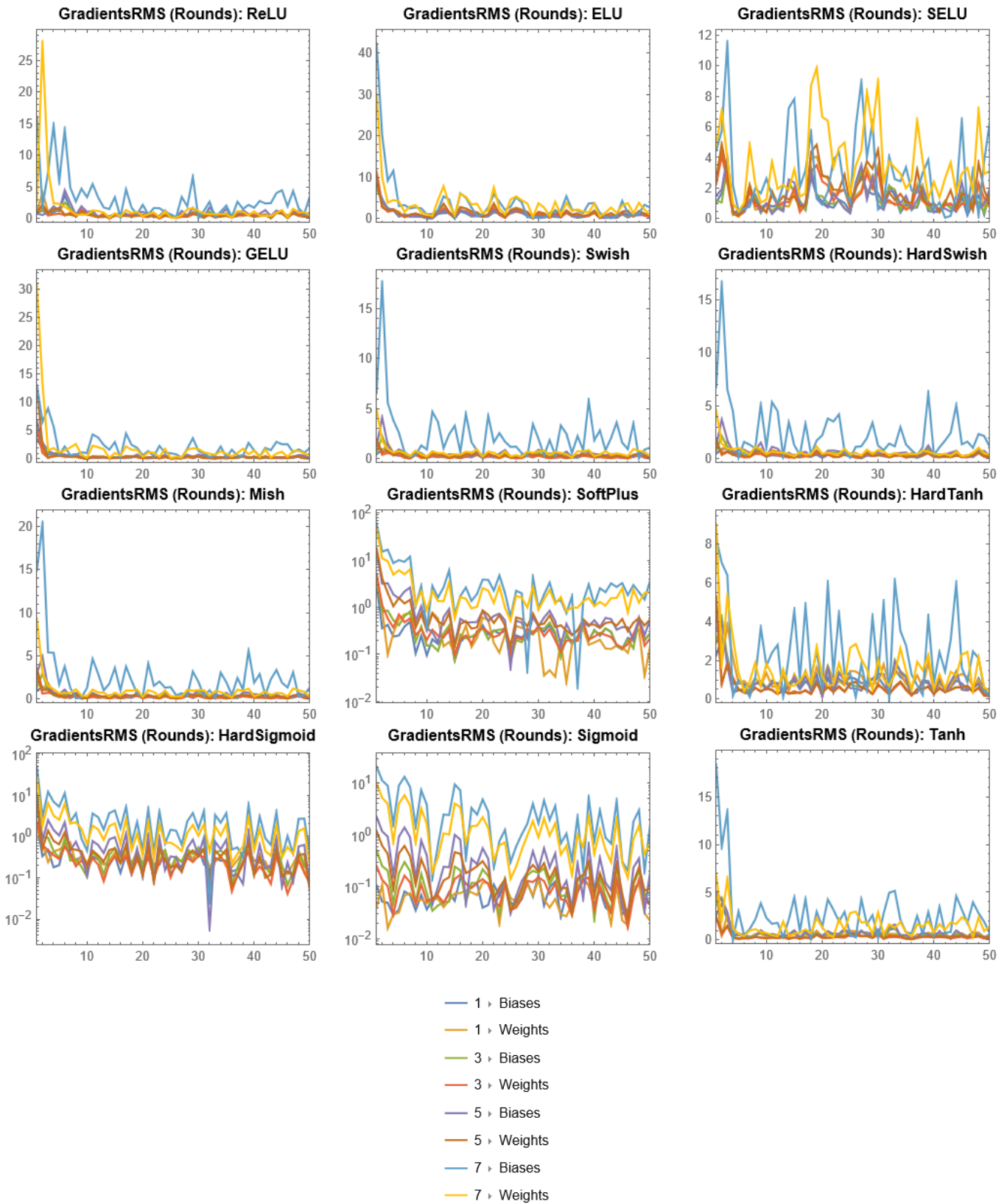

**Fig. 81. (Regression Experiment)** Similar to Fig.80 but each subplot presents the Gradients RMS monitored at each training round (epoch) over a maximum of 50 training rounds for each layer.

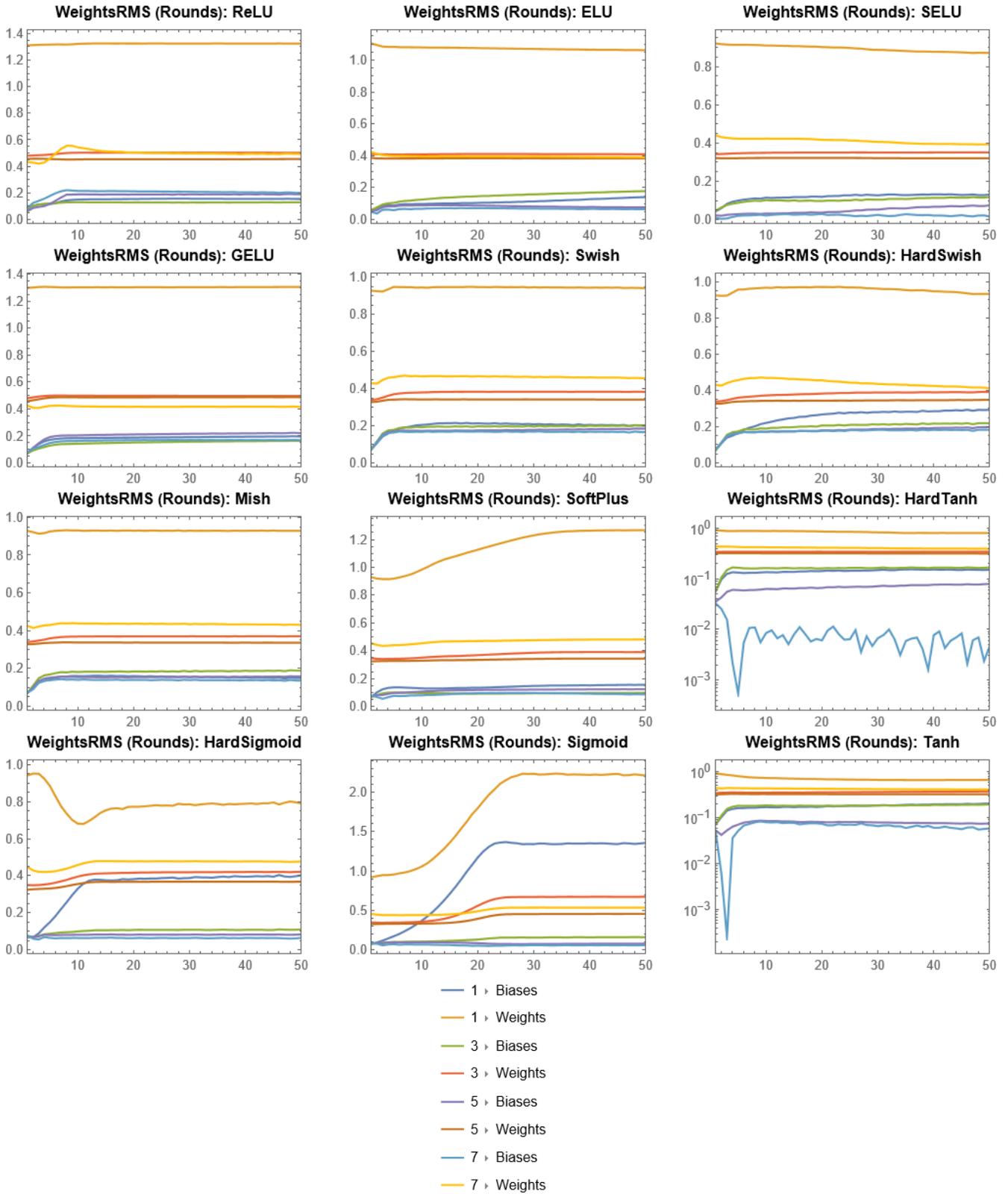

**Fig. 82. (Regression Experiment)** This set of subplots illustrates the weight RMS evolution for NNs trained with different AFs: ReLU, ELU, SELU, GELU, Swish, HardSwish, Mish, SoftPlus, HardTanh, HardSigmoid, Sigmoid, and Tanh. Each subplot presents the weight RMS monitored at each training round over a maximum of 50 training rounds (epochs) for each layer. The $x$-axis represents the training rounds, while the $y$-axis shows the weight RMS values.

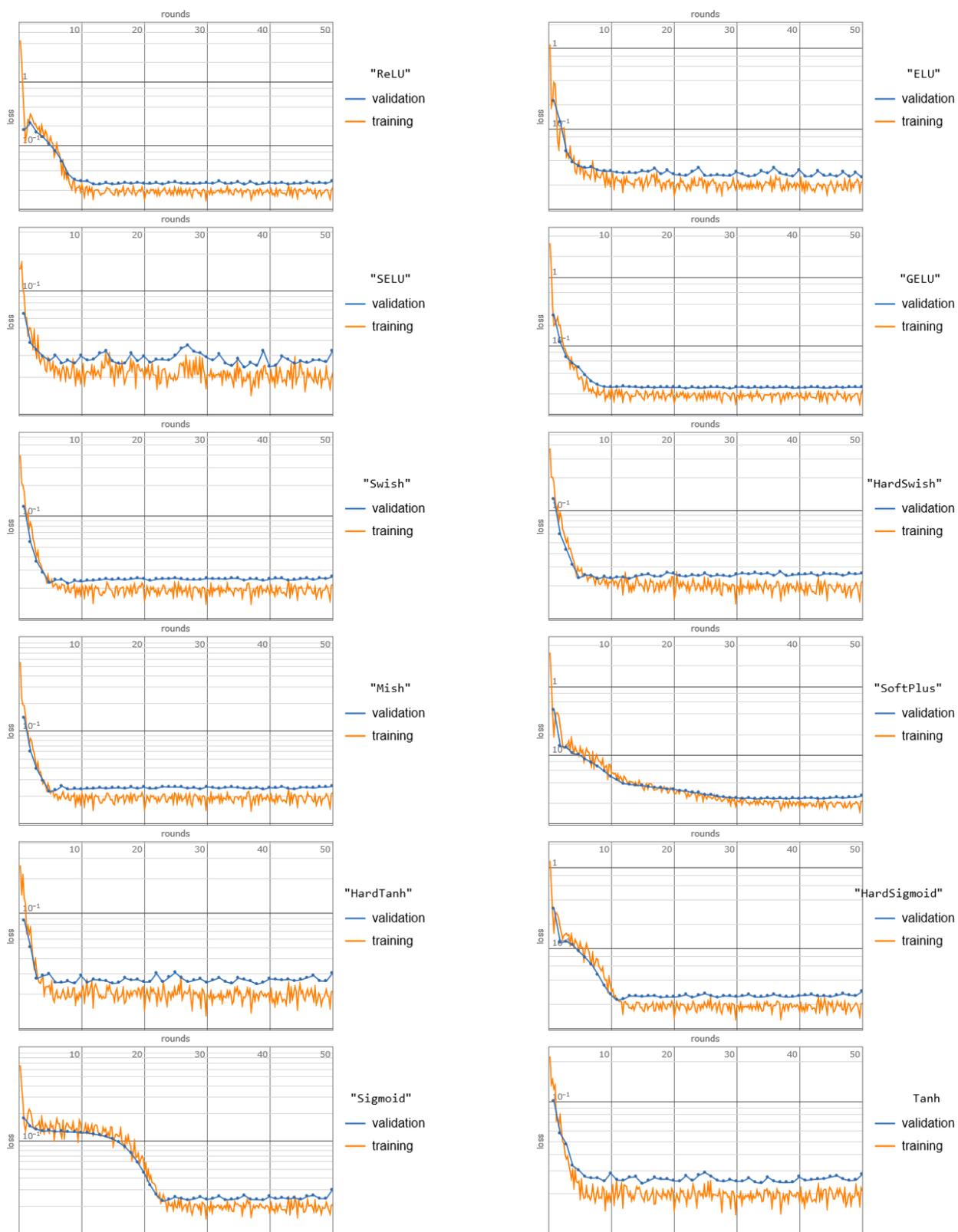

**Fig. 83. (Regression Experiment)** These plots show the training and validation loss (mean squared error) over 50 training rounds for NNs using various AFs. This visual representation allows for the comparison of training dynamics, highlighting how quickly and effectively each AF minimizes the training and validation loss.

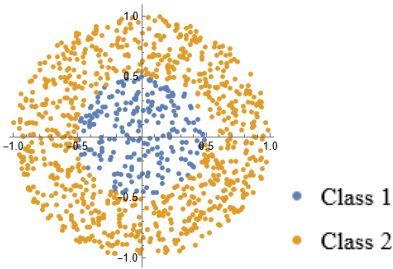

**Fig. 84.** The synthetic training data generated within a unit disk, labeled based on the distance from the origin. Points are colored differently to represent two classes: Class 1 (inside radius 0.5) and Class 2 (outside radius 0.5 but within the unit disk).

- Training Process: Mean Squared Loss (MSE) is used as the training objective, which measures the average squared difference between the predicted and actual values. The Adam optimizer is used for training, known for its adaptive learning rate capabilities and efficiency in handling sparse gradients. The learning rate is set to 0.01, a commonly used value for the Adam optimizer. The training data is processed in batches of 64 samples, balancing computational efficiency and gradient estimation accuracy. The training runs for a maximum of 50 epochs (rounds).
- Monitoring Metrics: During training, the Gradients RMS per batch, Fig. 80, Gradients RMS per epoch, Fig. 81, weights RMS per epoch, Fig. 82, round loss and validation loss, Fig. 83, are monitored to ensure proper training dynamics. The evolutions of metrics are plotted for each AF, showing how the metrics change over the training iterations.

*9.2 Classification Experiment*

*Objective*:
The primary objective of this experiment is to generate synthetic training data points within a unit disk, label them based on their distance from the origin, and evaluate the performance of various AFs in a NN. This experiment aims to provide insights into how different AFs impact the training dynamics and classification performance of the NN.

*Experiment Design*:

- Generating Training Data: 1000 random points are generated within a unit disk centered at the origin. Each point is labeled based on its distance from the origin. Points within a radius of 0.5 are labeled as class 1, and points outside this radius but within the unit disk are labeled as class 2. The synthetic training data is visualized using a scatter plot, Fig. 84, where points from class 1 and class 2 are shown in different colors. This visualization helps inspect the data distribution and verify the labeling process.
- AFs: The experiment evaluates 12 different AFs: ReLU, ELU, SELU, GELU, Swish, HardSwish, Mish, SoftPlus, HardTanh, HardSigmoid, Sigmoid, and Tanh. Each AF has unique properties affecting gradient flow, training stability, and classification performance.
- NN Architecture: The NN consists of two hidden layers, each with 5 neurons followed by one of the specified AFs. The output layer uses a Logistic Sigmoid AF for binary classification. The output layer is followed by a NetDecoder set to "Boolean" to decode the network output into binary labels.
- Training Process: The Adam optimizer is used for training, known for its adaptive learning rate capabilities and efficiency in handling sparse gradients. The learning rate is set to 0.01. The training data is processed in batches of 64 samples. The training runs for a maximum of 50 iterations (rounds).
- Monitoring Metrics: During training, the Gradients RMS per batch, Fig. 85, Gradients RMS per epoch, Fig. 86, weights RMS per epoch, Fig. 87, round loss and validation loss, Fig. 88, are monitored to ensure proper training dynamics. The evolutions of metrics are plotted for each AF, showing how the metrics change over the training iterations. Moreover, the experiment monitors training progress through confusion matrix plots, Fig. 89, compares the results, and visualizes the decision boundaries, Fig. 90, to understand the influence of each AF on training dynamics and classification performance.

This experiment provides a comprehensive analysis of the impact of different AFs on NN training and classification performance. By systematically evaluating and comparing the validation accuracy, confusion matrices, loss plots, and decision boundaries, the experiment highlights the strengths and weaknesses of each AF. These insights guide the selection of appropriate AFs for specific NN architectures and tasks, aiding in the design of more effective NNs.

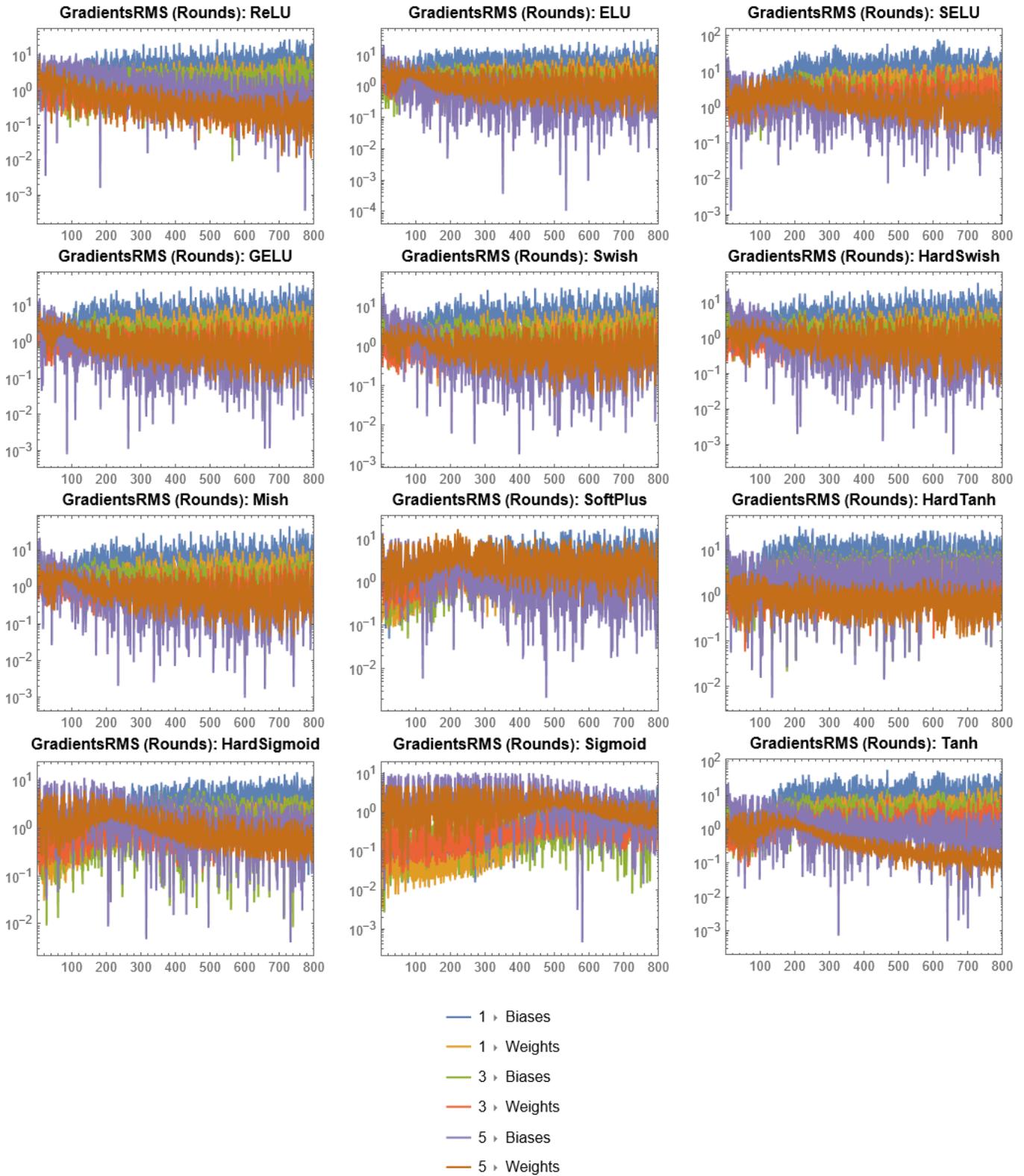

**Fig. 85. (Classification Experiment)** The figure consists of a series of 12 subplots, each depicting the Gradients RMS evolution for a NN trained with a distinct AF: ReLU, ELU, SELU, GELU, Swish, HardSwish, Mish, SoftPlus, HardTanh, HardSigmoid, Sigmoid, and Tanh. Each subplot presents the Gradients RMS monitored at each training batch over a maximum of 50 training rounds (epochs) for each layer. The $x$-axis represents the training batches, while the $y$-axis shows the Gradients RMS values. Monitoring Gradients RMS helps understand the stability and convergence of the training process.

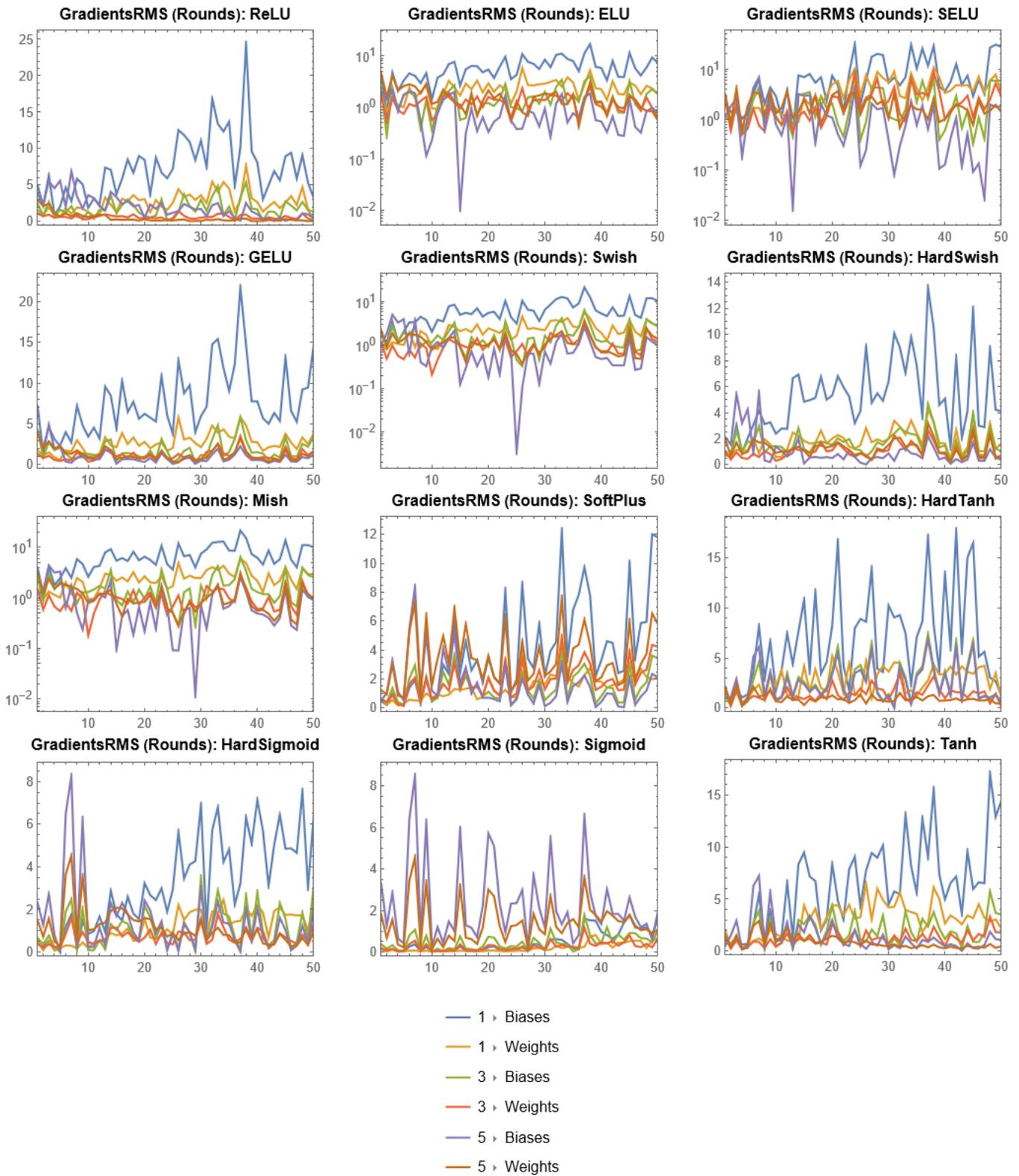

**Fig. 86. (Classification Experiment)** Similar to Fig.85 but each subplot presents the Gradients RMS monitored at each training round over a maximum of 50 training rounds for each layer.

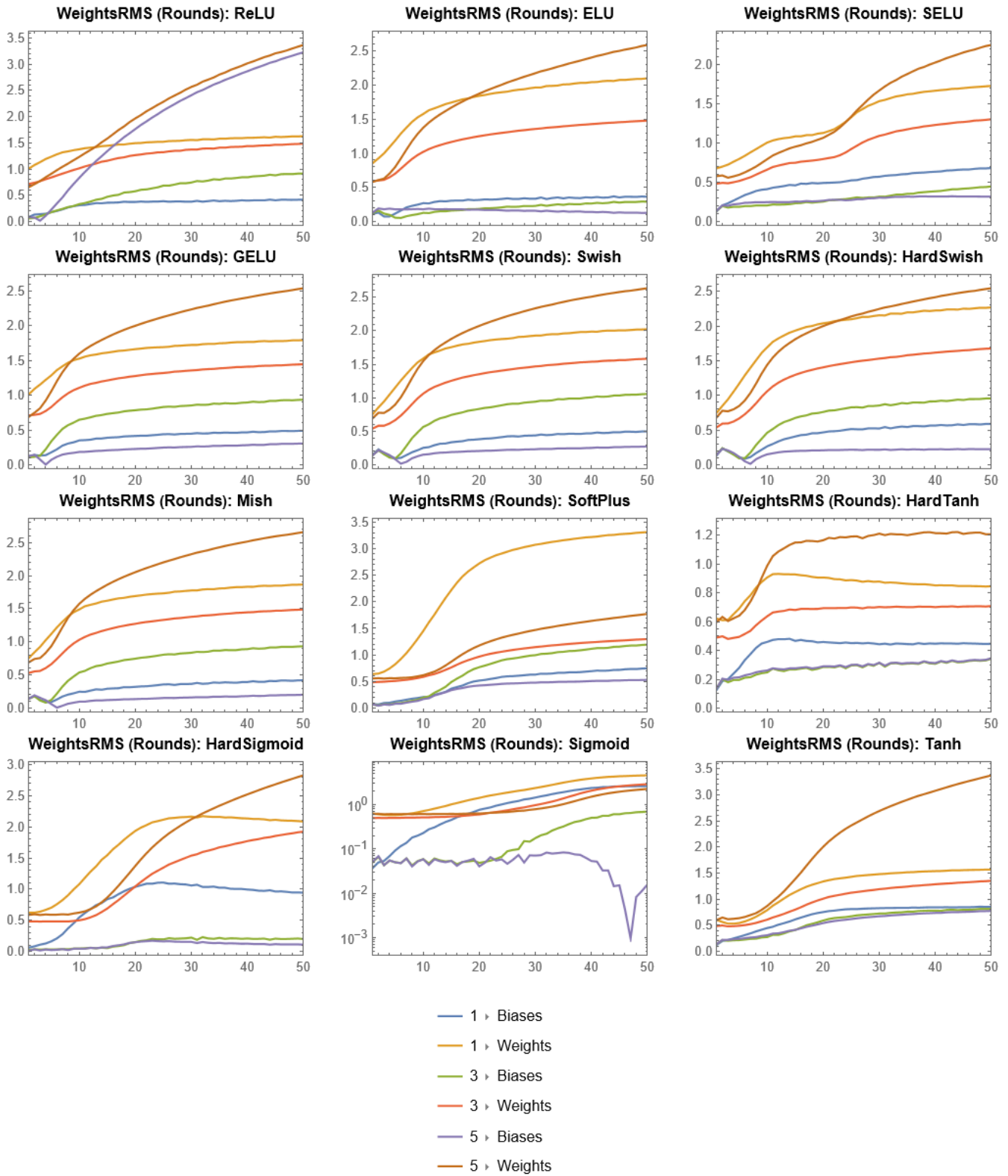

**Fig. 87. (Classification Experiment)** This set of subplots illustrates the weight RMS evolution for NNs trained with different AFs: ReLU, ELU, SELU, GELU, Swish, HardSwish, Mish, SoftPlus, HardTanh, HardSigmoid, Sigmoid, and Tanh. Each subplot presents the weight RMS monitored at each training round over a maximum of 50 training rounds (epochs) for each layer. The $x$-axis represents the training rounds, while the $y$-axis shows the weight RMS values.

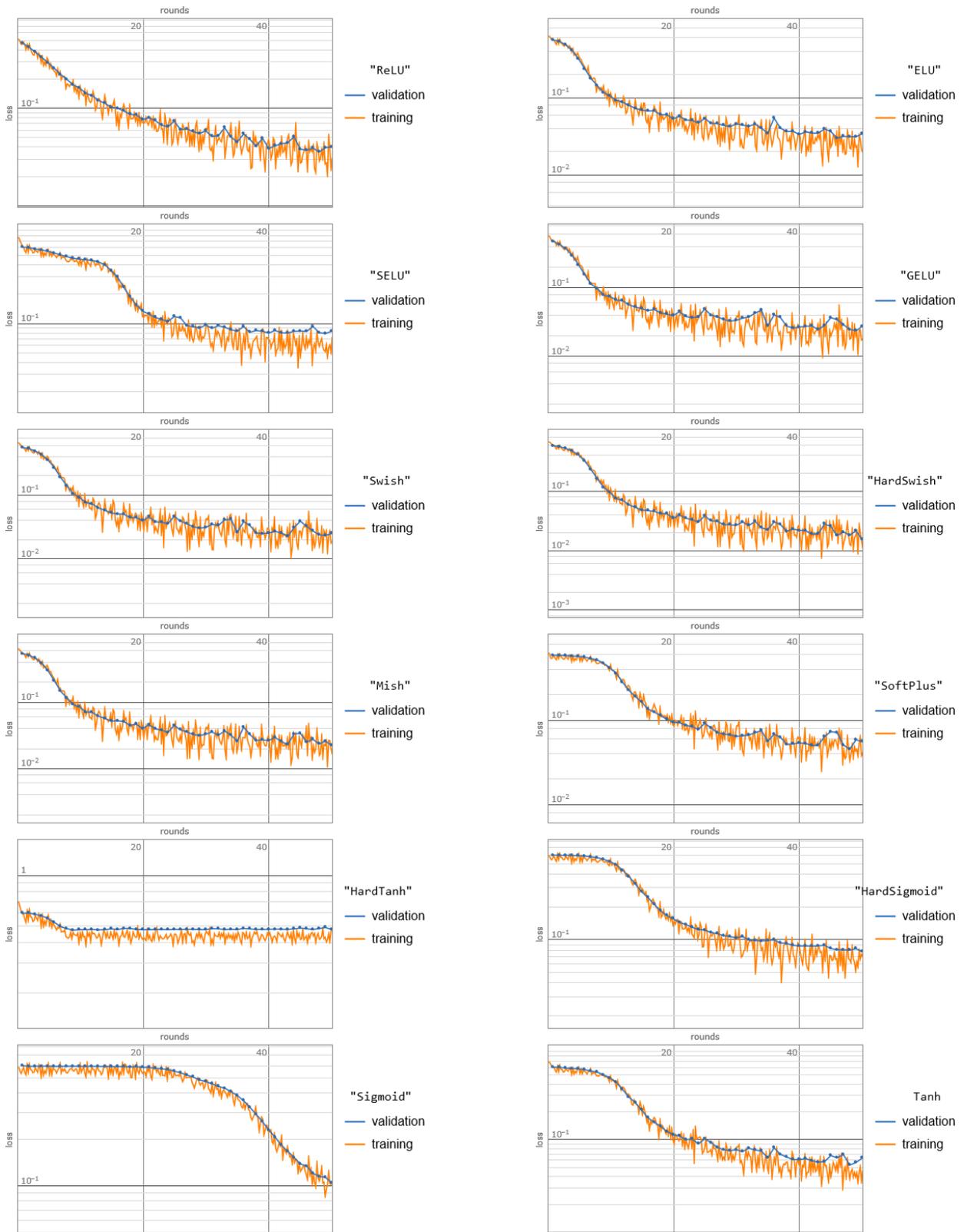

**Fig. 88. (Classification Experiment)** The figure consists of 12 plots, each depicting the training and validation loss evolution for a NN trained with a distinct AF for classification. The loss plots are essential for understanding how quickly and effectively each AF minimizes the loss during training and validation. Comparing these plots helps identify which AFs provide the fastest and most stable convergence. Functions that show consistent and rapid loss reduction are preferred for their efficiency in optimizing the NN.

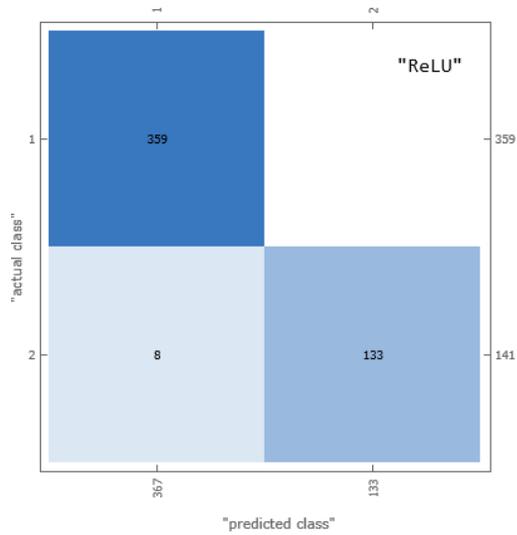
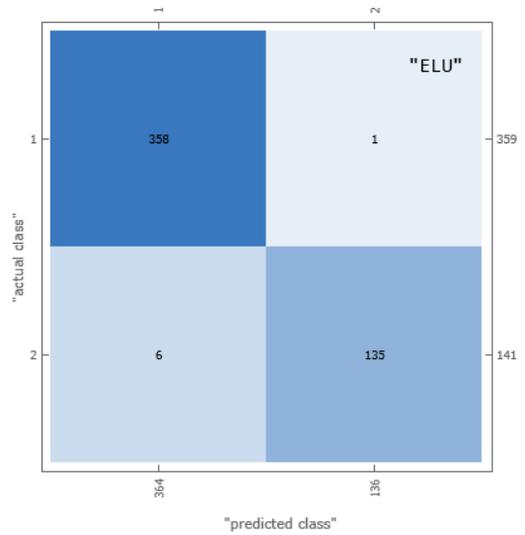
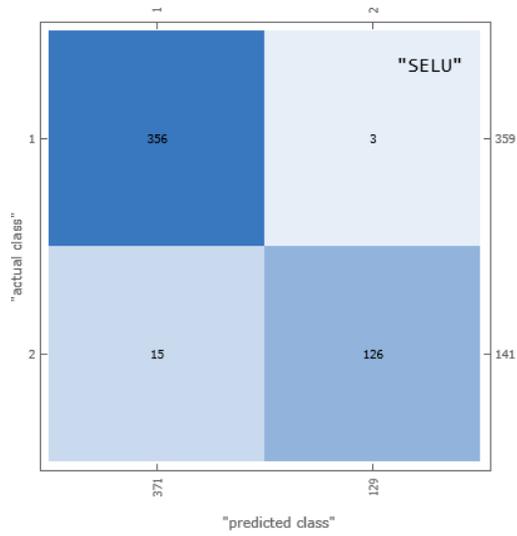
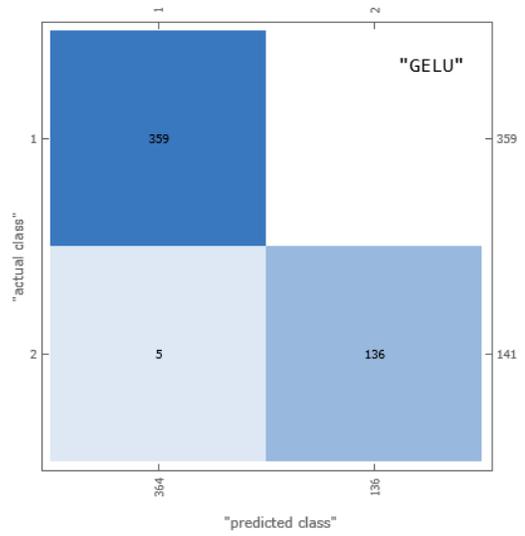
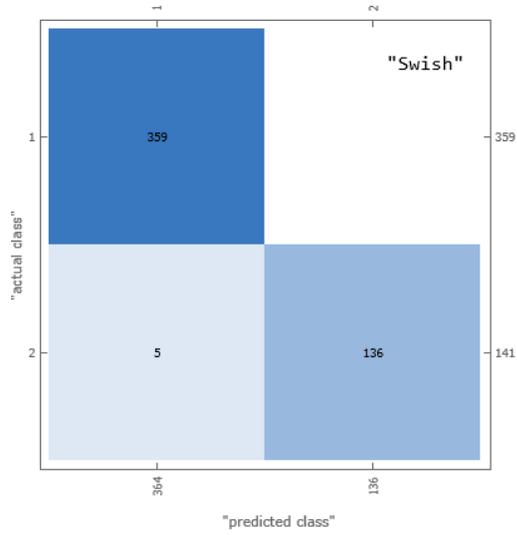
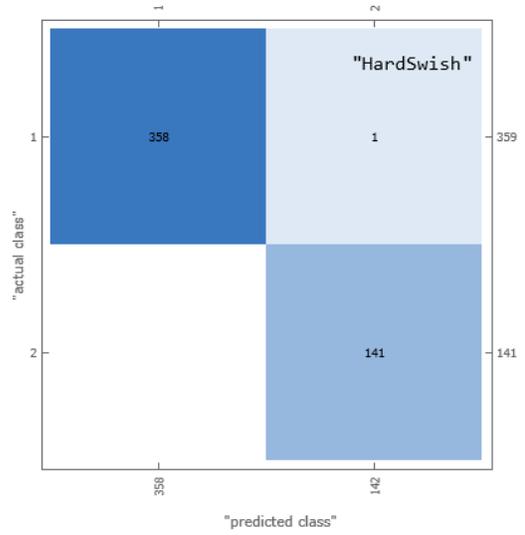

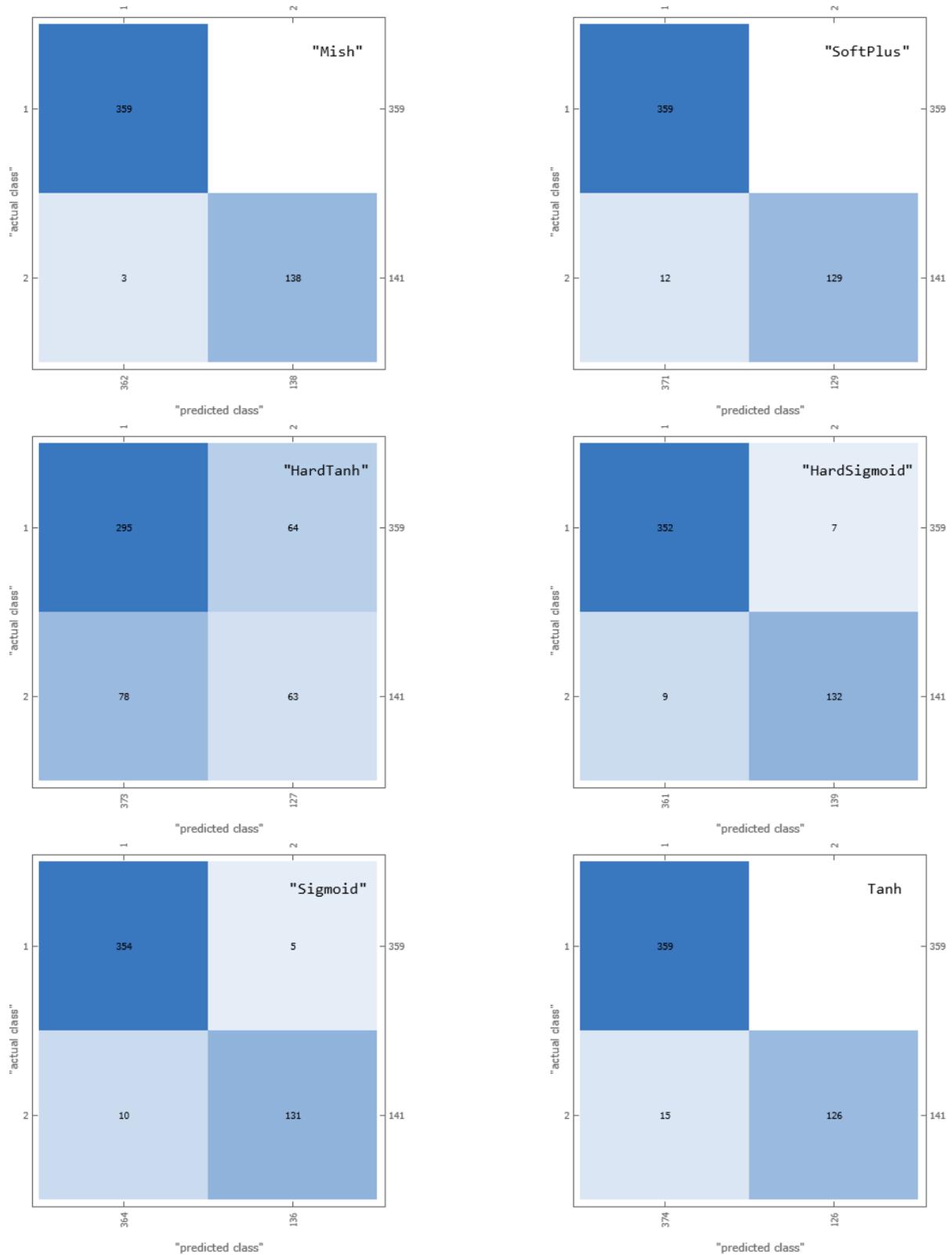

**Fig. 89. (Classification Experiment)** The figure consists of 12 plots, each depicting the confusion matrix for a NN trained with a distinct AF. The confusion matrix provides detailed insights into the classification performance. The *x*-axis and *y*-axis represent the predicted and actual classes, respectively. The plot title includes the AF used. The entries in the matrix show the counts of true positives, true negatives, false positives, and false negatives. A balanced and diagonal-heavy confusion matrix indicates good classification performance. Comparing these plots helps identify which AFs result in better discrimination between classes.

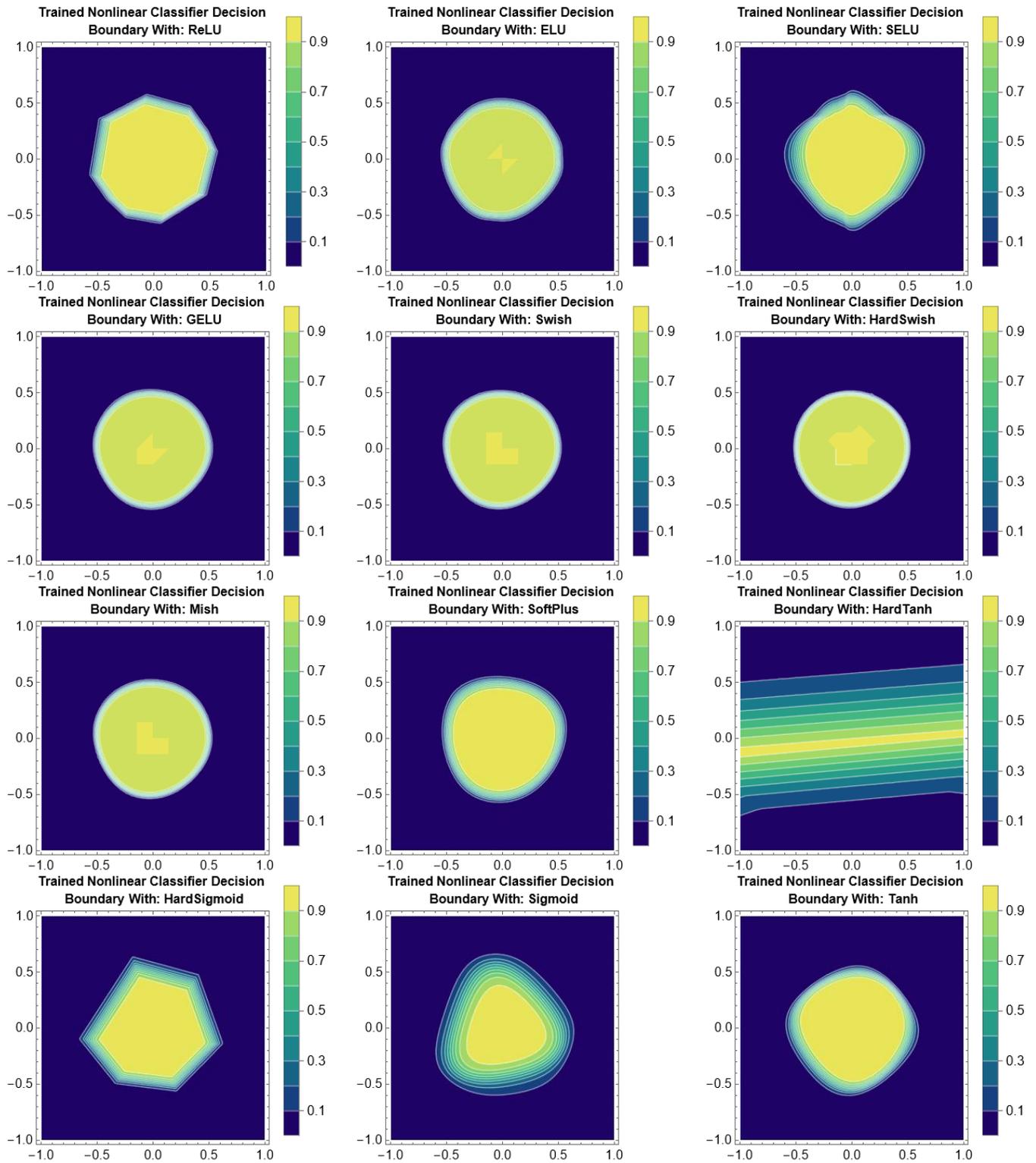

**Fig. 90. (Classification Experiment)** The figure consists of 12 plots, each depicting the decision boundary for a NN trained with a distinct AF. The decision boundary illustrates how the network classifies different regions of the input space. The $x$-axis and $y$-axis represent the coordinates of the input space within the unit disk. The plot title includes the AF used. Different colors in the plot represent different classes. A well-defined and smooth decision boundary indicates effective learning and classification by the network. Comparing these plots helps understand the influence of each AF on the network's decision-making process. The figure provides a comprehensive visualization of how different AFs affect the classification regions, aiding in the selection of suitable AFs for future NN designs.

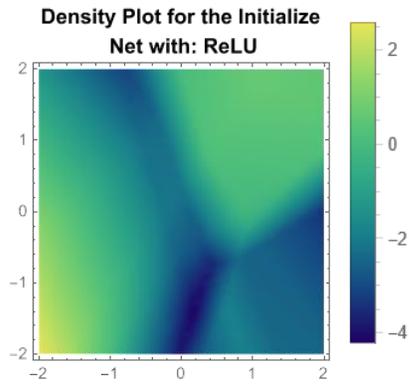
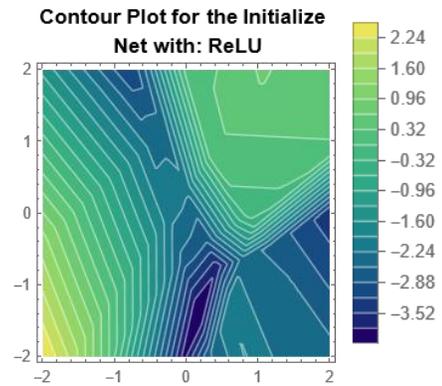
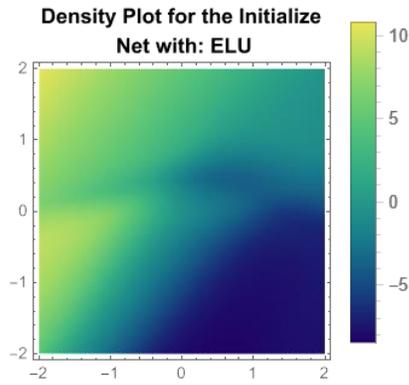
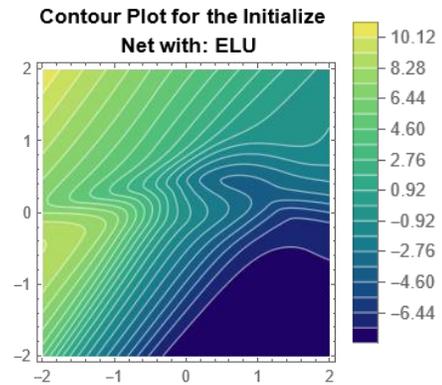
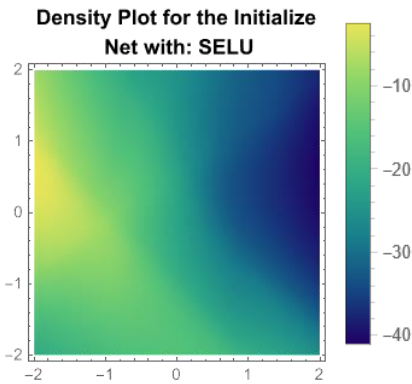
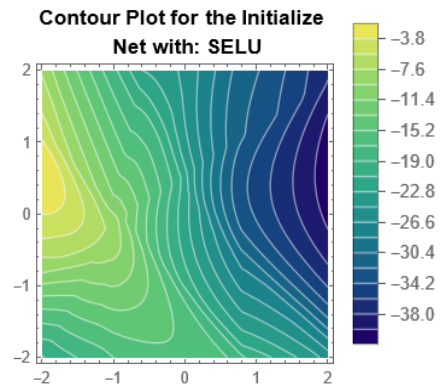
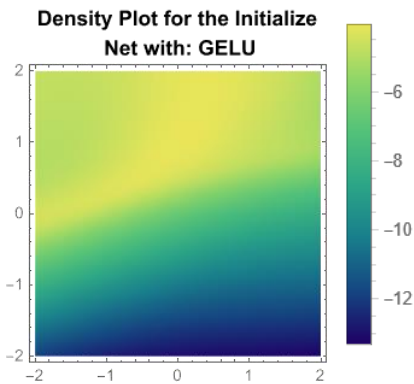
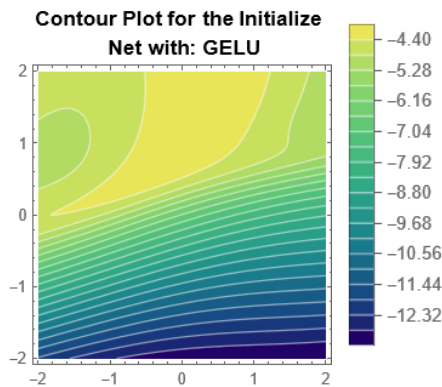

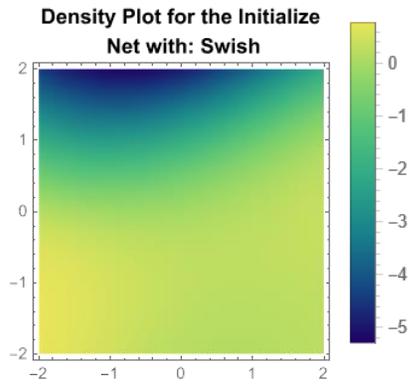
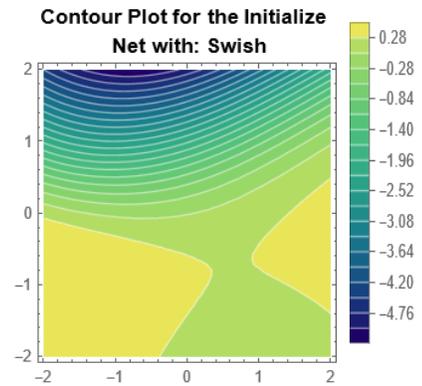
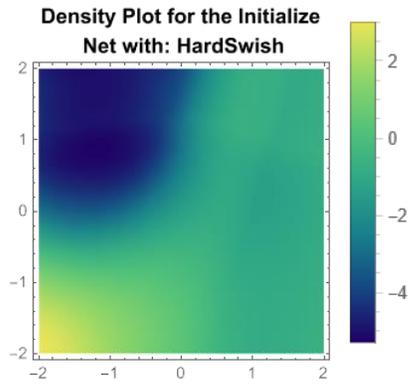
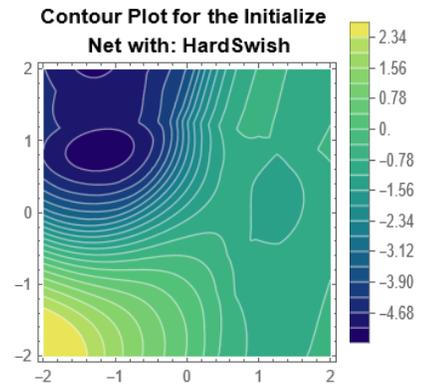
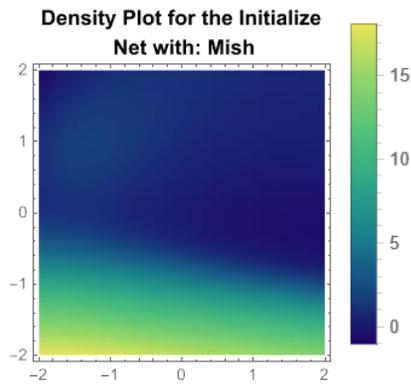
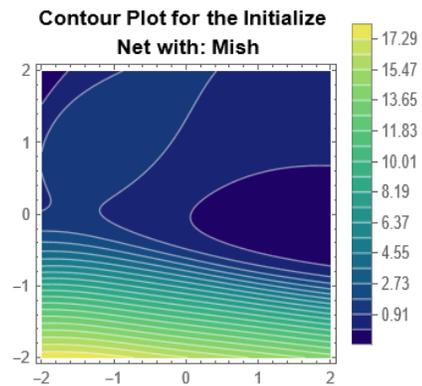
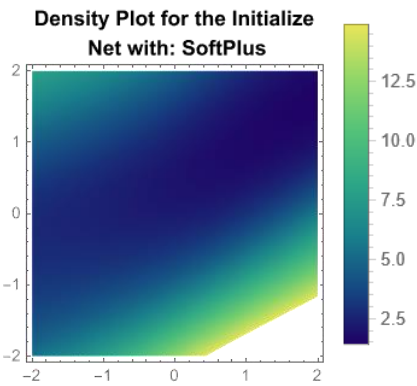
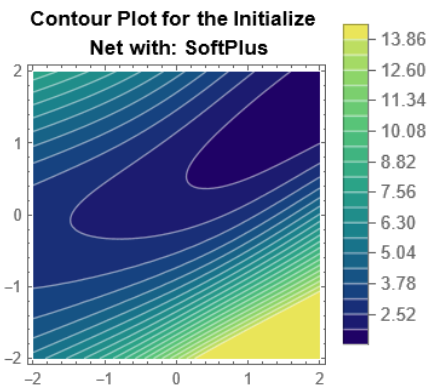

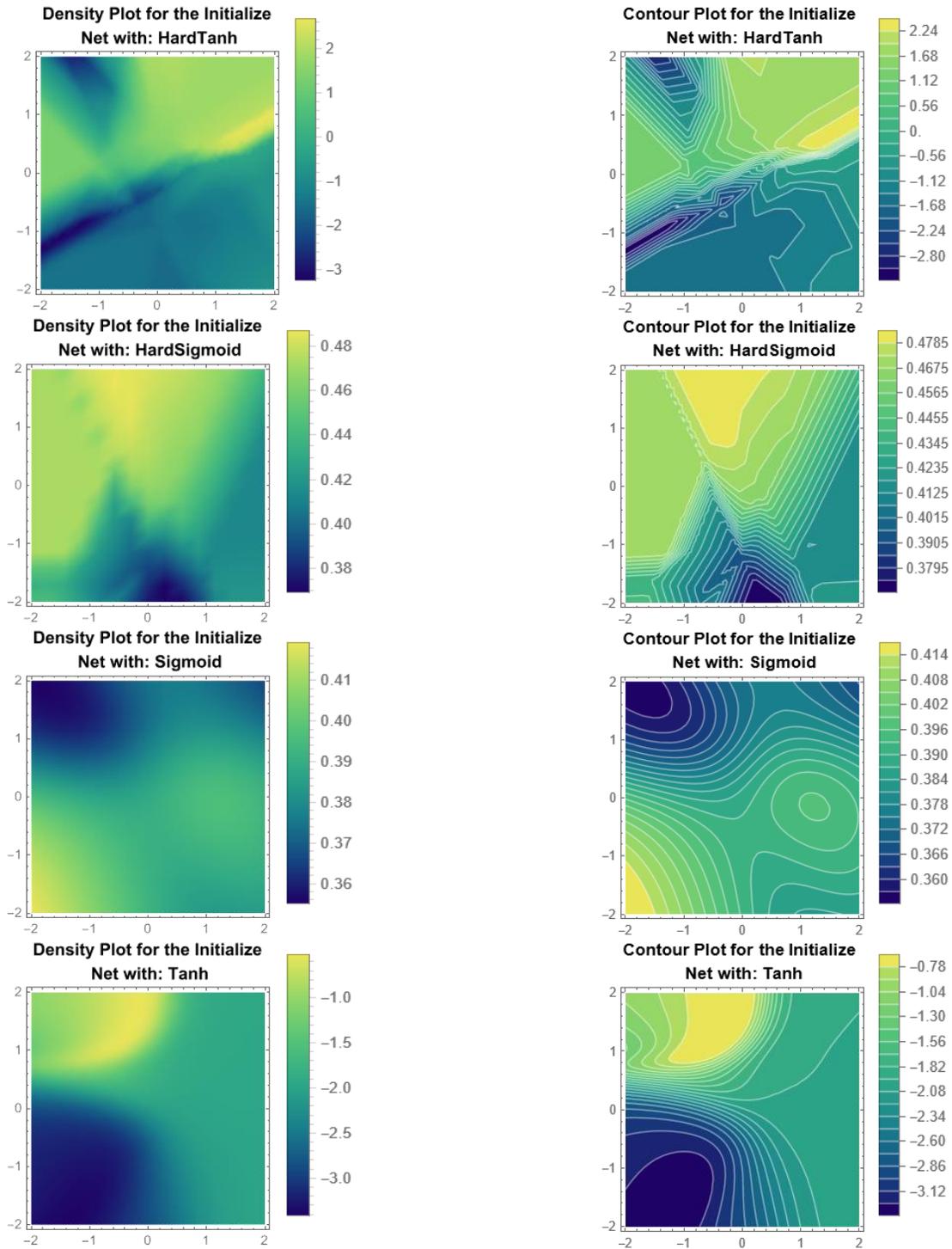

**Fig. 91. (Smooth Output Landscape)** Visualization of Output Landscapes for Various AFs. This figure showcases the output landscapes of a 3-layer randomly initialized NN, visualized using density and contour plots, for different AFs. The AFs tested include ReLU, ELU, SELU, GELU, Swish, HardSwish, Mish, SoftPlus, HardTanh, HardSigmoid, Sigmoid, and Tanh. Each subplot represents the network's scalar output evaluated over a 2-dimensional grid ranging from $-2$ to $2$ in both $x$ and $y$ dimensions. Left panels: (Density Plots) These plots display the network's output values using a gradient color scheme (Blue-Green-Yellow), highlighting regions of varying output intensity. Right panels: (Contour Plots) These plots present the network's output as contour lines, providing a clear view of the smoothness and sharpness in the output landscape. The visualizations illustrate the dramatic effect that different AFs have on the smoothness and structure of the output landscape. The smoother landscapes observed with functions like Swish contrast with the sharp and chaotic regions produced by ReLU, underscoring the importance of AF selection in NN optimization and generalization.

## 10. Conclusion

In this paper, we have provided a comprehensive review and systematic classification of AFs in NNs, emphasizing their essential role in enabling non-linearity, learning complex patterns, and enhancing the performance of deep learning models. Our exploration encompassed a diverse array of AFs, including fixed-shape, parametric, adaptive, stochastic/probabilistic, non-standard, and ensemble/combining types. We meticulously discussed their theoretical foundations, mathematical formulations, and practical implications, highlighting the distinct benefits and limitations of each. Through our detailed analysis, we identified the key attributes that influence the effectiveness of AFs, such as output range, monotonicity, and smoothness. We addressed critical challenges in NN optimization, such as the vanishing and exploding gradient problems, internal covariate shift, and the dying ReLU phenomenon.

Our survey also showcased recent advancements in developing adaptive and non-standard AFs that offer enhanced adaptability and performance. The adaptive nature of AFs can be realized through various approaches. While these adaptive activations involve higher computational costs, they frequently surpass the performance of their fixed counterparts. Our discussion has highlighted state-of-the-art adaptive AFs that dynamically adjust to specific requirements. We explored several methods for adapting AFs, including the introduction of tunable parameters and the application of ensemble techniques, where a combination of multiple AFs results in superior outcomes. Furthermore, the development of adaptive AFs can be achieved by incorporating stochastic parameters, offering a promising direction for enhancing NN performance.

The comparative evaluation of 12 state-of-the-art AFs, based on rigorous statistical and experimental methodologies, provided valuable insights into their efficacy. This evaluation assists practitioners in making informed decisions when selecting and designing AFs for specific deep learning tasks, thereby improving the efficiency and accuracy of their models.

We also visualized the output landscape of a 3-layer randomly initialized NN with various AFs such as ReLU, ELU, SELU, GELU, Swish, HardSwish, Mish, SoftPlus, HardTanh, HardSigmoid, Sigmoid, and Tanh for visual clarity, as depicted in Fig. 91. Specifically, we input the 2-dimensional coordinates of each position in a grid into the network and plotted the scalar network output for each grid point. Our observations indicate that AFs significantly influence the smoothness of output landscapes.

ReLU, HardTanh, HardSigmoid, and HardSwish each uniquely affect NN performance. ReLU enhances sparsity and efficiency by producing zero output for negative inputs, but its abrupt transition at zero causes sharp regions in the output landscape (see Fig. 91), posing challenges for gradient-based optimization. HardTanh, which approximates the Tanh function, provides some sparsity with constant gradients within the range of $-1$ to 1, but introduces sharp transitions at $-1$ and 1, potentially affecting optimization. HardSigmoid, a piecewise linear approximation of the Sigmoid function, offers smoother transitions with a constant gradient within the range of $-2.5$ to 2.5, yet still has sharp regions at the boundaries, impacting optimization. HardSwish, a piecewise linear approximation of Swish, maintains smoothness within $[-3, 3]$, promoting better gradient flow and mitigating the vanishing gradient problem while being computationally efficient. These four functions can create non-smooth regions in the output landscape (and loss landscape), potentially leading to issues in gradient-based optimization. Conversely, networks using ELU, SELU, GELU, Swish, Mish, SoftPlus, Sigmoid, and Tanh exhibit considerably smoother output landscapes. Smoother output landscapes lead to smoother loss landscapes, which are easier to optimize and result in improved training and test accuracy.